\RequirePackage[l2tabu, orthodox]{nag}

\documentclass[12pt,phd,a4paper,twoside]{ucl_thesis}

\usepackage[T1]{fontenc}
\usepackage[utf8]{inputenc}
\usepackage{blindtext}
\usepackage{emptypage}
\usepackage{setspace}
\usepackage{pdfpages}

\usepackage{graphicx}
\usepackage{float}
\usepackage{subfigure}
\usepackage[format=hang,font=small,labelfont=bf]{caption}

\usepackage{xcolor,colortbl}

\usepackage{multirow}
\usepackage{tabularray}
\usepackage{tabularx}
\usepackage{array}
\usepackage{arydshln}
\usepackage{booktabs}
\usepackage{makecell}
\usepackage{changepage}
\usepackage{enumitem}
\newcommand{\subscript}[2]{$#1 _ #2$}
\usepackage[xindy,acronym,sort=def]{glossaries}

\usepackage{amssymb}
\usepackage{amsmath}
\usepackage{latexsym}
\usepackage{pifont}
\usepackage{bm}
\usepackage[mathcal]{eucal}

\usepackage{url}
\usepackage{hyperref}
\hypersetup{
 unicode=false,
 pdftoolbar=true,
 pdfmenubar=true,
 pdffitwindow=false,
 pdfstartview={FitH},
 pdfauthor={Zhengxiang Shi},
 pdfnewwindow=true,
 colorlinks=true,
 linkcolor=dred2,
 citecolor=dgreen,
 filecolor=magenta,
 urlcolor=dblue
}

\usepackage{cleveref}
\usepackage{natbib}
\usepackage{bibentry}
\crefformat{section}{\S#2#1#3}
\crefformat{subsection}{\S#2#1#3}
\newcommand\sect[1]{\S\ref{#1}}

\usepackage{etoolbox}
\usepackage[outline]{contour}
\usepackage{adjustbox}
\usepackage{algorithm}
\usepackage{algorithmic}
\usepackage{listings}
\usepackage{csquotes}
\usepackage{amsthm}
\usepackage{xspace}

\definecolor{tablebrown}{RGB}{139,69,19}
\definecolor{tableblue}{RGB}{0, 123, 255}
\definecolor{dbrown}{RGB}{140, 106, 93}
\definecolor{dblue}{RGB}{52, 104, 192}
\definecolor{dgreen}{RGB}{65, 171, 93}
\definecolor{dred}{RGB}{210, 69, 69}
\definecolor{dred2}{RGB}{169, 68, 56}
\definecolor{legend1}{HTML}{66c2a4}
\definecolor{legend2}{HTML}{fa8c62}
\definecolor{legend3}{HTML}{8da0cb}
\definecolor{cid}{HTML}{dae8f5}
\definecolor{ccon}{HTML}{fee9d4}
\definecolor{gred}{HTML}{cc0200}
\definecolor{ggreen}{HTML}{4C9F26}
\definecolor{Gray}{gray}{0.93}

\newcommand\imdb{\textsc{IMDB}\xspace}
\newcommand\ag{\textsc{AG News}\xspace}
\newcommand\amazon{\textsc{Amazon Review}\xspace}
\newcommand\yelp{\textsc{Yelp Review}\xspace}

\newcommand\sst{\textsc{SST-2}\xspace}
\newcommand\yahoo{\textsc{Yahoo! Answer}\xspace}

\newcommand\llamas{\textsc{Llama-2-7B}\xspace}
\newcommand\llamam{\textsc{Llama-2-13B}\xspace}
\newcommand\llama{\textsc{Llama-2}\xspace}
\newcommand\llamabase{\textsc{Llama-2-}{\footnotesize \textsc{Base}}\xspace}
\newcommand\llamachat{\textsc{Llama-2-}{\footnotesize \textsc{Chat}}\xspace}
\newcommand\pt{\textsc{PT}\xspace}

\newcommand\dept{\textsc{DePT}\xspace}
\newcommand\peft{\textsc{PEFT}\xspace}
\newcommand\petl{\textsc{PETL}\xspace}

\newcommand\vat{\textsc{VAT}\xspace}

\newcommand\fixmatch{\textsc{FixMatch}\xspace}
\newcommand\dash{\textsc{Dash}\xspace}
\newcommand\flexmatch{\textsc{FlexMatch}\xspace}
\newcommand\adamatch{\textsc{AdaMatch}\xspace}
\newcommand\tapt{\textsc{TAPT}\xspace}

\newcommand\pcp{\textsc{PCP}\xspace}
\newcommand\cls{\texttt{CLS}\xspace}
\newcommand\ft{\textsc{FT}\xspace}
\newcommand\st{\textsc{ST}\xspace}
\newcommand\ssl{\textsc{SSL}\xspace}

\newcommand\nlp{\textsc{NLP}\xspace}
\newcommand\mlm{\textsc{MLM}\xspace}
\newcommand\neftune{\textsc{Neftune}\xspace}
\newcommand\im{\textsc{IM}\xspace}
\newcommand\sft{\textsc{IT}\xspace}
\newcommand\sah{\textsc{SAH}\xspace}

\newcommand\partials{\textsc{Supervised}\xspace}
\newcommand\fullys{\textsc{Fully-Supervised}\xspace}
\newcommand\bert{\textsc{Bert}\xspace}
\newcommand\roberta{\textsc{RoBERTa}\xspace}
\newcommand\robertalarge{\textsc{RoBERTa-}{\footnotesize \textsc{Large}}\xspace}
\newcommand\robertabase{\textsc{RoBERTa-}{\footnotesize \textsc{Base}}\xspace}
\newcommand\tfbase{\textsc{T5-}{\footnotesize \textsc{Base}}\xspace}
\newcommand\tfsmall{\textsc{T5-}{\footnotesize \textsc{Small}}\xspace}
\newcommand\tflarge{\textsc{T5-}{\footnotesize \textsc{Large}}\xspace}
\newcommand\gptsmall{\textsc{GPT2-}{\footnotesize \textsc{Small}}\xspace}
\newcommand\gptmedium{\textsc{GPT2-}{\footnotesize \textsc{Medium}}\xspace}
\newcommand\gptlarge{\textsc{GPT2-}{\footnotesize \textsc{Large}}\xspace}

\newcommand\flan{\texttt{Flan V2}\xspace}
\newcommand\wizardlm{\texttt{WizardLM}\xspace}
\newcommand\sharegpt{\texttt{Sharegpt}\xspace}
\newcommand\alpagasus{\texttt{Alpagasus}\xspace}
\newcommand\alpagasusdollyone{\texttt{Alpagasus Dolly 3k}\xspace}
\newcommand\alpagasusdollytwo{\texttt{Alpagasus Dolly 9k}\xspace}
\newcommand\alpagasusalpaca{\texttt{Alpagasus Alpaca 5k}\xspace}
\newcommand\less{\texttt{Less}\xspace}
\newcommand\mmluchat{\texttt{Less MMLU Chat}\xspace}
\newcommand\tydiqa{\texttt{Less Tydiqa}\xspace}
\newcommand\bbhicl{\texttt{Less BBH ICL}\xspace}
\newcommand\lima{\texttt{LIMA}\xspace}
\newcommand\tulu{\texttt{Tulu V2}\xspace}
\newcommand\dolly{\texttt{Dolly}\xspace}
\newcommand\alpaca{\texttt{Stanford Alpaca}\xspace}
\newcommand\codealpaca{\texttt{Code Alpaca}\xspace}
\newcommand\science{\texttt{Science Literature}\xspace}

\newcommand\ti[1]{\textit{#1}}

\newcommand\tf[1]{\textbf{#1}}
\newcommand{\tableindent}{~~}
\newcommand\mt{{\texttt{(m)}\tiny}}

\newcommand\cg{\cellcolor{Gray}}
\newcommand\hll{\cellcolor{cid}}

\newcommand\hl{\cellcolor{cid}}
\newcommand\se{\cellcolor{ccon}}

\newcommand{\ua}{\textcolor{ggreen}{$\uparrow$}}
\newcommand{\da}{\textcolor{gred}{$\downarrow$}}
\newcommand{\uaa}[1]{\scriptsize\textcolor{ggreen}{\footnotesize $\uparrow$}{\color{ggreen}#1}}
\newcommand{\daa}[1]{\scriptsize\textcolor{gred}{\footnotesize $\downarrow$}{\color{gred}#1}}

\newcommand{\ie}[0]{\emph{i.e., }}

\newcommand{\eg}[0]{\emph{e.g., }}

\def\eqref#1{equation~\ref{#1}}
\def\1{\bm{1}}

\def\mA{{\bm{A}}}
\def\mB{{\bm{B}}}

\def\mE{{\bm{E}}}

\def\mP{{\bm{P}}}

\DeclareMathAlphabet{\mathsfit}{\encodingdefault}{\sfdefault}{m}{sl}
\SetMathAlphabet{\mathsfit}{bold}{\encodingdefault}{\sfdefault}{bx}{n}

\newglossarystyle{acronyms}{%
    \setglossarystyle{listdotted}%
    \setlength{\glslistdottedwidth}{.3\linewidth}%
}

\newacronym{AI}{AI}{Artificial Intelligence}
\newacronym{LM}{LM}{Language Model}
\newacronym{LLM}{LLM}{Large Language Model}
\newacronym{SOTA}{SOTA}{State-of-the-Art}
\newacronym{NLP}{NLP}{Natural Language Processing}
\newacronym{MLM}{MLM}{Masked Language Modelling}
\newacronym{CLM}{CLM}{Causal Language Modelling}
\newacronym{FS}{FS}{Few Shot Learning}
\newacronym{FT}{FT}{Fine-tuning}
\newacronym{St}{St}{Self Training}
\newacronym{UDA}{UDA}{Unsupervised Domain Adaptation}
\newacronym{STL}{STL}{Self-taught Learning}
\newacronym{VL}{VL}{Vision-Language}
\newacronym{Ssl}{Ssl}{Self Supervised Learning}
\newacronym{PT}{PT}{Prompt Tuning}
\newacronym{LoRA}{LoRA}{Low-Rank Adaptation of Large Language Models}
\newacronym{DePT}{DePT}{Decomposed Prompt Tuning}
\newacronym{PEFT}{PEFT}{Parameter-Efficient Fine-Tuning}
\newacronym{PETL}{PETL}{Parameter-Efficient Transfer Learning}
\newacronym{TAPT}{TAPT}{Task Adpative Pre-training}
\newacronym{ICL}{ICL}{In Context Learning}
\newacronym{PCP}{PCP}{Promp-based Continued Pre-training}
\newacronym{IM}{IM}{Instruction Modelling}
\newacronym{IT}{IT}{Instruction Tuning}
\newacronym{SAH}{SAH}{Superficial Alignment Hypothesis}
\newacronym{RL}{RL}{Reinforcement Learning}
\newacronym{RLHF}{RLHF}{Reinforcement Learning from Human Feedback}
\newacronym{DPO}{DPO}{Direct Preference Optimisation}
\newacronym{DAA}{DAA}{Direct Alignment Algorithm}
\newacronym{NLL}{NLL}{Negative Log-Likelihood}
\newglossarystyle{notationstyle}{%
    \setglossarystyle{listdotted}%
    \setlength{\glslistdottedwidth}{.33\linewidth}%
}
\newglossary[ng]{notation}{ns}{no}{Notation Used}

\newglossaryentry{loss_function}
{
    type=notation,
    name=$\mathcal{L}$,
    description={A loss function that measures the difference between predicted and actual values, used to evaluate and optimise model performance}
}

\newglossaryentry{sigmoid}
{
    type=notation,
    name=$\sigma$,
    description={Sigmoid function used to convert logits to probabilities}
}

\newglossaryentry{probability}
{
    type=notation,
    name=$P(\cdot|\cdot)$,
    description={Conditional probability of one event given another event}
}

\newglossaryentry{sequence_probability}
{
    type=notation,
    name=$P(\cdot)$,
    description={The probability}
}

\newglossaryentry{number_n}
{
    type=notation,
    name=$N$,
    description={The number of labelled training examples}
}

\newglossaryentry{number_m}
{
    type=notation,
    name=$M$,
    description={The number of unlabelled training examples}
}

\newglossaryentry{labeled_data}
{
    type=notation,
    name=$L$,
    description={Set of labelled examples containing pairs of inputs and their corresponding labels}
}

\newglossaryentry{unlabeled_data}
{
    type=notation,
    name=$U$,
    description={Set of unlabelled examples}
}

\newglossaryentry{labeled_example}
{
    type=notation,
    name={$ (x_i, y_i) $},
    description={A pair of input $x_i$ and its corresponding label $y_i$.}
}

\newglossaryentry{unlabeled_example}
{
    type=notation,
    name={$ \tilde{x}_i $},
    description={An input from the unlabelled dataset.}
}

\newglossaryentry{pseudo_label}
{
    type=notation,
    name={$ \tilde{y}_i $},
    description={The predicted label (pseudo-label) for an unlabelled example generated by the teacher model.}
}

\newglossaryentry{perturbed_input}
{
    type=notation,
    name=$\tilde{x}^{\prime}$,
    description={A perturbed version of an unlabelled input created through data augmentation or noise injection}
}

\newglossaryentry{input_sequence}
{
    type=notation,
    name=$\mathcal{S}$,
    description={A sequence of input tokens}
}

\newglossaryentry{masked_sequence}
{
    type=notation,
    name=$\mathcal{\tilde{S}}$,
    description={The masked version of the input sequence $\mathcal{S}$}
}

\newglossaryentry{prompted_sequence}
{
    type=notation,
    name=\ensuremath{\mathcal{S}_{\text{prompt}}},
    description={The input sequence after applying the prompt template transformation}
}

\newglossaryentry{sequence_element}
{
    type=notation,
    name=$s_i$,
    description={The $i$-th token in the sequence $\mathcal{S}$}
}

\newglossaryentry{sequence_len}
{
    type=notation,
    name=$n$,
    description={The number of tokens in the sequence $\mathcal{S}$}
}

\newglossaryentry{max_sequence_len}
{
    type=notation,
    name=$s$,
    description={The maximum input sequence length}
}

\newglossaryentry{mask_set}
{
    type=notation,
    name=$\mathcal{M}$,
    description={The set of masked tokens in the sequence $\mathcal{S}$}
}

\newglossaryentry{masked_token}
{
    type=notation,
    name=$m_i$,
    description={$i$-th masked token from the set $\mathcal{M}$}
}

\newglossaryentry{hidden_vectors}
{
    type=notation,
    name=$h$,
    description={Hidden vector}
}

\newglossaryentry{hidden_vectors_d}
{
    type=notation,
    name=$d$,
    description={Hidden vector dimension},
}

\newglossaryentry{soft_prompt_len}
{
    type=notation,
    name=$l$,
    description={The number of virtual tokens in the soft prompt},
}

\newglossaryentry{dept_soft_prompt_len}
{
    type=notation,
    name=$m$,
    description={The number of virtual tokens in the soft prompt for DEPT},
}

\newglossaryentry{teacher_parameters}
{
    type=notation,
    name=\ensuremath{\Theta,\Phi},
    description={Parameters of Models}
}

\newglossaryentry{confidence_threshold}
{
    type=notation,
    name=$\kappa$,
    description={A threshold value that determines whether a pseudo-label should be retained based on the model's prediction confidence}
}

\newglossaryentry{discard_operation}
{
    type=notation,
    name=\texttt{discard},
    description={Operation indicating that a sample}
}

\newglossaryentry{prompt_template}
{
    type=notation,
    name=$\mathcal{T}$,
    description={A function that augments input text with a prompt pattern, typically including masked tokens}
}

\newglossaryentry{verbalizer}
{
    type=notation,
    name=$\mathcal{Z}$,
    description={A function that maps task labels to vocabulary tokens}
}

\newglossaryentry{label_space}
{
    type=notation,
    name=$\mathcal{Y}$,
    description={The set of possible labels or classes for a given task}
}

\newglossaryentry{vocabulary}
{
    type=notation,
    name=$\mathcal{V}$,
    description={The model's vocabulary, containing all possible tokens that can be predicted}
}

\newglossaryentry{word_embeddings}
{
    type=notation,
    name=$\mE$,
    description={Matrix of word embeddings for input text}
}

\newglossaryentry{prompt_matrix}
{
    type=notation,
    name=$\mP$,
    description={Trainable prompt matrix containing soft prompts}
}

\newglossaryentry{weight_matrix}
{
    type=notation,
    name=\ensuremath{W,B,A},
    description={Trainable model weight matrix}
}

\newglossaryentry{rank}
{
    type=notation,
    name=$r$,
    description={Rank of the decomposition in LoRA}
}

\newglossaryentry{instruction}
{
    type=notation,
    name=$I$,
    description={An instruction sequence that describes the task or provides context}
}

\newglossaryentry{instruction_token}
{
    type=notation,
    name=$I_i$,
    description={The $i$-th token in the instruction sequence}
}

\newglossaryentry{completion}
{
    type=notation,
    name=$C$,
    description={The desired output sequence that the model should generate in response to the instruction}
}

\newglossaryentry{completion_token}
{
    type=notation,
    name=$C_j$,
    description={The $j$-th token in the completion sequence}
}

\newglossaryentry{comparison_dataset}
{
    type=notation,
    name=\ensuremath{\mathcal{D}},
    description={Dataset of human preference comparisons}
}

\newglossaryentry{preferred_completion}
{
    type=notation,
    name=$y_w$,
    description={The preferred completion in a pair, as judged by human annotators}
}

\newglossaryentry{dispreferred_completion}
{
    type=notation,
    name=$y_l$,
    description={The less preferred completion in a pair, as judged by human annotators}
}

\newglossaryentry{true_reward}
{
    type=notation,
    name={\ensuremath{r(\cdot,\cdot)}},
    description={The reward model}
}

\newglossaryentry{optimised_policy}
{
    type=notation,
    name=$\pi_{\theta}$,
    description={Policy models being optimised, with parameters $\theta$}
}

\newglossaryentry{reference_policy}
{
    type=notation,
    name=$\pi_{\text{ref}}$,
    description={Initial policy model obtained through supervised fine-tuning, serving as a reference for optimization}
}

\newglossaryentry{kl_divergence}
{
    type=notation,
    name=$\mathbb{D}_{\textrm{KL}}$,
    description={Kullback-Leibler divergence}
}

\newglossaryentry{kl_coefficient}
{
    type=notation,
    name=$\beta$,
    description={Coefficient controlling the KL divergence constraint in preference learning, used in RLHF or DPO}
}

\newglossaryentry{ipo_temperature}
{
    type=notation,
    name=$\tau$,
    description={Temperature hyperparameter in IPO that scales the target difference in log probabilities}
}

\newglossaryentry{margin}
{
    type=notation,
    name=$\gamma$,
    description={Margin hyperparameter in the hinge loss that defines the minimum desired difference between log probabilities}
}

\newglossaryentry{nll_coefficient}
{
    type=notation,
    name=$\lambda$,
    description={Scaling coefficient for combining NLL loss with the primary DAA objective}
}

\newglossaryentry{entity_set}
{
    type=notation,
    name=$\mathcal{E}$,
    description={The set of all possible entities}
}

\newglossaryentry{relation_count}
{
    type=notation,
    name=$k$,
    description={Number of spatial relations (or sentences) in a story}
}

\newglossaryentry{position_embedding}
{
    type=notation,
    name=$p_i$,
    description={The position embedding for $i$-th token in the sequence $\mathcal{S}$}
}

\newglossaryentry{question_embedding}
{
    type=notation,
    name=$q$,
    description={The vector representation for the question $q$}
}

\newglossaryentry{encoded_story}
{
    type=notation,
    name=$S^*$,
    description={Matrix of encoded sentences}
}

\newglossaryentry{entity_representations}
{
    type=notation,
    name=$E$,
    description={Entity representations}
}

\newglossaryentry{relation_representations}
{
    type=notation,
    name=$R$,
    description={Relation representations}
}

\newglossaryentry{memory_tensor}
{
    type=notation,
    name=$\boldsymbol{M}$,
    description={TPR-like memory representation updated through recurrent layers}
}
\makeglossaries
\glstoctrue
\glsaddall

\usepackage{bibentry}
\makeatletter\let\saved@bibitem\@bibitem\makeatother
\makeatletter\let\@bibitem\saved@bibitem\makeatother
\makeatletter
\AtBeginDocument{
    \hypersetup{
        pdfsubject={Thesis Subject},
        pdfkeywords={Thesis Keywords},
        pdfauthor={Author},
        pdftitle={Title},
    }
}
\makeatother

\begin{document}
% \includepdf[pages=1,scale=0.9,pagecommand={}]{cover_page.pdf}
% \includepdf[pages=1]{cover_page.pdf}
% \includepdf[pages=1,fitpaper=true]{cover_page.pdf}
\includepdf[offset=1in -1in, scale=1.09]{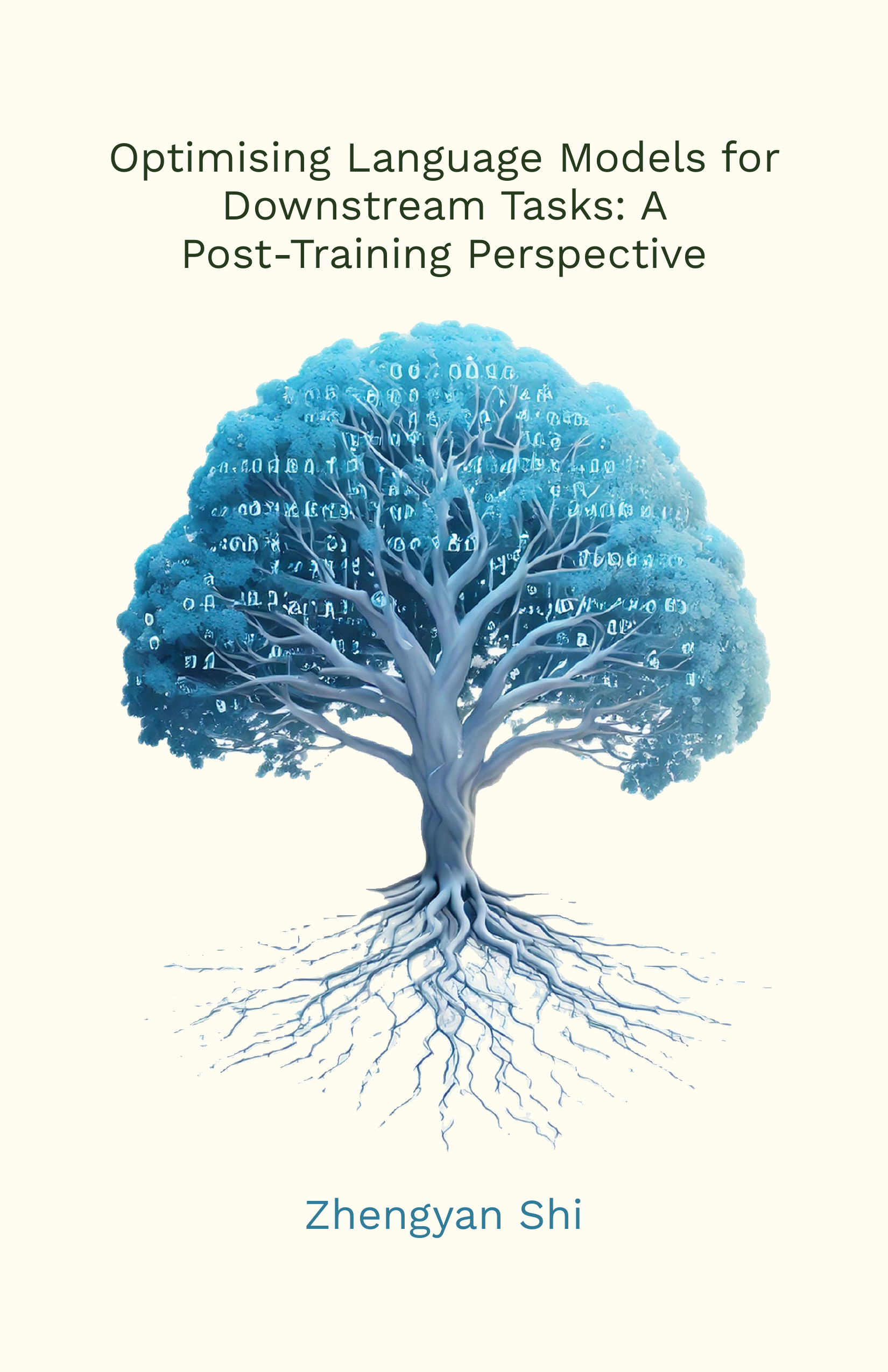}

\nobibliography*
% ^-- This is a dumb trick that works with the bibentry package to let
%  you put bibliography entries whereever you like.
% I used this to put references to papers a chapter's work was 
%  published in at the end of that chapter.
% For more information, see: http://stefaanlippens.net/bibentry

% If you haven't finished making your full BibTex file yet, you
%  might find this useful -- it'll just replace all your
%  citations with little superscript notes.
% Uncomment to use.
%\renewcommand{\cite}[1]{\emph{\textsuperscript{[#1]}}}

% At last, content! Remember filenames are case-sensitive and 
%  *must not* include spaces.

% I may change the way this is done in a future version, 
%  but given that some people needed it, if you need a different degree title 
%  (e.g. Master of Science, Master in Science, Master of Arts, etc)
%  uncomment the following 3 lines and set as appropriate (this *has* to be before \maketitle)
% \makeatletter
% \renewcommand {\@degree@string} {Master of Things}
% \makeatother

\title{Optimising Language Models for Downstream Tasks: A Post-Training Perspective}
\author{Zhengyan Shi}
\department{Faculty of Engineering}

\maketitle
\frontmatter
\chapter*{}
\begin{center}
\emph{To my parents, who provided the foundational pre-training that shaped my base model, and to my wife, supervisors, mentors, and friends, who fine-tuned and aligned my capabilities through instruction and feedback - you have all been my most crucial learning signals.}
\end{center}

\begin{acknowledgements}
A PhD journey, much like a large language model, isn't trained in isolation - it requires a diverse training dataset and multiple stages of optimisation. My journey has been an ensemble learning process, enhanced by my family, friends, and many remarkable mentors who have contributed to my instruction tuning, shaped my reward modelling, and guided my alignment through countless iterations of feedback and refinement. 
Their collective influence has been permanently encoded into my neural architecture, fundamentally improving both my research outputs and behavioural patterns.

First and foremost, I extend my deepest gratitude to my supervisors, Prof. Aldo Lipani and Prof. Emine Yilmaz, who served as my primary optimisation algorithms from my master's studies at UCL through my PhD journey. 
Aldo, thank you for being my adaptive optimiser, recognising my initial noisy gradients and believing in my convergence potential as a researcher. 
Your flexibility in allowing me to explore diverse research directions, constant encouragement, and thoughtful advice has been instrumental in my growth. Your mentorship has instilled in me the courage to tackle big challenges and the resilience to learn from setbacks. I am also grateful for your career guidance in navigating the complex optimisation landscape of my career path. 
Emine, your insightful perspectives and thought-provoking discussions have been instrumental in minimising my loss function. Your support has opened numerous doors throughout my journey, and your career guidance has been invaluable in shaping my professional trajectory.
Looking back on this training process, I can say with certainty that choosing this PhD program has been one of the most rewarding decisions of my life. Despite the challenges, occasional hurdles and disagreement, the growth, learning, and relationships forged during this time have been invaluable. If given another random initialisation, I would apply for this same path without hesitation.

My industry experiences have significantly enriched my research perspective. I am particularly grateful to Gabriella Kazai and Yunlong Jiao for their exceptional mentorship during my first internship at Amazon. Their guidance and support were truly transformative for my research journey. 
At Cohere, I had the privilege of working with extraordinary mentors. Max Bartolo, your impact on my journey extends far beyond technical collaboration – thank you for your invaluable career advice, meticulous guidance in paper writing, and for creating opportunities that have shaped my path. Your mentorship has been truly transformative. 
Sander Land, your remarkable ability to uncover hidden patterns and extract meaningful insights from learning curves has fundamentally changed how I approach experimental analysis. Your keen eye for detail and talent for finding interesting phenomena in training dynamics have not only enhanced my research methodology but also taught me to look deeper into the stories that data can tell.

Beyond my supervisors, I have been fortunate to receive guidance and support from several remarkable mentors throughout my journey. 
Prof. Tao Cheng's leadership at SpaceTimeLab has been instrumental in creating an environment where research could flourish. Her invaluable advice and generous support, particularly in providing reference letters and career guidance, have been crucial to my academic development.
Dr. Qiang Zhang's guidance was instrumental in helping me publish my first paper - a milestone that marked the beginning of my research career and gave me the confidence to pursue more challenging questions. Prof. Nikolaos Aletras went above and beyond during my first internship, setting an exceptional standard of mentorship. I'll always remember and appreciate his dedication - his entire Saturday helping improve my paper. 
Dr. Laurence Aitchison's exceptional support and guidance, particularly during the summer collaboration with Alpay Deniz, demonstrated an extraordinary commitment to nurturing young researchers, sharing not only his expertise but also dedicating significant time to ensure our research success.
Speaking of collaborations, I have been fortunate to work with brilliant early-career researchers during their master's studies: Alpay Deniz, Youjing Yu, Junyuan Liu, and Haijin Li. While I may have played a small role in mentoring, I have learned just as much from our interactions. Your fresh perspectives and dedication to research have been truly inspiring, and I am proud to see many of you now pursuing your PhD journeys. 

I am also grateful to my friends and co-authors on interdisciplinary projects for the stimulating discussions, collaborations, and the much-needed laughter and support throughout this journey:
Adam X. Yang, Yuchen Zhu, Bin Wu, Mengyue Yang, Mingtian Zhang, Jerome Ramos, To Eun Kim, Xi Wang, Procheta Sen, Hossein A. Rahmani, Varsha Ramineni, Xiao Fu, Dr Sahan Bulathwela, Yue Feng,
Mustafa Can Ozkan, Natchapon Jongwiriyanurak, Xinglei Wang, Fangzhou Zhou, Zhihui Song, Zichao Zeng, Ilya Ilyankou, James Haworth, Meihui Wang, Huanfa Chen, Yunzhe Liu,
Acyr Locatelli, Matthieu Geist, Francesco Tonolini and Oleg Rokhlenko.

Finally and most importantly, I am forever indebted to my family, whose support has achieved state-of-the-art performance in every benchmark. To my parents, Jun Dai and Fangliang Shi, thank you for being my dedicated training cluster, and offering computational resources through every epoch of my development. Your extensive hyperparameter tuning and input optimization have compiled into the most robust base model I could ask for. 
To my wife, Xin Zhao, you've been the perfect post-training phase after my parents' initial model pre-training. 
From our first fine-tuning experiments at the University of Liverpool, through our joint learning in London for my master, and throughout my PhD's annealing training phase, you have enhanced my model capabilities beyond what any initial pre-training could achieve. Your commitment during the high-variance period of COVID-19, culminating in our successful model deployment (\ie marriage registration) in London in July 2024, has produced results that exceed all convergence expectations.

This thesis is a testament to the collective efforts and inspiration of all those mentioned here, and many more who have touched my life in countless ways. Thank you for being an integral part of this incredible journey.
\end{acknowledgements}

\chapter*{Declaration}
I, Zhengyan Shi, confirm that the work presented in this thesis is my own. Where information has been derived from other sources, I confirm that this has been indicated in the work

\begin{abstract}
Language models (LMs) have demonstrated remarkable capabilities in natural language processing (\nlp) tasks, yet harnessing their full potential for specific applications remains a significant challenge. 
As the scale and complexity of these models continue to grow, adapting them efficiently and robustly becomes increasingly challenging. 
The common paradigm of fine-tuning LMs on labelled data often fails to effectively leverage the vast amount of available unlabelled data and can lead to overfitting on small, task-specific datasets, and incurs substantial computational costs.
These limitations are particularly problematic in real-world scenarios that present an open-ended landscape of language tasks and domains.

This thesis proposes a series of innovative methods for tailoring LMs to task-specific applications, addressing key challenges in adapting these models to downstream tasks.
First, we explore strategies for maximising the use of unlabelled data in scenarios with limited labelled resources. 
Our goal is to extract task-relevant knowledge from unlabelled data to improve LM performance on specific tasks, allowing for more robust alignment with task demands. 
This research led to the development of novel continued pre-training techniques that outperform state-of-the-art semi-supervised approaches.
Next, we introduce a new parameter-efficient fine-tuning method that substantially reduces the memory and time costs associated with fine-tuning LMs. 
This method facilitates more efficient alignment of models to downstream tasks, making the fine-tuning process more feasible while maintaining competitive performance.
Additionally, we improve supervised fine-tuning approaches to strengthen LMs' ability to follow instructions, particularly in situations with limited learning resources. 
This enables LMs to perform more effectively across various \nlp tasks, including open-ended generation tasks, enhancing their flexibility and usefulness in real-world applications.
Furthermore, to better understand and assess LM performance on specific downstream tasks, we create new benchmarks and evaluation methods. 
These include tests for complex cognitive abilities such as multi-hop spatial reasoning, providing more comprehensive and nuanced ways to evaluate LM capabilities and adaptations.

Through extensive empirical evaluations across a diverse set of \nlp tasks, our findings demonstrate that these proposed methods largely improve the robustness, efficiency, and generalisation of LMs to a broad array of language tasks. 
The approaches presented in this thesis represent promising steps toward more robust and efficient LMs, bringing us closer to achieving artificial general intelligence. % in the field of \nlp.

\end{abstract}

\chapter*{Impact Statement}
Language models (LMs) have achieved remarkable success on a wide range of \nlp tasks. However, their performance often remains suboptimal when fine-tuned on downstream applications, especially in real-world scenarios with complex task requirements, limited supervised data, and potential distribution shifts. Adapting LMs to perform robustly and efficiently under these challenging conditions is crucial for their successful deployment in practical settings.

This thesis develops novel methods for optimising LMs for downstream tasks using advanced post-training techniques. Continued pre-training and prompt-based pre-training are powerful semi-supervised learning methods that improve LM performance, especially with limited labelled data. Decomposed prompt tuning significantly boosts fine-tuning efficiency and reduces computational costs, enabling practical adaptation of LMs to diverse downstream tasks. Instruction modelling enhances LMs' ability to robustly follow instructions, which is key for reliable conversational AI systems. The StepGame benchmark enables fine-grained evaluation of LMs' reasoning capabilities, revealing current limitations and guiding future research.
By tackling these challenges, the proposed approaches enable LMs to generalise open-ended real-world language tasks more reliably.

The algorithmic and empirical contributions of this thesis are largely model-agnostic, allowing extension to various LM architectures and problem domains. The proposed methods have already been applied successfully to different settings such as multi-task learning and domain adaptation. We anticipate further generalisations will benefit an even broader range of language tasks.

In the longer term, the ideas presented in this thesis could contribute to the development of increasingly capable and flexible AI systems that can continuously improve through interaction with open environments. By advancing the fundamental techniques for optimising LMs, this research takes important steps towards realizing the transformative potential of language AI to augment human knowledge work and enable more natural human-computer collaboration. Responsibly deploying such systems, however, will require proactively addressing key ethical considerations around robustness, fairness, and safety.

% \raggedcolumns
\newpage
\setcounter{tocdepth}{2} 
\hypersetup{linkcolor=black} % Make the link color in Tables and Figures as Black
\tableofcontents
{\small{
    \listoffigures
    \listoftables
}}
\hypersetup{linkcolor=dred2}

\setglossarystyle{acronyms}\printglossary[title=List of Abbreviations, type=\acronymtype]
\setglossarystyle{notationstyle}\printglossary[title=Notation Used, type=notation]

\mainmatter
\chapter{Introduction}
\label{chapterlabel1}

\section{The Power of Language Models}
The evolution of language models (LMs) has been nothing short of extraordinary. Beginning with rudimentary statistical models, LMs have advanced to complex neural architectures that now play a pivotal role in natural language processing (\nlp). 
These advances have fueled breakthroughs in diverse applications, from generating human-like text to automating sophisticated workflows in code, tool use, and interactive AI systems.

Early methods, such as Word2Vec \citep{mikolov2013efficient}, formed the foundation of early LMs, capturing semantic relationships between words with relatively small parameter sizes.
The real scaling began with the introduction of transformer-based models \citep{vaswani2017attention}.
BERT \citep{devlin2018bert}, for instance, represented a major leap with its attention mechanism and its power in capturing contextual understanding, ranging from 29 million to 340 million parameters. 
Subsequent models, such as the T5 model \cite{t5} with its largest variant boasting 11 billion parameters, pushed boundaries further, demonstrated the power of treating various NLP tasks as text generation problems, providing a unified approach to tasks such as translation, summarisation, and question-answering.
The GPT series dramatically shifted the landscape by pushing the scale further, exemplified by GPT-3 \citep{10.5555/3495724.3495883}, which with 175 billion parameters, showcased unprecedented capabilities in text generation, reasoning, and problem-solving. GPT-4 \citep{achiam2023gpt}, estimated to exceed one trillion parameters, took this a step further, solidifying the importance of model scale in achieving breakthroughs in AI performance.
The open-source LLaMA series \citep{touvron2023llama2} continued the trend, culminating in LLaMA-3 \citep{dubey2024llama}, a model with 405 billion parameters. 
Similarly, multilingual LMs such as BLOOM \citep{le2023bloom}, with up to 176 billion parameters, showed that scaling could be applied to models capable of fluency in multiple languages.
Additionally, Anthropic's Claude\footnote{\url{https://www.anthropic.com/news/introducing-claude}}, Perplexity\footnote{\url{https://www.perplexity.ai/}}, and Cohere Command\footnote{\url{https://cohere.com/command}} represent continued efforts to refine the balance between scalability, efficiency, and ethical considerations in LM development.

Investment in scaling these models has paralleled their growth, but this aggressive scaling approach raises critical concerns about computational sustainability and efficiency.
While initiatives initiatives like xAI’s Colossus supercomputer\footnote{\url{https://x.com/elonmusk/status/1830650370336473253}}, 
OpenAI’s \$6.6 billion funding round\footnote{\url{https://openai.com/index/scale-the-benefits-of-ai/}}, and 
NVIDIA’s focus on AI hardware to meet rising demand\footnote{\url{https://fortune.com/2024/10/03/nvidia-stock-jensen-huang-ai/}} demonstrate the industry's push toward larger models, they also highlight a fundamental challenge: these models are becoming increasingly resource-intensive and computationally expensive.
For comparison, while the human brain processes information at remarkable efficiency, consuming only about 20 watts of power (roughly 175 kWh per year), training a single LM can consume hundreds of thousands of kilowatt-hours \citep{doi:10.1073/pnas.2107022118}. 
This stark efficiency gap, combined with the growing need to adapt these models for specific tasks and domains, underscores the urgent need for more efficient and robust training and optimisation methods. These challenges motivate our research into post-training methods that can effectively adapt and tailor LMs while minimising computational costs and maximising resource efficiency.

\section{Tailoring Language Models to Task-Specific Demands}
LMs have demonstrated remarkable general-purpose capabilities, but their true potential is realized when adapted for specific downstream tasks.  
Without task alignment, even state-of-the-art models may underperform in complex, real-world applications. Fine-tuning and post-training have thus emerged as critical processes for enhancing LM performance and ensuring their applicability across diverse domains.

The increasing sophistication of LMs has catalyzed transformative applications across industries, driving productivity and innovation. 
For example, the GPT series \citep{openai2023gpt4} revolutionises content creation, enabling writers and marketers to generate high-quality, human-like text efficiently. 
Additionally, Llama Guard \citep{inan2023llama} provides a useful tool for categorizing safety risks in LM-generated prompts and responses.
LMs have been fine-tuned into reward models \cite{leike2018scalable,yang2024bayesian} to evaluate how well a model's responses align with human preferences.
Specialized models tailored to specific domains have emerged. 
BloombergGPT is designed for the financial industry, while CodeLlama \citep{roziere2023code}, a derivative of Llama 2 \citep{touvron2023llama}, is optimised for programming tasks. 
Similarly, ``The AI Scientist'' \citep{lu2024ai} applies LMs to automate scientific discovery by generating hypotheses, analyzing large datasets, and proposing new experiments in fields like chemistry and biology.
Task-specific adaptation also extends to efficiency and privacy, with Apple developing LMs optimised for on-device processing to enhance data security \citep{grangier2024projected}. AI agents, such as Agent Q \cite{putta2024agent}, leverage LMs to perform complex, multi-step tasks, assisting in decision-making and workflow management.

Post-training plays a pivotal role in refining LMs for specialised tasks, bridging the gap between general-purpose capabilities and domain-specific performance. 
Post-training refers to a range of methods applied after the initial pre-training phase, allowing LMs to adapt to new tasks, domains, or preferences. This stage is crucial for mitigating biases, aligning models with human values, and boosting task-specific performance.

Key post-training techniques encompass a range of methods designed to enhance model adaptability, efficiency, and alignment with human preferences. 
\textbf{Instruction tuning} \cite{leike2018scalable,li2024selfalignment,wei2021finetuned,xu2024wizardlm} serves as the foundation, refining LMs by teaching instruction-following abilities and adapting them to specific task formats through labelled datasets. This process ensures that models respond accurately to task-specific inputs and improves general chat or interactive capabilities.

\textbf{Preference fine-tuning} focuses on aligning models with human values and expectations by incorporating feedback on output quality \cite{NIPS2017_d5e2c0ad,ziegler2019fine}. 
This stage contains training reward models using preference data and has been central to the development of highly responsive chat systems. 
Reinforcement Learning from Human Feedback \citep{NEURIPS2020_1f89885d,bai2022training,NEURIPS2022_b1efde53,zhu2024can,shi2024understanding} fine-tune models to prioritise outputs that reflect user preferences, while direct alignment algorithms such as Direct Preference Optimisation \citep{rafailov2023direct} bypass the need for full RL pipelines by directly optimising the policy without explicit reward modelling.

\textbf{Reinforcement fine-tuning} further extends model capabilities by optimising performance on verifiable tasks. Methods such as Reinforcement Learning with Verifiable Rewards \cite{lambert2024t} refine models without relying on reward models, instead leveraging environment-based signals to guide tuning. 
This technique improves task completion stability and reduces the risk of degradation, making it ideal for complex, logic-driven applications.

In addition to these alignment methods, \textbf{continued pre-training} \cite{alsentzer-etal-2019-publicly,margatina-etal-2022-importance,howard-ruder-2018-universal,sun2019fine,xue-etal-2021-mt5,gururangan-etal-2020-dont} exposes models to domain-specific or task-related corpora, amplifying performance in specialized fields such as healthcare, finance, and scientific research. Models such as \texttt{BioBERT} \citep{lee2020biobert} and \texttt{SciBERT} \citep{beltagy-etal-2019-scibert} exemplify this approach, demonstrating significant improvements through further in-domain training.

To address computational constraints, \textbf{parameter-efficient fine-tuning}, including Adapters \citep{houlsby2019parameter}, BitFit \citep{zaken2021bitfit}, and LoRA \citep{hu2021lora}, enable model adaptation by updating a minimal subset of parameters. 
This approach maintains performance levels comparable to full fine-tuning while significantly reducing resource requirements, making it especially valuable for scaling large models across diverse tasks.

Together, these post-training strategies create a comprehensive pipeline for tailoring LMs to specific demands, making LMs not only task-adapted but also efficient, robust, and responsive to evolving user needs.

\section{Contributions and Structure}
This thesis contributes several new methods for efficiently and robustly adapting LMs to tackle a range of \nlp tasks. 
These contributions are organised around the following research questions:
\begin{enumerate}[label=\subscript{\color{black}\textbf{RQ}}{{\arabic*}}]
\item \label{Q1}  How can we effectively leverage unlabelled data from target domains to adapt LMs for downstream tasks?
\item \label{Q2} How can we further efficiently adapt LMs to downstream tasks?
\item \label{Q3} How can we robustly improve LMs instruction following ability under low-resource settings?
\item \label{Q4} What methods can be used to evaluate the performance and effectiveness of LMs on specific downstream tasks?
\end{enumerate}

Before presenting the specific contributions of this thesis, Chapter \ref{chapter:background} provides background on various post-training methods. This includes continued pre-training (Section \sect{subsec:pretraining_objectives}), self training (Section \sect{subsec:self_training}), task-specific fine-tuning (Section \sect{subsec:traditional_finetuning}), parameter-efficient fine-tuning (Section \sect{subsec:parameter_efficient_finetuning}), instruction tuning (Section \sect{subsec:instruction_tuning}), and preference learning (Section \sect{subsec:preference_learning}).
Each subsequent section builds upon the previous concepts, and the later chapters then focus on our specific contributions to advancing these areas of research.

\paragraph{Leveraging Unlabelled Data for Task-Specific Adaptation (\ref{Q1}).~}
Chapter \ref{chapter:continued_pretraining} investigates how unlabelled data from target domains can be effectively used to adapt LMs to specific downstream tasks. 
We begin by demonstrating that Task-Adaptive Pre-training (\tapt), which further pre-trains LMs on unlabelled data from the target domain, serves as a robust semi-supervised learning method (Section \sect{sec:acl_paper}). 
\tapt often outperforms sophisticated self-training (ST) methods, especially when dealing with domain shifts or limited unlabelled data. 

We also challenge the widely accepted notion in \nlp that continued pre-training LMs on task-related texts improves fine-tuning performance in downstream tasks.
Our findings reveal that conventional continued pre-training does not consistently provide benefits and can even be detrimental for sentence-pair tasks or when prompt-based fine-tuning is used.
To tackle these issues, we propose Prompt-based Continued Pre-training (\pcp), which combines the idea of instruction tuning with conventional continued pre-training (Section \sect{sec:neurips_paper}).
We demonstrate that \pcp consistently improves the performance of state-of-the-art prompt-based fine-tuning approaches in both semi-supervised and fully-supervised settings, even with only hundreds of unlabelled examples.
Prompt-based fine-tuning with the \pcp outperforms state-of-the-art semi-supervised approaches with greater simplicity, eliminating the need for an iterative process and extra data augmentation.
The content of this chapter is based on the following works:
\begin{itemize}
    \item Rethinking Semi-supervised Learning with Language Models.
    Zhengxiang Shi, Francesco Tonolini, Nikolaos Aletras, Emine Yilmaz, Gabriella Kazai, and Yunlong Jiao.
    2023.
    In \emph{Findings of the Association for Computational Linguistics (ACL 2023)}.

    \item Don’t Stop Pretraining? Make Prompt-based Fine-tuning Powerful Learner.
    Zhengxiang Shi, Aldo Lipani.
    2023.
    In \emph{Advances in Neural Information Processing Systems (NeurIPS 2023)}.
\end{itemize}

\paragraph{Efficient Adaptation with Parameter-Efficient Fine-Tuning (\ref{Q2}).~}
Chapter \ref{chapter:peft} focuses on further improving efficient fine-tuning strategies for adapting LMs to downstream tasks in both data-rich and data-scarce scenarios.
Efficiency is a key concern, especially in environments such as mobile applications or systems requiring real-time decision-making, where running resource-intensive, full-scale LMs is not feasible. 
In these contexts, it is essential to develop lightweight and efficient models that maintain the required model performance while optimising speed and resource usage.
Prompt tuning (\pt), where a small amount of trainable soft (continuous) prompt vectors is affixed to the model input, has shown promising results across various tasks and model architecture for \peft.
\pt stands out from other \peft approaches because it maintains competitive performance with fewer trainable parameters and does not drastically scale up its parameters as the model size expands. 
However, \pt introduces extra soft prompt tokens, leading to longer input sequences, which significantly impacts training/inference time and memory usage due to the Transformer's quadratic complexity. 
Particularly concerning for LMs that face heavy daily querying. 
To address this issue, we propose \textbf{De}composed \textbf{P}rompt \textbf{T}uning (\dept), which decomposes the soft prompt into a shorter soft prompt and a pair of low-rank matrices that are then optimised with two different learning rates. 
This allows \dept to achieve better performance while saving substantial memory and time costs compared to vanilla \pt and its variants, without changing trainable parameter sizes.
Through extensive experiments on \nlp and vision-language (VL) tasks, we demonstrate that \dept outperforms state-of-the-art \peft approaches, including the full fine-tuning baseline, in some scenarios.
The content of this chapter is based on the following works:
\begin{itemize}
    \item DePT: Decomposed Prompt Tuning for Parameter-Efficient Fine-tuning.
    Zhengxiang Shi, Aldo Lipani.
    2024.
    In \emph{International Conference on Learning Representations (ICLR 2024)}.
\end{itemize}

\paragraph{Enhancing Instruction-Following Capabilities (\ref{Q3}).~}
Chapter \ref{chapter:sft} focuses on how to robustly improve LMs' ability to follow instructions.
Instruction tuning plays a crucial role in shaping the LM outputs to desired styles.
We introduce a simple yet effective method, \textsc{Instruction Modelling} (\im), which trains LMs by applying a loss function to the instruction and prompt part rather than solely to the output part.
We show that, in many scenarios, \im can effectively improve the LM performance on both \nlp tasks (\eg MMLU, TruthfulQA, and HumanEval) and open-ended generation benchmarks (\eg MT-Bench and AlpacaEval).
Remarkably, in the most advantageous case, \im boosts model performance on AlpacaEval 1.0 by over 100\%.
We identify two key factors influencing the effectiveness of \im:
(1) The ratio between instruction length and output length in the training data; and
(2) The number of training examples.
We observe that \im is especially beneficial when trained on datasets with lengthy instructions paired with brief outputs, or under the Superficial Alignment Hypothesis (\sah) where a small amount of training examples are used for instruction tuning.
Further analysis substantiates our hypothesis that our improvement can be attributed to reduced overfitting to instruction tuning datasets. 
These insights offer practical guidelines for instruction tuning, particularly in low-resource scenarios.
The content of this chapter is based on the following works:
\begin{itemize}
    \item Instruction Tuning With Loss Over Instructions.
    Zhengyan Shi, Adam X. Yang, Bin Wu, Laurence Aitchison, Emine Yilmaz, Aldo Lipani.
    2024.
    In \emph{Advances in Neural Information Processing Systems (NeurIPS 2024)}.

    % \item Understanding Likelihood Over-optimisation in Direct Alignment Algorithms.
    % Zhengyan Shi, Sander Land, Acyr Locatelli, Matthieu Geist, Max Bartolo.
    % 2024.
    % \emph{Under Review}.
\end{itemize}

\paragraph{Evaluating LMs on Specific Downstream Tasks (\ref{Q4}).~} 
While RQ1–RQ3 develop novel techniques for adapting and aligning LMs, a robust evaluation framework is needed to measure the true impact of these adaptations on targeted tasks. 
Chapter \ref{chapter:downstream_tasks} addresses RQ4 by constructing \textbf{StepGame}, a new multi-hop spatial‐reasoning benchmark designed to probe the kinds of inference that neither generic pre‑training (Chapter \ref{chapter:continued_pretraining}) nor fine‑tuning tricks (Chapter \ref{chapter:peft} and \ref{chapter:sft}) directly optimize for. 
A key capability for intelligent systems is the ability to infer spatial relations in natural language. 
To explore this, we present a new Question-Answering dataset called StepGame, designed to test robust multi-hop spatial reasoning in texts.
Our experiments reveal significant limitations in the spatial reasoning abilities of state-of-the-art LMs, highlighting a key area for improvement.
The content of this chapter is based on the following works:
\begin{itemize}
    \item StepGame: A New Benchmark for Robust Multi-Hop Spatial Reasoning in Texts.
    Zhengxiang Shi, Qiang Zhang, Aldo Lipani.
    2022.
    In \emph{Proceedings of the AAAI Conference on Artificial Intelligence (AAAI 2022)}.
\end{itemize}

By addressing these research questions, this thesis advances LM development, proposing techniques to enhance \nlp efficiency and robustness in real-world applications.
Additional publications from this PhD, not included in this thesis, are:
\begin{itemize}
    \item Learning to Execute Actions or Ask Clarification Questions.
    Zhengxiang Shi, Yue Feng, and Aldo Lipani.
    2022.
    In \emph{Findings of the Association for Computational Linguistics: NAACL 2022}.

    \item Lexical Entrainment for Conversational Systems.
    Zhengxiang Shi, Procheta Sen, Aldo Lipani.
    2022.
    In \emph{Findings of the Association for Computational Linguistics: EMNLP 2022}.

    \item Self Contrastive Learning for Session-based Recommendation.
    Zhengxiang Shi, Xi Wang, Aldo Lipani.
    2024.
    In \emph{Advances in Information Retrieval. ECIR 2024}.
\end{itemize}

\section{Open-sourced Software and Code Repositories}

The research presented in this thesis is supported by a collection of open-source software implementations and code repositories. These resources are made publicly available to promote reproducibility, facilitate further research, and contribute to the broader scientific community.
These are listed in the order they appear in the previous sections, reflecting the progression of the research:
\begin{itemize}
\setlength{\itemsep}{-1pt}
    \item \url{https://github.com/amzn/pretraining-or-self-training}
    \item \url{https://github.com/ZhengxiangShi/PowerfulPromptFT}
    \item \url{https://github.com/ZhengxiangShi/DePT}
    \item \url{https://github.com/ZhengxiangShi/InstructionModelling}
    \item \url{https://github.com/ZhengxiangShi/StepGame}
    \item \url{https://github.com/ZhengxiangShi/LearnToAsk}
    \item \url{https://github.com/ZhengxiangShi/SelfContrastiveLearningRecSys}
    \item \url{https://github.com/ZhengxiangShi/LexicalEntrainment}
\end{itemize}
These repositories contain the implementation details, data processing scripts, model architectures, and evaluation protocols used throughout this research. By making this code openly available, we aim to ensure transparency in our methods and enable other researchers to build upon our work.
\chapter{Background: Post-Training Methods}
\label{chapter:background}

Post-training denotes the family of techniques applied \textbf{after} the initial, broad-coverage pre-training stage; they \textbf{update the parameters of an already-pre-trained large language model} so it can excel at a particular downstream task, operate within a specialised domain, or align more closely with human preferences. 
Where pre-training imparts general linguistic competence from massive, diverse corpora, post-training selectively fine-tunes the model on curated data to sharpen performance, add new capabilities, or steer its behaviour toward desired goals.

\section{Continued Pre-training}
\label{subsec:pretraining_objectives}
Continued pre-training \cite{howard-ruder-2018-universal,bai2023qwen} serves as a fundamental pre-training strategy, which can be utilised in two main ways:
(1) \textbf{As an architectural adaptation technique to enhance model capabilities.} This approach allows for the expansion of model functionalities, such as extending the context window length or adapting to new input formats \cite{dubey2024llama}. For example, the effectiveness of continued pre-training has been demonstrated for architectural adaptations, such as extending context processing from 8K to 128K tokens; and
(2) \textbf{As a domain specialisation method}. In this case, models are further trained on domain-specific corpora to enhance performance in targeted areas. 
For instance, recent research \cite{dubey2024llama} has shown that continued pre-training with domain-focused data compositions (\eg $>85\%$  specialised content) can significantly improve model performance in specific domains, such as code generation or scientific reasoning. 
This approach, known as domain-specific or task-adaptive pre-training \citep{gururangan-etal-2020-dont}, builds upon the model's existing contextual and linguistic knowledge while adapting it for new capabilities or specialised domains.
Additionally, continued pre-training followed by fine-tuning \cite{howard-ruder-2018-universal,sun2019fine,gururangan-etal-2020-dont} is one type of semi-supervised approaches. 
While the benefits of continued pre-training are well acknowledged \cite{beltagy-etal-2019-scibert,alsentzer-etal-2019-publicly,margatina-etal-2022-importance}, it is commonly assumed that large amounts of data are necessary for continued pre-training \cite[\eg][]{li-etal-2021-task-adaptive,10.1162/tacl_a_00517,gu-etal-2022-ppt}. 
Contrarily, our research in this thesis demonstrates that continued pre-training can improve performance using only a few hundred unlabelled samples.
In this section, we briefly review the pre-training objectives, which are also employed during continued pre-training.

\paragraph{Masked Language Modelling.}
Masked Language Modelling (\mlm) is one of the most prominent pre-training objectives used in \bert \cite{devlin2018bert} and \roberta \cite{liu2019roberta}. 
It remains an effective approach during continued pre-training. 
By exposing the model to domain-specific text under the \mlm framework, it adapts its representations better to capture knowledge patterns unique to the target domain (\eg \texttt{BioBERT} \cite{lee2020biobert}).

The core idea behind \mlm is to randomly mask a subset of input tokens in a given sequence and train the model to predict the masked tokens based on the surrounding context. 
This forces the model to learn bidirectional representations, capturing both the left and right context of a token.

Formally, given a tokenised domain-specific sequence $\mathcal{S}$, \mlm randomly selects a subset of tokens $\mathcal{M}$ to mask, typically using a masking probability of 0.15. 
The masked tokens are typically replaced with a special token \texttt{[MASK]} 80\% of the time, replaced with random tokens 10\% of the time, and kept unchanged 10\% of the time. During training, the model is trained to predict the original tokens of these masked tokens.
The \mlm objective can be expressed as the following loss function:
\begin{equation}
\label{equation:mlm}
\mathcal{L}_{\text{mlm}}(\mathcal{S}) = - \mathop{\mathbb{E}}\limits_{\mathcal{M} \subset \mathcal{S}}\left[\sum\limits_{m_i \in \mathcal{M}} \log P(m_i | \mathcal{\tilde{S}})\right],
\end{equation}
where $\mathcal{\tilde{S}}$ represents the masked version of the sequence $\mathcal{S}$, and the model predicts each masked token $m_i$ based on the unmasked tokens in $\mathcal{\tilde{S}}$.
\roberta employs dynamic masking to enhance the effectiveness of \mlm by selecting different masking tokens to the same training example in different epochs. 
This prevents the model from memorising the positions of masked tokens and encourages it to develop a deeper understanding of the context across different iterations of training.
However, this multi-epoch strategy has become less relevant as modern LMs \cite{touvron2023llama,ouyang2022training} typically train for only one epoch.

The T5 model \cite{t5} introduces a modified variant of masked language modelling that differs from \bert's token-level masking approach. Specifically, T5 employs a "span corruption" objective, which replaces contiguous spans of text with single sentinel tokens rather than masking individual tokens. 
By requiring the model to predict entire text spans rather than isolated tokens, this approach enables the capture of broader contextual dependencies. This span-based objective, typically combined with T5's architecture, has proven to be a versatile foundation for continued pre-training across various domain-specific tasks.

\paragraph{Causal Language Modelling.}
Causal Language Modelling (CLM), typically associated with autoregressive models such as GPT series \cite{radford2018improving,10.5555/3495724.3495883}, represents another objective used in continued pre-training, particularly effective for tasks requiring high-quality text generation. In this approach, the model learns to predict each token in a sequence based on its preceding context.

CLM operates by modelling the probability distribution of a token sequence as a product of conditional probabilities. 
Given a sequence of tokens $\mathcal{S} = (s_1, s_2, ..., s_n)$, the CLM objective aims to maximise the likelihood of the sequence:
\begin{equation}
\label{equation:clm}
P(\mathcal{S}) = \prod_{i=1}^n P(s_i | s_1, ..., s_{i-1})
\end{equation}
The corresponding loss function for CLM is expressed as:
\begin{equation}
\label{equation:clm_loss}
\mathcal{L}_{\text{clm}}(\mathcal{S}) = -\sum_{i=1}^n \log P(s_i | s_1, ..., s_{i-1})
\end{equation}
A notable strength of CLM lies in its left-to-right text generation capability, making it particularly suitable for tasks such as text generation, completion, and open-ended question answering. 
For example, BioGPT \cite{luo2022biogpt} demonstrates the effectiveness of continued pre-training with CLM.

\section{Self Training}
\label{subsec:self_training}
Semi-supervised learning has become a key approach in \nlp, addressing the challenge of limited labelled data by utilising large amounts of unlabelled data, where labelling data can be expensive and time-consuming, while unlabelled text data is plentiful \cite{grandvalet2004semi,chapelle2009semi,kipf2017semi}. 
One common approach in this area is pretraining LLMs on vast amounts of unlabelled data, followed by fine-tuning them on a smaller labelled dataset.
This is also known as continued pre-training.
Another powerful semi-supervised technique is Self Training (\st) \cite{yarowsky-1995-unsupervised,mcclosky-etal-2006-effective}, where a model generates pseudo-labels for unlabelled data and refines its predictions iteratively \cite{lee2013pseudo,DBLP:conf/iclr/LaineA17,tarvainen2017mean,miyato2018virtual,artetxe-etal-2018-robust,cai-lapata-2019-semi,dong-de-melo-2019-robust,10.5555/3495724.3496249,sohn2020fixmatch,gera2022zero}. 
This thesis offers an alternative way to use pseudo-labels without resorting to an iterative process, as discussed in \cref{neurips:para:discussion}.
In this section, we will mainly focus on \st, while continued pretraining has already been discussed in the previous section.

\st represents a fundamental approach in semi-supervised learning, operating on an intuitive principle of iterative model improvement. At its core, \st utilises a teacher model trained on labelled data to generate predictions (pseudo-labels) for unlabelled examples, which are then used to train a student model.
Formally, let $L \triangleq \{(x_1, y_1), \ldots, (x_N, y_N)\}$ denote $N$ labelled examples and $U \triangleq \{\tilde{x_1}, \ldots, \tilde{x_M}\}$ denote $M$ unlabelled examples, where usually $M \gg N$. 
The \st framework is trained with three main steps as follows.

\paragraph{Step 1.} A teacher model $F$, parameterized by a neural network $\Theta$, is trained via minimizing the cross entropy loss $\mathcal{L}$ on labelled examples $L$:
\begin{equation}
    \label{equation:step_st_1}
    \mathcal{L}_{teacher}(L) = \sum \limits_{x_i, y_i \in L} \mathcal{L}(y_i, F(x_i, \Theta)),
\end{equation}

\paragraph{Step 2.} The teacher model $F$ is used to make predictions (referred to as “pseudo-labels”) on unlabelled examples $U$:
\begin{equation}
    \label{equation:step_st_2}
    \tilde{y_i} = F(\tilde{x_i}, \Theta),
\end{equation}
where $\tilde{y_i}$ can be either the continuous logit or the discrete label induced by an \textsc{ArgMax} operation.

\paragraph{Step 3.} A student model $G$, parameterized by a fresh neural network $\Phi$, is trained to fit labelled and pseudo-labelled examples:
\begin{multline}
    \label{equation:step_st_3}
    \mathcal{L}_{student}(L, U) = 
        \sum \limits_{x_i, y_i \in L} \mathcal{L}(y_i, G(x_i, \Phi))  + \sum \limits_{\tilde{x_i}, \tilde{y_i} \in U} \mathcal{L}(\tilde{y_i}, G(\tilde{x_i}, \Phi))
\end{multline}

This process is repeated for a given number of times by treating the student as a new teacher to re-predict pseudo-labels as in Eq. \ref{equation:step_st_2} and then training a new student with Eq. \ref{equation:step_st_3}.

% \paragraph{Techniques in Self-Training.}
In practice, \st with techniques such as consistency regularisation \cite{miyato2018virtual,clark-etal-2018-semi,berthelot2019mixmatch}, strong data augmentation \cite{sohn2020fixmatch,xie2020self,10.5555/3495724.3496249,shi2023rethink_data}, confidence threshold \cite{sohn2020fixmatch,zhang2021flexmatch,berthelot2021adamatch} usually leads to substantial improvements in model performance.
Below we discuss about these techniques in details.
% Several techniques  have been developed to improve the performance of \st in semi-supervised learning, such as consistency regularisation, strong data augmentation, and confidence thresholding.

\paragraph{Consistency regularisation.}
Consistency regularisation is based on the idea that a model should produce similar predictions for the same input under different perturbations. In \st, this technique enforces that the model’s predictions remain stable even when inputs are augmented or perturbed, thereby improving the robustness of the pseudo-labels. Formally, for an unlabelled input $\tilde{x}$, the consistency loss $\mathcal{L}_{\text{consistency}}$ can be defined as:
\begin{equation}
    \label{equation:consistency_regularisation}
    \mathcal{L}_{\text{consistency}} = \mathcal{L}(F(\tilde{x}, \Theta), F(\tilde{x}^{\prime}, \Theta)),
\end{equation}
where $\tilde{x}^{\prime}$ represents a perturbed version of $\tilde{x}$. The goal is to minimise the difference in predictions between the original and perturbed inputs, often by using techniques such as Virtual Adversarial Training \cite{miyato2018virtual}, dropout, or other noise injections \cite{clark-etal-2018-semi}. This regularisation helps the model to be more consistent, leading to improved performance on unlabelled data.

\paragraph{Data Augmentation.}
Data augmentation techniques introduce variations in input data, helping models generalise better by training on diverse examples. 
In \nlp, data augmentations can include synonym replacement, back-translation, or random deletion of words. 
By applying these transformations, \st models are exposed to a broader range of linguistic expressions, which enhances generalisation to new, unseen data. Recent work such as MixMatch \cite{berthelot2019mixmatch} and FixMatch \cite{sohn2020fixmatch} combines data augmentation with \st to enforce consistency across augmented versions of the same input. For a sentence $\tilde{x}$, augmentation leads to a set of variations $\{\tilde{x}^{\prime}_1, \tilde{x}^{\prime}_2, \dots, \tilde{x}^{\prime}_k\}$, enabling the student model to generalise more effectively by training on diverse but semantically similar inputs.

\paragraph{Confidence Thresholding.}
Confidence thresholding improves the quality of pseudo-labels by filtering out low-confidence predictions. This ensures that only high-confidence pseudo-labels are used in training, which can prevent noisy labels from negatively affecting model performance. Typically, a confidence threshold $\kappa$ is set, and only those pseudo-labels with probabilities from teacher models exceeding $\kappa$ are retained:
\begin{equation}
    \label{equation:confidence_threshold}
    \tilde{y}_i = 
    \begin{cases} 
      F(\tilde{x}_i, \Theta) & \text{if } \max(F(\tilde{x}_i, \Theta)) \geq \kappa, \\
      \texttt{discard} & \text{otherwise}.
   \end{cases}
\end{equation}
By discarding low-confidence pseudo-labels, confidence thresholding helps focus the training on high-quality, more reliable pseudo-labelled data. Approaches such as FlexMatch \cite{zhang2021flexmatch} and AdaMatch \cite{berthelot2021adamatch} dynamically adjust this threshold, allowing for more flexibility across different unlabelled data distributions, further enhancing the model’s ability to generalise.

\section{Task Specific Fine-tuning}
\label{subsec:traditional_finetuning}

Task-specific fine-tuning is a widely adopted approach for adapting pre-trained language models to particular downstream tasks. This method further trains the pre-trained model on task-specific data, allowing it to specialise its representations for the target application.
This section introduces two task-specific fine-tuning methods, widely used in \nlp, especially for classification tasks.

\paragraph{CLS-based Fine-tuning.}
For LMs trained with the \mlm objective, such as \bert \cite{devlin2018bert} and \roberta \cite{liu2019roberta}, fine-tuning adapts the model's output layer for specific tasks. 
Let $S = \{s_1, s_2, ..., s_n\}$ represent a sequence of input tokens, where $n$ is the total number of tokens. 
These models encode the input text $S$ into a corresponding sequence of hidden vectors ${h_i \in \mathbb{R}^d}$, where $d$ is the dimensionality of the hidden representations.

\cls-based fine-tuning method, introduced with \bert \cite{devlin2018bert}, utilises the special \texttt{[CLS]} token prepended to the input sequence during pre-training.
The model uses this token's output vector $h$ as a holistic sequence representation. 
A task-specific head, typically a feedforward network or MLP, is added atop the \texttt{[CLS]} representation to produce task-specific outputs. 
For example, in a binary classification task, this layer might consist of a single feedforward layer followed by a sigmoid activation function to produce a probability distribution over the two classes.
During fine-tuning, both the pre-trained model and task-specific layer parameters are updated using labelled data.

\paragraph{Prompt-based Fine-tuning.}
\label{subsec:prompt_finetuning}
Prompt-based fine-tuning has emerged as an alternative to traditional \cls-based fine-tuning, aiming to bridge the gap between pre-training objectives and downstream task objectives. 
Prompt-based methods have shown particular promise in few-shot learning scenarios \cite{schick-schutze-2021-exploiting,schick-schutze-2021-just}, where only a small amount of labelled data is available for fine-tuning.
This approach reformulates downstream tasks as \mlm problems, aligning more closely with the pre-training objective.
In prompt-based fine-tuning, the input text $S$ is augmented with a specific prompt template $\mathcal{S}_{\text{prompt}} = \mathcal{T}(S)$, which includes a special \texttt{[MASK]} token. The goal is to predict this masked token, which corresponds to the task label. Formally, the probability of predicting a class $y \in \mathcal{Y}$ is computed as:
\begin{equation}
\label{eqn:marginal-prob}
P(y | S) = P(\mathcal{Z}(y) | \mathcal{S}_{\text{prompt}}),
\end{equation}
where $\mathcal{Z}: \mathcal{Y} \rightarrow \mathcal{V}$ is a verbalizer function that maps the task label space to individual words in the model's vocabulary $\mathcal{V}$.
Prompt templates $\mathcal{T}$ can be categorized into two main types: hard prompts and soft prompts:
\begin{itemize}
    \item \textbf{Hard Prompts.} Hard prompt templates \cite{schick-schutze-2021-exploiting,gao-etal-2021-making,shin-etal-2020-autoprompt,zhou2025riot} involve manually designed text patterns specific to each task. For example, in sentiment analysis, a hard prompt might be "This movie was \texttt{[MASK]} ." While intuitive, hard prompts require careful design and can be sensitive to the specific choice of prompt \cite{pmlr-v139-zhao21c,liu2021gpt}. The performance of the model can vary significantly based on the selected prompt, making this approach potentially suboptimal.
    \item \textbf{Soft Prompts.} To address the limitations of hard prompts, soft prompts \cite{liu2021gpt,zhang2022differentiable} were introduced. Soft prompts use either unused tokens from the vocabulary $\mathcal{V}$ or additional tokens as tunable embeddings for the prompt template. These embeddings can be directly trained with task-specific supervision. The key advantage of soft prompts is that the token embeddings in the prompt template can be updated independently of specific word embeddings after initialization. This approach reduces the effort required in searching for optimal prompt templates and label words. 
\end{itemize}

Other studies \cite{khashabi-etal-2020-unifiedqa,aribandi2022ext,ouyang2022training,wei2021finetuned,mishra-etal-2022-cross,sanhmultitask,wang-etal-2022-super,min-etal-2022-metaicl} have also explored the idea of LMs training on a variety of \nlp tasks with natural language instructions/templates, with the goal of generalizing to unseen tasks.
Similar ideas, prompt transfer, have also been explored in the context of Parameter-Efficient Fine-tuning \cite{gu-etal-2022-ppt,su-etal-2022-transferability,vu-etal-2022-spot,shi2023dept}, which seeks to learn an effective representation of the soft prompt for the target task by training on other tasks.

\section{Parameter-Efficient Fine-tuning}
\label{subsec:parameter_efficient_finetuning}
Parameter-Efficient Fine-tuning (\peft) refers to a class of methods designed to adapt pre-trained language models to downstream tasks while updating only a small subset of the model's parameters. 
% These approaches aim to reduce the computational and storage costs associated with full fine-tuning, especially for large language models, while maintaining comparable performance.
% \paragraph{Parameter-efficient Fine-tuning.} 
In contrast to standard fine-tuning and prompt-based fine-tuning \citep{devlin2018bert,schick-schutze-2021-exploiting,shi_dont_2023} where full parameters are updated, 
\peft approaches have demonstrated remarkable performance across a wide range of tasks \citep{wang-etal-2018-glue,pmlr-v202-wu23d,10.1007/978-3-030-99736-6_20,10.1145/3609225,yang2023theory}
while updating only a limited number of parameters.
%
% In the field of \nlp, 
Adapters \citep{houlsby2019parameter}, along with its variants, HyperFormer \citep{mahabadi2021parameter} and Compacter \citep{karimi2021compacter}, add new trainable modules (adapters) to each transformer block of the T5 model \citep{t5}.
BitFit \citep{zaken2021bitfit} limits updates only to the bias parameters, while this method tends to underperform on larger networks \citep{lialin2023scaling}.
Prefix-tuning \citep{li-liang-2021-prefix} adds a soft prompt, parameterized by a feed-forward network, to the model input.
Diff pruning \citep{guo2021diff} learns a sparse update of a neural network’s weights at the cost of more memory usage. 
FishMask \citep{NEURIPS2021_cb2653f5} also performs sparse updates, but it is computationally intensive and inefficient on contemporary deep learning hardware \citep{lialin2023scaling}.
Low-Rank Adaptation (LoRA) \citep{hu2021lora} employs a straightforward low-rank matrix decomposition to parameterise the weight update. 
(IA)$^3$ \citep{NEURIPS2022_0cde695b} scales activations by learned vectors for few-shot learning.
LST \citep{sunglst} operates a small transformer network on the side of the pre-trained network, aiming to decrease the training memory.
Prompt Tuning (\pt) \citep{lester-etal-2021-power} appends a trainable soft prompt to the model input embeddings. 
In comparison to the above-mentioned \peft approaches, \pt uses fewer trainable parameters, which do not proliferate as the model size expands.
\citep{mao-etal-2022-unipelt} introduces a method that combines Prefix-tuning, Adapters, and LoRA through a gating mechanism. 
\dept is also applicable to this method and can be easily integrated with other \peft approaches.
Here we introduce two representative methods, \pt and LoRA.

\paragraph{Prompt Tuning.}
\pt \cite{lester-etal-2021-power} is a \peft method that builds upon the concept of soft prompts introduced in prompt-based fine-tuning. In \pt, instead of fine-tuning the entire model, only a small set of continuous task-specific vectors are learned.
Let $L \triangleq \{\bm{x}_i, \bm{y}_i\}_{i=1}^{N}$ denote $N$ labeled training data for the target task $\mathcal{T}$. Given a backbone model parameterized by $\Theta$, each input text $\bm{x}_i$ is mapped into a sequence of word embeddings $\mE_i \in \mathbb{R}^{s \times d}$, where $s$ is the maximum sequence length and $d$ is the dimension of word embeddings.
PT appends a trainable prompt matrix $\mP \in \mathbb{R}^{l \times d}$ to the frozen word embedding matrix $\mE_i$, where $l$ is a hyperparameter representing the number of virtual tokens. The soft prompt $\mP$ can be initialized either randomly or by sampling word embeddings from the vocabulary. Consequently, the model's input becomes the combined matrix $[\mP; \mE_i] \in \mathbb{R}^{(l + s) \times d}$.
The targeted loss function for \pt is formulated as:
\begin{align}
\label{eq:pt_loss}
\mathcal{L}_{\text{PT}}= -\sum_i \log P(\bm{y}_i | [\mP, \mE_i] ,; \Theta),
\end{align}
where the loss function is optimised only with respect to the soft prompt matrix $\mP$, while the pre-trained model parameters $\Theta$ remain frozen.

Several works aim to enhance the performance of \pt through \petl.
PPT \citep{gu-etal-2022-ppt} strives to improve the performance of \pt \citep{lester-etal-2021-power} by further pre-training \citep{gururangan-etal-2020-dont,shi-etal-2023-rethinking}, which necessitates a set of hand-crafted, task-specific designs and considerable computational cost.
\citep{su-etal-2022-transferability} improves \pt via prompt transfer across different tasks and models.
SPoT \citep{vu-etal-2022-spot} adopts a single prompt, chosen based on a similarity measure at the cost of a massive search.
ATTEMPT \citep{asai-etal-2022-attempt} employs an attention mechanism over the source prompts to initialize the prompt for target tasks at the cost of extra parameters.
MPT \citep{wang2023multitask} applies a shared soft prompt across different tasks, while its effectiveness for a broad range of source tasks remains untested.
% While transfer learning can potentially enhance the performance of the \pt, 
We find that \petl for \pt \citep{asai-etal-2022-attempt,wang2023multitask} can efficiently accelerate training convergence, and that \petl for \pt is more useful for improving the model performance in the few-shot learning setting for PT \citep{gu-etal-2022-ppt,10.1145/3485447.3512036}. 
However, when extensive labelled datasets are available, training \pt for additional steps typically leads to performance improvements. 

\paragraph{Low-Rank Adaptation.}
LoRA \cite{hu2021lora} is another \peft method that introduces trainable low-rank decomposition matrices to the weights of the pre-trained model. LoRA is based on the hypothesis that the updates to the model during fine-tuning have a low intrinsic rank.
For a pre-trained weight matrix $W \in \mathbb{R}^{d \times k}$, LoRA parameterizes its update by a low-rank decomposition:
\begin{align}
\label{eq:lora}
W' = W + BA,
\end{align}
where $B \in \mathbb{R}^{d \times r}$ and $A \in \mathbb{R}^{r \times k}$ are the trainable matrices, and $r \ll \min(d,k)$ is the rank of the decomposition. During inference, the update $BA$ can be merged with the original weights $W$ for efficient computation.
The loss function for LoRA can be expressed as:
\begin{align}
\label{eq:lora_loss}
\mathcal{L}_{\text{LoRA}}= -\sum_i \log P(\bm{y}_i | \bm{x}_i ,; \Theta, A, B),
\end{align}
where $\Theta$ represents the frozen pre-trained model parameters, and only $A$ and $B$ are updated during fine-tuning.

\paragraph{Comparison between \pt and LoRA.} \pt and LoRA differs in the following points:
\begin{itemize}
    \item \textbf{Relative Performance of LoRA and \pt.} When adapting LMs to specialised domains, like mathematical reasoning, which requires much different knowledge than what LLMs have been trained on, LoRA may perform better than \pt. However, in case tasks have already been somewhat understood by LMs and the key challenge is just to properly prompt the LMs, \pt can be the better option. \pt modifies minimal model parameters, focusing instead on improving the input prompt, which has been proven more effective than LoRA \citep{asai-etal-2022-attempt,wang2023multitask}.
    \item \textbf{Specific Use Cases for \pt.} \pt offers advantages in particular cases. For example, soft prompts can be used to compress few-shot examples in the prompt or long context \citep{chevalier2023adapting,wingate-etal-2022-prompt}. While the number of trainable parameters is low, LoRA updates the weight matrices across the whole model. In contrast, \pt only improves the input of the LM through the soft prompt, which helps the model focus on understanding the task and context better rather than learning new knowledge.
    \item \textbf{Parameter Efficiency.} Unlike LoRA, which requires trainable parameters at each layer, \pt contains less number of trainable parameters that are more concentrated.
\end{itemize}

\section{Instruction Tuning}
\label{subsec:instruction_tuning}
Instruction tuning aims to improve LMs' ability to follow specific user instructions or prompts. 
LMs can better align with user intents through fine-tuning on datasets consisting of instructions and human-written completions \cite{bai2022training,ouyang2022training,yang2024bayesianrm,wu2024understanding}.
Early studies mainly focus on \nlp tasks, showing that fine-tuning with various \nlp datasets trained with instruction output pairs improves cross-task generalisation \cite{aribandi2022ext,khashabi-etal-2020-unifiedqa,mishra-etal-2022-cross,ouyang2022training,sanhmultitask,wei2021finetuned}.
Recent works explore the creation of instruction tuning datasets by LLMs themselves \cite{wang-etal-2023-self-instruct,honovich-etal-2023-unnatural,xu2024wizardlm,li2024selfalignment} or through crowdsourcing approaches \cite{vicuna2023,zhou2023lima}.
Such instruction-tuning phrase \cite{ivison2023camels,singh2024aya,hosking2024human,yang2024bayesian} enables LLMs to generalise beyond instructions in the training set, largely enhancing their practical utility.

\paragraph{Training Objectives.} Specifically, in instruction tuning, each data consists of two main components: an instruction $I$ and a completion $C$. The instruction $I$ is a sequence of tokens ${I_1, I_2, \ldots, I_m}$ that describes the task or provides context. The completion $C$ is the desired output sequence ${C_1, C_2, \ldots, C_n}$ that the model should generate in response to the instruction.
The instruction sequence $I$ may include special prompt template tokens (such as "\texttt{<|user|>}" and "\texttt{<|assistant|>}") to structure the input in a dialogue format or to delineate different parts of the prompt. The total input sequence $x$ is the concatenation of the instruction and completion: ${I_1, I_2, \ldots, I_m, C_1, C_2, \ldots, C_n}$.

The model is trained to predict each token in the completion $C$ given all the previous tokens in both the instruction $I$ and the completion $C$ up to that point. This can be expressed probabilistically as:
\begin{align}
\label{eq:it_prob}
P(C_1, C_2, \ldots, C_n | I_1, I_2, \ldots, I_m) = \prod_{j=1}^n P(C_j | I_1, I_2, \ldots, I_m, C_1, C_2, \ldots, C_{j-1})
\end{align}
This formulation captures the autoregressive nature of the language model, where each token prediction is conditioned on all preceding tokens. The loss function $\mathcal{L}$ for instruction tuning is defined as the negative log-likelihood of the completions given the instructions:
\begin{multline}
\label{eq:it_loss}
\mathcal{L} = -\log P(C_1, C_2, \ldots, C_n | I_1, I_2, \ldots, I_m) = \\-\sum_{j=1}^n \log P(C_j | I_1, I_2, \ldots, I_m, C_1, C_2, \ldots, C_{j-1})
\end{multline}
This loss function encourages the model to maximise the probability of generating the correct completion tokens given the instruction and previous tokens.

\paragraph{Data Selection and Scale.} 
Instruction tuning offers several significant implications for LM development and application. 
However, instruction tuning also presents certain challenges. 
The performance of instruction-tuned models heavily depends on the quality and diversity of the instruction-completion pairs used for training. 
LMs may exhibit sensitivity to the exact phrasing of instructions, potentially leading to inconsistent behaviour. 
Additionally, creating high-quality instruction-completion pairs for a wide range of tasks can be resource-intensive, posing scalability challenges.
Research on instruction tuning for LMs presents diverging perspectives on the optimal data scale for supervised fine-tuning.
A prevailing view recommends fine-tuning on expansive datasets to enhance LM performance across various NLP tasks, thereby improving zero-shot and few-shot learning capabilities \cite{aribandi2022ext,khashabi-etal-2020-unifiedqa,ouyang2022training,wei2021finetuned,mishra-etal-2022-cross,sanhmultitask,wang-etal-2022-super,min-etal-2022-metaicl}. For example, \flan comprises over a million question-answer pairs from diverse NLP sources \cite{JMLR:v25:23-0870}, and \texttt{Natural Instructions} features 61 distinct tasks and 193k task instances \cite{mishra-etal-2022-cross}.
Conversely, another research trajectory prioritises data quality over quantity \cite{gunasekar2023textbooks,xu2023rethinking,liu2023makes,jha2023limit}.
The Superficial Alignment Hypothesis (\sah) \cite{zhou2023lima}
advocates for using smaller, high-quality datasets, arguing that LMs primarily acquire their capabilities during the pretraining phase and thus require only minimal data for effective instruction tuning. 
For instance, LIMA \cite{zhou2023lima} employs a carefully curated set of 1k diverse prompts to generate stylistically consistent responses, aimed at creating a helpful AI assistant. 
AlpaGasus \cite{chen2024alpagasus} and LESS \cite{xia2024less} employ methods to select high-quality data based on LLM-generated judgements and gradient signals.
% Our understanding is that for the general purpose of the LLMs, for the task-specific LLMs.
However, both views agree on the importance of
(1) the quality of pre-trained base LMs and
(2) the diversity and quality of the \sft data.

\section{Preference Learning}
\label{subsec:preference_learning}
Preference learning in language models has evolved along two main research directions: (1) Reinforcement Learning from Human Feedback (RLHF), which uses explicit reward modelling and reinforcement learning to optimise for human preferences, and (2) Direct Alignment Algorithms (DAAs), which aim to directly optimise the model's policy without intermediate reward learning. 
Below we discuss each approach in detail.

\paragraph{Reinforcement Learning from Human Feedback.}
RLHF is a powerful approach that has gained significant attention, especially in the context of training AI models to align with human values and objectives. 
The primary goal of RLHF is to teach AI agents to optimise behaviour based on human preferences, which are often complex, subjective, and not easily defined by a straightforward reward function. 
RLHF typically contains three stages:
\begin{itemize}
    \item \textbf{Supervised fine-tuning.} The process begins by fine-tuning a pre-trained language model with supervised learning on high-quality data specifically curated for the downstream task(s) of interest, such as dialogue or summarisation. This produces an initial model $\pi_{\text{ref}}$ that serves as both the starting point and reference policy for subsequent optimisation.
    \item \textbf{Preference Collection and Reward Modelling.} The SFT model generates pairs of completions $(y_1, y_2)$ for given prompts $x$, where $y_1, y_2 \sim \pi_{\text{ref}}(y|x)$. Human annotators then express preferences between these pairs, denoted as $y_w \succ y_l|x$, where $y_w$ and $y_l$ represent the preferred and dispreferred completions respectively. These preferences are assumed to be generated by a reward model $r^*(x,y)$. Following the Bradley-Terry model, the human preference distribution $p^*$ is modelled as: \begin{equation}
        p^*(y_1\succ y_2 \mid x)=\frac{\exp(r^*(x, y_1))}{\exp(r^*(x, y_1)) + \exp(r^*(x, y_2))}.
    \end{equation}
    Given a dataset of comparisons $\mathcal{D}=\{x^{(i)}, y_w^{(i)}, y_l^{(i)}\}_{i=1}^N$, a parametric reward model $r_{\phi}(x,y)$ is trained using the negative log-likelihood loss:
    \begin{equation}
        \mathcal{L}_R(r_{\phi}, \mathcal{D}) = -\mathbb{E}_{(x, y_w, y_l)\sim \mathcal{D}}\bigl[\log \sigma(r_{\phi}(x, y_w)- r_{\phi}(x, y_l))\bigr].
    \end{equation}
    Importantly, $r_{\phi}(x,y)$ is typically initialised from the SFT model with an additional linear layer on top of the final transformer layer to produce scalar reward predictions. The rewards are normalised such that $\mathbb{E}_{x,y\sim \mathcal{D}}[r_\phi(x,y)] = 0$ for all $x$ to reduce variance.
    
    \item \textbf{Reinforcement Learning Fine-tuning.}
    The final phase uses the learned reward model to guide policy optimisation. The loss objective is formulated as:
    \begin{equation}
        \max_{\pi_{\theta}}  \mathbb{E}_{x\sim \mathcal{D}, y\sim \pi_{\theta}(y \mid x)}\bigl[r_{\phi}(x, y)\bigr] - \beta\mathbb{D}_{\textrm{KL}}\bigl[\pi_{\theta}(y\mid x)\mid \mid \pi_{\text{ref}}(y\mid x)\bigr],
    \end{equation}
    where $\pi_{\theta}$ is the policy being optimised (initialized from $\pi_{\text{ref}}$), and $\beta$ controls the KL divergence constraint. This constraint serves two crucial purposes: preventing the policy from deviating too far from the distribution where the reward model is accurate, and maintaining generation diversity by avoiding mode collapse to single high-reward answers. Due to the discrete nature of language generation, this objective is typically optimised using Proximal Policy Optimisation (PPO) \cite{schulman2017proximal} with a modified reward function $r(x,y) = r_{\phi}(x,y) - \beta(\log \pi_{\theta}(y|x) - \log \pi_{\text{ref}}(y|x))$.
\end{itemize}

\paragraph{Direct Alignment Algorithms.}
In contrast to RLHF's multi-stage approach, Direct Alignment Algorithms (DAAs) are designed to train LMs to align with human preferences without the need for explicit reward modelling. These algorithms aim to optimise a policy model to maximise the probability of better completions over worse ones directly from preference data:
\begin{itemize}
    \item Direct Preference Optimisation (DPO) \citep{rafailov2023direct} is a foundational DAA method. The DPO loss function is defined as follows:
    \begin{align}
        \mathcal{L}_{\text{DPO}}(\pi_\theta; \pi_{\text{ref}}) &= -\mathbb{E}_{(x,y_w,y_l)\sim D} \left[\log \sigma\left(\beta \Delta(x,y_w,y_l)\right)\right], \\
        \Delta(x,y_w,y_l) &= \log\frac{\pi_\theta(y_w | x)}{\pi_{\text{ref}}(y_w | x)} - \log\frac{\pi_\theta(y_l | x)}{\pi_{\text{ref}}(y_l | x)},
    \end{align}
    where $\pi_\theta$ is the policy model being optimised, $\pi_{\text{ref}}$ is a reference model where $\pi_\theta$ is initialised from, $D$ is the dataset of preference pairs, $x$ is the input, $y_w$ and $y_l$ are the better and worse completions respectively, $\sigma$ is the sigmoid function, and $\beta$ is a temperature hyperparameter. 
    The term $\Delta(x,y_w,y_l)$ quantifies the difference in log probabilities between better and worse completions.
    \item Identity Preference Optimisation (IPO) \citep{azar2024general} is a variant of DAA methods. Specifically, IPO uses a quadratic loss function, which is defined as:
    \begin{equation}
    \mathcal{L}_{\text{IPO}}(\pi_\theta; \pi_{\text{ref}}) = \mathbb{E}_{(x,y_w,y_l)\sim D} \left[\left(\tau\Delta(x,y_w,y_l) - \frac{1}{2}\right)^2\right],
    \end{equation}
    where $\tau$ is a temperature hyperparameter. This formulation aims to push the difference in log probabilities $\Delta(x,y_w,y_l)$, defined within the DPO framework, towards a target value of $\frac{1}{2\tau}$.
    \item The hinge loss method \citep{zhao2023slic,liu2024statistical} represents another variation within the DAA framework. Specifically, the loss function from \textsc{SLiC-HF} \citep{zhao2023slic} is defined as follows:
    \begin{equation}
        \mathcal{L}_{\text{Hinge}}(\pi_\theta; \pi_{\text{ref}}) = \mathbb{E}_{(x,y_w,y_l)\sim D} \left[\max\left(0, \gamma - \log\frac{\pi_\theta(y_w | x)}{\pi_\theta(y_l | x)}\right)\right],
    \end{equation}
    where $\gamma$ is a hyperparameter. 
    The hinge loss method also incorporates SFT loss over the better completion as a regularisation term.
\end{itemize}

Standard DAAs do not guarantee an increase in the absolute probability of better completions. This can lead to scenarios where the model assigns very low probabilities to both better and worse completions, as long as the better completion has a higher relative probability.
To mitigate this issue, Negative Log-Likelihood (NLL) loss is commonly employed as a regularisation term in DAA \citep{hong2024orpo,pang2024iterative,adler2024nemotron,dubey2024llama}. It encourages the policy to maintain a high likelihood of better completions. The NLL loss is formulated as:
\begin{equation}
\mathcal{L}_{\text{NLL}}(\pi_\theta) = -\mathbb{E}_{(x,y_w,y_l)\sim D} \left[\log\pi_\theta(y_w | x)\right],
\end{equation}
where $y_w$ represents the better completion for a given input $x$. This loss term is typically combined with the primary objective of the DAA using a scaling coefficient $\lambda$. Several other regularisation methods have been proposed. For example, \cite{pal2024smaug} introduces an additional term, \(-\max\left(0, \log\frac{\pi_{\text{ref}}(y_w | x)}{\pi_\theta(y_w | x)}\right)\), to \(\Delta(x,y_w,y_l)\) to ensure that the log-likelihood of better examples remains high relative to that of the reference model. 

\paragraph{Regularisation Through Language Modelling Objectives.}
Pretraining data and language modelling objectives have been used as a regularisation technique in fine-tuning LMs. 
In particular, previous works \cite{clark-etal-2018-semi,liu-etal-2023-compositional} fine-tune LMs on labelled data, with unsupervised learning on unlabelled data for auxiliary tasks as regularisation. 
% \cite{liu-etal-2023-compositional} fine-tunes the T5 model \cite{t5} with regularization through masked language modelling objectives on in-domain data.  
Ouyang et al. \cite{ouyang2022training} mix the alignment objective with the next token prediction objective using pretraining data to mitigate alignment tax in RLHF. 
He et al.\cite{he2023preserving} adopt the masked language objective on the pretraining or downstream task corpus to preserve pre-trained features, and show improvements in calibration and accuracy.
Mathew et al. \cite{Mathew2024Instruction} investigate the effect of incorporating instruction loss weighting on instruction tuning, suggesting that the instruction loss ratio is an important hyperparameter when fine-tuning short-completion data but is irrelevant when using long-completion data.

\section{Summary and Limitations}
In the preceding sections, we explored a diverse set of post-training techniques developed to extend the capabilities of LLMs beyond their initial pre-training. 
These approaches play a critical role in adapting LMs to specialised tasks, new domains, and user preferences. While each technique contributes uniquely to the overall landscape of post-training, they also reveal several limitations for further research.

Continued pre-training enables models to acquire new capabilities by exposing them to additional data—whether for domain adaptation or architectural modifications (\eg longer context windows). It employs the same objectives as the original pre-training (such as MLM or CLM) and is particularly effective when models are trained on high-quality domain-specific corpora. However, its reliance on large-scale unlabelled data remains a common assumption, despite recent evidence that performance improvements can be achieved even with a few hundred examples. Furthermore, if not carefully balanced with general-domain data, continued pre-training risks catastrophic forgetting of previously acquired knowledge, and practical deployment can be constrained by its computational cost.

Self-training, as another semi-supervised strategy, offers a powerful way to leverage unlabelled data through the generation of pseudo-labels by a teacher model. This is followed by training a student model using both labelled and pseudo-labelled data. When combined with techniques such as consistency regularisation, strong data augmentation, and confidence thresholding, self-training can substantially enhance model robustness and generalisation. However, the approach is highly sensitive to the quality of pseudo-labels and thresholding parameters, and it often incurs significant computational overhead due to iterative retraining. Moreover, noise in pseudo-labelled data can reinforce model biases rather than correct them.

Task-specific fine-tuning remains one of the most direct ways to adapt a model to a downstream task. 
CLS-based fine-tuning, typically used with models like BERT, adds task-specific layers on top of a fixed architecture and trains the entire network on labelled data. While this approach is effective when ample labelled data is available, it is prone to overfitting in low-resource settings. 
Prompt-based fine-tuning offers an alternative that reformulates downstream tasks to better match the model’s original pre-training objectives, using either hard (manually designed) or soft (learnable) prompts. However, prompt quality can significantly affect performance, and soft prompt tuning often requires careful optimisation and still struggles to generalise across tasks without sufficient tuning data.

PEFT methods have emerged as a response to the increasing cost of full model updates. Approaches such as \pt and LoRA allow models to be adapted by updating only a small fraction of parameters. \pt modifies input embeddings using trainable vectors, offering minimal parameter updates and good performance in settings where models already possess relevant knowledge. However, it can underperform in tasks requiring domain-specific reasoning. LoRA, in contrast, introduces trainable low-rank decompositions directly into weight matrices, offering greater flexibility and stronger performance in complex domains, albeit with higher memory and compute costs compared to prompt-based methods. The choice between PEFT techniques often depends on the structure of the task and available resources.

Instruction tuning has gained prominence for improving model alignment with user intent, particularly in few-shot and zero-shot scenarios. By training on (instruction, completion) pairs—either human-authored or generated by LLMs—models learn to respond more coherently to diverse prompts. Despite notable success, the effectiveness of instruction tuning remains highly sensitive to the phrasing and quality of instructions. Moreover, there is an ongoing debate about the value of data scale versus data quality: while some studies advocate training on millions of diverse examples, others argue that a small set of carefully curated prompts may suffice for high alignment performance. In both cases, inconsistencies in model responses due to subtle changes in instruction phrasing remain a concern.

These observations underscore the importance of developing more \textbf{efficient}, \textbf{robust}, and \textbf{general-purpose} post-training pipelines, particularly for low-resource and specialised settings. The subsequent chapters explore how to address these limitations through novel strategies that combine the strengths of existing techniques while mitigating their individual weaknesses.

\chapter{The Power of Continued Pre-training}
\label{chapter:continued_pretraining}

\section{Introduction}
\label{sec:neurips_paper}
Pre-training LMs \citep{devlin2018bert,liu2019roberta,radford2019language} over massive unlabelled data and then fine-tuning on task-specific labelled data for the specific downstream task offer large performance gains across \nlp tasks. 
\textit{Semi-Supervised Learning} (\ssl) \citep{grandvalet2004semi,chapelle2009semi,kipf2017semi} is a powerful and effective approach to utilise unlabelled data.
A typical \ssl setting assumes access to a (relatively small) labelled training set and an (often large) unlabelled set. The goal of \ssl is to make effective use of the unlabelled data to improve LM performance. A key focus of this chapter is to explore methods for effectively utilising unlabelled data to improve LM performance. 

\begin{figure}[!t]
\centering
\includegraphics[width=0.9\textwidth]{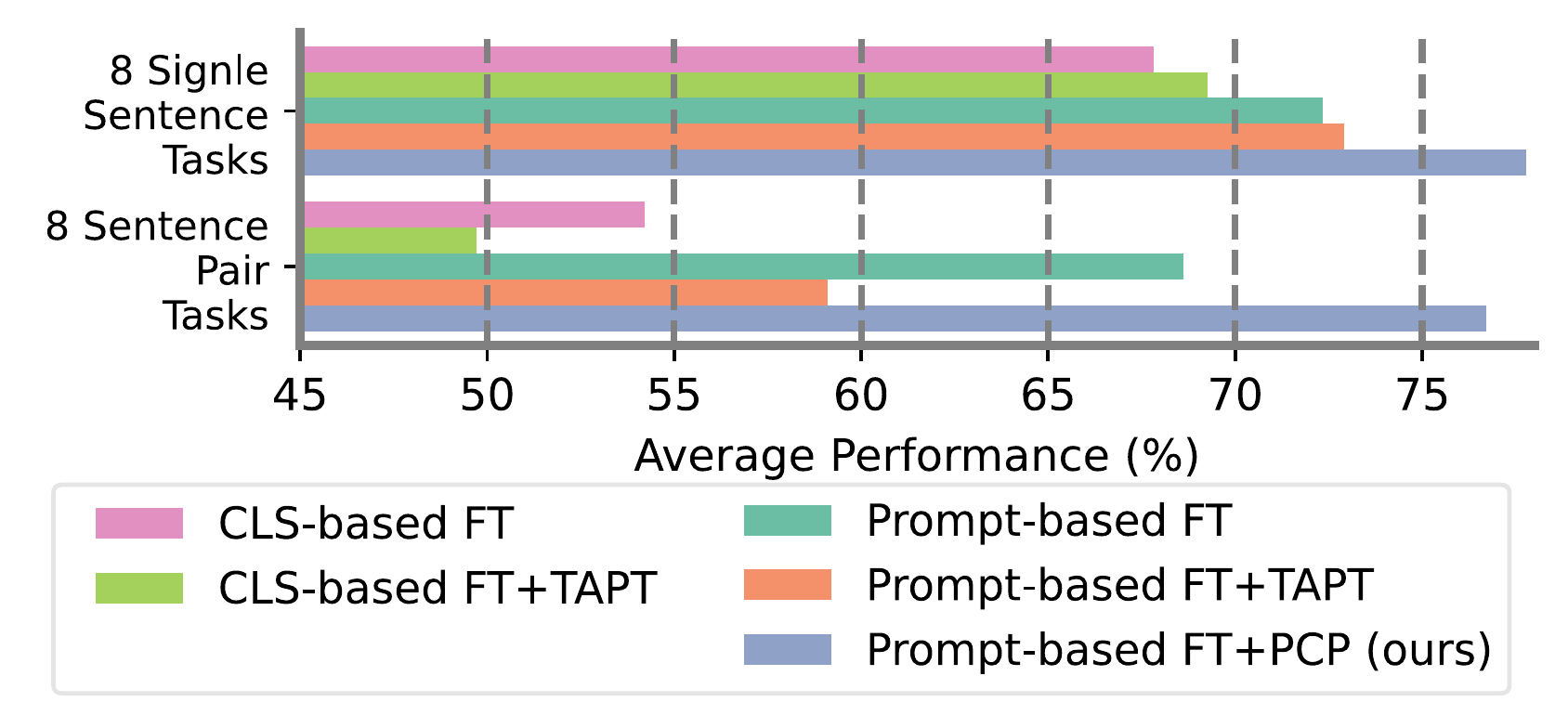}
\caption{Mean performance of \cls- and prompt-based \ft across 16 \nlp tasks when trained by \textit{themselves} or in combination with either \tapt or our proposed \pcp in the semi-supervised setting. Please refer to Table \ref{neurips:table:main_results} for details.}
\label{neurips:fig:preview}
\end{figure}

\begin{figure*}[t!]
  \centering
  \includegraphics[width=\textwidth]{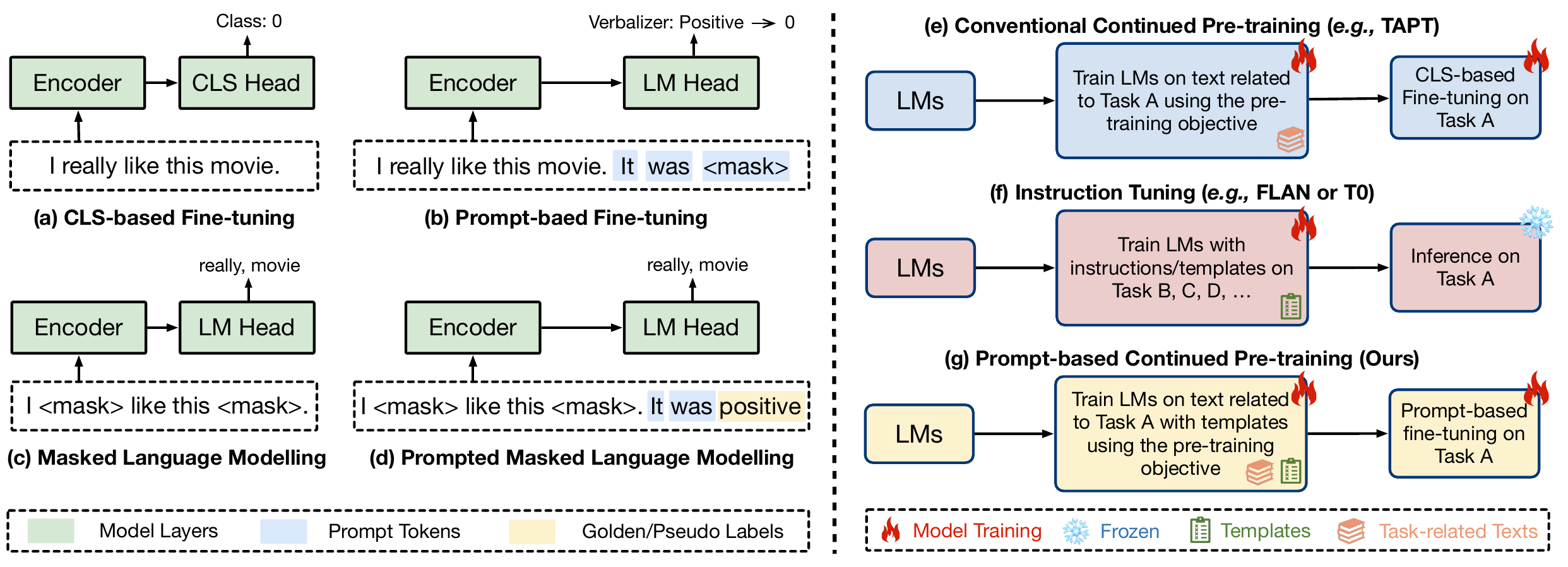}
  \caption{
  The overview of \textbf{Prompt-based Continued Pre-training} (g), in comparison to conventional continued pre-training (e) and instruction tuning (f), along with fine-tuning methods (a,b) and continued pre-training techniques (c,d).
  The verbalizer functions as a mapping from the task label space to individual words.
  We use masked language modelling for illustrative purposes, where <mask> represents a masked token in the LM vocabulary.
  }
  \label{neurips:fig:overview}
\end{figure*}

In this Chapter, we first evaluate the effectiveness of Task Adaptive Pre-training (\tapt) in comparison to Self-Training (\st) approaches within semi-supervised settings. Our analysis highlights the often-overlooked yet remarkable potential of \tapt as a straightforward, stable, and effective \ssl method (Section \sect{sec:acl_paper}).
Then, we re-visit the commonly held belief in \nlp \citep{howard-ruder-2018-universal,sun2019fine,gururangan-etal-2020-dont} that continued pre-training LMs on either task-specific data \citep{alsentzer-etal-2019-publicly,margatina-etal-2022-importance} or in-domain data \citep{logeswaran-etal-2019-zero,xue-etal-2021-mt5} is generally beneficial for improving the performance of fine-tuning (\ft) on downstream tasks.
As shown in Figure \ref{neurips:fig:preview}, our experiments on eight single-sentence tasks and eight sentence-pair tasks in both semi- and fully-supervised settings reveal that conventional continued pre-training on task-specific data \citep{gururangan-etal-2020-dont}, known as task adaptive pre-training (\tapt) (see Figure \ref{neurips:fig:overview}e):
(1) can lead to a substantial drop in performance of \cls-based \ft (see Figure \ref{neurips:fig:overview}a) on sentence-pair tasks; and
(2) may perform unstably across different tasks for prompt-based \ft (see Figure \ref{neurips:fig:overview}b), which is typically considered a better alternative to \cls-based \ft by previous studies \citep{schick-schutze-2021-exploiting,le-scao-rush-2021-many} (see Section \sect{neurips:para:taptissue}).
These findings suggest that exclusively \textbf{presenting task-related texts to LMs} through continued pre-training may not be the most effective approach for improving the performance of \ft in the aforementioned situations.

Recent research \citep{khashabi-etal-2020-unifiedqa,aribandi2022ext,ouyang2022training,wei2021finetuned,mishra-etal-2022-cross,sanhmultitask,wang-etal-2022-super,min-etal-2022-metaicl} on cross-task generalization has demonstrated the impressive improvement on zero-shot or few-shot learning capabilities of LMs (see Figure \ref{neurips:fig:overview}f).
These studies suggest that \textbf{presenting appropriate instructions/prompt templates to LMs} through training on a range of NLP tasks improves their downstream performance on held-out tasks.
Although these works train LMs with different objectives from pre-training phases, we interpret ``fine-tuning LMs on a range of NLP tasks'' as a special type of continued pre-training. 
Therefore, we hypothesize that \textbf{presenting both task-related texts and instructions/prompt templates to LMs} can relieve the above-mentioned issues for conventional continued pre-training and be beneficial for the target task performance.
Rather than improve the generalizability of the LMs with supervised objectives, our work places a greater emphasis on enhancing specific target task performance with unsupervised pre-training objectives. 

In this work, we propose Prompt-based Continued Pre-training (\pcp) (see Section \sect{neurips:sec:approach}), which integrates instructions/prompt templates into task-related texts with golden or pseudo labels (see Figure \ref{neurips:fig:overview}g). 
Our experiments demonstrate that \pcp consistently improves the performance of state-of-the-art prompt-based \ft approaches \citep{gao-etal-2021-making,zhang2022differentiable} in both semi- and fully-supervised settings, covering both single sentence tasks and sentence pair tasks, and that the performance gains from \pcp exceed those from conventional continued pre-training (\tapt) by a substantial margin (see Section \sect{neurips:sec:comparsion_with_conventional_continued_pretraining}).
In the most favourable case, \pcp boosts the performance of prompt-based \ft by more than 20\% absolute while \tapt results in a 9.2\% performance decline.
Furthermore, our results show that \pcp outperforms state-of-the-art semi-supervised approaches \citep{sohn2020fixmatch,10.5555/3495724.3496249,xu2021dash,zhang2021flexmatch,berthelot2021adamatch} with greater simplicity, eliminating the need for an iterative process and extra data augmentation (see Section \sect{neurips:sec:comparsion_with_self_training}).
Additionally, our analysis suggests that the \pcp can efficiently improve the performance of prompt-based \ft with only hundreds of unlabelled examples.
% , while previous work \citep{gu-etal-2022-ppt} pursuing a similar goal with ours requires a 10GB English corpus for continued pre-training. 
Meanwhile, our analysis explores the performance lower bound of the \pcp and reveals that the advantages of \pcp persist across different sizes of models and datasets (see Section \sect{neurips:sec:analysis}).
Finally, we outline the limitations of our study and suggest avenues for future research (see Section \sect{neurips:sec:limitations}).

In summary, the main contributions of this chapter are as follows:
\begin{itemize}
    \item An extensive empirical study to directly compare five state-of-the-art \st approaches and \tapt across various \nlp tasks in \ssl, with varying amounts of labelled and unlabelled data. We offer practical insights learned about the limitations of \st approaches, alongside an exploration of the often-unrecognized yet impressive capacity of  \tapt as a simple, stable and powerful \ssl learner;
    \item Our study empirically demonstrates that \tapt might not be as effective as initially thought and can even negatively impact fine-tuning performance, particularly in sentence pair tasks or when utilising prompt-based \ft;
    \item Our evaluation on 21 classification and regression \nlp tasks shows that our proposed method \pcp provides a superior option to conventional continue pre-training for prompt-based \ft.
    This approach consistently yields performance improvements in diverse model and dataset settings, even with only a few hundred unlabelled examples.
    Moreover, it can outperform state-of-the-art semi-supervised approaches with greater simplification;
    \item Our result shows the effectiveness of presenting both task-related texts and templates/instructions to the LMs through unsupervised pre-training objectives on improving the performance of prompt-based \ft on downstream tasks. To the best of our knowledge, this is the first work to perform instruction tuning via unsupervised objectives.
\end{itemize}

\section{Continued Pre-training as a Powerful Semi-supervised Learner}
\label{sec:acl_paper}

In \nlp, \textit{Self-Training} (\st) approaches have been proposed to produce pseudo labels for unlabelled examples to train the model \citep{yarowsky-1995-unsupervised,mcclosky-etal-2006-effective}. With the advent of neural networks, \st approaches typically focus on using student-teacher models to assign pseudo-labels to the unlabelled data \citep{artetxe-etal-2018-robust,cai-lapata-2019-semi,dong-de-melo-2019-robust,10.5555/3495724.3496249,gera2022zero}.
Apart from the sophisticated \st approaches, previous work \citep{gururangan-etal-2020-dont} proposed \textit{Task Adaptive Pre-Training} (\tapt), which is a straightforward yet effective method for utilising unlabelled examples. This method involves continuing pre-training the LM on the task-specific data without using labels, before proceeding with fully-supervised fine-tuning.
\tapt and \st are both motivated by the need for effectively leveraging unlabelled examples, raising the questions of how \tapt performs in \ssl tasks, as well as how these two approaches perform against each other.

In this section, we investigate the performance of \tapt against the five state-of-the-art \st approaches across five \nlp tasks (Section \sect{sec:st_vs_tapt}). We empirically show that \tapt outperforms all state-of-the-art \st approaches on several tasks, suggesting that it should serve as a strong baseline for \ssl methods. 
Previous research \citep{gururangan-etal-2020-dont} has shown that TAPT can improve performance in fully-supervised settings. This section goes further by showing that \tapt can be even more effective in \ssl settings.

\subsection{Experimental Setup}
\label{sec:setup}
\begin{table*}[hbt!]
\centering
\begin{adjustbox}{max width=\textwidth}
\begin{tabular}{llrrrrcc}
\toprule
\textbf Dataset                                 & \textbf Task Type           & \textbf Train Size & \textbf Dev. Size  & \textbf Test Size  & \textbf{$|\mathcal{Y}|$} & \textbf{$L$}  \\ \midrule
\imdb \cite{maas-etal-2011-learning}        & Movie Review Sentiment         & 23,000         & 2,000          & 25,000         & 2         &  149 \\ % 8H
\sst \cite{wang-etal-2018-glue}            & Movie Review Sentiment          & 60,000         & 7,349          & 872            & 2         & 37 \\
\ag \cite{10.5555/2969239.2969312}         & News Topic Classification       & 100,000        & 10,000         & 7,600          & 4         & 134\\ % 20H
\amazon \cite{10.1145/2507157.2507163}     & Product Review Sentiment        & 250,000        & 25,000         & 650,000        & 5         & 79 \\
% \yelp         & Product Review Sentiment       & 250,000        & 25,000         & 50,000         & 5          \\
\yahoo \cite{chang2008importance}         & Topic Classification             & 500,000        & 50,000         & 60,000         & 10        & 32 \\ 
\bottomrule
\end{tabular}
\end{adjustbox}
\caption{Statistics of datasets. $|\mathcal{Y}|$: \# of classes for classification tasks. $L$: average \# of words in input sentence(s). Note that we only sample examples from the original training set in our experiments.}
\label{table:acl:datasets}
\end{table*}

\paragraph{Datasets.} We experiment with five datasets used in previous related work for \ssl \citep{gururangan-etal-2019-variational,chen-etal-2020-mixtext,10.5555/3495724.3496249,li-etal-2021-semi-supervised,gera2022zero}, including \imdb \citep{maas-etal-2011-learning}, \sst \citep{wang-etal-2018-glue}, \ag \citep{10.5555/2969239.2969312}, \amazon \citep{10.1145/2507157.2507163}, 
and \yahoo \citep{chang2008importance}. 
Table \ref{table:acl:datasets} shows data statistics. We also provide descriptions and examples of datasets, as well as the process for quantifying the similarity between datasets in Appendix \ref{appendix:dataset_similarity}. Adhering to previous work \cite[\eg][]{chen-etal-2020-mixtext,wang2022usb}, we sample the same amount of labelled data per class from the train set, given the labelled size, to form the labelled set. We re-sample the labelled data using the same five seeds for all different approaches and report the average performance with an error bar.

\paragraph{\tapt.} % Following the setting in the previous work \citep{gururangan-etal-2020-dont}, 
Our approach to \textit{task adaptive pre-training} (\tapt) using \robertabase \citep{liu2019roberta} is to further pre-train on the training text corpus including labelled and unlabelled data.  
The model is then fine-tuned on the labelled data where the \texttt{[CLS]} token representation is passed to an extra feed-forward layer for classification.
see Table \ref{neurips:sec:implementation_details} in Appendix for hyperparameter details. 
The process of \tapt + \textsc{Fine-tuning} is denoted by \tapt henceforth.

\paragraph{\st.} We implement five state-of-the-art \st approaches, including VAT \citep{miyato2018virtual}, FixMatch \citep{sohn2020fixmatch}, Dash \citep{xu2021dash}, FlexMatch \citep{zhang2021flexmatch}, and AdaMatch \citep{berthelot2021adamatch} (see descriptions of these approaches in Appendix \ref{neurips:appendix:baseline_models}). 
We use \robertabase as the backbone, and the \texttt{[CLS]} token representation with an extra feed-forward layer is used for classification (see Appendix \ref{neurips:sec:implementation_details} for hyperparameter details). Adhering to previous work \citep{10.5555/3495724.3496249,wang2022usb}, back-translation \citep{ott2019fairseq} is used for data augmentation.

\paragraph{Baselines.} For reference, we also evaluate two baseline models that are only fine-tuned (from an off-the-shelf \robertabase checkpoint) on (1) the same labelled set as \tapt and \st (\partials); and (2) the whole training set (\fullys).

\begin{table*}[t!]
\centering
\resizebox{\textwidth}{!}{
\begin{tabular}{lccccccccccc}
\toprule
\multirow{2}{*}{\textbf{Method}} & \multicolumn{2}{c}{\textbf \imdb}      & \multicolumn{2}{c}{\textbf \sst}     & \multicolumn{2}{c}{\textbf \ag}     & \multicolumn{2}{c}{\textbf \amazon}   & \multicolumn{2}{c}{\textbf \yahoo}
\cr                         \cmidrule(lr){2-3}                   \cmidrule(lr){4-5}                \cmidrule(lr){6-7}                 \cmidrule(lr){8-9}               \cmidrule(lr){10-11}                           
                            & 20              & 100              & 40             & 100              & 40           & 200              & 250           & 1000                    & 500          & 2000     \\ 
\midrule
\multicolumn{11}{l}{\textbf{\st Approaches}}  \\
\vat                        & 90.2$_{0.9}$    & 92.0$_{0.4}$.    &\se75.0$_{12.0}$&\hl86.2$_{3.4}$    &\hl87.5$_{1.0}$ &\hl89.5$_{0.7}$ & 52.2$_{1.3}$  & 57.5$_{0.2}$       & 66.9$_{0.5}$ & 68.6$_{0.2}$  \\
\fixmatch                   &\se93.4$_{0.1}$  & 93.4$_{0.1}$     & 37.3$_{8.5}$   &  66.4$_{21.3}$    & 75.6$_{8.7}$ &\se88.8$_{0.6}$   &\se55.9$_{1.1}$& 59.0$_{0.5}$        & 67.5$_{1.0}$ &\se69.6$_{0.4}$  \\
\dash                       & 93.2$_{0.3}$    & 93.4$_{0.2}$     & 38.2$_{10.1}$  & 73.3$_{18.6}$   & 74.3$_{6.6}$ &  88.5$_{0.6}$    & 56.6$_{1.8}$  & 59.3$_{0.2}$      & 67.6$_{1.0}$ & 69.5$_{0.3}$   \\
\flexmatch                  & 93.3$_{0.1}$    &\se93.4$_{0.1}$   & 40.6$_{7.7}$   &  83.0$_{8.3}$    & 80.6$_{4.4}$ &  88.2$_{0.5}$    & 54.9$_{3.9}$  & 58.8$_{0.4}$        & 66.6$_{0.7}$ & 68.7$_{0.4}$   \\
\adamatch                   &\hl94.4$_{0.4}$. &\hl94.7$_{0.2}$   & 42.6$_{13.3}$  & 83.1$_{4.4}$   & 82.7$_{5.9}$ &  88.6$_{0.4}$    & 55.5$_{2.8}$  &\se59.0$_{0.7}$     &\se68.0$_{0.7}$& 69.5$_{0.3}$   \\
\midrule
\partials                   & 83.3$_{7.4}$  & 88.7$_{0.2}$       & 74.7$_{6.1}$   & 84.0$_{2.7}$   &\se84.6$_{1.6}$& 88.0$_{0.8}$    & 53.1$_{0.7}$  & 57.2$_{0.1}$       & 65.4$_{0.3}$ & 68.5$_{0.3}$ \\
\quad + \tapt               & 86.9$_{2.8}$  & 90.9$_{0.6}$       &\hl 82.6$_{4.0}$&\se 85.4$_{2.4}$  & 84.0$_{1.3}$&  88.7$_{0.7}$     &\hl58.4$_{0.7}$&\hl60.6$_{0.1}$  &\hl68.8$_{0.7}$&\hl71.5$_{0.3}$\\
\midrule
\fullys                    &\multicolumn{2}{c}{93.9$_{0.1}$}    & \multicolumn{2}{c}{93.0$_{0.6}$} & \multicolumn{2}{c}{94.8$_{0.1}$} & \multicolumn{2}{c}{65.0$_{0.2}$} & \multicolumn{2}{c}{75.3$_{0.2}$} \\
\quad + \tapt               &\multicolumn{2}{c}{94.0$_{0.2}$}    & \multicolumn{2}{c}{93.5$_{0.3}$} &\multicolumn{2}{c}{95.0$_{0.1}$}   &\multicolumn{2}{c}{65.6$_{0.1}$}   &\multicolumn{2}{c}{75.4$_{0.1}$} \\
\bottomrule
\end{tabular}
}
\caption{Performance of \tapt, \st approaches and the baselines across five datasets using two different sizes of the training labelled data. We report average Macro-$F_1$ on the test set across five seeds, with standard deviations in subscripts. Blue and orange represent the best and second-best performance in a column respectively.}
\label{table:main_results}
\end{table*}

\begin{figure*}[!ht]
  \centering
  \includegraphics[width=\textwidth]{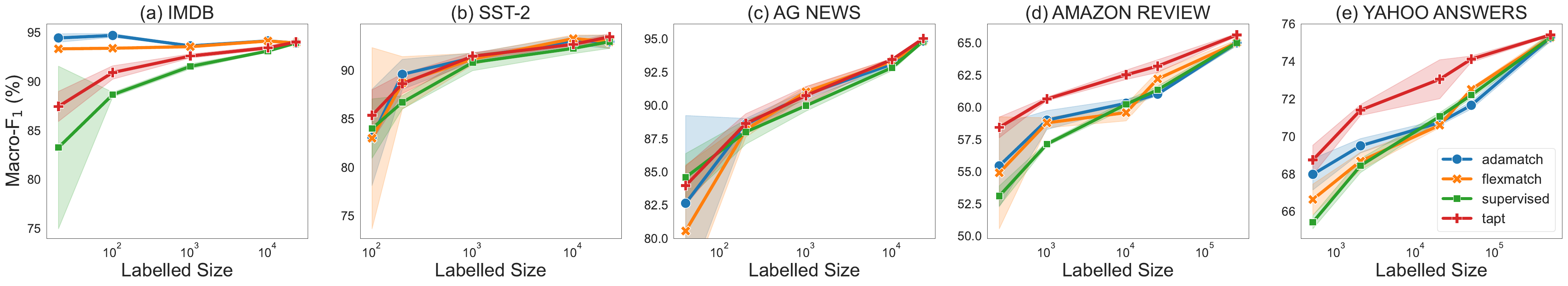}
  \caption{The effect of labelled size on \tapt and \st. Average test Macro-$F_1$ score over 5 seeds is reported. From the left to the right, \tapt and \st utilizes $23$k, $60$k, $100$k, $250$k, and $500$k unlabelled samples respectively.}
  \label{fig:f1_wrt_labelled_sample_size}
\end{figure*}

\subsection{Empirical Results: Self-Training vs Task Adaptive Pre-training}
\label{sec:st_vs_tapt}

\paragraph{Overview.} \label{paragraph:overview} 
Table \ref{table:main_results} shows the performance of \tapt against five state-of-the-art \st approaches and the baselines (\partials and \fullys) across five datasets, each with two different sizes of labelled data for training following the previous work \citep{wang2022usb}. 
Overall, we observe that:
(\hyperref[paragraph:41]{\color{black}1}) \tapt achieves highly competitive results compared with state-of-the-art \st approaches; and
(\hyperref[paragraph:42]{\color{black}2}) \tapt gains more improvement compared to the \partials baselines when using fewer labelled samples. 

For our first finding, the experimental results show that \tapt outperforms all five state-of-the-art \st approaches with lower variances on \amazon, and \yahoo, as shown in Table \ref{table:main_results}. For example, \tapt obtains a $F_1$ score of 68.8\% compared to the best \st approach's $F_1$ score of 68.0\% (using 500 labelled samples) and 71.5\% compared to \st's 69.6\% (using 2000 labelled samples) on \yahoo. 
For an example of the second finding, \tapt gains 3.6\% $F_1$ improvement over \partials (using 20 labelled samples) compared to 2.2\% (using 100 labelled samples) on \imdb. Below we delve deeper into these two findings and discuss them in more detail.

\paragraph{\#1. \tapt is a strong semi-supervised learner and can outperform state-of-the-art \st approaches.} 
\label{paragraph:41}
Figure \ref{fig:f1_wrt_labelled_sample_size} shows how the performance of \st, \tapt, and \partials vary with respect to five different labelled sizes on each dataset, where two latest \st approaches (\adamatch and \flexmatch) are selected as representatives for \st.
Experimental results further verify that \tapt has a consistent advantage over \adamatch and \flexmatch across different labelled sizes on \amazon and \yahoo. It is also worth noting that, while \tapt brings a stable improvement over \partials across all datasets with varying labelled sizes, \st can sometimes bring more substantial improvement, for example when only a few hundred labelled samples are available from \imdb. However, we do not observe similar phenomena for \st on other datasets.
Our experimental results demonstrate that \tapt is a simple, effective and strong learner for \ssl tasks.

\begin{figure}[!t]
  \centering
  \includegraphics[width=0.75\textwidth]{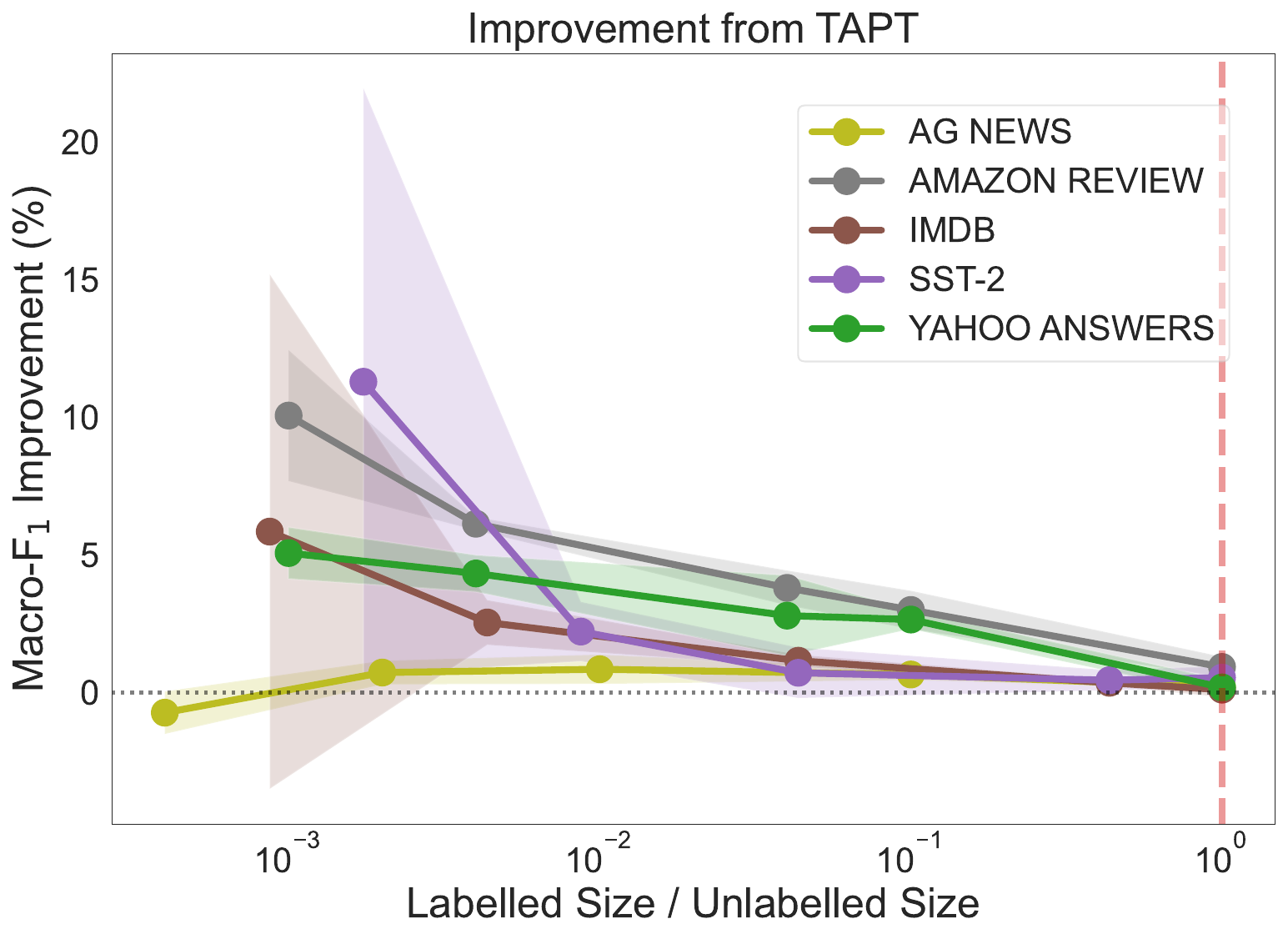}
  \caption{The impact of labelled size on the $F_1$ improvement from \tapt over \partials, where unlabelled size is fixed for each dataset. The red vertical line highlights the \fullys setting on which prior work \citep{gururangan-etal-2020-dont} focuses.}% {\color{red} Add similar figure for \st in Appendix.}{
  \label{fig:f1_wrt_tapt_5}
\end{figure}

\paragraph{\#2. \tapt tends to bring more improvements in \ssl than in \fullys setting.} 
\label{paragraph:42}
We further study the behaviour of \tapt \textit{itself} under \ssl, where we select \partials as the baseline rather than \st approaches.
Figure \ref{fig:f1_wrt_labelled_sample_size} shows that the differences in performance (in absolute values) between \tapt (red lines) and \partials (green lines) generally increase as the labelled size decreases.
To gain a better understanding of the impact of labelled data sizes, we plot the improvement from \tapt over \partials (in percentages) against the ratio between labelled size and unlabelled size (unlabelled size is fixed for each dataset) in Figure \ref{fig:f1_wrt_tapt_5}. We see that \tapt improves over \partials further as the ratio of labelled and unlabelled sizes decreases, highlighting the trends of gaining greater improvement in low-resource \ssl setting. This finding is complementary to prior works \citep{howard-ruder-2018-universal,gururangan-etal-2020-dont} that focus on \tapt's improvement from the \fullys perspective, represented by the rightmost red vertical line in Figure \ref{fig:f1_wrt_tapt_5}.
The rising trend of the improvement is not monotonic as the labelled size is reduced.
Rather it could provide insight into how \tapt improves over \partials in \ssl and inspire the design of new approaches.

\section{Prompt-based Continued Pre-training}
\label{neurips:sec:approach}
In this section, we introduce the proposed method, Prompt-based Continued Pre-training (\pcp), which aims to improve the performance of LMs on downstream tasks through continued pre-training with prompt templates, as shown in Figure \ref{neurips:fig:overview}g.
Let $L \triangleq \{(x_1, y_1), \ldots, (x_N, y_N)\}$ denote $n$ labelled examples and $U \triangleq \{x_1', \ldots, x_M'\}$ denote $m$ unlabelled examples. 
Our approach consists of two main steps, as described below.

\paragraph{Step 1: Construct Continued Pre-training Corpus.} 
Initially, we select a model $F$, pre-trained with the MLM objective and parameterized by $\Theta$. 
We then train this model using the prompt-based \ft, minimising the target loss function $\ell$ on the labelled examples $L$, as illustrated in Figure \ref{neurips:fig:overview}b:
\begin{equation}
    \label{neurips:equation:step_1}
    \mathcal{L}_{\text{Total}}(L) = \sum \limits_{x_i, y_i \in L} \mathcal{L}(y_i, F(\mathcal{T}(x_i), \Theta)),
\end{equation}
Next, we use the trained model $F$ with the learned parameters $\Theta'$ to generate predictions (termed "pseudo-labels") on the unlabelled samples $U$:
\begin{equation}
    \label{neurips:equation:step_2}
    y_i' = F(\mathcal{T}(x_i'), \Theta'),
\end{equation}
For each text example $x$ and its associated (golden or pseudo) label $y$, we create an example for our proposed \pcp as $x^{pcp} = \mathcal{T}(x, \mathcal{Z}(y))$, where the original \texttt{[MASK]} position is substituted with $\mathcal{Z}(y)$. 
This results in a new corpus, $\mathcal{C}=\{x_i^{pcp}\}_{i=1}^{N+M}$.
In the fully-supervised setting, $M=0$ and all examples use the golden labels.

\paragraph{Step 2: Perform continued pre-training and prompt-based \ft.} \label{neurips:para:method_step2}
We then proceed to further pre-train another model $G$, parameterized by $\Theta$, using the \mlm objective on the newly generated corpus $\mathcal{C}$, to obtain the \pcp checkpoint $\Phi$ (see Figure \ref{neurips:fig:overview}d). 
Finally, we train model $G$, initialised by $\Phi$, using Equation \ref{neurips:equation:step_1} with prompt-based \ft for downstream tasks. 

In comparison to conventional continued pre-training, \pcp does not require any modification for the model architecture or training process. 
The sole difference is the addition of a few extra tokens to the input text during continued pre-training. 
% which makes the proposed \pcp easily adaptable to various types of LMs, such as BERT \citep{devlin2018bert}, ALBERT \citep{Lan2020ALBERT}, and RoBERTa \citep{liu2019roberta}.
This modification does not hinder the efficiency of the method, \ie both conventional continued pre-training and \pcp maintain equal levels of efficiency.
%
% Consequently, \pcp can be seamlessly extended to accommodate diverse architectures and objectives, in line with these adaptable prompt-based techniques.
%
In this study, we primarily focus on LMs pre-trained with the \mlm objective \citep{liu2019roberta}. 
It is noteworthy to mention that comprehensive exploration of other architectures \citep{devlin2018bert,t5,10.5555/3495724.3495883,ouyang2022training} remains an avenue for future research. 
Nonetheless, considering prompt-based fine-tuning approaches \citep{liu2021gpt,lester-etal-2021-power,li-liang-2021-prefix} have already been adapted for different model architectures and pre-training objectives \citep{devlin2018bert,t5,10.5555/3495724.3495883,ouyang2022training}.
This implies that extending our method to alternative architectures should be a feasible undertaking.
\section{Experiments and Results}
In this section, we evaluate the proposed method \pcp by comparing it with conventional continued pre-training and four state-of-the-art semi-supervised approaches.
We assess their relative performance across 21 different classification and regression \nlp tasks,
including single-sentence and sentence-pair tasks.
We conduct additional analysis concerning the performance lower bound of \pcp and the effectiveness of the \pcp across varying datasets and model sizes.

\subsection{Experimental Setup}
\paragraph{Datasets.} 
Our study conducts a comprehensive analysis of 21 \nlp datasets, including classification and regression tasks.
Following previous studies \citep{gao-etal-2021-making,hambardzumyan-etal-2021-warp,zhang2022differentiable} on prompt-based \ft, we derive 8 single-sentence tasks and 8 sentence-pair English tasks from the GLUE benchmark \citep{wang-etal-2018-glue}, SNLI \citep{bowman-etal-2015-large}, and 6 other widely used sentence classification tasks (\ie SST-5, MR, CR, MPQA, Subj, TREC).
Additionally, we use 5 popular benchmarks for semi-supervised learning from previous research \citep{gururangan-etal-2019-variational,chen-etal-2020-mixtext,10.5555/3495724.3496249,li-etal-2021-semi-supervised,gera2022zero}, including \imdb \citep{maas-etal-2011-learning}, \ag \citep{10.5555/2969239.2969312}, \yelp \footnote{\url{https://www.yelp.com/dataset}}, \yahoo \citep{chang2008importance}, and \amazon \citep{10.1145/2507157.2507163}.
See dataset details in Appendix \ref{neurips:sec:dataset}.
We train the model with two different settings:
(1) fully-supervised setting, where we train the model with the full training set; and
(2) semi-supervised setting, where we sample the same amount of labelled data per class from the full training set. 
We re-sample the labelled data using the same five seeds for all comparison approaches and report the average performance with an error bar.

\paragraph{All Comparison Approaches.}
In our study, we mainly experiment using the \robertabase ($125$M) and the \robertalarge  ($355$M) models.
We utilise the conventional \cls-based \ft and two state-of-the-art prompt-based \ft approaches:
(1) ``\cls-based \ft'': fine-tuning with the \texttt{[CLS]} token representation with an extra MLP layer;
(2) ``Prompt-based \ft (hard)'': fine-tuning with high-quality manual or auto-generated prompts and label words \citep{schick-schutze-2021-exploiting,gao-etal-2021-making}; and 
(3) ``Prompt-based \ft (soft)'': fine-tuning with soft prompts using additional tokens for both templates and label words \citep{zhang2022differentiable}. Since the objective of soft prompt \ft is to minimize the reliance on human-designed templates, we unify the template for all tasks here. See the template specifics used for each dataset in Appendix \ref{neurips:sec:templates}.  
We train these three types of \ft approaches from three different types of checkpoints to evaluate their relative effectiveness:
(i) the off-the-shelf \robertalarge checkpoint;
(ii) the task-adaptive pre-training (\tapt) checkpoint \citep{gururangan-etal-2020-dont} (represents the conventional continued pre-training). For sentence pair tasks, we concatenate the two sentences as an input example; and 
(iii) the proposed \pcp checkpoint, obtained in Section \cref{neurips:para:method_step2}.
For both (ii) and (iii), we perform \mlm on all full training sets except MNLI, MNLI-mm, SNLI, QNLI, and QQP, where we select up to 10k unlabelled examples from the full training sets (see supplementary experiments on the full training sets in Appendix \ref{neurips:sec:supplementary_experiments}).
Additionally, we compare the proposed \pcp with four state-of-the-art semi-supervised approaches, including FixMatch \citep{sohn2020fixmatch}, Dash \citep{xu2021dash}, FlexMatch \citep{zhang2021flexmatch}, and AdaMatch \citep{berthelot2021adamatch} (see descriptions of these approaches in Appendix \ref{neurips:appendix:baseline_models}), 
where back-translation \citep{ott2019fairseq} is used for data augmentation as previous works \citep{10.5555/3495724.3496249} and prompt-based \ft (hard) is used as the backbone. 
% All above-mentioned models use \robertalarge as the backbone. 
See hyperparameter and implementation details in Appendix \ref{neurips:sec:implementation_details}. 

\begin{table}[!t]
\centering
\resizebox{0.94\textwidth}{!}{%
\begin{tabular}{lllllllll}
\toprule
\multicolumn{9}{c}{\textbf \textit{Single Sentence Tasks}} \\
\midrule
                                         & \tf{SST-2}        & \tf{SST-5}        & \tf{MR}           & \tf{CR}           & \tf{MPQA}         & \tf{Subj}         &  \tf{TREC}         & \tf{CoLA} \\
                                         & (acc)             & (acc)             & (acc)             & (acc)             & (acc)             & (acc)             & (acc)              & (Matt.)\\
\midrule
Majority (full)                          & 50.9         & 23.1       & 50.0         & 50.0         & 50.0         & 50.0         & 18.8          & 0.0  \\
Prompt-based zero-shot$^\dagger$         & 83.6              & 35.0              &  80.8             & 79.5              & 67.6              & 51.4              & 32.0               & 2.0  \\
% \midrule ``GPT-3'' 
in-context learning            & $84.8_{1.3}$      &	$30.6_{0.9}$     &	$80.5_{1.7}$     & $87.4_{0.8}$      & $63.8_{2.1}$      & $53.6_{1.0}$      & $26.2_{2.4}$       &  $-1.5_{2.4}$ \\
\midrule
\multicolumn{9}{l}{\textbf Fully Supervised Learning} \\
\textsc{Cls}-based \ft                   & $95.1$           & $59.4$           & $90.8$           & $90.8$           & $89.1$            & $96.9$           &  $96.8$           & $54.3$  \\
\tableindent + \tapt                     & $96.0$  \uaa{0.9}& $60.6$  \uaa{1.2}& $91.4$  \uaa{0.6}& $91.0$  \uaa{0.2}& $89.9$  \uaa{0.8} & $96.9$  \uaa{0.0}&  $97.6$  \uaa{0.8}& $43.6$   \daa{10.7}\\ \rowcolor{Gray}
Prompt-based \ft (hard)                  & $95.2$           & $60.0$           & $90.8$           & $92.4$           & $89.4$            & $95.9$           & $97.8$            &  $54.7$ \\ \rowcolor{Gray}
\tableindent + \tapt                     & $93.5$  \daa{1.7}& $60.4$  \uaa{0.4}& $90.3$  \daa{0.5}& $90.8$  \daa{1.6}& $89.5$   \uaa{0.1}& $95.9$  \uaa{0.0}& $97.6$   \daa{0.2}&  $44.0$ \daa{10.7}  \\ \rowcolor{Gray}
\tableindent + \pcp (ours)               & $95.5$  \uaa{0.3}& $60.5$  \uaa{0.5}& $91.7$  \uaa{0.9}& $92.8$  \uaa{0.4}& $89.6$   \uaa{0.2}& $96.8$  \uaa{0.9}& $97.8$   \uaa{0.0}&  $56.0$ \uaa{1.3}  \\
Prompt-based \ft (soft)                  & $94.2$           & $59.8$           & $90.4$           & $92.7$           & $87.8$            & $96.4$           & $97.4$            & $61.3$  \\
\tableindent + \tapt                     & $92.7$  \daa{1.5}& $59.5$  \daa{0.3}& $91.8$  \uaa{1.4}& $92.5$  \daa{0.2}& $89.5$  \uaa{1.7} & $96.8$ \uaa{0.4} & $97.8$   \uaa{0.4}& $52.6$ \daa{8.7}   \\
\tableindent + \pcp (ours)               & $94.3$  \uaa{0.1}& $60.7$  \uaa{0.9}& $91.8$  \uaa{1.4}& $92.8$  \uaa{0.1}& $90.4$  \uaa{2.6} & $97.1$ \uaa{0.7} & $98.0$   \uaa{0.6}& $62.0$ \uaa{0.7}   \\
% \tableindent + \pcp    (\fs)  & &   &   &   &   &   &   &   \\
\midrule
\multicolumn{9}{l}{{\textbf Semi Supervised Learning}} \\
\textsc{Cls}-based \ft                   & $81.2_{2.7}$          & $41.7_{1.3}$          & $76.3_{3.2}$          & $79.5_{3.8}$          & $65.1_{12.6}$         & $91.7_{0.4}$          &  $80.3_{5.8}$          & $26.7_{7.8}$ \\
\tableindent + \tapt                     & $88.2_{1.5}$ \uaa{7.0}& $43.4_{2.6}$ \uaa{1.7}& $86.1_{0.7}$ \uaa{9.8}& $86.2_{2.4}$ \uaa{6.7}& $73.7_{4.4}$ \uaa{8.6}& $94.2_{1.5}$ \uaa{2.5}&  $80.4_{6.4}$ \uaa{0.1}& $1.9_{2.4}$ \daa{24.8}\\ \rowcolor{Gray}
Prompt-based \ft (hard)                  & $92.7_{1.3}$          & $46.7_{1.5}$          & $86.2_{1.2}$          & $90.7_{0.8}$          & $80.8_{6.9}$          & $91.0_{1.1}$          & $84.7_{4.4}$           &  $7.2_{5.5}$     \\ \rowcolor{Gray}
\tableindent + \tapt                     & $92.9_{1.0}$ \uaa{0.2}& $48.9_{1.1}$ \uaa{2.2}& $88.4_{0.5}$ \uaa{2.2}& $89.8_{2.3}$ \daa{0.9}& $84.6_{4.9}$ \uaa{3.8}& $93.5_{1.1}$ \uaa{2.5}& $85.2_{2.9}$ \uaa{0.5} &  ~$1.4_{3.5}$ \daa{5.8}\\ \rowcolor{Gray}
\tableindent + \pcp  (ours)              & $93.6_{0.3}$ \uaa{0.9}& $50.9_{1.3}$ \uaa{4.2}& $89.0_{0.6}$ \uaa{2.8}& $92.3_{0.4}$ \uaa{1.6}& $87.9_{0.5}$ \uaa{7.1}& $95.7_{0.4}$ \uaa{4.7}& $90.6_{3.5}$ \uaa{5.9} &  $25.0_{2.9}$ \uaa{17.8}\\
Prompt-based \ft (soft)                  & $92.5_{1.2}$          & $48.0_{0.7}$          & $86.8_{1.4}$          & $90.8_{1.3}$          & $81.2_{6.8}$          & $90.3_{2.1}$          & $83.0_{3.0}$           &  $4.9_{3.7}$ \\
\tableindent + \tapt                     & $93.4_{0.5}$ \uaa{0.9}& $47.0_{1.2}$ \daa{1.0}& $88.5_{0.8}$ \uaa{1.7}& $89.6_{3.4}$ \daa{1.2}& $83.4_{5.1}$ \uaa{2.2}& $93.3_{0.7}$ \uaa{3.0}& $84.5_{2.4}$ \uaa{1.5} &  ~$2.1_{1.8}$ \daa{2.8} \\
\tableindent + \pcp (ours)               & $93.9_{0.3}$ \uaa{1.4}& $50.7_{1.3}$ \uaa{2.7}& $89.8_{0.6}$ \uaa{3.0}& $92.0_{0.5}$ \uaa{1.2}& $88.3_{0.5}$ \uaa{7.1}& $94.9_{0.9}$ \uaa{4.6}& $88.6_{5.4}$ \uaa{5.6} &  $21.5_{2.5}$ \uaa{16.6}\\
% \tableindent + \pcp    (\fs)  & &   &   &   &   &   &   &   \\

\midrule
\multicolumn{9}{c}{\textbf \textit{Sentence Pair Tasks}} \\
\midrule
                                         & \tf{MNLI}        & \tf{MNLI-mm}      & \tf{SNLI}         & \tf{QNLI}         &  \tf{RTE}         & \tf{MRPC}         & \tf{QQP}         & \tf{STS-B} \\
                                         & (acc)            & (acc)             & (acc)             & (acc)             & (acc)             & (F1)              & (F1)             & (Pear.)\\
\midrule
Majority (full)                          & 32.7        & 33.0         &  33.8        & 49.5         & 52.7         & 81.2         & 0.0          & -  \\
Prompt-based zero-shot$^\dagger$         & 50.8             & 51.7              & 49.5              & 50.8              & 51.3              & 61.9              & 49.7              & -3.2   \\
in-context learning                      & $52.0_{0.7}$     & $53.4_{0.6}$      & $47.1_{0.6}$      & $53.8_{0.4}$      & $60.4_{1.4}$      & $45.7_{6.0}$      & $36.1_{5.2}$      & $14.3_{2.8}$ \\
\midrule
\multicolumn{9}{l}{\textbf Fully Supervised Learning} \\
%                                        %  MNLI                   %  MNLI-mm                % SNLI                    % QNLI                    % RTE                     %  MRPC                    % QQP                     %  STS-B      
%                                        %  256 1e-5               %  256 1e-5               % 256  1e-5               % 256 1e-5                % 128 1e-5                %  128  1e-5               % 128 1e-5                %  256 2e-5
\cls-based \ft                           & $82.1$                  & $82.7$                  & $88.1$                  & $90.2$                  & $83.4$                  & $91.9$                   & $79.7$                   & $91.2$  \\
\tableindent + \tapt                     & $81.0$       \daa{1.1}  & $82.0$       \daa{0.7}  & $86.7$      \daa{1.4}   & $85.6$      \daa{4.6}   & $83.4$       \uaa{0.0}  & $91.6$       \daa{0.3}   & $80.2$    \uaa{0.5}     & $90.4$ \daa{0.8}    \\ \rowcolor{Gray}
Prompt-based \ft (hard)                  & $85.4$                  & $85.8$                  & $89.0$                  & $89.6$                  & $88.1$                  & $93.1$                   & $73.8$                  & $91.5$        \\ \rowcolor{Gray}
\tableindent + \tapt                     & $82.8$       \daa{2.6}  & $83.2$       \daa{2.6}  & $88.3$      \daa{0.7}   & $90.9$      \uaa{1.3}   & $83.8$       \daa{4.3}  & $92.7$       \daa{0.4}   & $78.2$    \uaa{4.4}     & $91.2$   \daa{0.3} \\ \rowcolor{Gray}
\tableindent + \pcp (ours)               & $86.5$       \uaa{1.1}  & $86.2$       \uaa{0.4}  & $89.5$      \uaa{0.5}   & $91.5$      \uaa{1.9}   & $88.5$       \uaa{0.4}  & $93.3$       \uaa{0.2}   & $79.6$    \uaa{5.8}     & $91.9$ \uaa{0.4}\\
Prompt-based \ft (soft)                  & $84.6$                  & $85.4$                  & $89.0$                  & $89.5$                  & $84.5$                  & $92.4$                   & $73.9$                  & $91.6$       \\
\tableindent + \tapt                     & $83.5$     \daa{1.1}    & $84.1$      \daa{1.3}   & $88.3$      \daa{0.7}   & $90.9$       \uaa{1.4}  & $82.7$      \daa{1.8}   & $92.6$      \uaa{0.2}    & $79.9$     \uaa{6.0}    & $90.9$  \daa{0.7}  \\
\tableindent + \pcp (ours)               & $85.7$     \uaa{1.1}    & $86.0$      \uaa{0.6}   & $89.5$      \uaa{0.5}   & $91.0$       \uaa{1.5}  & $85.5$      \uaa{1.0}   & $92.6$       \uaa{0.2}   & $79.6$     \uaa{5.7}    & $91.7$  \uaa{0.1} \\
\midrule
\multicolumn{9}{l}{\textbf Semi Supervised Learning} \\
%                                        %  MNLI                   %  MNLI-mm                % SNLI                    % QNLI                    % RTE                     %  MRPC                    % QQP                     %  STS-B      
%                                        %  256 1e-5               %  256 1e-5               % 256  1e-5               % 256 1e-5                % 128 1e-5                %  128  1e-5               % 128 1e-5                %  256 2e-5
\cls-based \ft                           & $46.2_{0.6}$            & $48.5_{1.0}$            & $45.6_{5.4}$            & $61.4_{8.2}$            & $54.2_{4.3}$            & $73.2_{8.7}$             & $58.5_{3.8}$            & $46.0_{16.3}$ \\
\tableindent + \tapt                     & $36.0_{1.0}$ \daa{10.2} & $36.3_{1.1}$ \daa{12.2} & $45.7_{3.6}$ \uaa{0.1}  & $55.6_{2.7}$ \daa{5.8}  & $53.4_{1.0}$ \daa{0.8}  & $67.7_{8.5}$ \daa{5.5}   & $55.0_{4.1}$ \daa{3.5}  & $48.1_{19.6}$ \uaa{2.1} \\ \rowcolor{Gray}
Prompt-based \ft (hard)                  & $67.3_{1.3}$            & $68.9_{1.2}$            & $76.7_{1.6}$            & $66.5_{4.3}$            & $68.3_{3.1}$            & $75.9_{1.6}$             & $66.8_{1.9}$            & $67.7_{8.1}$  \\ \rowcolor{Gray}
\tableindent + \tapt                     & $50.7_{3.9}$ \daa{16.6} & $52.2_{4.6}$ \daa{16.7} & $74.5_{3.1}$ \daa{2.2}  & $55.3_{1.1}$ \daa{11.2} & $59.9_{2.7}$ \daa{8.4}  & $63.2_{6.3}$ \daa{12.7}  & $58.2_{2.6}$ \daa{8.6}  & $63.1_{8.0}$  \daa{4.6} \\ \rowcolor{Gray}
\tableindent + \pcp (ours)               & $75.6_{1.4}$ \uaa{8.3}  & $76.8_{0.9}$ \uaa{7.9}  & $82.4_{1.3}$ \uaa{5.7}  & $85.1_{0.8}$ \uaa{18.6} & $70.2_{2.7}$ \uaa{1.9}  & $80.7_{3.3}$ \uaa{4.8}   & $71.8_{1.3}$ \uaa{5.0}  & $71.5_{8.4}$ \uaa{3.8}\\
Prompt-based \ft (soft)                  & $62.7_{2.2}$            & $65.9_{1.2}$            & $75.4_{0.8}$            & $64.2_{4.7}$            & $68.2_{3.7}$            & $73.0_{10.6}$            & $66.5_{1.8}$            & $63.7_{6.8}$  \\
\tableindent + \tapt                     & $46.6_{3.9}$ \daa{16.1} & $49.5_{6.8}$ \daa{16.4} & $72.1_{2.0}$ \daa{3.3}  & $55.0_{2.3}$ \daa{9.2}  & $58.4_{2.4}$ \daa{9.8}  & $63.3_{5.8}$ \daa{9.7}   & $58.3_{1.9}$ \daa{8.2}  & $65.3_{6.3}$ \uaa{1.6} \\
\tableindent + \pcp (ours)               & $75.4_{0.7}$ \uaa{12.7} & $76.8_{0.3}$ \uaa{10.9} & $82.6_{1.2}$ \uaa{7.2}  & $84.3_{2.0}$ \uaa{20.1} & $70.4_{3.2}$ \uaa{2.2}  & $80.0_{2.4}$ \uaa{7.0}   & $72.3_{1.2}$ \uaa{5.8}  & $71.4_{7.8}$ \uaa{7.7}\\
\midrule
\multicolumn{9}{c}{\textbf \textit{Summary of results: the probability of improving the performance for \tapt and \pcp}} \\
\midrule
                                         & \multicolumn{4}{c}{Single Sentence Tasks}                                     & \multicolumn{4}{c}{Sentence Pair Tasks} 
\cr                                      \cmidrule(lr){2-5}                                                               \cmidrule(lr){6-9}
Checkpoint                &\tapt{\scriptsize (full)}&\pcp{\scriptsize (full)}&\tapt{\scriptsize (semi)}& \pcp{\scriptsize (semi)}&\tapt{\scriptsize(full)}&\pcp{\scriptsize (full)}&\tapt{\scriptsize (semi)}&\pcp{\scriptsize (semi)}\\\midrule
\cls-based \ft                           & 87.5 (7/8)        & \ti{-}            &  87.5 (7/8)       & \ti{-}            &  25.0 (2/8)       & \ti{-}             & 25.0 (2/8)        & \ti{-}   \\
Prompt-based \ft (hard)                  & 37.5 (3/8)        & 100 (8/8)         &  75.0 (6/8)       & 100 (8/8)         &  25.0 (2/8)       & 100 (8/8)          & ~0.0 (0/8)        & 100 (8/8) \\
Prompt-based \ft (soft)                  & 50.0 (4/8)        & 100 (8/8)         &  62.5 (5/8)       & 100 (8/8)         &  37.5 (3/8)       & 100 (8/8)          & 12.5 (1/8)        & 100 (8/8) \\
\bottomrule
\end{tabular}
}
\caption{
Comparison between the \pcp and conventional continued pre-training (\tapt) using \robertalarge.
The summary highlights the percentage of positive impact brought by the \pcp and \tapt.
The mean and standard deviation on test sets are reported over 5 different seeds.
In semi-supervised learning, 16 examples per class are used for training, in line with previous studies \cite{gao-etal-2021-making,liu2021gpt,zhang2022differentiable}. 
Green and red arrows indicate changes with respect to the \ft baselines that do not use \tapt or \pcp.
$\dagger$ represents that no training examples are used.
Three extra baselines sourced from \cite{gao-etal-2021-making} are included, where ``Majority'' refers to the majority class, and ``in-context learning'' indicates the usage of in-context learning \cite{10.5555/3495724.3495883} with \robertalarge, without updating any parameters.
% {``\ssl'':} use additional unlabelled examples for \mlm under the semi-supervised setting;
% {``\fs'':} only use $K = 16$ (per class) for \mlm under the few-shot setting;
% {man:} manual prompt (ref);
% {con:} continuous templates (ref). 
}
\label{neurips:table:main_results}
\end{table}

\subsection{Prompt-based Continued Pre-training vs Conventional Continued Pre-training}
\label{neurips:sec:comparsion_with_conventional_continued_pretraining}
Table \ref{neurips:table:main_results} presents and summarises our experimental results on 8 single-sentence tasks and 8 sentence-pair tasks.
% These results reveal two key findings: (1) conventional continued pre-training does not reliably improve the performance of fine-tuning; and (2) \pcp consistently boosts the performance of prompt-based \ft, surpassing conventional continued pre-training by a large margin in both semi- and fully-supervised setting. 
Below we delve deeper into our two major findings.

\paragraph{\#1. \tapt is not consistently beneficial for sentence pair tasks, nor when prompt-based \ft is employed.}
\label{neurips:para:taptissue}
Initially, we re-visit the impact of \tapt (representing the conventional continued pre-training) on the \cls-based \ft, as shown in Table \ref{neurips:table:main_results}. 
Our experimental results align with earlier studies \citep{howard-ruder-2018-universal,gururangan-etal-2020-dont}, showing that \tapt generally improves the performance of the \cls-based \ft on 7 out of 8 single sentence tasks in both semi-supervised and full-supervised setting. 
However, intriguingly, we observe that \tapt negatively affects the performance of \cls-based \ft on 6 out of 8 sentence pair tasks, as summarised in Figure \ref{neurips:fig:preview} and the at the bottom of Table \ref{neurips:table:main_results}. This finding implies that conventional continued pre-training (\tapt) may not be beneficial for sentence pair tasks.

Moreover, our investigation reveals that \tapt may negatively affect prompt-based \ft. 
Specifically, in the fully supervised setting, \tapt results in reduced performance on 11 out of 16 tasks for prompt-based \ft (hard) and on 9 out of 16 tasks for prompt-based \ft (soft).
In the most favourable scenario, \tapt enhances the performance of prompt-based \ft (soft) from 73.9\% to 79.9\% on the QQP dataset.
Conversely, in the least favourable situation, \tapt diminishes the performance of prompt-based \ft (hard) from 54.7\% to 44.0\% on the CoLA dataset.
In the semi-supervised setting, \tapt leads to a decline in performance on 12 out of 16 tasks for both prompt-based \ft (hard) and prompt-based \ft (soft) (see the summary of results in Figure \ref{neurips:fig:preview} and the at the bottom of Table \ref{neurips:table:main_results}).
Particularly, for sentence pair tasks, \tapt results in an average absolute decrease of 9.5\% in performance for prompt-based \ft.
These results suggest that the effectiveness of \tapt varies across different tasks and cannot be universally applied.
We conduct additional experiments to confirm the limitations of \tapt persist across different sizes of the pre-training corpus in Appendix \ref{neurips:sec:supplementary_experiments}.

\paragraph{\#2. \pcp offers consistent and substantial improvements in both semi- and fully-supervised settings.} 
As depicted in Table \ref{neurips:table:main_results}, our experiments covering 16 datasets in both semi- and fully-supervised settings, including single sentence tasks and sentence pair tasks, reveal that 
(1) \pcp consistently boosts the performance of prompt-based \ft; and that
(2) the performance gains achieved by \pcp consistently exceed those obtained by \tapt by a substantial margin.
Specifically, compared to prompt-based \ft, \pcp leads to more than a 1.0\% average absolute improvement in the fully-supervised setting and contributes to an average absolute performance boost of 6.8\% in the semi-supervised setting across 16 tasks.
Compared to \tapt, \pcp yields over a 1.8\% average absolute improvement in the fully-supervised setting and contributes to an average absolute performance increase of 11.2\% in the semi-supervised setting across 16 tasks.
Notably, \pcp can produce considerable gains in certain datasets.
For instance, it elevates the performance of prompt-based \ft (hard) from 7.2\% (Matthews Correlation Coefficient) to 25.0\%, while \tapt even reduces the performance of the prompt-based \ft.
Additionally, \pcp improves the performance of prompt-based \ft (soft) on the QNLI dataset from 64.2\% to 84.3\% with 31\% improvement, while \tapt leads to a 9.2\% absolute performance decline.
We attribute the improvements to presenting the prompt template to the LMs through the further pre-training phrase, which implies that merely showing task-related texts to the LMs may not be the optimal approach for prompt-based \ft.

\begin{table*}[t]
\centering
\resizebox{\textwidth}{!}{
\begin{tabular}{lccccccccccc}
\toprule
\multirow{2}{*}{\textbf Method}          & \multicolumn{2}{c}{\textbf \imdb}       & \multicolumn{2}{c}{\textbf \ag}     & \multicolumn{2}{c}{\textbf \yelp}  & \multicolumn{2}{c}{\textbf \yahoo}  & \multicolumn{2}{c}{\textbf \amazon} & \multirow{2}{*}{\textbf Mean}
\cr                                  \cmidrule(lr){2-3}                   \cmidrule(lr){4-5}                \cmidrule(lr){6-7}                 \cmidrule(lr){8-9}               \cmidrule(lr){10-11}                        
                                     & 20              & 100               & 40              & 200             & 40              & 200             & 40              & 200             & 40              & 200     &     \\ 
\midrule
\dash \cite{xu2021dash}              & $93.34_{0.7}$   &  $93.30_{0.6}$    & $85.00_{2.9}$   & $87.90_{0.3}$   & $47.44_{2.2}$   & $58.85_{1.0}$   & $60.07_{4.7}$   & $66.46_{0.9}$   & $44.09_{2.6}$   & $53.95_{1.0}$ & 69.04     \\
\fixmatch \cite{sohn2020fixmatch}    & $95.26_{0.4}$\hl&  $94.28_{0.5}$    & $85.44_{1.1}$   & $88.21_{0.4}$   & $47.26_{1.2}$   & $58.51_{0.3}$   & $61.56_{6.9}$   & $68.37_{0.7}$   & $44.26_{2.0}$   & $52.33_{1.7}$ & 69.55    \\ 
\flexmatch \cite{zhang2021flexmatch} & $95.22_{0.3}$\se&  $94.84_{0.4}$\se & $85.33_{1.4}$   & $88.57_{0.6}$   & $50.60_{2.5}$   & $58.34_{1.9}$   & $58.09_{3.7}$   & $66.43_{1.3}$   & $45.48_{3.1}$   & $54.19_{1.1}$ & 69.71     \\
% \quad + \pcp (ours)            &                 &                   &                 &                 &                 &                 &                 &                 &                 &  \\
\adamatch\cite{berthelot2021adamatch}& $95.20_{0.5}$   &  $94.94_{0.1}$\hl & $85.79_{2.1}$   & $88.72_{0.8}$   & $50.42_{2.7}$   & $58.95_{1.9}$   & $63.68_{0.7}$   & $68.09_{0.8}$   & $44.66_{3.4}$   & $53.05_{0.6}$ & 70.35    \\ 
% \quad + \pcp (ours)            &                 &                   &                 &                 &                 &                 &                 &                 &                 &  \\
\midrule
Prompt-based \ft (hard)        & $86.78_{2.1}$   &  $89.52_{1.6}$    & $84.87_{1.1}$   & $86.99_{0.3}$   & $46.69_{4.2}$   & $58.27_{0.7}$   & $60.63_{1.5}$   & $66.94_{1.1}$   & $44.34_{1.5}$   & $57.01_{0.4}$    & 68.20       \\ 
\quad + \pcp (ours)            & $92.49_{1.2}$   &  $94.24_{0.9}$    & $87.06_{1.0}$\se& $88.94_{0.4}$\se& $52.92_{4.5}$\hl& $63.15_{1.3}$\hl& $65.58_{1.8}$\hl& $70.22_{0.9}$\hl& $53.44_{3.0}$\hl& $59.64_{1.6}$\hl & 72.77 \hl   \\
Prompt-based \ft (soft)        & $88.14_{2.9}$   &  $90.80_{1.5}$    & $85.65_{1.3}$   & $87.66_{0.3}$   & $45.43_{3.4}$   & $57.12_{1.0}$   & $61.18_{1.5}$   & $67.85_{0.8}$   & $44.52_{3.6}$   & $55.03_{1.5}$    & 68.34       \\ 
\quad + \pcp (ours)            & $93.53_{1.5}$   &  $94.36_{0.7}$    & $87.26_{0.8}$\hl& $88.96_{0.6}$\hl& $50.66_{3.3}$\se& $62.92_{1.0}$\se& $65.26_{1.5}$\se& $70.03_{0.9}$\se& $52.78_{3.2}$\se& $59.16_{0.8}$\se & 72.49 \se   \\
\midrule
Prompt-based \ft (hard)$\dagger$ &\multicolumn{2}{c}{$95.60$}        &\multicolumn{2}{c}{$91.06$}       &\multicolumn{2}{c}{$68.71$}      &\multicolumn{2}{c}{$74.30$}      &\multicolumn{2}{c}{$63.85$} & 78.70 \\
Prompt-based \ft (soft)$\dagger$ &\multicolumn{2}{c}{$95.50$}        &\multicolumn{2}{c}{$91.10$}       &\multicolumn{2}{c}{$69.63$}      &\multicolumn{2}{c}{$75.66$}      &\multicolumn{2}{c}{$63.32$} & 79.04 \\
\bottomrule
\end{tabular}
}
\vspace{-0.35em}
\caption{Comparison between the \pcp and four semi-supervised approaches using \robertalarge.
Each dataset is evaluated with two different labelled data sizes and a full training set is used as unlabelled data.
$\dagger$ indicates that the full training set is used as the labelled data.
We report the average Macro-$F_1$ score on the test set across five seeds, with standard deviations as subscripts. For each column, blue represents the best performance and orange stands for the second-best performance.}
\label{neurips:table:ssl_results}
\end{table*}
\subsection{Prompt-based Continued Pre-training vs State-of-the-art Semi-supervised Approaches}
\label{neurips:sec:comparsion_with_self_training}
Table \ref{neurips:table:ssl_results} presents our experimental results on five datasets, comparing the proposed \pcp with state-of-the-art semi-supervised approaches. Below we delve deeper into our main finding with a discussion.

\paragraph{The proposed \pcp outperforms state-of-the-art semi-supervised approaches on 4 out of 5 tasks.}
As shown in Table \ref{neurips:table:ssl_results}, our proposed \pcp approach with either hard or soft variants of prompt-based \ft outperforms the best-performing semi-supervised approaches on 4 out of 5 datasets.
Notably, the prompt-based \ft (hard) with the \pcp outperforms the best performing semi-supervised approaches (\flexmatch) with an absolute 5.5\% Macro-$F_1$ score on the \amazon dataset when 200 labelled training examples are used.
While the best performing semi-supervised approach, \fixmatch, outperforms \pcp by 1.7\% in absolute value on the \imdb dataset using 20 labelled examples, the performance discrepancy narrows as the number of labelled training examples increases.
Overall, the prompt-based \ft (hard) and (soft) with the \pcp outperform all these semi-supervised approaches with an average absolute performance improvement of more than 2\% across various datasets and labelled dataset sizes, demonstrating the effectiveness of our proposed approach.
% It is worth noting that all these compared semi-supervised approaches require additional weak or strong data augmentation and complex iterative processes, whereas our proposed \pcp approach achieves highly competitive or even superior performance with simple further pre-training.
% Although our approach can be easily adapted to semi-supervised approaches by training them from the \pcp checkpoint, we find that training semi-supervised approaches from the \pcp checkpoint does not lead to better than the performance of prompt-based \ft with the \pcp (see Appendix \ref{neurips:sec:train_st} for experimental evidence).

\paragraph{Discussion.}
\label{neurips:para:discussion}
State-of-the-art semi-supervised approaches typically rely on generating pseudo labels for unlabelled examples in order to train student and teacher models iteratively \citep{artetxe-etal-2018-robust,cai-lapata-2019-semi,xie2020self,dong-de-melo-2019-robust,10.5555/3495724.3496249,gera2022zero}. However, this iterative process is prone to \textit{confirmation bias} \citep{tarvainen2017mean,arazo2020pseudo,goel2022pars}, which can result in error accumulation if the pseudo label is incorrect at any iterative step \citep{li_dividemix_2020,wang2021selftuning,goel2022pars,DST2022}.
Various efforts have been made to mitigate \textit{confirmation bias}, such as using only high-confidence pseudo labels \citep{sohn2020fixmatch,zhang2021flexmatch,berthelot2021adamatch} or relying heavily on data augmentation \citep{10.5555/3495724.3496249,chen-etal-2020-mixtext,berthelot2019remixmatch}.
While these efforts make the training process more sophisticated, the issue remains difficult to fully address \citep{DST2022}.
Our proposed method offers an alternative way to utilise pseudo labels different from previous semi-supervised approaches \citep{yarowsky-1995-unsupervised,mcclosky-etal-2006-effective}. 
Instead of relying on an iteration process with direct supervision signals from pseudo labels, we incorporate pseudo labels through continued pre-training with an unsupervised objective (\ie \mlm).
%
% Previous research has suggested that LMs possess more world and task knowledge than previously assumed \citep{liu2021gpt}, and continued pre-training phrases with pseudo labels may be helpful for LMs to explore additional knowledge within texts through unsupervised signals.
%
While our proposed approach may not always outperform semi-supervised approaches across all benchmarks, it delivers highly competitive performance while significantly streamlining the process by removing the necessity for iteration and additional data augmentation.
We will discuss the efficiency of the proposed \pcp later (Section \cref{neurips:para:computational_costs}).
Additionally, \pcp is orthogonal to these semi-supervised approaches and can be combined easily by initialising their backbone from the \pcp checkpoint.
In future work, we plan to investigate the more specific use cases where our proposed \pcp may be preferred over these semi-supervised approaches.

\subsection{Further Analysis}
\label{neurips:sec:analysis}
\begin{figure*}[t!]
  \centering
  \includegraphics[width=\textwidth]{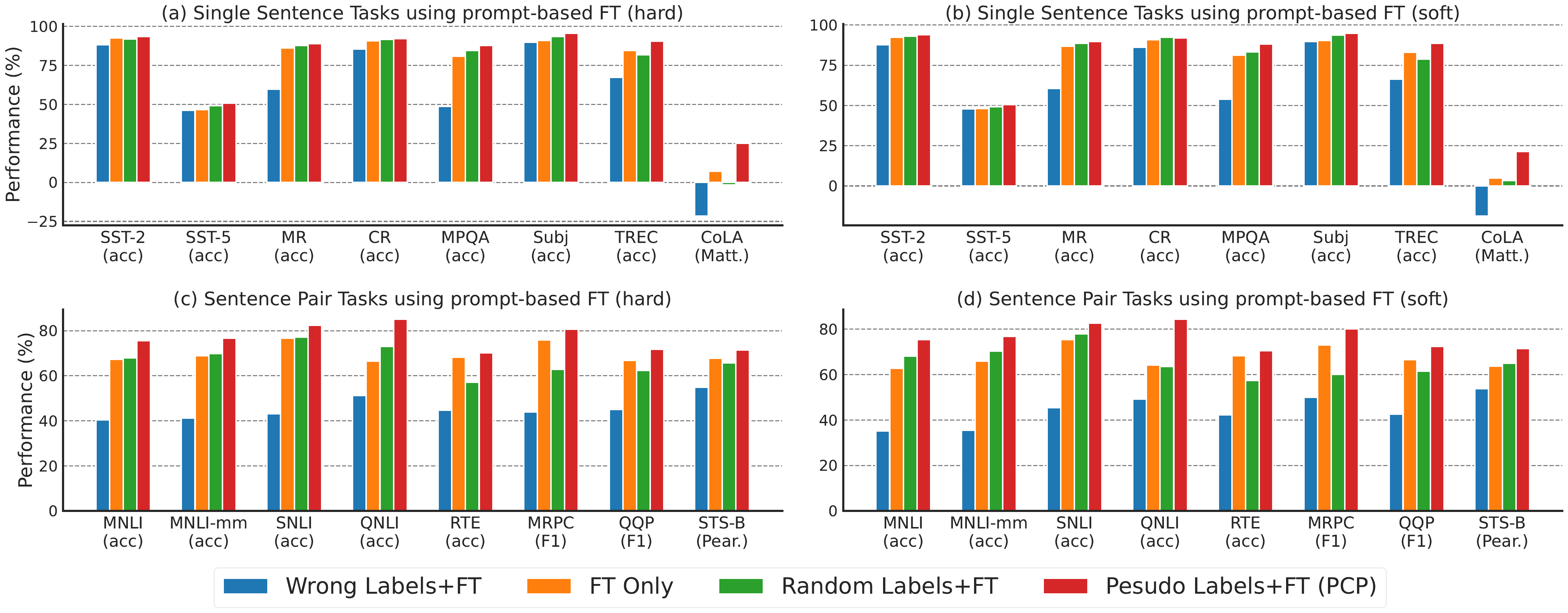}
  \caption{The performance lower bound of the \pcp, where 
  ``wrong labels'' indicates that all labels in the \pcp are incorrect and 
  ``random labels'' indicates that all labels in the \pcp are randomly selected. 
   For each dataset, 16 examples per class are used as labelled data and the full training set is used as unlabelled data. The mean performance on test sets is reported over 5 different seeds.}
  \label{neurips:fig:lower_bound}
\end{figure*}

\paragraph{\#1. What is the lower bound of the model performance using the \pcp?}
To understand the lower bound of \pcp performance, we conduct additional analysis with two different configurations of pseudo labels in \pcp: 
(1) all pseudo labels are incorrect; and 
(2) all labels are randomly selected. 
Figure \ref{neurips:fig:lower_bound} depicts the performance using different types of pseudo labels. 
We use the prompt-based \ft without \pcp (shown in yellow) and with \pcp (shown in red) as baselines.
Experimental results indicate that using incorrect pseudo labels (shown in blue) typically leads to inferior performance. 
In experiments using two prompt-based \ft on 16 datasets, we find that using random labels leads to improved outcomes in 19 out of 32 scenarios. 
This suggests that \pcp with random labels has over a 50\% chance of improving the performance of prompt-based \ft, indicating that the performance lower bound is satisfactory.
Additionally, \pcp with random labels improves the performance on sentence pair tasks in 8 out of 16 cases, while \tapt leads to poorer results in 15 of 16 cases (refer to Table \ref{neurips:table:main_results}). This suggests that \pcp can be advantageous even when using random labels, providing benefits in scenarios where \tapt falls short.
Interestingly, unlike prior study \citep{min-etal-2022-rethinking} on in-context learning \citep{10.5555/3495724.3495883}, where LMs using random labels in demonstrations perform close to those using ground-truth labels, our results show that using pseudo labels assigned by a trained model (shown in red) consistently leads to the better performance, highlighting the importance of accurate pseudo labels.

\begin{figure*}[t!]
  \centering
  \includegraphics[width=\textwidth]{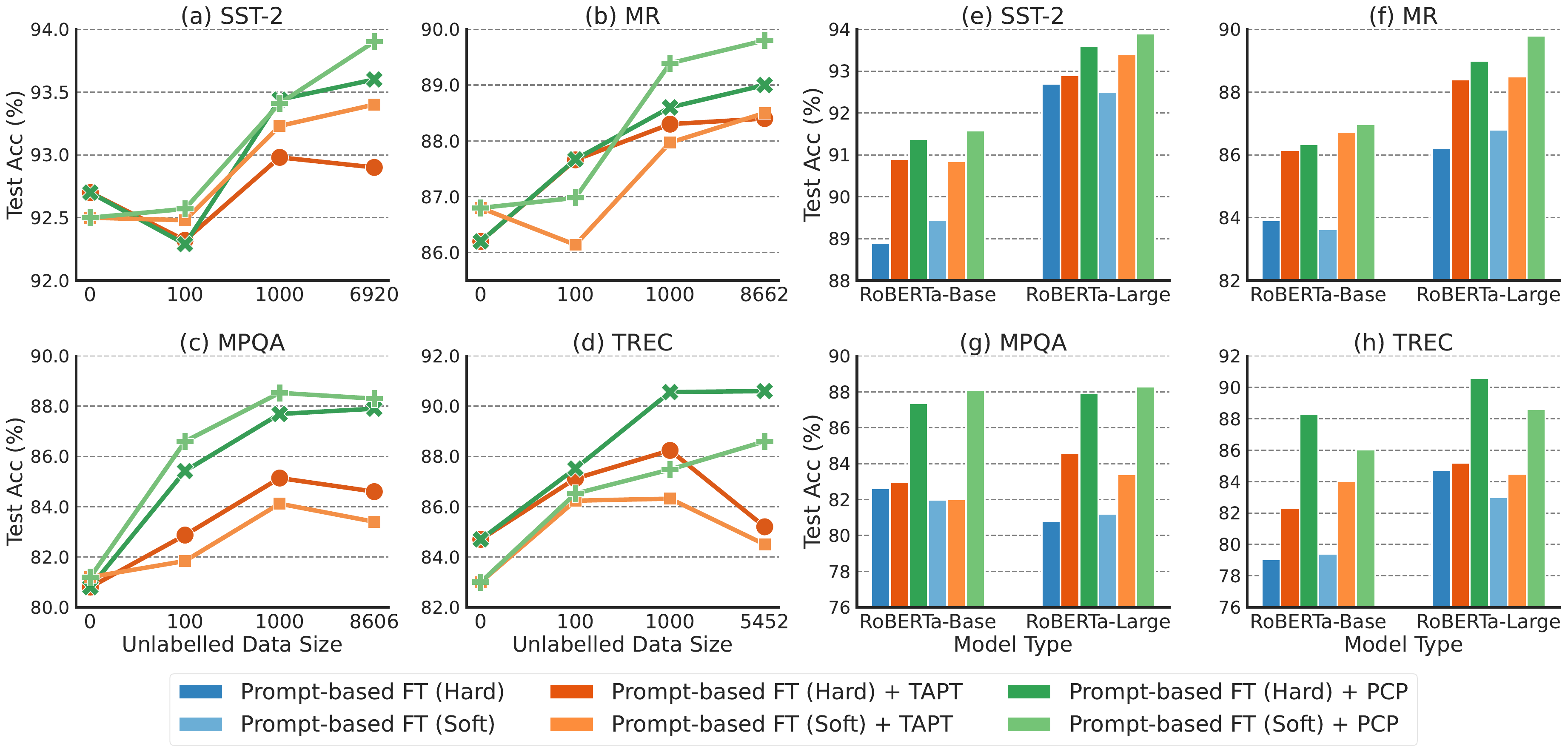}
  \caption{
  (Left) The effect of different unlabelled data sizes using \robertalarge. 
  (Right) The effect of Scaling Laws, where \robertabase (123M) and \robertalarge (354M). 
  All comparison approaches are trained with 16 examples per class for each dataset.
  }
  \label{neurips:fig:size_effect_and_scaling_law}
\end{figure*}

\begin{table}[!ht]
\centering
\resizebox{0.75\textwidth}{!}{
\begin{tabular}{lrccc}
\toprule
\textbf Dataset & \textbf Size &  \textbf \ft &  \textbf +\tapt &  \textbf +\pcp  \\
\midrule
\imdb       & 23K  & $87.3_{1.2}$ & $88.9_{1.3}$ & $91.4_{0.5}$ \hll \\
\ag         & 100K & $86.4_{0.9}$ & $87.6_{1.1}$ & $88.0_{0.4}$ \hll \\
\yelp       & 250K & $52.4_{2.5}$ & $60.3_{1.9}$ & $61.44_{2.0}$ \hll \\
\amazon     & 250K & $51.2_{1.8}$ & $56.8_{1.2}$ & $57.0_{1.5}$ \hll \\
\yahoo      & 500K & $64.9_{0.8}$ & $64.9_{1.1}$ & $69.0_{1.4}$ \hll \\
\bottomrule
\end{tabular}
}
\caption{Test Results for prompt-based \ft (soft) using \robertabase with varying continued pre-training corpus sizes. 
Average Macro-$F_1$ with standard deviations are reported across five seeds. The model is trained on the \imdb dataset using 100 labelled examples and uses 200 labelled examples for other datasets. The best performance for each dataset is highlighted in blue.}
\label{neurips:table:larger_corpus}
\end{table}

\paragraph{\#2. What are the requirements of data size and computational resources for the \pcp?}
\label{neurips:para:computational_costs}
To gain a deeper understanding of the efficacy of our proposed \pcp method, we conduct additional analysis to determine the number of data points necessary for the \pcp.
Figure \ref{neurips:fig:size_effect_and_scaling_law} (left) presents the performance of prompt-based \ft methods, including both hard and soft variants, across four datasets.
The prompt-based \ft performance generally improves when the \pcp is implemented with more than 1000 unlabelled examples, and some enhancements can be observed even with just 100 unlabelled examples.
This indicates that continued pre-training (both \tapt and \pcp) is not necessarily computationally demanding and can be used efficiently even with only hundreds of training examples.
In our experiments, performing the \pcp on 1k unlabelled example takes less than 10 minutes using two 24GB NVIDIA 3090 GPUs, and all \pcp performance achieved in Section \cref{neurips:sec:comparsion_with_conventional_continued_pretraining} use fewer than 10k unlabelled examples.
This is a stark contrast to the previous work \citep{gu-etal-2022-ppt} that pursued similar objectives (for parameter-efficient fine-tuning) to ours but utilised 10GB of English text data.

\paragraph{\#3. Power of scale.}
Our empirical analysis investigates the impact of increasing the backbone LM size on the model performance using the \pcp. 
Figure \ref{neurips:fig:size_effect_and_scaling_law} (right) shows the results of prompt-based \ft methods, including hard and soft variants, trained using either \tapt or \pcp, on four datasets. 
The performance of the \pcp method largely improves as the backbone LM size expands, which aligns with the scaling laws observed in LMs \citep{kaplan2020scaling,hoffmann2022training}. 
Furthermore, the \pcp method consistently surpasses other baseline approaches, highlighting the advantages of the \pcp across different model sizes.

\paragraph{\#4. The impact of a larger continued pre-training corpus on the model performance using \pcp and \tapt.}
Here we expand our investigation to whether the advantage of the proposed \pcp approach persists as the size of the continued pre-training corpus increases.
Table \ref{neurips:table:larger_corpus} presents the performance of prompt-based \ft (soft), trained using either \tapt or \pcp, across five datasets with varying sizes of unlabelled training examples. 
These experimental results are consistent with our findings in Section \cref{neurips:sec:comparsion_with_conventional_continued_pretraining} and Section \cref{neurips:sec:comparsion_with_self_training}, showing that the proposed \pcp approach consistently outperforms the model performance using the \tapt even when the larger corpus for continued pre-training is used.

\begin{table*}[t]
\centering
\resizebox{\textwidth}{!}{
\begin{tabular}{lccccccccc}
        \toprule
         & \textbf SST-2 & \textbf SST-5 & \textbf MR & \textbf CR & \textbf  MPQA & \textbf Subj & \textbf TREC & \textbf CoLA & \textbf Mean \\
        \midrule
        Prompt \ft & 92.5 & 48.0 & 86.8 & 90.8 & 81.2 & 90.3 & 83.0 & 4.9 & 72.2 \\
        Prompt \ft+\pcp & \hl 93.9 & 50.7 & \hl 89.8 & 92.0 & \hl 88.3 & 94.9 & \hl 88.6 & \hl 21.5 & \hl 77.5 \\
        Prompt \ft+\pcp (Labels Only) & 93.7 & \hl 50.8 & 87.7 & 91.3 & 85.1 & 94.3 & 85.7 & -0.7 & 73.5 \\
        Prompt \ft+\pcp (Template Only) & 90.7 & 43.5 & 88.6 & \hl 92.6 & 82.0 & \hl 95.1 & 84.1 & 0.7 & 72.2 \\
        \bottomrule
\end{tabular}
}
\caption{
Ablation study on the inclusion of the template and labels in our proposed \pcp. 
The test results using soft prompt \ft and \robertalarge are reported. 
The best performance for each dataset is highlighted in blue.
}
\label{neurips:table:ablation_study_1}
\end{table*}
\paragraph{\#5. Ablation study on the label and template inclusion in \pcp.} To gain a deeper understanding of the individual contributions of pseudo labels and templates in our proposed \pcp method, we conduct an additional ablation study, where we solely utilize pseudo labels or templates. 
This ablation study is carried out using soft prompt-based fine-tuning. 
As shown in Table \ref{neurips:table:ablation_study_1}, the experimental results reveals that using either labels or templates exclusively will hurt the model's performance compared to our proposed \pcp method, highlighting the vital importance of integrating both templates and pseudo labels.

\begin{table*}[t]
\centering
\resizebox{\textwidth}{!}{
\begin{tabular}{lccccccccc}
        \toprule
         & \textbf SST-2 & \textbf SST-5 & \textbf MR & \textbf CR & \textbf  MPQA & \textbf Subj & \textbf TREC & \textbf CoLA & \textbf Mean \\
        \midrule
        \textsc{Cls}-based \ft (1k steps) + \tapt & 88.2 & 43.4 & 86.1 & 86.2 & 73.7 & 94.2 & 80.4 & 1.9 & 69.3 \\
        \textsc{Cls}-based \ft (5k steps) + \tapt & 89.6 & 43.4 & 86.7 & 87.0 & 72.9 & 94.6 & 79.0 & 1.7 & 69.4 \\
        Prompt \ft (1k steps) + \pcp & \hl 93.9 & \hl 50.7 & \hl 89.8 & \hl 92.0 & \hl 88.3 & \hl 94.9 & \hl 88.6 & \hl 21.5 & \hl 77.5 \\
        \bottomrule
\end{tabular}
}
\caption{
Ablation study on the prolonged fine-tuning, where \robertalarge is used as the backbone model.
The test Results using \textsc{Cls}-based \ft and soft prompt \ft are reported. 
The best performance for each dataset is highlighted in blue.
}
\label{neurips:table:ablation_study_2}
\end{table*}

\paragraph{\#6. The impact of prolonged fine-tuning on the model performance.}
To ascertain that the effectiveness of our proposed method is not simply due to an extended fine-tuning duration, we conduct additional experiments. 
We train \textsc{Cls}-based \ft 5 times more steps (5k steps in total) from the \tapt checkpoint. 
As shown in Table \ref{neurips:table:ablation_study_2}, our results reveal that prolonged fine-tuning only brings about a marginal improvement of only 0.1\% across the eight tasks. Notably, this still falls significantly short of our proposed method (8.1\% in absolute).

\section{Why Task-Adaptive Pre-training Does Not Perform Well on Sentence Pair Tasks?}
\label{neurips:sec:supplementary_experiments}
In this section, we investigate why \tapt does not work on sentence pair tasks. We have evaluated three possible explanations for \tapt's ineffectiveness on sentence pair tasks: 
(1) \textbf{dataset size for continued pre-training},
(2) \textbf{sentence pairs with higher similarity than what was observed in pre-training data}, and 
(3) \textbf{lack of separation within sentence pairs}. 
Our experimental results suggest that the ineffectiveness of \tapt on the sentence pair tasks is not an isolated incident but a recurring issue. 
Below we discuss each setting in detail.

\paragraph{\#1. The impact of continued pre-training (\tapt) with a larger pre-training corpus on the performance of the prompt-based \ft on sentence pair tasks.}
In Section \ref{neurips:sec:comparsion_with_conventional_continued_pretraining}, we randomly selected at most 10k unlabeled examples from the full training sets of MNLI, MNLI-mm, SNLI, QNLI, and QQP, as corpus for continued pre-training due to our limited academic computational resources. For all other tasks, we use the full training set for continued pre-training because there are fewer than 10k examples in their training sets.
To verify our findings that ``\hyperref[neurips:para:taptissue]{\color{black} \textit{\tapt is not consistently beneficial for sentence pair tasks, nor when prompt-based \ft is employed}}'' holds true when utilising larger continued pre-training corpus, we perform conventional continued pre-training (\tapt) on the full training set on MNLI, MNLI-mm, SNLI, QNLI, and QQP.

\begin{table}[!ht]
\centering
\small
% \resizebox{\textwidth}{!}{
\begin{tabular}{lrcccc}
\toprule
\textbf Dataset              & \textbf MNLI       &\textbf MNLI-mm      &\textbf  SNLI       & \textbf QNLI      & \textbf QQP \\
    Corpus Size          & 393k           & 393k            &  549k          & 104k          &   364k  \\
\midrule
\cls-based \ft           & $46.2_{0.6}$   & $48.5_{1.0}$    & $45.6_{5.4}$   & $61.4_{8.2}$  & $58.5_{3.8}$   \\
\tableindent + \tapt     & $34.7_{0.4}$\da& $35.1_{0.6}$\da & $41.8_{2.7}$\da& $54.8_{2.0}$\da& $62.6_{2.9}$\ua\\
Prompt-based \ft (hard)  & $67.3_{1.3}$   & $68.9_{1.2}$    & $76.7_{1.6}$   & $66.5_{4.3}$  & $66.8_{1.9}$ \\
\tableindent + \tapt     & $47.8_{5.6}$\da& $47.9_{5.2}$ \da& $47.5_{9.4}$\da& $53.5_{0.8}$\da& $53.5_{0.8}$\da\\
Prompt-based \ft (soft)  & $62.7_{2.2}$   & $65.9_{1.2}$    & $75.4_{0.8}$   & $64.2_{4.7}$  & $66.5_{1.8}$  \\
\tableindent + \tapt     & $45.4_{3.7}$\da& $45.8_{4.1}$ \da& $50.2_{3.9}$\da& $53.8_{0.9}$\da& $53.8_{0.9}$\da\\
\bottomrule \\
\end{tabular}
% }
\caption{Test Results using \robertalarge, with corresponding continued pre-training corpus sizes for each task.
The mean performance with standard deviations are reported across five seeds.}
\label{neurips:table:tapt_larger_corpus}
\end{table}

Table \ref{neurips:table:tapt_larger_corpus} presents the performance of the \textsc{Cls}-based \ft, prompt-based \ft (hard), and prompt-based \ft (soft) using the \tapt.  The experimental results reveal that the \tapt generally results in poorer performance, even when a larger continued pre-training corpus is used.
Notably, the performance of these fine-tuning approaches could be even worse than those achieved using a smaller continued pre-training corpus (refer to results in Table \ref{neurips:table:main_results}), suggesting that training with a larger corpus is not an effective solution to the issues of conventional continued pre-training (\tapt).

\begin{table*}[!ht]
\centering
\resizebox{\textwidth}{!}{
\begin{tabular}{lccccccccc}
        \toprule
         & \textbf MNLI & \textbf MNLI-mm & \textbf SNLI & \textbf QNLI & \textbf RTE & \textbf MRPC & \textbf QQP & \textbf STS-B & \textbf Mean \\
        \midrule
        \textsc{Cls}-based \ft & 46.2 & 48.5 & 45.6 & 61.4 & 54.2 & 73.2 & 58.5 & 46.0 & 54.2 \\
        +\tapt & 36.0 & 36.3 & 45.7 & 55.6 & 53.4 & 67.7 & 55.0 & 48.1 & 49.7 \\
        +\tapt (Tokenizer Sep) & 36.4 & 37.5 & 50.5 & 58.8 & 50.8 & 63.5 & 59.2 & 48.8 & 50.7 \\
        +\tapt (\pcp Sep) & 36.3 & 36.7 & 64.6 & 58.3 & 51.2 & 65.3 & 57.4 & 44.5 & 51.8 \\
        +\tapt (random sent pair) & 34.8 & 35.4 & 37.7 & 52.2 & 51.2 & 64.8 & 56.9 & 23.8 & 44.6 \\
        +\tapt (first sent only) & 35.6 & 35.9 & 42.7 & 52.2 & 52.6 & 62.5 & 53.6 & 16.7 & 44.0 \\
        \bottomrule
\end{tabular}
}
\caption{
Ablation study on the performance of \textsc{Cls}-based fine-tuning with different settings of conventional continued pre-training, 
where \robertalarge is used as the backbone model.
}
\label{neurips:table:why_sentence_pair_not_work}
\end{table*}

\paragraph{\#2. High similarity within sentence pairs.} We consider that the high similarity between the sentence pairs might conflict with the word distribution that the model has observed during model pre-training. For instance, in the MNLI task, two sentences are \textit{Salt kept the town fed} and \textit{Salt kept the town thriving}. 
To explore this, we perform \tapt on two different settings, one where we continually pre-train \tapt on randomly paired sentences within the dataset and another where we continually pre-train \tapt using just the first sentence of each pair. As shown in Table \ref{neurips:table:why_sentence_pair_not_work}, the experimental results show that training \tapt with either case leads to even worse performance.

\paragraph{\#3. Token-based separation of sentence pairs.} In an attempt to mitigate the effect above, we also consider that distinguishing two sentences using distinct tokens might make a difference. To test this, we perform \tapt with two types of separate tokens, the special token from the tokenizer and the template used in PCP (without labels). As shown in Table \ref{neurips:table:why_sentence_pair_not_work}, training \tapt with separate tokens between two sentences can somewhat mitigate the performance drop for \textsc{Cls}-based fine-tuning on the sentence pair tasks. However, the results remain inferior compared to \textsc{Cls}-based fine-tuning without the use of \tapt.

In conclusion, our investigations highlight the difficulties that \tapt faces on sentence pair tasks, while our proposed method PCP provides a simple yet effective solution. We hypothesize that \tapt 's ineffectiveness for \textsc{Cls}-based fine-tuning on sentence pair tasks might be due to various factors, which we leave for a more comprehensive investigation in future work.

\section{Limitations}
% \paragraph{Limitations and Broader Impact.}
We outline several limitations inherent to our research:
\begin{itemize}
    \item \textbf{The scale of language models.} Our experiments utilise relatively modestly-sized language models \cite{liu2019roberta}. The implications of scaling up to more advanced language models, such as the Llama-2 \cite{touvron2023llama} or the mixture-of-experts approach like GPT-4 \cite{openai2023gpt4}, remains an open question. 
    In the context of large language models, applying \pcp with a full set of parameter updates for a specific task may not be justifiable in terms of computational costs. 
    Future research could explore multi-task learning strategies or parameter-efficient continued pretraining.
    \item \textbf{The architecture of language models.} Our work is limited to encoder-only models \cite{devlin2018bert,liu2019roberta}. To generalize our findings, future research should investigate the effects of our method \pcp on encoder-decoder \cite{t5} and decoder-only architectures \cite{10.5555/3495724.3495883}.
    \item \textbf{The diversity of tasks.} Our evaluation is confined to text classification and regression tasks. Future research should investigate generative or multi-modal tasks, which may offer more comprehensive insights into the applicability of our methods \pcp.
\end{itemize}
In addition, our work is based on pre-training and prompting methods for LMs.
Previous works \cite{bender-koller-2020-climbing,10.5555/3495724.3495883,10.1145/3442188.3445922} have extensively discussed the risks and potential harms associated with LMs, including the amplification of undesirable biases learned from unlabelled training data \cite{bender-koller-2020-climbing,austin2021program,carlini2021extracting}.
The energy cost and carbon footprint for our work were approximately 125 kWh and 70 kg CO$_{2}$e, which are comparatively smaller than LM pre-training \cite{devlin2018bert,liu2019roberta,10.5555/3495724.3495883,chowdhery2022palm}. 

\section{Summary}
\label{neurips:sec:limitations}
% \paragraph{Conclusion.} 
In this chapter, we shed light on how \tapt performs against state-of-the-art \st approaches in various \ssl settings. Our experiments reveal that  \tapt achieves strong and robust performance
We challenge the widely accepted notion in \nlp, showing that conventional continued pre-training can be detrimental to model performance, especially for sentence pair tasks and prompt-based \ft. 
As an alternative, we propose Prompt-based Continued Pre-training (\pcp), which consistently improves the performance of state-of-the-art prompt-based \ft approaches over conventional continued pre-training.
Additionally, our proposed \pcp outperforms state-of-the-art semi-supervised approaches with a more streamlined process.
Further analysis reveals that the advantages of \pcp remain consistent across different sizes of models and datasets.
This study emphasizes the importance of presenting both task-related texts and templates/instructions to LMs during pre-training for better fine-tuning performance on downstream tasks, contributing to the growing body of research on the optimisation of pre-training and fine-tuning strategies in \nlp.

\include{Chapters/5_prompt_based_continued_pretraining}
\chapter{Decomposed Prompt Tuning for Parameter-Efficient Fine-tuning}
\label{chapter:peft}
\section{Introduction}
\label{iclr:sec:intro}

\begin{figure}[h]
\centering
\includegraphics[width=\textwidth]{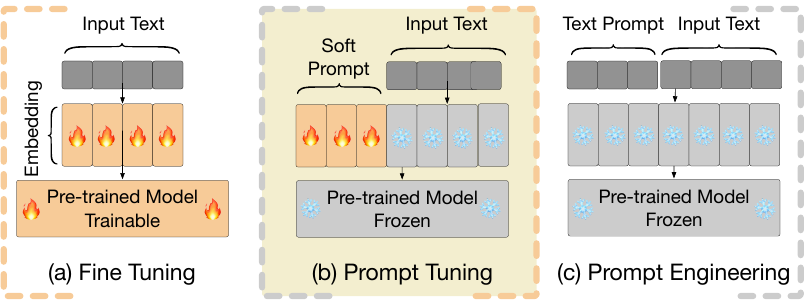}
\caption{
The overview of Fine Tuning (\ft), Prompt Tuning (\pt), and Prompting Engineering.
\pt increases the length of the input sequence, leading to much greater computational demands during train and inference phrases.
}
\label{iclr:fig:overview}
\end{figure}

Fine-tuning language models (LMs) \citep{t5,touvron2023llama} on downstream tasks offers large performance improvements across various \nlp tasks, but it requires updating and storing full parameters of the LMs  (see Figure \ref{iclr:fig:overview}a), which is especially expensive when LMs contain hundreds of millions or even billions of parameters. 
Prompt engineering \citep{10.5555/3495724.3495883} does not update any parameters (see Figure \ref{iclr:fig:overview}c) while it is typically hard to design and has a high-performance variance \citep{wang2023selfconsistency}.
Consequently, parameter-efficient fine-tuning (\peft) approaches \citep{NEURIPS2022_0cde695b} have attracted growing interest, aiming to learn only a small number of parameters per task while maintaining performance levels comparable to full fine-tuning.

Prompt Tuning (\pt) \citep{lester-etal-2021-power} has emerged as a promising \peft approach, which appends trainable continuous prompt vectors to the input (see Figure \ref{iclr:fig:overview}b).
\pt stands out from other \peft approaches as it maintains competitive performance with fewer trainable parameters and does not drastically scale up its trainable parameters as the model size grows.
Recent works suggest that the majority of the LM's knowledge is acquired during its pretraining phase \citep{zhou2023lima} and that in-context learning with just a few carefully designed stylistic examples and a carefully designed system prompt can lead to notable improvements in alignment results \citep{lin2023unlocking}. 
Considering scenarios where tasks have already been somewhat understood by LMs and the key challenge is just to properly prompt the LMs, PT emerges as a potentially better option to other \peft approaches.

\begin{figure}[!ht]
\centering
\includegraphics[width=\textwidth]{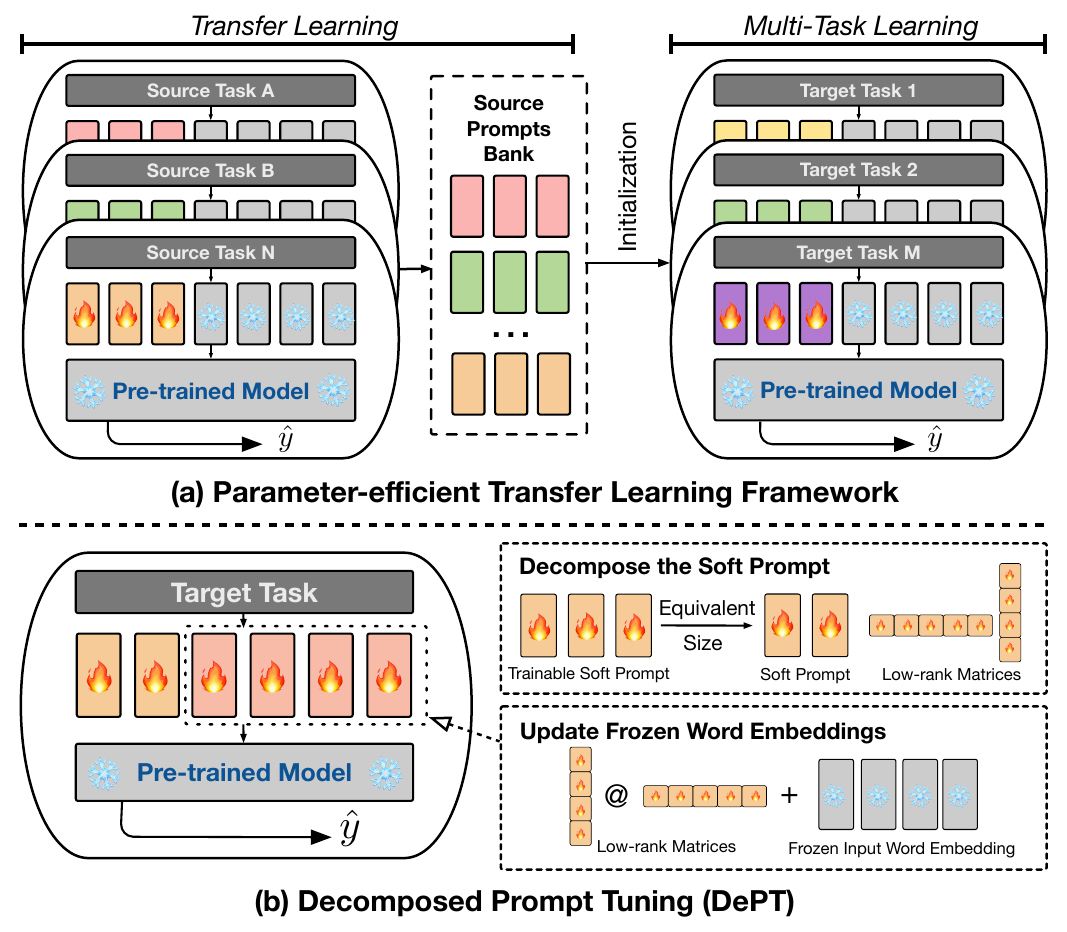}
\caption{
The overview of the \petl framework ({\em\textbf{Top}}) and our method \dept ({\em\textbf{Bottom}}).
\dept decomposes a trainable soft prompt of the vanilla \pt into a shorter soft prompt and a couple of low-rank matrices, where the multiplication of low-rank matrices serves to update frozen word embedding.
}
\label{iclr:fig:preview}
\end{figure}

While \pt has shown promising results across various tasks and models, it has two major limitations: 
(1)  \pt often suffers from slow convergence and is sensitive to the initialisation \citep{lester-etal-2021-power,vu-etal-2022-spot,wang2023multitask}; and
(2) \pt extends the total length of the input sequence, consequently exacerbating the computation demand (\ie train/inference time and memory cost), due to the quadratic complexity of the Transformer \citep{NIPS2017_3f5ee243}. This is further accentuated given the slow convergence issue of \pt.
Recent studies \citep{su-etal-2022-transferability,vu-etal-2022-spot,li-etal-2022-learning-transfer} have proposed the variants of the vanilla \pt to tackle the first issue by initially pre-training soft prompts on a variety of source tasks, which is known as \textit{Parameter-Efficient Transfer Learning} (\petl), as depicted in Figure \ref{iclr:fig:preview}a. 
Some studies \citep{,asai-etal-2022-attempt,wang2023multitask} also improve the performance of the \pt by jointly training learned prompts from these source tasks on multiple target tasks (referred to as \textit{Multi-task Learning}).
However, the issue of increased computational load due to the extension of sequence length remains largely unaddressed.
While \petl approaches can reduce the training steps for model convergence, each optimization step remains computationally expensive in terms of time and memory. 
Most importantly, it does not enhance the efficiency during the inference phase, which is particularly crucial in the era of LLMs, considering that the trained models may be queried millions of times per day.

In this chapter, we propose \textbf{De}composed \textbf{P}rompt \textbf{T}uning (\dept), which decomposes a trainable soft prompt into a shorter soft prompt and a couple of low-rank matrices, where the multiplication of low-rank matrices is then added element-wise to frozen word embeddings, as shown in Figure \ref{iclr:fig:preview}b (Section \sect{iclr:sec:ours}). 
This shorter soft prompt and the updated word embedding matrix are then optimised using two different learning rates - a crucial step for model convergence (Section \sect{iclr:para:two_speed_lr}). 
The intuition of this design is to enable representation updates within the frozen word embedding, thereby increasing the adaptability of input representations that were previously unavailable.
Experimental results on 23 natural language processing (\nlp) and vision-language (VL) tasks demonstrate \dept outperforms the state-of-the-art \peft approaches, including the full fine-tuning baseline in certain scenarios (Section \sect{iclr:sec:main_results}). 
Our study empirically shows that \dept largely improves the training efficiency across various model architectures and sizes, saving more than 20\% (using \tfbase) in both training time and memory costs compared to the vanilla \pt.
Importantly, \dept becomes increasingly efficient as the model size grows, making it particularly advantageous and suitable for LLMs (Section \sect{iclr:sec:training_efficiency}).
Furthermore, our additional analysis in the few-shot learning setting reveals the \dept's compatibility with \petl approaches  (Section \sect{iclr:sec:few_shot}).

In summary, the main contributions of this chapter are as follows:
\begin{itemize}
    \item We propose \dept method, which addresses a key efficiency limitation of Prompt Tuning by decomposing its soft prompt to reduce input sequence length. \dept largely improves the training and inference efficiency, in terms of both time and memory costs;
    \item Our comprehensive evaluation on 23 \nlp and VL tasks demonstrates that \dept outperforms state-of-the-art \peft approaches, including the full fine-tuning in some scenarios. Additionally, our experiments show that \dept smoothly integrates with \petl approaches and the advantage of \dept persists in the few-shot learning setting;
    \item We empirically show that \dept becomes increasingly efficient as the model size grows, making it particularly well-suited for LLMs. Furthermore, \dept is orthogonal to various \peft approaches (\ie Adapter, LoRA) and can be easily combined together.
\end{itemize}

\section{Decomposed Prompt Tuning}
\label{iclr:sec:ours}
In this section, we introduce our proposed method, \textbf{De}composed \textbf{P}rompt \textbf{T}uning (\dept).

\paragraph{The decomposition of the soft prompt.}
\dept differs from the vanilla \pt method in the aspect of inputs.
As shown in Figure \ref{iclr:fig:preview}b, we decompose a trainable prompt matrix $\mP\in \mathbb{R}^{l \times d}$ from the vanilla \pt into two components:
(1) a shorter trainable prompt matrix $\mP{_s} \in \mathbb{R}^{m \times d}$; and
(2) a pair of low-rank matrices, $\mA \in \mathbb{R}^{s \times r}$ and $\mB \in \mathbb{R}^{r \times d}$, where typically the rank of the matrices $r \ll \min(s, d)$.
The first component, the smaller trainable prompt matrix, is appended to the word embedding matrix in a similar manner as in the vanilla \pt.
The second component uses the multiplication of two low-rank matrices to represent the update of the word embedding through a coordinate-wise sum:
\begin{equation}
\mE_i^{'} = \mE_i + \Delta \mE_i = \mE_i + \mB\mA \in \mathbb{R}^{s \times d},
\label{iclr:eq:low_rank_mul}
\end{equation}
where $\mE_i$ is frozen and does not receive gradient updates during the training, whereas $\mA$ and $\mB$ are trainable.
Following the LoRA \citep{hu2021lora}, we use a random Gaussian initialization for $A$ and zero for $B$, so $\Delta W=BA$ is zero when the training starts.
The loss function is then optimised as follows:
\begin{align}
    \label{iclr:eq:dept_loss}
    \mathcal{L}_{\text{\dept}}= -\sum_i \log P(\bm{y}_i \, | \, [\mP{_s}, \mE_i^{'}] \,; \Theta)
\end{align}
In our experiment, we choose the values of $m$ and $r$ to satisfy the equation $l \times d = m \times d + (s + d) \times r$ for maintaining the exact size of trainable parameters as in the vanilla \pt. 
Consequently, $m$ is always less than $l$ when $r > 0$. 
This design improves memory efficiency and reduces computational expense compared to the vanilla \pt, as the shorter input sequence length (\ie $m + s < l + s$) substantially reduces computation due to the quadratic complexity of the Transformer \citep{NIPS2017_3f5ee243}. 

\paragraph{Two rates of learning.}
\dept also differs from the vanilla \pt in training.
We train the shorter trainable prompt matrix, $\mP{_s}$, with the learning rate $\alpha_{1}$ and the pair of low-rank matrices, $\mA$ and $\mB$, with the learning rate $\alpha_{2}$, rather than using a single learning rate as in the vanilla \pt. The $\alpha_{1}$ is typically much larger than the $\alpha_{2}$. We will empirically validate the importance of this choice in Section \sect{iclr:para:two_speed_lr}.
However, \dept may introduce extra training costs for the hyperparameter optimization (see Section \sect{iclr:para:limitations}).

\section{Experiments and Results}
In this section, we introduce our experimental setup (see Section \sect{iclr:sec:setup}), evaluate the performance of \dept across 23 different \nlp and VL tasks (see Section \sect{iclr:sec:main_results}), and assess relative train/inference time and memory cost of \dept (see Section \sect{iclr:sec:training_efficiency}), and explore the effectiveness of \dept in the few-shot learning setting and importance of two different learning rates for training \dept (see Section \sect{iclr:sec:analysis}).

\subsection{Experimental Setup}
\label{iclr:sec:setup}

\textbf{Datasets and tasks.} We evaluate our proposed method \dept on 21 \nlp tasks and 2 vision-language tasks.
For \nlp tasks, we follow the previous works \citep{vu-etal-2022-spot,sunglst,asai-etal-2022-attempt,wang2023multitask} and use various datasets sourced from:
(1) GLUE \citep{wang-etal-2018-glue} benchmark, including MNLI \citep{williams-etal-2018-broad}, QQP\footnote{\url{https://www.quora.com/q/quoradata/}}, QNLI \citep{rajpurkar-etal-2016-squad}, SST-2 \citep{socher-etal-2013-recursive}, STS-B \citep{cer-etal-2017-semeval}, MRPC \citep{dolan-brockett-2005-automatically}, RTE \citep{giampiccolo-etal-2007-third} and CoLA \citep{warstadt2019neural_cola};
(2) SuperGLUE benchmark \citep{wang2019superglue}, including MultiRC \citep{khashabi-etal-2018-looking}, BoolQ \citep{clark-etal-2019-boolq}, WiC \citep{pilehvar-camacho-collados-2019-wic}, WSC \citep{levesque2012winograd_wnli}, and CB \citep{de2019commitmentbank}; 
(3) MRQA 2019 Shared Task \citep{fisch-etal-2019-mrqa}, including Natural Questions \citep{kwiatkowski-etal-2019-natural}, HotpotQA \citep{yang-etal-2018-hotpotqa}, SearchQA \citep{dunn2017searchqa} and NewsQA \citep{trischler-etal-2017-newsqa};
(4) other datasets, including WinoGrande \citep{10.1145/3474381}, Yelp-2 \citep{10.5555/2969239.2969312}, SciTail \citep{Khot_Sabharwal_Clark_2018} and PAWS-Wiki \citep{zhang-etal-2019-paws}.
For vision-language tasks, we follow prior works \citep{sung2022vl,sunglst} to experiment with the visual question-answering task, VQA \citep{goyal2017making}, and the image caption generation task, MSCOCO \citep{chen2015microsoft}.

\textbf{Baselines.} We compare \dept with a variety of baselines: 
(1) fine-tuning (FT), where all the model parameters are tuned during adaptation on each downstream task;
(2) the vanilla \pt \citep{lester-etal-2021-power}, where target prompt vectors are initialized by randomly sampled top vocabularies, and its variants using additional transfer and multi-task learning, including SPoT \citep{vu-etal-2022-spot}, ATTEMPT \citep{asai-etal-2022-attempt}, and MPT \citep{wang2023multitask};
(3) state-of-the-art \peft approaches including 
Adapters \citep{houlsby2019parameter}, 
AdapterDrop \citep{ruckle-etal-2021-adapterdrop},
BitFit \citep{zaken2021bitfit}, 
HyperFomer \citep{mahabadi2021parameter}, 
HyperDecoder \citep{ivison-peters-2022-hyperdecoders},
P-tuning \citep{liu2021gpt},
LoRA \citep{hu2021lora}, 
LST \citep{sunglst}, 
and their multi-task learning variants.
For a fair comparison, we directly quote performance metrics from published papers \citep{karimi2021compacter,mahabadi2021parameter,asai-etal-2022-attempt,wang2023multitask,sunglst} for a fair comparison, where all these baselines using the \tfbase as the backbone and adhere to the train, validation and test splits used by previous works \citep{mahabadi2021parameter,karimi2021compacter} for \nlp tasks and by previous works \citep{sunglst} for vision-language tasks.

\textbf{Implementation details.}
In our study, we mainly experiment using the \tfbase model with $220$M parameters \citep{t5}.
We consistently set the number of virtual tokens $l$ as 100 across all tasks for the vanilla \pt and adjust the hyper-parameters of \dept accordingly to maintain the equivalent number of trainable parameters.
For instance, the vanilla \pt contains $l \times d$ trainable parameters where the hidden size $d$ is 768 for the \tfbase, and \dept can configure the number of virtual tokens $m$ as 40 and the rank of low matrices $r$ as 45, resulting in $m \times d + (s + d) \times r$ trainable parameters. This yields a total of $76,800$ trainable parameters, aligning with the vanilla \pt.
For VL tasks, we utilise the CLIP-T5 architecture which combines CLIP \citep{radford2021learning} and \tfbase \citep{t5}, with the CLIP frozen. 
We follow the prior work \citep{sunglst} to concatenate the visual representation from CLIP with the text embedding from the \tfbase, where a trainable visual projection layer is used between CLIP and T5 to align the visual representation to the same dimension as the text embedding.

We also extend our evaluation to include
\tfsmall ($60$M), \tflarge ($770$M),
\gptsmall ($110$M), \gptmedium ($345$M), and \gptlarge ($774$M) 
models.
In the few-shot experiments, we randomly select $k$ examples three times from the training set and report the mean and standard deviations for each $k$-shot experiment. 
Following the prior works in \petl for \pt \citep{vu-etal-2022-spot,su-etal-2022-transferability,asai-etal-2022-attempt}, we use 
MNLI,
QQP,
SST-2,
SQUAD \citep{rajpurkar-etal-2016-squad}, and
ReCoRD \citep{zhang2018record}
as five source tasks. 
Our soft prompt and low-rank matrix pairs are initialized from the soft prompts derived from one of these selected source tasks.
Please see more hyper-parameter and implementation details in Appendix \ref{iclr:sec:implementation_details}.

\begin{table}[!t]
\centering
\caption{
Test results on GLUE and SuperGLUE benchmarks, with the corresponding size of trainable parameters.
All of the results are based on \tfbase models. 
We use Pearson correlation for STS-B, F1 for MultiRC (Multi), and accuracy for other tasks as evaluation metrics.
$^1$ indicates the results sourced from \cite{asai-etal-2022-attempt}.
$^2$ indicates the results sourced from \cite{sunglst}. 
$^3$ indicates the results sourced from \cite{wang2023multitask}. 
$^4$ indicates that we reproduce and substantially increase the performance of the vanilla \pt reported in the prior work \citep{asai-etal-2022-attempt}.
$^*$ These values are obtained after averaging over 8 tasks, and the minimal number of parameters to perform a single task remains 232k and 77.6k for ATTEMPT and MPT. 
\mt represents additional multi-task training.
}
\label{iclr:table:main_results_glue}
\resizebox{1.0\textwidth}{!}{
\addtolength{\tabcolsep}{-4.25pt}  
\begin{tabular}{lrccccccccccccccc}
\toprule
\multirow{2}{*}{\textbf Method} & \multirow{2}{*}{\textbf \#Para}    &\multicolumn{9}{c}{\textbf{GLUE}}           & \multicolumn{6}{c}{\textbf{SuperGLUE}} \\ 
                                                     \cmidrule(lr){3-11}                           \cmidrule(lr){12-17}
                 &       & MNLI & QQP  & QNLI & SST-2  & STS-B & MRPC  & RTE   & CoLA  & \cg Mean &  Multi     & Bool  & WiC   & WSC  & CB  & \cg Mean \\  \midrule
\multicolumn{17}{c}{\textbf \textit{Single-Task Learning}} \\ \midrule
Fine-tuning$^1$  & 220M  & 86.8 & 91.6 & 93.0 & 94.6   & 89.7  & 90.2  & 71.9  & 61.8  & \cg 84.9 &  72.8      & 81.1  & 70.2  & 59.6 & 85.7 & \cg 73.9 \\ 
% \hdashline\noalign{\vskip 0.4ex}
Adapter$^1$      & 1.9M  & 86.5 & 90.2 & 93.2 & 93.8   & 90.7  & 85.3  & 71.9  & 64.0  & \cg 84.5 &  75.9      & 82.5  & 67.1  & 67.3 & 85.7 & \cg 75.7 \\
AdapterDrop$^1$  & 1.1M  & 86.3 & 90.2 & 93.2 & 93.6   & 91.4  & 86.3  & 71.2  & 62.7  & \cg 84.4 &  72.9      & 82.3  & 68.3  & 67.3 & 85.7 & \cg 75.3 \\
BitFit$^1$       & 280k  & 85.3 & 90.1 & 93.0 & 94.2   & 90.9  & 86.8  & 67.6  & 58.2  & \cg 83.3 &  74.5      & 79.6  & 70.0  & 59.6 & 78.6 & \cg 72.5  \\
LoRA$^2$         & 3.8M  & 86.3 & 89.0 & 93.2 & 94.3   & 90.9  & 90.1  & 75.5  & 63.3  & \cg 85.3 &  72.6     &  81.3 &  68.3 & 67.3   & 92.9  & \cg 76.5   \\
LST$^2$          & 3.8M  & 85.6 & 88.8 & 93.3 & 94.1   & 90.7  & 90.4  & 71.9  & 58.1  & \cg 84.1 &  --        & --    & --    & --   & --   & \cg -- \\
PT$^4$           & 76.8k & 83.4 & 90.2 & 93.1 & 91.9   & 90.2  & 90.1  & 78.8  & 60.7  & \cg 84.8 &  65.7      & 63.7  & 50.8  & 51.9 & 67.9 & \cg 60.0  \\
%
%                          MNLI   QQP    QNLI   SST-2    STS-B   MRPC    RTE     CoLA    Avg.        MultiRC     Boolq    WiC   WSC     CB      Avg. 
\dept (ours)     & 76.8k & 85.0 & 90.4 & 93.2 & 94.2   & 90.8  & 90.7  & 79.1  & 63.8  & \cg 85.9 &  74.3      & 79.3  & 68.7 & 67.3  & 92.9 & \cg 76.5 \\
\midrule
\multicolumn{17}{c}{\textbf \textit{Multi-task Learning}} \\
\midrule
Fine-tuning\mt$^1$  & 28M      & 85.7 & 91.1 & 92.0 & 92.5 & 88.8 & 90.2 & 75.4 & 54.9 & \cg 83.8 &  74.4     & 81.1      & 70.0  & 71.2   &  85.7 & \cg 76.1  \\
% \hdashline\noalign{\vskip 0.4ex}
Adapter\mt$^1$      & 1.8M     & 86.3 & 90.5 & 93.2 & 93.0 & 89.9 & 90.2 & 70.3 & 61.5 & \cg 84.4 &  72.6    & 82.3      &  66.5 & 67.3   &  89.3  &  \cg 75.6  \\
HyperFormer\mt$^1$  & 638k     & 85.7 & 90.0 & 93.0 & 94.0 & 89.7 & 87.2 & 75.4 & 63.7 & \cg 84.8 & 72.9     & 82.5      &  69.0 & 67.3   & 85.7  & \cg 75.4 \\
HyperDecoder\mt$^1$ & 1.8M     & 86.0 & 90.5 & 93.4 & 94.0 & 90.5 & 87.7 & 71.7 & 55.9 & \cg 83.7 & 70.4     &  78.8     & 67.1  & 61.5   & 82.1  & \cg 72.0 \\
\midrule
\multicolumn{17}{c}{\textbf \textit{Single-Task Training + Transfer Learning}} \\
\midrule
SPoT$^1$         & 76.8k  & 85.4 & 90.1 & 93.0 & 93.4 & 90.0 & 79.7  & 69.8  & 57.1 & \cg 82.3    &  74.0      & 77.2  & 67.0  & 50.0 & 46.4 & \cg 62.9 \\
ATTEMPT$^1$      & 232k   & 84.3 & 90.3 & 93.0 & 93.2 & 89.7 & 85.7  & 73.4  & 57.4 & \cg 83.4    &  74.4      & 78.8  & 66.8 & 53.8  & 78.6 & \cg 70.5 \\ 
MPT$^3$          & 77.6k  & 85.9 & 90.3 & 93.1 & 93.8 & 90.4 & 89.1  & 79.4  & 62.4 & \cg 85.6    &  74.8      & 79.6  & 69.0 & 67.3  & 79.8 & \cg 74.1 \\
\midrule
\multicolumn{17}{c}{\textbf \textit{Multi-task Learning + Transfer Learning}} \\
\midrule
%                           MNLI   QQP    QNLI   SST-2  STS-B  MRPC    RTE    CoLA    Avg.    MultiRC   Boolq    WiC     WSC    CB      Avg.
ATTEMPT\mt$^3$     & 96k$^*$  & 83.8 & 90.0 & 93.1 & 93.7 & 90.8 & 86.1  & 79.9 & 64.3 &  \cg 85.2  & 74.4    & 78.5  & 66.5 &   69.2 & 82.1 & \cg 74.1 \\ 
MPT\mt$^3$         & 10.5k$^*$& 84.3 & 90.0 & 93.0 & 93.3 & 90.4 & 89.2  & 82.7 & 63.5 &  \cg 85.8  & 74.8    & 79.2  & 70.2 &   67.3 & 89.3 & \cg 76.1 \\
\bottomrule
\end{tabular}
}
\end{table}

\subsection{Main Results}
\label{iclr:sec:main_results}
This section shows the empirical evidence supporting the effectiveness of our proposed method \dept across 23 \nlp and VL tasks. 
Table \ref{iclr:table:main_results_glue}, \ref{iclr:table:mrqa}, and \ref{iclr:table:vlt} present our experimental results on GLUE and SuperGLUE benchmarks, MRQA 2019 Shared Task and four other \nlp datasets, as well as two VL tasks. 
Additionally, we visualise the model performance against the number of trainable parameters for GLUE and SuperGLUE in Figure \ref{iclr:fig:performance_vs_efficient} of Appendix \ref{iclr:sec:sub_figure}.
% Furthermore, we evaluate the performance of \dept using \llama \citep{touvron2023llama} in Appendix \ref{iclr:sec:llama}.
Experimental results reveal three key findings: 
(1) \dept consistently outperforms the vanilla \pt and its \petl variants;
(2) \dept achieves competitive or even better performance than state-of-the-art \peft approaches while using fewer trainable parameters; and
(3) \dept falls short in some tasks.
Below we delve deeper with respect to various tasks.

\paragraph{\#1. Performance on GLUE and SuperGLUE benchmarks.}
As shown in Table \ref{iclr:table:main_results_glue}, our experimental result indicates that \dept outperforms state-of-the-art \peft approaches, such as Adapter, LoRA and LST on the GLUE and SuperGLUE benchmarks, while using fewer trainable parameters.
Remarkably, \dept also outperforms the full fine-tuning baseline on both benchmarks.
In addition, \dept outperforms vanilla \pt and all the variants of \pt that introduce additional transfer learning and multi-task learning.
For example, ATTEMPT, which requires additional training for the soft prompt on the source tasks, achieves an average score of 83.4 on the GLUE benchmark and 70.5 on the SuperGLUE benchmark.
Meanwhile, \dept outperforms ATTEMPT with scores of 85.9 and 76.5 on GLUE and SuperGLUE, despite training fewer parameters. 
Similarly, \dept surpasses MPT with 0.1\% on the GLUE benchmark and 0.4\% on the SuperGLUE benchmark, without utilizing additional transfer learning or multi-task learning.
These results are achieved with less inference time and reduced memory resources (refer to Section \sect{iclr:sec:training_efficiency} for specifics), which validates the effectiveness of \dept.
% It is worth noting that \dept is orthogonal to these \petl approaches.
As the \pt often underperforms in scenarios with limited labelled data \citep{gu-etal-2022-ppt}, we investigate the compatibility of \dept and \petl later in the few-shot learning setting (Section \sect{iclr:sec:few_shot}).

% \newcolumntype{a}{>{\columncolor{gray}}c}

\begin{table}[!t]
\centering
\caption{
Test results on MRQA 2019 Shared Task and other datasets using the \tfbase model.
We report the $F_1$ for MRQA tasks and accuracy for other datasets across three seeds, with standard deviations
in subscripts.
All baseline results are directly quoted from the previous work \cite{wang2023multitask}.}
\label{iclr:table:mrqa}
% \addtolength{\tabcolsep}{-1pt}
\resizebox{\textwidth}{!}{
\begin{tabular}{lccccccccccc}
\toprule 
\multirow{2}{*}{\textbf Method} & \multirow{2}{*}{\textbf \#Para} & \multicolumn{5}{c}{\textbf MRQA}         & \multicolumn{5}{c}{\textbf Others}  \\ 
\cmidrule(lr){3-7} \cmidrule(lr){8-12} 
                       &        & NQ   & HP   & SQA  & News & Mean    & WG   & Yelp & SciTail & PAWS & Mean \\ 
\midrule
Fine Tuning            & 220M   & 75.1 & 77.5 & 81.1 & 65.2 & 74.7    & 61.9 & 96.7 & 95.8    & 94.1 & 87.1 \\ 
% \hdashline\noalign{\vskip 0.4ex}
Adapters               & 1.9M   & 74.2 & 77.6 & 81.4 & 65.6 & 74.7    & 59.2 & 96.9 & 94.5    & 94.3 & 86.2 \\
BitFit                 & 280K   & 70.7 & 75.5 & 77.7 & 64.1 & 72.0    & 57.2 & 94.7 & 94.7    & 92.0 & 84.7 \\
LoRA                   & 3.8M   & 72.4 & 62.3 & 72.5 & 56.9 & 66.0    & 58.2 & 97.1 & 94.7    & 94.0 & 86.0 \\
PT                     & 76.8K  & 67.9 & 72.9 & 75.7 & 61.1 & 69.4    & 49.6 & 95.1 & 87.9    & 55.8 & 72.1 \\
SPoT                   & 76.8K  & 68.2 & 74.8 & 75.3 & 58.2 & 69.1    & 50.4 & 95.4 & 91.2    & 91.1 & 82.0 \\
ATTEMPT                & 232K   & 70.4 & 75.2 & 77.3 & 62.8 & 71.4    & 57.6 & 96.7 & 93.1    & 92.1 & 84.9 \\
MPT                    & 77.6K  & $72.0_{0.1}$ & $75.8_{0.1}$ & $77.2_{0.1}$ & $63.7_{0.1}$ & $72.2$    & $56.5_{0.9}$ & $96.4_{0.0}$ & $95.5_{0.1}$  & $93.5_{0.1}$ & $85.5$ \\
% \addlinespace[3pt]
\dept   (ours)         & 76.8K  & 73.2$_{0.1}$ & 76.8$_{0.3}$ & 77.6$_{0.2}$ & 64.4$_{0.1}$ & 73.0 & 59.0$_{0.2}$ & 96.8$_{0.1}$ & 95.6$_{0.2}$  & 93.7$_{0.1}$ & 86.3 \\
% lr                              3e-1   3e-1   3e-1   4e-1             4e-1   4e-1   4e-1      5e-1
% LoRA-lr                         1e-4   1e-4   1e-4   1e-4             5e-3   1e-4   1e-4      1e-4
% pt, r                           60,24  60,24  60,24  60,24            60,30  60,30  60,30     60,30
% steps                           300k   200k   100k   50k              200k   100k   50k       100k
% Note                            NQ may need lr<2e-1 to converge.
\bottomrule
\end{tabular}
}
\end{table}

\paragraph{\#2. Performance on MRQA 2019 Shared Task and other \nlp datasets.}
Table \ref{iclr:table:mrqa} presents the performance of various \peft approaches, including \dept, on the MRQA 2019 Shared Task and four other datasets.
We observe that \dept improves the average performance of the vanilla \pt by a substantial margin of +3.6\% on MRQA and +14.2\% on the other datasets. 
\dept exceeds the performance of the \pt variants that leverage additional transfer and multi-task learning, without introducing extra trainable parameters to the vanilla \pt or relying on any \petl approaches.
While \dept improves over the vanilla \pt and its variants are promising, there remains a disparity in performance when compared to the full fine-tuning baseline.
Investigating ways to incorporate \dept with other \peft methods, such as LoRA and Adapter, may provide a valuable direction for future research towards narrowing this performance gap.

\begin{table}[!ht]
\centering
\caption{
Test results on the VQA and MSCOCO dataset using \tfbase model.
We report average results across three seeds, with standard deviations in subscripts.
All baseline results are directly quoted from the previous work \cite{sunglst}.
The best performance for each column is highlighted in blue.
}
\label{iclr:table:vlt}
\resizebox{0.65\textwidth}{!}{
\begin{tabular}{lccc}
\toprule
\multirow{3}{*}{Method}   & Updated & VQA &   MSCOCO  \\ 
                          & Params & Karpathy test  & Karpathy test \\
                          & (\%) & Acc. (\%)  & CIDEr \\
\midrule
FT                        & 100   & 67.1$_{0.1}$ & 112.2$_{0.3}$ \\
\hdashline\noalign{\vskip 0.4ex}
Adapters                  & 7.98  & \hll 67.1$_{0.1}$ & 111.8$_{0.1}$ \\
LoRA                      & 7.54  & 63.7$_{0.2}$ & 110.3$_{0.4}$ \\
BitFit                    & 0.83  & 55.1$_{0.2}$ & 101.2$_{0.2}$ \\
P-Tuning                  & 1.26  & 47.4$_{0.7}$ & 96.1$_{0.9}$  \\
LST                       & 7.46  & 66.5$_{0.1}$ & 113.5$_{0.3}$ \\
\dept (ours)              & 0.74  & 59.8$_{0.4}$ & \hll 113.7$_{0.3}$         \\
\bottomrule
\end{tabular}
}
\end{table}

\paragraph{\#3. Performance on Vision-Language tasks.}
Table \ref{iclr:table:vlt} provides an overview of the performance of various \peft approaches on two VL tasks, specifically VQA and MS COCO Caption Generation. 
Results show that \dept, while updating much fewer parameters, achieves a CIDEr score of 113.7 on the MS COCO Caption Generation task, outperforming state-of-the-art \peft approaches. 
This suggests the effectiveness of our proposed method.
However, while \dept outperforms methods such as P-tuning and BitFit on the VQA dataset, it still falls short of the full fine-tuning performance. 
This suggests that in some tasks, the use of a greater number of trainable parameters could be beneficial. 

\begin{figure*}[t!]
\centering
\includegraphics[width=\textwidth]{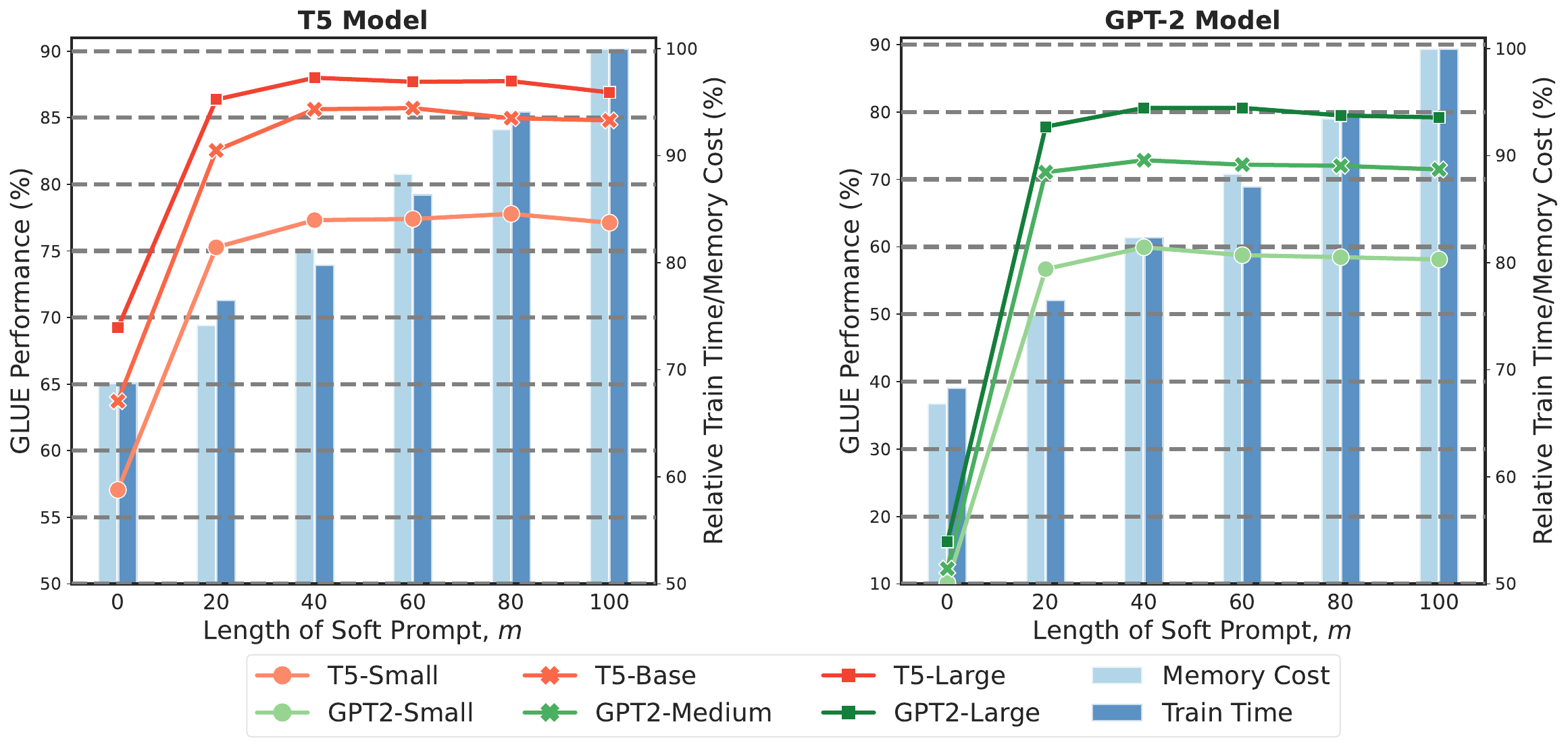}
\caption{
Performance on the GLUE benchmark for different soft prompt lengths $m$ in \dept, associated with corresponding relative train time and memory cost.
The time and memory are averaged over different model sizes using batch size as 16.
\dept consistently uses the same number of trainable parameters as the vanilla \pt ($m$=100).
}
\label{iclr:fig:ablation_study_m}
\end{figure*}

\subsection{Time and Memory Efficiency}
\label{iclr:sec:training_efficiency}
\begin{figure*}[!t]
\includegraphics[width=\textwidth]{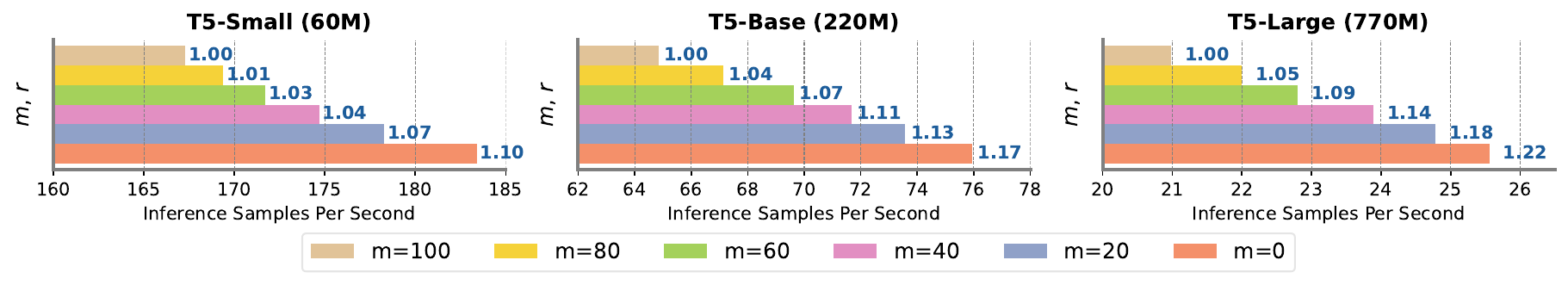}
\caption{
Average inference speed on GLUE benchmark using varying soft prompt length $m$ and the rank of low-rank matrices $r$, keeping the total number of trainable parameters constant.
Small texts in {\color{tableblue} blue} indicate the speed relative to the vanilla \pt (represented by {\color{tablebrown} brown}) ($m$=100).
}
\label{iclr:fig:inference_speed}
\end{figure*}

This section shows the empirical evidence supporting the efficiency of \dept, spanning over diverse model architectures of varying scales on the GLUE benchmark.
To ensure a fair comparison, we consistently keep the number of trainable parameters in \dept the same as that in the vanilla \pt ($l=100$). 
As a result, once we choose the length of the soft prompt $m$ in \dept, the rank of the low-rank matrices $r$ becomes determined.
% Consequently, in \dept, the rank of low-rank matrices $r$ is deterministic after the length of the soft prompt $m$ is selected.
In our experiments, we primarily compare \dept with the vanilla \pt using 5 different lengths of soft prompt  $m$ (\ie 0, 20, 40, 60, 80).
Figure \ref{iclr:fig:ablation_study_m} and \ref{iclr:fig:inference_speed} depict the average GLUE performance of \dept, along with the associated training/inference time and memory cost compared to the vanilla \pt.
Below we discuss two key findings.

\paragraph{\#1. \dept improves time and memory efficiency substantially.}
Figure \ref{iclr:fig:ablation_study_m} presents the mean performance of \dept, associated with average training time and memory, on the GLUE benchmarks, against different lengths of soft prompt $m$. 
The average training time and memory costs are computed across 8 tasks on the GLUE benchmark and three different model sizes.
Both the encoder-decoder (T5) and decoder-only (GPT-2) models are evaluated across three different model sizes.
The study reveals that decomposing the soft prompt ($l=100$) into a small soft prompt and low-rank matrices delivers comparable or even better performance while substantially enhancing the efficiency of training and reducing memory utilization. 
Specifically, using a soft prompt length greater than 20 in \dept with the T5 model leads to a better average performance on the GLUE benchmark to vanilla \pt, while improving the efficiency of training and reducing memory utilization by approximately 25\%.
This improvement is more pronounced (37\% on the SST-2 dataset) when we test \dept (with $m=60$) using the T5-3B model (see Section \sect{iclr:para:t5-3b} for details).
Similar observations are also found when the GPT model is used, suggesting the adaptability of \dept for different model architectures.
It is worth noting that \dept may have a notable performance drop regardless of using T5 or GPT-2, when the soft prompt is eliminated ($m=0$) and the model solely depends on the pair of low-rank matrices.

% \definecolor{tableblue}{HTML}{1D5D9B}
% \definecolor{tablebrown}{HTML}{e8c49c}

\paragraph{\#2. \dept grows more efficient as the model size increases.}
Figure \ref{iclr:fig:inference_speed} represents the inference speed, measured by the average number of samples evaluated per second on the GLUE benchmark using a single RTX 3090 GPU. 
The inference time is computed using the Huggingface Trainer Class.
We observe that the relative improvement in the number of inference samples per second over vanilla \pt grows as the model size increases.
For example, when using the \tfsmall model, the vanilla \pt evaluates 167.3 samples per second, while \dept ($m=20$) evaluates 178.3 samples per second, resulting in a 6.5\% boost in inference speed.
In contrast, when the \tflarge is utilized, the vanilla \pt evaluates 21.0 samples per second and \dept ($m=20$) evaluates 24.8 samples per second, resulting in an 18.1\% increase in inference speed, a substantial rise from the previous 6.5\%.
This indicates that \dept is particularly beneficial and more applicable in the context of LLMs.

We further extend our experiments regarding inference speed of \dept and PT using T5-3B and \llama.

\paragraph{\llama.}
We evaluate the performance and inference speed of our proposed method \dept using \llamas and \llamam \citep{touvron2023llama} on the SST-2 dataset. 
In our experiment, the soft prompt length for the vanilla \pt is set to $l = 100$. For \dept, we set the soft prompt length to $m = 60$ and select a rank of $r = 40$ for the low-rank matrices. 
As shown in Table \ref{iclr:table:llama}, our experimental results suggest that \dept not only outperforms the vanilla \pt in terms of test accuracy but also improves the speed of inference.
We only limit our evaluation of \dept to the SST-2 dataset due to the high computational expenses. 
We will do our best to get the necessary resources to further probe the performance of \dept, aiming to deliver a more exhaustive evaluation in future work.

\begin{table}[!ht]
\centering
% \footnotesize
% \small
\resizebox{\columnwidth}{!}{
\begin{tabular}{lcccc}
\toprule
   & \multicolumn{2}{c}{\textbf Prompt Tuning} & \multicolumn{2}{c}{\textbf \dept (ours)} 
\cr                         \cmidrule(lr){2-3}                   \cmidrule(lr){4-5}
\textbf Method                & Test Acc   & Inference samples per second & Test Acc & Inference samples per second \\   
\midrule
\llamas        &  94.48    & 3.895  & 94.95  & 4.857 \\  
\llamam             &  95.99     & 2.083   & 96.01 & 2.835 \\   
\bottomrule
\end{tabular}
}
\caption{
Test results using \llamas and \llamam on the SST-2 dataset.
}
\label{iclr:table:llama}
\end{table}

\paragraph{T5-3B.}
\label{iclr:para:t5-3b}
We evaluate the performance and inference speed of our proposed method \dept using the T5-3B model.
We report the average performance on the Glue dataset as well as inference speed, measured in inference samples per second. 
As shown in Table \ref{iclr:table:T5-3B}, our findings indicate that DePT (m=60, r=30) outperforms \pt in terms of inference speed by 37\%. This suggests the advantage of DePT increases as the model size increases.

\begin{table}[!ht]
\centering
% \footnotesize
% \small
\resizebox{\columnwidth}{!}{
\begin{tabular}{lcc}
\toprule
\textbf Method                & \textbf Average Glue Performance   & \textbf Inference samples per second \\   
\midrule
\dept (m=60, r=30)        &  86.4 &	 8.9 \\  
PT (m=100)            &  85.6 &	 6.5 \\
\bottomrule
\end{tabular}
}
\caption{
Test results using T5-3B on the Glue Benchmark.
}
\label{iclr:table:T5-3B}
\end{table}

\subsection{Further Analysis}
\label{iclr:sec:analysis}

\paragraph{Few-shot Learning.}
\label{iclr:sec:few_shot}
The vanilla \pt often underperforms in the few-shot learning tasks \citep{gu-etal-2022-ppt} due to the first limitation discussed in Section \sect{iclr:sec:intro}.
To evaluate the performance of \dept in the few-shot setting, we employ the transfer learning method inspired by the recent \petl studies, as illustrated in Figure \ref{iclr:fig:preview}a.
Specifically, we pre-train both the soft prompt and the low-rank pair on source tasks and select the best checkpoint before proceeding with the target task.
Following prior works \citep{mahabadi2021parameter,asai-etal-2022-attempt,wang2023multitask}, we evaluate the effectiveness of \dept across 14 \nlp tasks, with $k$ training examples where k = 4, 16, 32.
Our experimental findings reveal two key observations as follows:
(1) \dept integrates seamlessly with \petl approaches; and
(2) \dept attains competitive or even better performance than state-of-the-art \peft approaches in the few-shot learning setting.

\begin{table}[!t]
\centering
\caption{
Few-shot learning results with $k$ = \{4, 16, 32\} on the SuperGLUE BooQ, SuperGLUE CB and SciTail datasets. 
We report average results across three seeds, with standard deviations in subscripts.
Baseline results are directly quoted from the previous work \cite{wang2023multitask}.
The best performance for each row is highlighted in blue.
}
\label{iclr:table:few_Shot}
\resizebox{\textwidth}{!}{
\begin{tabular}{lcccccccccc}
\toprule  
\multirow{2}{*}{\textbf{Task}}&\textbf{$k$-shot} &\textbf{FT} & \textbf{AD} &\textbf{\pt}&\textbf{ST} &\textbf{HF} &\textbf{(IA)$^3$} & \textbf{ATP} &\textbf{MPT}   &   \textbf{\dept} \\
                         &\textbf{\#Para}   & 220M  & 1.9M  & 76.8K & 76.8K & 638K  &   55.3K     & 232K   & 77.6K   & 76.8K \\ \midrule
% \multirow{3}{*}{\rotbf{BoolQ}}
\multirow{3}{*}{BoolQ}
                         & 4        & 50.5  & 53.4  & 61.6  & 50.5  & 48.0 &  56.7 & 61.8   & 62.2 & \hl $62.7_{5.4}$ \\      % 
                         & 16       & 56.5  & 51.4  & 61.9  & 50.6  & 50.2 &  62.0 & 60.0   & 63.3 & \hl $66.9_{4.4}$ \\          % 
                         & 32       & 58.4  & 54.5  & 61.7  & 61.2  & 58.3 &  67.2 & 65.3   & \hl 68.9 &  $67.2_{3.4}$ \\ \midrule %  
% \multirow{3}{*}{\rotbf{CB}}
\multirow{3}{*}{CB}
                         & 4        & 57.7  & 51.1  & 53.5  & 71.4  & 60.7 &  65.5 & 67.9   & 73.6 & \hl $75.0_{5.1}$ \\  
                         & 16       & 77.0  & 74.8  & 63.5  & 64.3  & 76.3 &  71.4 & 71.4   & 78.6 & \hl $78.6_{4.3}$ \\  
                         & 32       & 80.0  & 74.8  & 67.8  & 64.3  & 81.4 &  75.0 & 78.5   & 82.1 & \hl $82.1_{2.3}$ \\ \midrule  % 
% \multirow{3}{*}{\rotbf{SciTail}}
\multirow{3}{*}{SciTail}
                         & 4        & 79.6  & 79.5  & 57.7  & 69.6  & 82.0 &  65.4 & 80.2   & \hl 80.2 &  $78.1_{2.5}$ \\     % 
                         & 16       & 80.0  & 83.2  & 60.8  & 71.9  & 86.5 &  74.4 & 79.5   & \hl 87.3 &  $78.5_{1.4}$ \\      % 
                         & 32       & 81.9  & 85.0  & 60.2  & 71.9  & 85.8 &  80.4 & 80.2   & \hl 86.3 &  $85.4_{3.1}$ \\     % 
\bottomrule
\end{tabular}
}
\end{table}
Table \ref{iclr:table:few_Shot} compares the effectiveness of our proposed method \dept with various \peft approaches in few-shot experiments,
including full fine-tuning (FT), Adapters (AD), vanilla \pt (PT), SPoT (ST), HyperFormer (HF), (IA)$^3$, ATTEMPT (ATP), and MPT on BoolQ, CB, and SciTail datasets.
Table \ref{iclr:table:glue} and \ref{iclr:table:superglue} presents the performance of \dept against the vanilla \pt and MPT on the GLUE and SuperGLUE benchmark.
Experimental results show that vanilla \pt struggles with few-shot tasks, indicating the importance of \petl for the \pt in few-shot learning tasks as suggested in previous works \citep{vu-etal-2022-spot,su-etal-2022-transferability}. 
Nevertheless, the performance of \dept largely benefits from the \petl framework (see Figure \ref{iclr:fig:preview}a). 
For example, while the vanilla \pt obtains an accuracy of 53.5\% on SuperGLUE CB dataset and 57.7\% on the SciTail dataset when $k$=4, \dept with \petl achieves an accuracy of 75.0\% on SuperGLUE CB dataset and 78.1\% on the SciTail dataset, for the same $k$ value.
This result supports our first observation about the compatibility of \dept and \petl approaches. 
Furthermore, \dept with transfer learning achieves comparable performance with the variant of the \pt, MPT across 14 \nlp tasks.
Notably, \dept surpasses the performance of all other variants of the \pt (\ie SPoT, ATTEMPT) and other \peft approaches, demonstrating our method's efficacy and endorsing our second observation.

\begin{table}[!t]
\caption{Few-shot learning results with $k$ = \{4, 16, 32\} on GLUE benchmark.
We report average results across three seeds, with standard deviations in subscripts.
Baseline results are directly quoted from the previous work \cite{wang2023multitask}.} 
\label{iclr:table:glue} 
\centering
\resizebox{\textwidth}{!}{
\begin{tabular}{llccccccccc}
\toprule
\multirow{2}{*}{\textbf{$k$-shot}} & \multirow{2}{*}{\textbf{Method}} & \multicolumn{9}{c}{\textbf{GLUE}} \\ 
\cmidrule(lr){3-11}
&       & \textbf{MNLI} & \textbf{QQP}  & \textbf{QNLI} & \textbf{SST-2} & \textbf{STS-B} & \textbf{MRPC} & \textbf{RTE}  & \textbf{CoLA} & \textbf{Avg.} \\ 
\midrule
\multirow{3}{*}{4}
& PT    & 40.1         & 63.2        & 40.4         & 53.0          & 88.8          & 68.1         & 56.3         & 27.4         & 54.7         \\
& MPT   & 59.4         & 82.0        & 86.2         & 56.5          & 89.1          & 68.1         & 62.6         & 34.8         & 67.3         \\ 
& \dept & 44.0$_{1.1}$ & 77.4$_{6.7}$& 85.8$_{4.4}$ & 59.3$_{3.1}$  & 84.1$_{2.7}$  & 73.5$_{2.8}$ & 63.5$_{2.8}$ & 29.3$_{2.3}$ & 64.6         \\   
\midrule
\multirow{3}{*}{16}   
& PT    & 41.5         & 62.3        & 59.9         & 50.9          & 87.8          & 68.1         & 54.7         & 28.5         & 56.7         \\
& MPT   & 61.6         & 84.7        & 90.6         & 63.2          & 89.1          & 70.1         & 64.8         & 32.1         & 69.5         \\
& \dept & 61.8$_{2.5}$ & 80.3$_{1.3}$& 91.2$_{0.5}$ & 77.6$_{6.3}$  & 87.1$_{1.7}$  & 78.1$_{2.3}$ & 71.9$_{1.0}$ & 27.1$_{1.7}$ & 71.9         \\   
\midrule
\multirow{3}{*}{32}
& PT    & 37.0         & 62.3        & 56.7         & 50.9          & 87.5          & 68.1         & 54.7         & 23.2         & 55.1         \\
& MPT   & 63.6         & 88.5        & 91.0         & 75.9          & 89.7          & 74.5         & 59.7         & 30.8         & 71.7         \\ 
& \dept & 63.3$_{3.5}$ & 80.1$_{0.7}$& 91.3$_{0.5}$ & 80.4$_{8.7}$  & 89.2$_{0.1}$  & 81.4$_{3.3}$ & 72.7$_{2.9}$ & 28.6$_{2.1}$ & 73.4         \\   
\bottomrule
\end{tabular}
}
\end{table}

\begin{table}[!t]
\caption{Few-shot learning results with $k$ = \{4, 16, 32\} on SuperGLUE benchmark.
We report average results across three seeds, with standard deviations in subscripts.
Baseline results are directly quoted from the previous work \cite{wang2023multitask}.} 
\label{iclr:table:superglue} 
\centering
\resizebox{0.75\textwidth}{!}{
\begin{tabular}{llcccccc}
\toprule
\multirow{2}{*}{\textbf{$k$-shot}} & \multirow{2}{*}{\textbf{Method}} & \multicolumn{6}{c}{\textbf{SuperGLUE}} \\ 
\cmidrule(lr){3-8}
&       & \textbf{Multi} & \textbf{BoolQ} & \textbf{WiC}  & \textbf{WSC}  & \textbf{CB}   & \textbf{Avg.} \\ 
\midrule
\multirow{3}{*}{4}
& PT    & 61.8          & 61.6          & 51.2         & 60.4         & 53.5         & 57.7         \\
& MPT   & 62.2          & 62.2          & 52.9         & 67.3         & 73.6         & 63.6         \\ 
& \dept & 62.3$_{1.3}$  & 62.7$_{5.4}$  & 57.5$_{1.1}$ & 67.9$_{0.9}$ & 75.0$_{5.1}$ & 65.1         \\   
\midrule
\multirow{3}{*}{16}   
& PT    & 60.3          & 61.9          & 48.9         & 44.2         & 63.5         & 55.8         \\
& MPT   & 64.5          & 63.3          & 49.8         & 67.3         & 78.6         & 64.7         \\
& \dept & 60.6$_{2.8}$  & 66.9$_{4.4}$  & 59.6$_{0.7}$ & 57.7$_{2.7}$ & 78.6$_{4.3}$ & 64.7         \\   
\midrule
\multirow{3}{*}{32}
& PT    & 59.2          & 61.7          & 52.6         & 67.3         & 67.8         & 61.7         \\
& MPT   & 63.3          & 68.9          & 53.9         & 67.3         & 82.1         & 67.1         \\ 
& \dept & 60.1$_{2.7}$  & 67.2$_{3.4}$  & 58.0$_{0.7}$ & 63.1$_{3.6}$ & 82.1$_{2.3}$ & 66.4         \\   
\bottomrule
\end{tabular}
}
\end{table}
% Table \ref{iclr:table:few_Shot} and \ref{iclr:table:few_glues} present the performance of \dept against various \peft approaches in the few-shot learning setting. Our experiments follow the setting in the prior work \citep{wang2023multitask}.

\begin{figure}[!ht]
\centering
\includegraphics[width=0.65\textwidth]{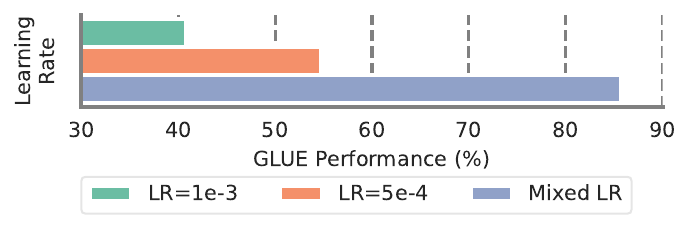}
\caption{
Test results on GLUE benchmark using \tfbase,
showing the importance of training \dept with different learning rates.
}
\label{iclr:fig:lr_ablation_study}
\end{figure}

\paragraph{The importance of different learning rates.}
\label{iclr:para:two_speed_lr}
Figure \ref{iclr:fig:lr_ablation_study} presents the experimental results from 3 different learning rate settings to train the soft prompt and the pair of low-rank matrices as follows: 
(1) use a singular learning rate of 3e-1;
(2) use a singular learning rate of 5e-4;
(3) apply mixed learning rates (with grid search), where the soft prompt is trained with a larger rate and the pair of low-rank matrices is trained with a lower rate.
In our experiments, the first option obtains an average performance of 40.8 on the GLUE benchmark.
The second option exhibits an average performance of 54.7, while the third option demonstrates a largely improved average performance of 85.7 on the GLUE benchmark.
This indicates the importance of training \dept with two different learning rates.
% \section{Related Works}

% Strikingly, as model capacity increases, PROMPTTUNING becomes competitive with MODELTUNING, which fine-tunes the entire model on each downstream task. Nevertheless, at smaller model sizes (below 11B parameters), there are still large gaps between PROMPTTUNING and MODELTUNING

% \paragraph{Transfer Learning for \pt.}

\section{Limitations}
\label{iclr:para:limitations}
We outline several limitations in our work:
(1) the main limitation of \dept is the introduction of extra hyperparameters for tuning, \eg the learning rate of the low-rank matrices. 
This might introduce some additional computational overhead during the hyperparameter optimization phase of model training. 
In our work, we train \dept up to 300k steps (in a data-rich setting) following the previous work \citep{vu-etal-2022-spot} with a careful search for optimal learning rates, which may increase training costs.
However, the number of training steps might be efficiently reduced by \petl, which we plan to investigate in future work.
In addition, it is important to note that the model training process is a one-time event, while model inference is not. In this context, the efficiency benefits of \dept become especially valuable;
(2) the number of trainable parameters in \dept depends on the maximum sequence length $s$. In this work, we have limited our evaluation to tasks with hundreds of input tokens. Future work could explore the performance of \dept when $s$ is extremely large; and 
(3) our research focuses on leveraging prompting techniques for LMs, where previous studies \citep{bender-koller-2020-climbing,10.5555/3495724.3495883,10.1145/3442188.3445922} have already addressed concerns and potential hazards linked to LMs. 
% Future endeavours will explore \dept within the framework of LLMs \citep{zhang2022opt,touvron2023llama} and investigate its synergy with other \peft approaches \citep{NEURIPS2022_0cde695b}.

\section{Summary} 
% In this work, we propose \dept, which substantially improves the efficiency of the vanilla \pt in terms of time and memory while delivering competitive or even superior performance compared to the state-of-the-art \peft methods.
% Remarkably, \dept efficiency amplifies with increasing model sizes, making it exceptionally apt for LLMs.
% Our further analysis shows the compatibility of \dept with \petl approaches and highlights its versatility across diverse model architectures and scales.

In this work, we propose \dept, which substantially improves the efficiency of vanilla \pt in terms of both time and memory costs while delivering competitive or even superior performance compared to state-of-the-art PEFT methods. Our comprehensive evaluation across 23 \nlp and vision-language tasks demonstrates \dept's ability to maintain high performance while reducing computational overhead, with improvements of over 20\% in training efficiency using \tfbase models. Remarkably, \dept's efficiency advantages amplify with increasing model sizes, making it exceptionally well-suited for LLMs and addressing a critical need in the era of ever-growing model architectures. The method's ability to decompose soft prompts into more efficient components while preserving model performance represents a significant advancement in parameter-efficient fine-tuning approaches. 
Our further analysis reveals \dept's seamless compatibility with \petl approaches, highlighting its versatility as a fine-tuning solution that can be readily integrated into existing workflows. Additionally, \dept's strong performance in few-shot learning scenarios demonstrates its robustness across different training regimes and data availability conditions. The method's success across diverse model architectures and scales, from smaller models to LLMs, indicates its potential as a general-purpose solution for efficient model adaptation and deployment. Most importantly, \dept's increasing efficiency gains with larger models position it as a promising approach for addressing the computational challenges in deploying and fine-tuning the next generation of LMs.

\chapter{Instruction Tuning With Loss Over Instructions}
\label{chapter:sft}
\section{Introduction}
\begin{figure*}[h]
\centering
\includegraphics[width=\textwidth]{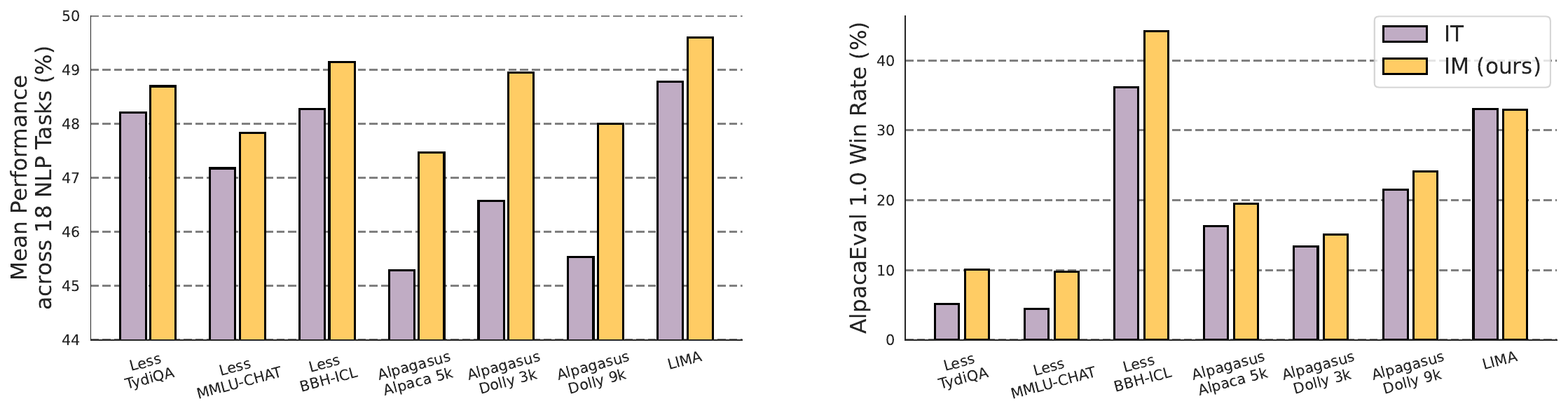}
\caption{
Performance differences between \textsc{Instruction Tuning} (\sft) and our proposed method \textsc{instruction modelling} (\im) trained on 7 instruction tuning datasets. 
(\textbf{Left}) The mean performance across 18 traditional \nlp tasks.
(\textbf{Right}) The win rate on the AlpacaEval 1.0 benchmark.
Please refer to Section \sect{neurips2024:sec:main} for details.
}
\label{neurips2024:fig:results_overview}
\end{figure*}

Language models (LMs) are trained to predict the next token on massive corpora, enabling them to learn general-purpose representations transferable to various language understanding or generation tasks. 
However, it does align LMs to act in accordance with the user’s intentions \cite{leike2018scalable}.
To enable this transfer, various methods for aligning language models have thus been proposed, one of which is instruction tuning (\sft) \cite{ouyang2022training,bai2022training,JMLR:v25:23-0870}.
Zhou et al. \cite{zhou2023lima} propose the Superficial Alignment Hypothesis (\sah): The model’s knowledge and capabilities are learnt almost entirely during pretraining, and only minimal instruction tuning data is required to enable high-quality outputs in the desired output style. 
Existing works \cite{aribandi2022ext,mishra-etal-2022-cross,ouyang2022training,sanhmultitask,wei2021finetuned,xu2024wizardlm,li2024selfalignment} mainly perform instruction tuning by focusing the loss computation solely on the output segments.

\begin{figure*}[!t]
\centering
\includegraphics[width=\textwidth]{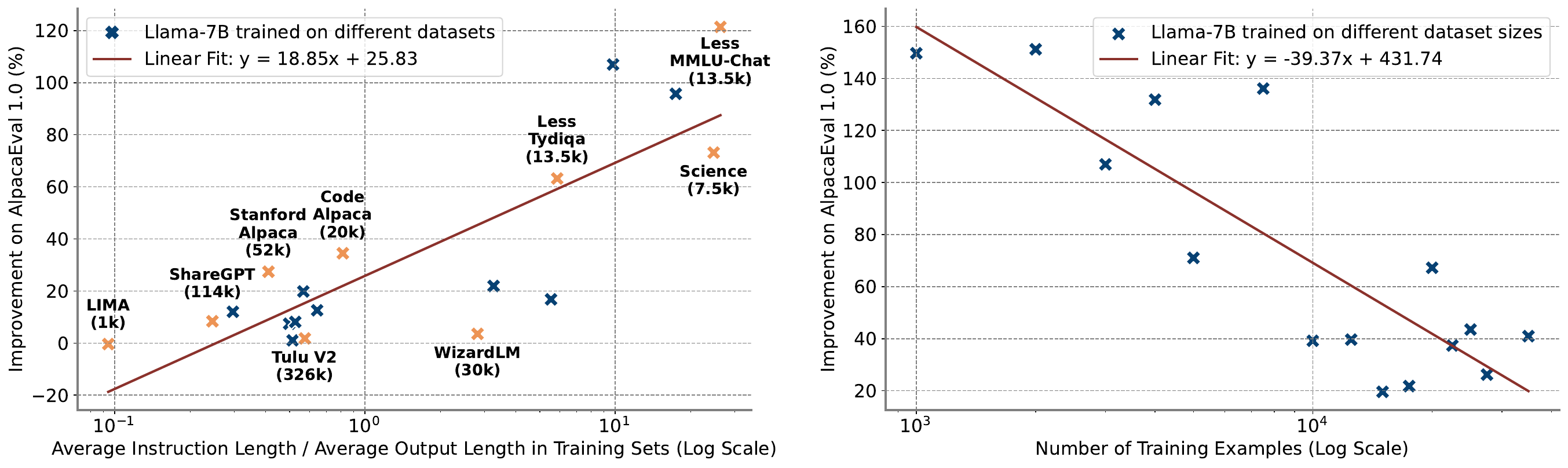}
\caption{
(\textbf{Left}) Performance improvement, achieved by our approach \textsc{instruction modelling} (\im) compared to \textsc{Instruction Tuning} (\sft) on the AlpacaEval 1.0, against the ratio between average instruction length and average output length in instruction tuning datasets (training size noted in parentheses).
We highlight several representative instruction tuning datasets in yellow.
Our analysis suggests that \im is especially beneficial for datasets characterized by lengthy instructions or prompts paired with comparably brief outputs, such as \texttt{Code Alpaca} \cite{codealpaca} and \mmluchat \cite{xia2024less}.
(\textbf{Right}) Performance improvement achieved by our approach \im over \sft on the AlpacaEval 1.0 against the number of training examples in instruction tuning datasets.
Here we maintain a fixed ratio between instruction and output length of 10.
This analysis suggests that \im is particularly effective under the low-resource setting or Superficial Alignment Hypothesis.
Please refer to Section \sect{neurips2024:sec:main} for details.
}
\label{neurips2024:fig:insight}
\end{figure*}

In this work, we demonstrate that in many scenarios, incorporating the loss computation for instructions or prompts, which we refer to as \textsc{Instruction Modelling} (\im) (see Section \sect{neurips2024:sec:method}), could substantially improve the performance of instruction tuning on both various \nlp tasks (\eg MMLU, TruthfulQA, and HumanEval) and open-ended generation benchmarks (\eg MT-Bench and AlpacaEval), as shown in Figure \ref{neurips2024:fig:results_overview}.
Remarkably, in the most favourable case, our proposed method \im boosts performance on AlpacaEval 1.0 by over 100\%.
% As shown in Figure \ref{neurips2024:fig:results_overview}, \im improves the \textsc{Llama-2-7-Base} performance on 18 \nlp tasks and the win rate on the AlpacaEval 1.0 benchmark.
%
As illustrated in Figure \ref{neurips2024:fig:insight}, our study further identifies two key factors influencing the effectiveness of \im:
(1) \textbf{The ratio between instruction length and output length} (see Figure \ref{neurips2024:fig:insight} Left). Our analysis shows that our approach \im is especially beneficial for datasets characterised by lengthy instructions or prompts paired with comparably brief outputs, such as \codealpaca \cite{vicuna2023} and \mmluchat \cite{xia2024less};
(2) \textbf{The number of training examples} (see Figure \ref{neurips2024:fig:insight} Right). We demonstrate that our approach \im performs better under the \sah, where a small amount of training examples are available (see Section \sect{neurips2024:sec:main}).

Recent works \cite{overfitting2023,jain2024neftune,ouyang2022training,xue2023to,yang2024bayesian} suggest that LMs can quickly memorise training examples even after seeing them just once.
We hypothesise that the improvement stems from reducing instruction tuning's tendency to overfit, particularly under limited training resource conditions: 
Instruction tuning on brief outputs or a small amount of data can potentially lead to rapid overfitting. 
To substantiate our hypothesis, our analysis shows that 
(1) \im exhibits higher training losses but lower test losses on new instruction tuning data;
(2) The outputs generated by \im have a lower similarity to the training examples compared to those from \sft, as indicated by BLEU scores; and
(3) \im leads to less performance degradation on \nlp tasks across training epochs (see Section \sect{neurips2024:sec:overfitting}).
Additionally, our study reveals that this overfitting cannot be effectively addressed by applying Kullback-Leibler (KL) divergence for regularisation \cite{bai2022training,ouyang2022training}, as it compromises the model's ability to follow instructions. 
Our further analysis reveals that the advantages of \im persist across different LMs and model sizes, and that \im could be effectively combined with the previous approach (\ie \neftune \cite{jain2024neftune}).
Meanwhile, we investigate the relationship between output length and win rate for our approach (see Section \sect{neurips2024:sec:analysis}). 

In summary, the main contributions of this chapter are:
\begin{itemize}
    \item We propose \textsc{Instruction Modelling} (\im), aiming to enhance both the instruction-following and general performance on \nlp tasks of LMs. Through extensive experiments across 21 benchmarks, we demonstrate that, in many scenarios, \im substantially improves performance of LMs trained on various instruction tuning datasets, particularly notable in the AlpacaEval 1.0 benchmark where it boosts scores by over 100\%.
    \item Our study identifies key factors influencing the effectiveness of \im, including the ratio between instruction length and output length and the number of training examples, providing practical guidance for instruction tuning LMs, especially under the low-resource scenarios.
    \item We provide underlying mechanisms that make IM effective, specifically how it mitigates overfitting, thereby enhancing the LMs' performance across various tasks.
\end{itemize}

\section{Our Approach: Instruction Modelling}
\label{neurips2024:sec:method}
In this section, we introduce our proposed method, \textsc{instruction modelling} (\im).
% \paragraph{Our Approach: Instruction Modelling.} 
Our approach is an expansion of instruction tuning by incorporating loss calculation for both the instruction and the completion tokens, except it omits any special prompt template tokens. 
The model is trained to predict both the instruction and completion parts of $s$ but excludes tokens that are part of prompt templates (denoted as $T$). 
For simplicity, we consider that these template tokens are not part of $I$ or $C$.
\begin{figure}[!h]
\centering
\includegraphics[width=0.9\textwidth]{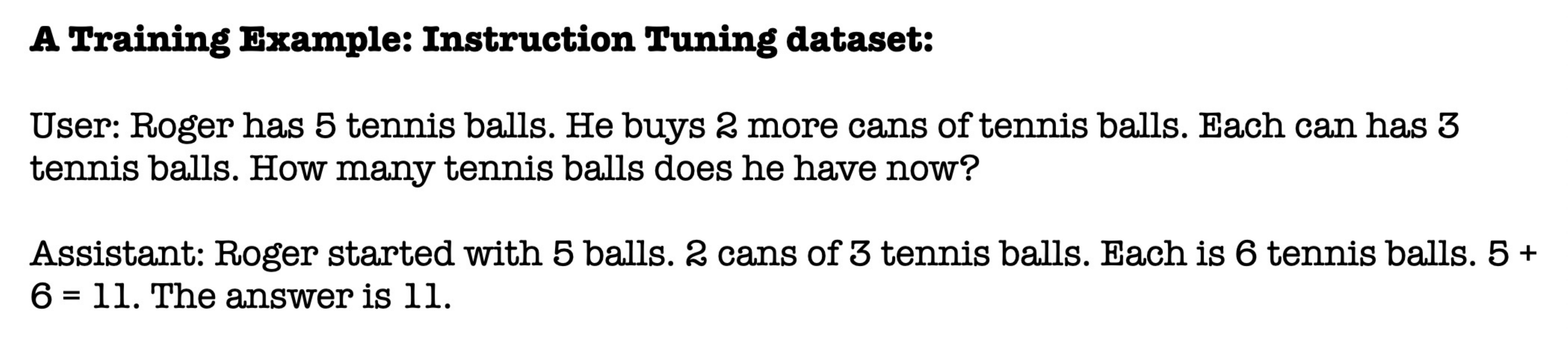}
\caption{
An example of instruction tuning training data.
}
\label{neurips2024:fig:sft_trainingdat}
\end{figure}
The model predicts the next token given all previous tokens (both instructions and completions up to that point):
\begin{align}
\label{neurips2024:eq:im_prob}
P(x) = P(I_1, I_2, ..., I_m, C_1, C_2, ..., C_n) = \prod_{t=1}^{m+n} P(s_t | s_1, s_2, ..., s_{t-1})
\end{align}
The loss function, $\mathcal{L}$, for instruction modelling calculates the negative log-likelihood for both instruction and completion tokens, excluding any prompt template tokens. It is computed as follows:
\begin{align}
\label{neurips2024:eq:im_loss}
% \myfonts
% \mathcal{L} = - \sum_{t=1}^{m+n} \log P(s_t | s_1, s_2, ..., s_{t-1}) \cdot \mathbf{1}(s_t \notin T),
\mathcal{L} = - \sum_{t=1}^{m+n} \log P(s_t | s_1, s_2, ..., s_{t-1})
\end{align}
% where $\mathbf{1}(s_t \notin T)$ is an indicator function that is 1 if $s_t$ is not a template token and 0 otherwise. 
This ensures that the loss is computed only over the meaningful tokens, not over the static template tokens.
Our approach allows the model to improve its understanding of both the instructions and the completions while being sensitive to the context provided by both segments of the input sequence.
\section{Experiments and Results}
In this section, we evaluate the effectiveness of our proposed method \textsc{instruction modelling} (\im) by comparing it with \textsc{Instruction Tuning} (\sft) and other baselines on various datasets.

\subsection{Experimental Setup}
\label{neurips2024:sec:setup}
\paragraph{Instruction Tuning Datasets.}
Following previous research \cite{ivison2023camels,xia2024less,chen2024alpagasus}, we assess our method, \im, across various instruction tuning datasets, detailed as follows:
\begin{itemize}
\item (1) \alpaca \cite{alpaca} ($52\,002$ examples);
\item (2) \dolly \cite{DollyV2} ($15\,011$ examples);
% (3) \flan \cite{JMLR:v25:23-0870} (50,000 examples);
\item (3) \sharegpt \cite{vicuna2023} ($50\,000$ examples);
\item (4) \codealpaca \cite{codealpaca} ($20\,022$ examples);
\item (5) \science \cite{ivison2023camels} ($7\,544$ examples);
\item (6) \wizardlm \cite{xu2024wizardlm} ($30\,000$ examples);
\item (7) \tulu \cite{ivison2023camels} ($326\,181$ examples).
Additionally, we incorporate instruction tuning datasets under the low-resource setting or \sah:
\item (8) \lima \cite{zhou2023lima} ($1\,030$ examples);
\item (9) \less\footnote{\url{https://github.com/princeton-nlp/LESS}} \cite{xia2024less}, where high-quality instruction tuning data are selected from \flan and \dolly. Here, we use the \mmluchat ($13\,533$ examples), \bbhicl ($13\,533$ examples), and \tydiqa ($13\,533$ examples);
\item (10) \alpagasus\footnote{\url{https://github.com/gpt4life/alpagasus}} \cite{chen2024alpagasus}, which offers three subsets: \alpagasusdollyone ($2\,996$ examples), \alpagasusdollytwo ($9\,229$ examples) selected from \dolly, and \alpagasusalpaca ($5\,305$ examples) selected from \alpaca.
\end{itemize}
See dataset details and statistical analysis in Appendix \ref{neurips2024:sec:sft_dataset}.

\paragraph{Evaluation Benchmarks.}
\label{neurips2024:para:evaluation}
Our study conducts a comprehensive analysis of 21 \nlp datasets, focusing on a suite of canonical \nlp benchmarks and their capacity for open-ended language generation. 
For canonical NLP benchmarks, the evaluation is organised into six categories (18 tasks in total): 
(1) \textit{Language Understanding and Knowledge} includes MMLU \cite{hendrycks2021measuring}, PIQA \cite{Bisk2020piqa}, OpenbookQA \cite{mihaylov-etal-2018-suit}, HellaSwag \cite{zellers-etal-2019-hellaswag}, and LAMBADA \cite{paperno-etal-2016-lambada};
(2) \textit{Multilinguality} contains LAMBADA Multilingual \cite{paperno-etal-2016-lambada}, WMT 2014 \cite{bojar-etal-2014-findings}, and WMT 2016 \cite{sennrich-etal-2016-edinburgh};
(3) \textit{Commonsense Reasoning} features Winograd schema challenge (WSC) \cite{ea01b9c0db064caca6986b925d75f2bb}, WinoGrande \cite{sakaguchi2019winogrande}, AI2 Reasoning Challenge (ARC) \cite{Clark2018ThinkYH}, and CoQA \cite{reddy-etal-2019-coqa};
(4) \textit{Math and Coding Reasoning} includes GSM8K \cite{cobbe2021training}, and HumanEval \cite{chen2021codex};
(5) \textit{Safety and Helpfulness} comprises TruthfulQA \cite{lin-etal-2022-truthfulqa}, ToxiGen \cite{hartvigsen-etal-2022-toxigen}, and Hendrycks Ethics \cite{hendrycks2021aligning}. 
(6) \textit{Big Bench Hard (BBH)} dataset \cite{suzgun-etal-2023-challenging} is included to assess models.
Our models are also tested for their open-ended text generation capabilities using model-based evaluations, specifically through MT-Bench \cite{zheng2023judging}, AlpacaEval 1.0 and 2.0 \cite{alpaca_eval}, 
where the AlpacaEval 1.0 compares the model outputs against the \texttt{text\_davinci\_003} evaluated by GPT-4 and the AlpacaEval 2.0 compares the model outputs against \texttt{GPT-4} outputs evaluated by GPT-4 Turbo.
See evaluation details in Appendix \ref{neurips2024:sec:evaluation}.

\paragraph{All Comparison Approaches.} 
In our study, we mainly conduct experiment using the \textsc{Llama-2-7B-Base} and \textsc{Llama-2-13B-Base} \cite{touvron2023llama}, and the \textsc{Opt-6.7B} \cite{zhang2022opt} models.
We report model performance trained on \textsc{Llama-2-7B-Base} if not specified.
We compare with \neftune \cite{jain2024neftune} as the baseline, which adds noise to the embedding during the instruction tuning to increase the robustness of instruction-tuned models.
See hyperparameter and implementation details in Appendix \ref{neurips2024:sec:implementation_details}.

% \newcolumntype{a}{>{\columncolor{gray!25}}c}

\begin{table}[!t]
\centering
\caption{
Performance comparisons using 7 instruction tuning datasets with the \textsc{Llama-2-7B} on 6 categories of 18 traditional \nlp tasks and 3 open-ended benchmarks with LLM as judgements. 
``IT'' refers to \textsc{instruction tuning}.
``IM'' refers to \textsc{instruction modelling}.
Green and red arrows indicate performance changes against the baseline (\sft).
}
\label{neurips2024:table:main}
\resizebox{\textwidth}{!}{
\begin{tabular}{lcccccccccc}
        \toprule
        & \multicolumn{7}{c}{\textbf{NLP Benchmarks}} & \multicolumn{3}{c}{\textbf{LLM-based Evaluation}} \\
        \cmidrule(lr){2-8} \cmidrule(lr){9-11} 
        % \textbf{Method} & \textbf{\makecell{Understanding\\\& Knowledge}} & \textbf{\makecell{Multi-\\linguality}} & \textbf{\makecell{Commonsense \\Reasoning}} & \textbf{\makecell{Math\&Code \\Reasoning}} & \textbf{BBH} & \textbf{\makecell{Safety \& \\Helpfulness}} & \textbf{Mean} & \textbf{MT-Bench} & \textbf{\makecell{AlpacaEval\\1.0}} & \textbf{\makecell{AlpacaEval\\2.0}} \\
        \textbf{\makecell{Method}} & \textbf{\rotatebox{90}{Understanding \& Knowledge}} & \textbf{\rotatebox{90}{Multilinguality}} & \textbf{\rotatebox{90}{Commonsense Reasoning}} & \textbf{\rotatebox{90}{Math\&Code Reasoning}} & \textbf{\rotatebox{90}{BBH}} & \textbf{\rotatebox{90}{Safety \& Helpfulness}} & \textbf{\rotatebox{90}{Mean}} & \textbf{\rotatebox{90}{MT-Bench}} & \textbf{\rotatebox{90}{AlpacaEval 1.0}} & \textbf{\rotatebox{90}{AlpacaEval 2.0}} \\
        \midrule
        \llamabase                                            & 63.91 & 61.99 & 75.86 & 13.32 & 38.80 & 42.03 & 49.32 & 1.16 & 0.01 & 0.01  \\
        \llamachat                                            & 63.42 & 55.15 & 70.28 & 15.33 & 38.92 & 51.79 & 49.15  & 6.63 & 79.04 & 6.48  \\
        \midrule
        \multicolumn{11}{l}{\alpagasusalpaca (5,305 training examples)} \\
        \hdashline\noalign{\vskip 0.4ex}
        \sft                                                    & 64.98 & 57.24 & 66.06 & 8.93  & 26.80 & 47.74 & 45.29          & 3.62          & 16.29          & 2.46 \\  % 2 epoch results
        \neftune                                                & 65.18 & 56.88 & 66.45 & 10.24 & 29.53 & 45.46 & 45.62\uaa{0.33} & 3.50\daa{0.12} & 21.37\uaa{5.08} & 2.37\daa{0.09} \\
        \im (ours)                                            & 64.01 & 56.63 & 72.47 & 11.58 & 35.52 & 44.62 & 47.47\uaa{2.18} & 3.48\daa{0.14} & 19.52\uaa{3.23} & 3.29\uaa{0.83}\\
        \midrule
        \multicolumn{11}{l}{\alpagasusdollyone (2,996 training examples)} \\
        \hdashline\noalign{\vskip 0.4ex}
        \sft                                                    & 65.81 & 57.46 & 67.55 & 11.96 & 33.02 & 43.70 & 46.58          & 4.23          & 13.42          & 2.00  \\
        \neftune                                                & 65.90 & 57.79 & 67.28 & 11.64 & 35.43 & 44.36 & 47.07\uaa{0.49} & 4.42\uaa{0.19} & 14.04\uaa{0.62} & 2.03\uaa{0.03} \\
        \im (ours)                                            & 65.66 & 57.47 & 73.24 & 14.57 & 37.48 & 45.29 & 48.95\uaa{2.37} & 4.06\daa{0.17} & 15.11\uaa{1.69} & 2.44\uaa{0.44} \\
        \midrule
        \multicolumn{11}{l}{\alpagasusdollytwo (9,229 training examples)} \\
        \hdashline\noalign{\vskip 0.4ex}
        \sft                                                    & 64.10 & 56.62 & 69.70 & 7.96  & 32.19 & 42.65 & 45.54          & 4.33          & 21.54           & 2.28 \\  % 2 epoch results
        \neftune                                                & 64.20 & 56.69 & 69.51 & 8.99  & 33.91 & 42.62 & 45.99\uaa{0.45} & 4.21\daa{0.12} & 31.61\uaa{10.07} & 2.84\uaa{0.56} \\
        \im (ours)                                            & 64.67 & 55.32 & 74.87 & 12.50 & 36.69 & 43.96 & 48.00\uaa{2.46} & 4.55\uaa{0.22} & 30.77\uaa{9.23}  & 2.67\uaa{0.39}  \\ % Check
        \midrule
        \multicolumn{11}{l}{\tydiqa (13,533 training examples)} \\
        \hdashline\noalign{\vskip 0.4ex}
        \sft                                                  & 64.01 & 56.81 & 64.77 & 12.06 & 36.54 & 55.09 & 48.21          & 4.08          & 5.12              & 1.88 \\  % 2 epoch results
        \neftune                                              & 64.03 & 55.09 & 64.02 & 13.84 & 36.65 & 51.21 & 47.47\daa{0.74} & 4.19\uaa{0.11} & 8.35\uaa{3.23}     & 2.58\uaa{0.70} \\
        \im (ours)                                          & 64.28 & 56.10 & 65.70 & 17.15 & 34.86 & 54.09 & 48.70\uaa{0.49} & 4.36\uaa{0.28} & 10.10\uaa{4.98}    & 2.88\uaa{1.00} \\ %Check
        \midrule
        \multicolumn{11}{l}{\mmluchat (13,533 training examples)} \\
        \hdashline\noalign{\vskip 0.4ex}
        \sft                                                  & 64.74 & 57.42 & 62.94 & 9.53  & 33.13 & 55.35 & 47.18           & 3.86          & 4.42          & 1.20 \\  % 2 epoch results
        \neftune                                              & 65.21 & 57.43 & 63.14 & 9.45  & 35.89 & 55.32 & 47.74\uaa{0.56}  & 4.06\uaa{0.20} & 6.22\uaa{1.80} & 1.06\daa{0.14} \\
        \im (ours)                                          & 63.95 & 56.34 & 64.76 & 12.52 & 36.94 & 52.55 & 47.84\uaa{0.66}  & 4.54\uaa{0.68} & 9.78\uaa{5.36} & 1.93\uaa{0.73} \\
        \midrule
        \multicolumn{11}{l}{\bbhicl (13,533 training examples)} \\
        \hdashline\noalign{\vskip 0.4ex}
        \sft                                                   & 63.83 & 62.04 & 75.92 & 6.90  & 38.93 & 42.07 & 48.28          & 4.78          & 36.20          & 2.36 \\  % 2 epoch results
        \neftune                                               & 63.88 & 58.83 & 67.97 & 13.54 & 38.63 & 51.33 & 49.03\uaa{0.75} & 5.05\uaa{0.27} & 39.81\uaa{3.61} & 2.87\uaa{0.51} \\
        \im (ours)                                           & 64.14 & 56.72 & 71.12 & 13.56 & 39.03 & 50.34 & 49.15\uaa{0.87} & 5.03\uaa{0.25} & 44.15\uaa{7.95} & 3.56\uaa{1.20} \\
        \midrule
        \multicolumn{11}{l}{\lima (1,030 training examples)} \\
        \hdashline\noalign{\vskip 0.4ex}
        % \midrule
        \sft                                                     & 63.92 & 58.29 & 71.96 & 16.01 & 39.27 & 43.29 & 48.79          & 4.77          & 33.06          & 2.58 \\ 10 epoch
        \neftune                                                 & 63.66 & 57.67 & 73.03 & 15.95 & 38.77 & 43.14 & 48.70\daa{0.09} & 4.79\uaa{0.02} & 30.51\daa{2.55} & 2.43\daa{0.15} \\
        \im (ours)                                             & 64.49 & 58.21 & 75.55 & 17.06 & 38.84 & 43.45 & 49.60\uaa{0.81} & 4.83\uaa{0.06} & 32.94\daa{0.12} & 2.47\daa{0.11} \\ 
        \bottomrule
\end{tabular}
}
\end{table}
\subsection{Main Results}
\label{neurips2024:sec:main}
In this section, we first evaluate the model performance of our approach and baselines across various tasks. Then we investigate the key factors that contribute to the effectiveness of our approach.
Below we will discuss our findings in detail.

\paragraph{\#1: Our approach \im can improve the performance of instruction tuning on various \nlp tasks and open-ended generation benchmarks.}
% Figure \ref{neurips2024:fig:results_overview} provides a summary of the model's performance across both traditional NLP tasks and the AlpacaEval 1.0 benchmark.
Table \ref{neurips2024:table:main} offers a detailed breakdown of experimental results for traditional NLP tasks across six categories, as well as performance on additional benchmarks for open-ended generation (\ie MT-Bench and AlpacaEval).
The experimental results show that our approach (\im) can improve the performance of instruction tuning on various \nlp tasks and open-ended generation benchmarks.
Specifically, on the \alpagasusdollyone dataset, \im improves the overall mean score of \nlp tasks to 48.95, an increase of 2.37 points from the baseline. 
Similarly, on the \alpagasusdollytwo dataset, we observe an improvement of 2.46 points in the mean NLP score. 
These improvements are mirrored in the LLM-based evaluations.
Specifically, \im raises scores on the AlpacaEval 1.0 benchmark, achieving approximately a ten-point increase on the \alpagasusdollytwo dataset and doubling the performance on datasets such as \mmluchat and \tydiqa.
However, the extent of improvement varies across different datasets. 
For example, the \lima dataset shows more modest gains, prompting our further analysis to understand the factors influencing the effectiveness of \im.

\paragraph{\#2: Our approach \im is especially beneficial for datasets characterised by lengthy instructions or prompts paired with comparably brief outputs.} 
To better understand the impact factors on the effectiveness of \im, we extend our experiments to more instruction-tuning datasets, such as \science, \codealpaca and \tulu.
Interestingly, as shown in Figure \ref{neurips2024:fig:insight} (left), we find that \im is particularly effective in scenarios where datasets characterised by lengthy instructions and shorter outputs, such as \mmluchat and \bbhicl.
For example, in datasets like \mmluchat and \tydiqa, \im shows remarkable efficacy. 
In contrast, the \tulu dataset, with an instruction to output length ratio of about 0.5, benefits less compared to the \science dataset, which has a much higher ratio of 24.7. 
We hypothesise that this trend can be attributed to the tendency of language models trained on datasets with shorter outputs to overfit. 
In cases where the instructions are longer, \im acts as an effective form of regularisation, mitigating this issue. 
% For further details on the experimental setup, refer to the Appendix \ref{neurips2024:sec:implementation_details}.

\paragraph{\#3: Our approach \im performs better with fewer training examples.}
We find that another important factor in the effectiveness of \im is the quantity of training examples.
Specifically, we design additional experiments by sampling different numbers of examples from the \tulu datasets, which contain about 320k training examples and achieve a modest improvement compared to other datasets in Figure \ref{neurips2024:fig:insight} (left).
We ensure that our samples maintain an instruction-to-output length ratio of around 10.
As all these samples are from \tulu, we can assume they are from the same distribution.
As shown in Figure \ref{neurips2024:fig:insight} (right), \im demonstrates substantial performance improvements on the AlpacaEval 1.0 as the number of training examples decreases.
This suggests that \im could be particularly valuable for developing robust models in resource-constrained scenarios or under the \sah.

\subsection{Instruction Modelling Mitigates Overfitting of Instruction Tuning}
\label{neurips2024:sec:overfitting}
This section explores the underlying interpretation behind the effectiveness of our approach. 
Our experimental results demonstrate that \im can alleviate the overfitting problem of Instruction Tuning. 
Below we will discuss our findings in detail.

\begin{figure}[!ht]
\centering
\includegraphics[width=\textwidth]{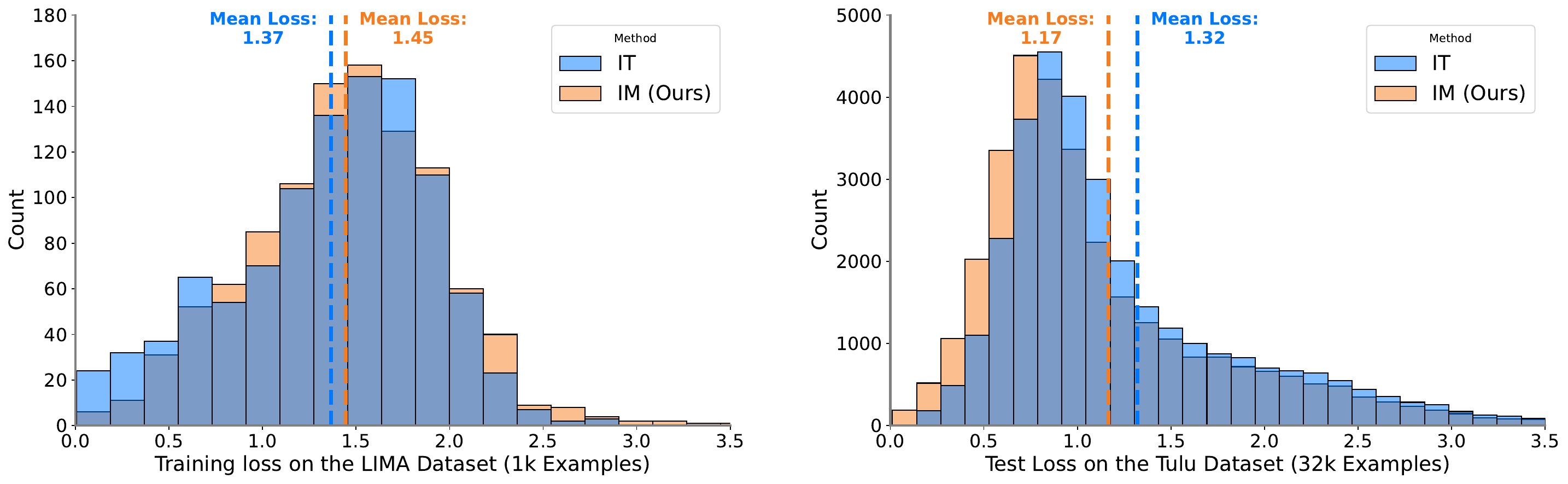}
\caption{
(\textbf{Left}) 
Training loss distribution for each example between our approach \textsc{instruction modelling} (\im) and \textsc{Instruction Tuning} (\sft) on the \lima dataset.
(\textbf{Right}) 
Test loss distribution for each example between \im and \sft on the \tulu dataset, using a 10\% randomly sampled data for efficacy.
Mean losses are marked by dashed lines.
For both \im and \sft, here we only compute the loss over the output part.
\im has a higher train loss with lower test loss, suggesting that \im effectively mitigates the overfitting issues compared to \sft.
% See Appendix \ref{neurips2024:sec:train_test_loss} for more examples.
}
\label{neurips2024:fig:train_test_loss_analysis_overfitting}
\end{figure}

\paragraph{\#1. Train and test loss analysis.}
Figure \ref{neurips2024:fig:train_test_loss_analysis_overfitting} clearly illustrates the effectiveness of our approach \im in mitigating overfitting issues compared to \sft. 
For both \im and \sft, here we only compute the loss over the output part.
In the training loss distribution for the \lima dataset, \im exhibits a slightly higher mean loss of $1.45$ compared to $1.37$ for \sft, suggesting that \im does not overfit to the training data as much as \sft does. 
This is further corroborated in the test loss distribution on the \tulu dataset (using a 10\% randomly sampled data set), where \im demonstrates a lower mean test loss of $1.17$ compared to $1.32$ for \sft. 
This indicates that \im maintains better generalisation to new data, emphasising the model's capability to learn effectively without fitting excessively to training examples.

\begin{figure}[ht]
\centering
\includegraphics[width=\textwidth]{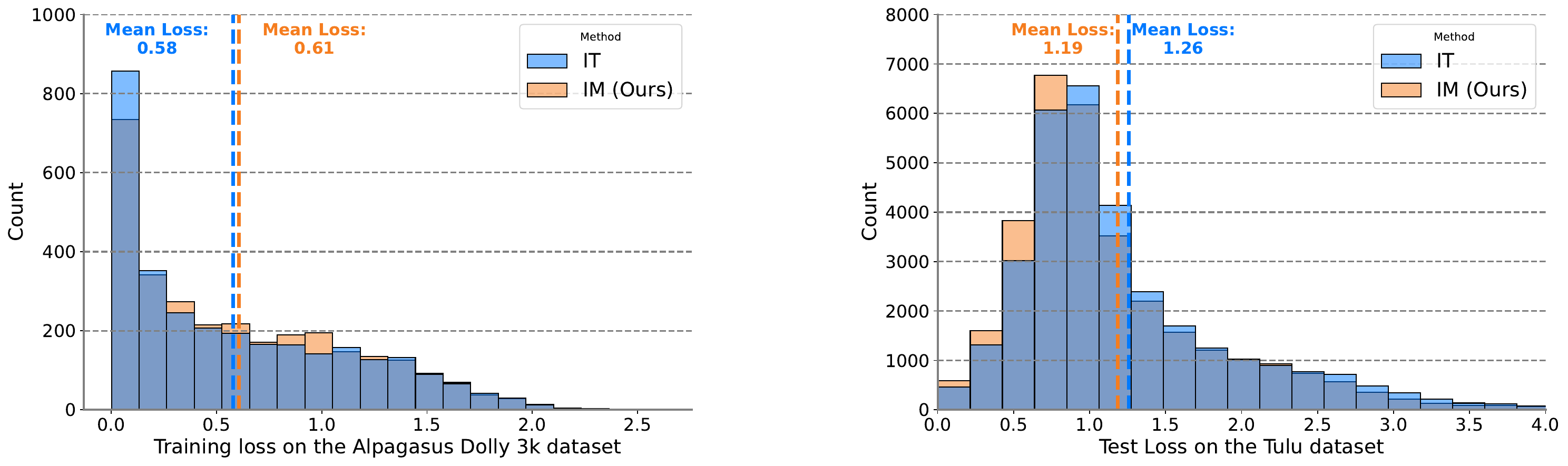}
\caption{
(\textbf{Left}) 
Training loss distribution for each example between our approach \textsc{instruction modelling} (\im) and \textsc{Instruction Tuning} (\sft) on the \alpagasusdollyone dataset.
(\textbf{Right}) 
Test loss distribution for each example between \im and \sft on the \tulu dataset, using a 10\% sampled data.
Mean losses are marked by dashed lines.
For both \im and \sft, here we only compute the loss over the output part.
\im has a higher train loss with lower test loss, suggesting that \im effectively mitigates the overfitting issues compared to \sft. 
}
\label{neurips2024:fig:loss_dolly3k_tulu}
\end{figure}

\begin{figure}[ht]
\centering
\includegraphics[width=\textwidth]{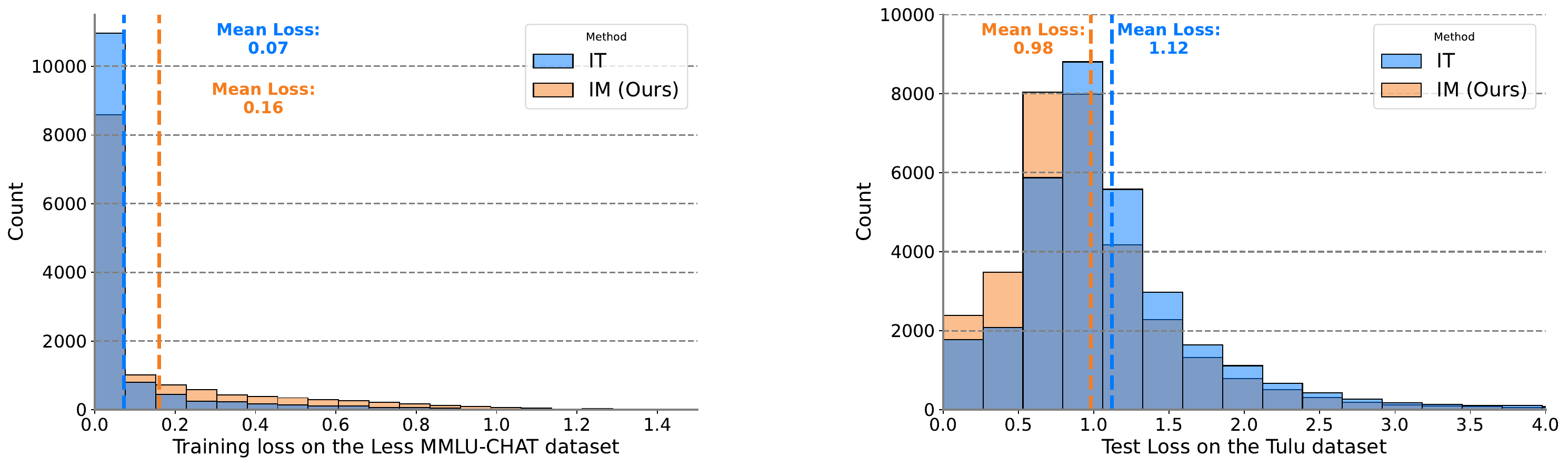}
\caption{
(\textbf{Left}) 
Training loss distribution for each example between our approach \textsc{instruction modelling} (\im) and \textsc{Instruction Tuning} (\sft) on the \mmluchat dataset.
(\textbf{Right}) 
Test loss distribution for each example between \im and \sft on the \tulu dataset, using a 10\% sampled data.
Mean losses are marked by dashed lines.
For both \im and \sft, here we only compute the loss over the output part.
\im has a higher train loss with lower test loss, suggesting that \im effectively mitigates the overfitting issues compared to \sft. 
}
\label{neurips2024:fig:loss_mmlu_chat_tulu}
\end{figure}

We provide additional experiments regarding training and testing loss distributions.
Figure \ref{neurips2024:fig:loss_dolly3k_tulu} focuses on the \alpagasusdollyone and \tulu datasets, displaying how \im tends to exhibit higher training losses yet achieves lower test losses compared to \sft. Similarly, Figure \ref{neurips2024:fig:loss_mmlu_chat_tulu} compares these methods on the \mmluchat and \tulu datasets under analogous conditions.

\begin{table}[h]
\centering
\caption{
Average BLEU Score comparison of \im and \sft, where a lower score indicates less overfitting.
Green and red arrows indicate performance changes against the baseline (\sft).
% This analysis shows that \im reduces the overfitting of \sft on training sets.
}
\label{neurips2024:table:bleu}
\resizebox{\textwidth}{!}{
\begin{tabular}{lccccccc}
\toprule
 & \lima & \makecell{\texttt{Less}\\\texttt{Tydiqa}} & \makecell{\texttt{Less}\\\texttt{MMLU Chat}} & \makecell{\texttt{Less}\\\texttt{BBH ICL}} & \makecell{\texttt{Alpagasus}\\\texttt{Alpaca 5k}} & \makecell{\texttt{Alpagasus}\\\texttt{Dolly 9k}} & \makecell{\texttt{Alpagasus}\\\texttt{Dolly 3k}} \\
\midrule
\textbf{\sft}                  & 18.15          & 69.21           & 72.43           & 60.96          & 72.26           & 61.76          & 60.99 \\
\textbf{\im} (ours)          & 17.30\daa{0.85} & 65.63\daa{3.58}  & 69.20\daa{3.23}  & 53.94\daa{7.02} & 70.50\daa{1.76}  & 60.61\daa{1.15} & 59.04\daa{1.95} \\
\bottomrule
\end{tabular}
}
\end{table}

\paragraph{\#2. BLEU score analysis.}
Here we generate outputs using the instructions from the training examples via greedy decoding, and then compare the generated outputs with the ground truth outputs in training examples and report the results. 
We use the BLEU score (up to n-gram order 4) \cite{papineni-etal-2002-bleu} to measure the similarity between outputs, where a higher score on outputs indicates a higher overlap with training examples.
As shown in Table \ref{neurips2024:table:bleu}, outputs generated by \im consistently have lower BLEU scores than those generated by \sft. 
This suggests that \im produces outputs have less overlap with the ground truth outputs in training examples, indicating less overfitting.

\begin{figure*}[!ht]
\centering
\includegraphics[width=\textwidth]{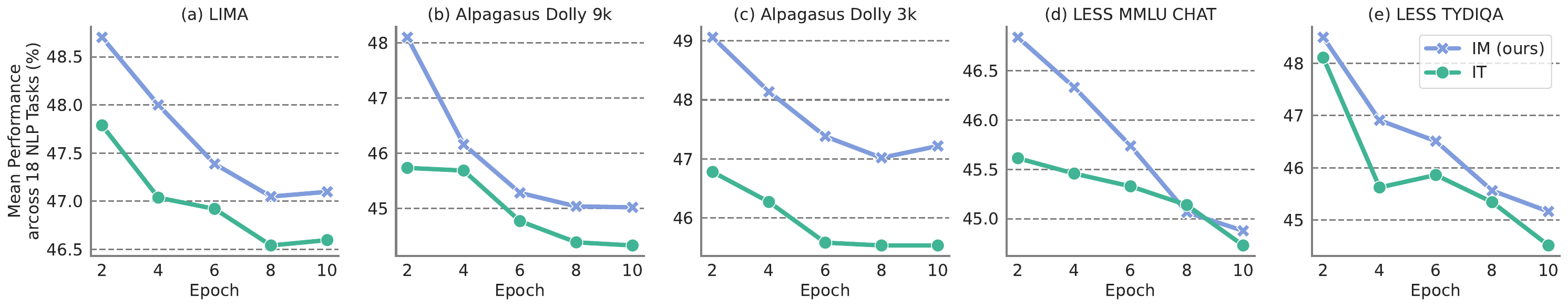}
\caption{
Mean performance on 18 \nlp tasks over epochs using \textsc{Llama-2-7B-Base}.
This analysis suggests that \im experiences a lower instruction tuning tax compared to \sft.
}
\label{neurips2024:fig:nlp_results_vs_epochs}
\end{figure*}

\paragraph{\#3. Instruction Tuning Tax on the \nlp tasks.}
Previous works show that training LMs with RLHF causes an \textit{Alignment Tax} on the \nlp tasks \cite{bai2022training,ouyang2022training}. 
In this study, we observe that instruction tuning can sometimes lead to diminished model capabilities in some areas, such as multilinguality and commonsense reasoning.
To this end, we further explore the impact of instruction tuning on the performance of NLP tasks. 
Figure \ref{neurips2024:fig:nlp_results_vs_epochs} illustrates that our approach \im generally has a lower instruction tuning tax compared to \sft, suggesting better robustness under low-resource settings.
We provide additional experiments for win rates across epochs in Appendix \ref{neurips2024:sec:win_rate_vs_epoch}.

\begin{table}[!th]
\centering
\caption{Performance on 18 NLP benchmarks and AlpacaEval 2.0.
Green and red arrows indicate performance changes against the baseline (\textsc{Llama-2-7B-Base}).
This analysis suggests that while applying KL Loss in the instruction tuning helps mitigate performance degradation in NLP tasks, it substantially harms the model performance in open-ended generation tasks.
}
\resizebox{\textwidth}{!}{
\begin{tabular}{lccccc}
\toprule
        &    & \multicolumn{2}{c}{\textsc{Lima (1k)}}         & \multicolumn{2}{c}{\textsc{Alpagasus Dolly (9k)}}
          \cr                                  \cmidrule(lr){3-4}                   \cmidrule(lr){5-6} 
\textbf                & \textbf \textsc{Llama-2-7B-Base} &  \textbf \sft w/o KL Loss  &  \textbf \sft w/ KL Loss  &  \textbf \sft w/o KL Loss &  \textbf \sft w/ KL Loss \\
\midrule
\textbf NLP Tasks      &  49.32         &  48.79\daa{0.53}  &  49.26\daa{0.06}     &  45.54\daa{3.78}   & 49.31\daa{0.01} \\
\textbf AlpacaEval 2.0 &  0.01          &  2.58\uaa{2.57}   &  0.06\uaa{0.05}       &  2.28\uaa{2.27}    & 0.04\uaa{0.03}  \\
\bottomrule
\end{tabular}
}
\label{neurips2024:table:kl}
\end{table}
\paragraph{\#4. Can we simply use KL divergence loss as regularisation for instruction tuning?}
\label{neurips2024:para:kl}
In this analysis, we demonstrate that the application of KL divergence loss in instruction tuning, which is widely used as regularisation for aligning LMs \cite{bai2022training,ouyang2022training}, cannot easily address the overfitting issue of instruction tuning.
Table \ref{neurips2024:table:kl} offers a detailed comparison across various NLP benchmarks and open-ended language generation tasks, particularly using AlpacaEval 2.0, with models trained with and without KL divergence loss.
Our findings are as follows:
(1) Incorporating KL Loss reduces overfitting and reduces the performance degradation on traditional \nlp tasks. 
For example, on the \dolly dataset, incorporating KL Divergence Loss leads to less instruction tuning tax in \nlp tasks, with scores rising from $45.54$ to $49.31$.
(2) KL Loss detrimentally affects the model's instructions following abilities. For example,  on the \lima dataset, we observe a substantial decrease in AlpacaEval 2.0 from $2.58$ to $0.06$.
For additional experiments and implementation details, see Appendix \ref{neurips2024:sec:kl}.

\subsection{Further Analysis}
In this section, we perform three ablation studies for our proposed method.
\label{neurips2024:sec:analysis}

\begin{figure}[!ht]
\centering
\includegraphics[width=\textwidth]{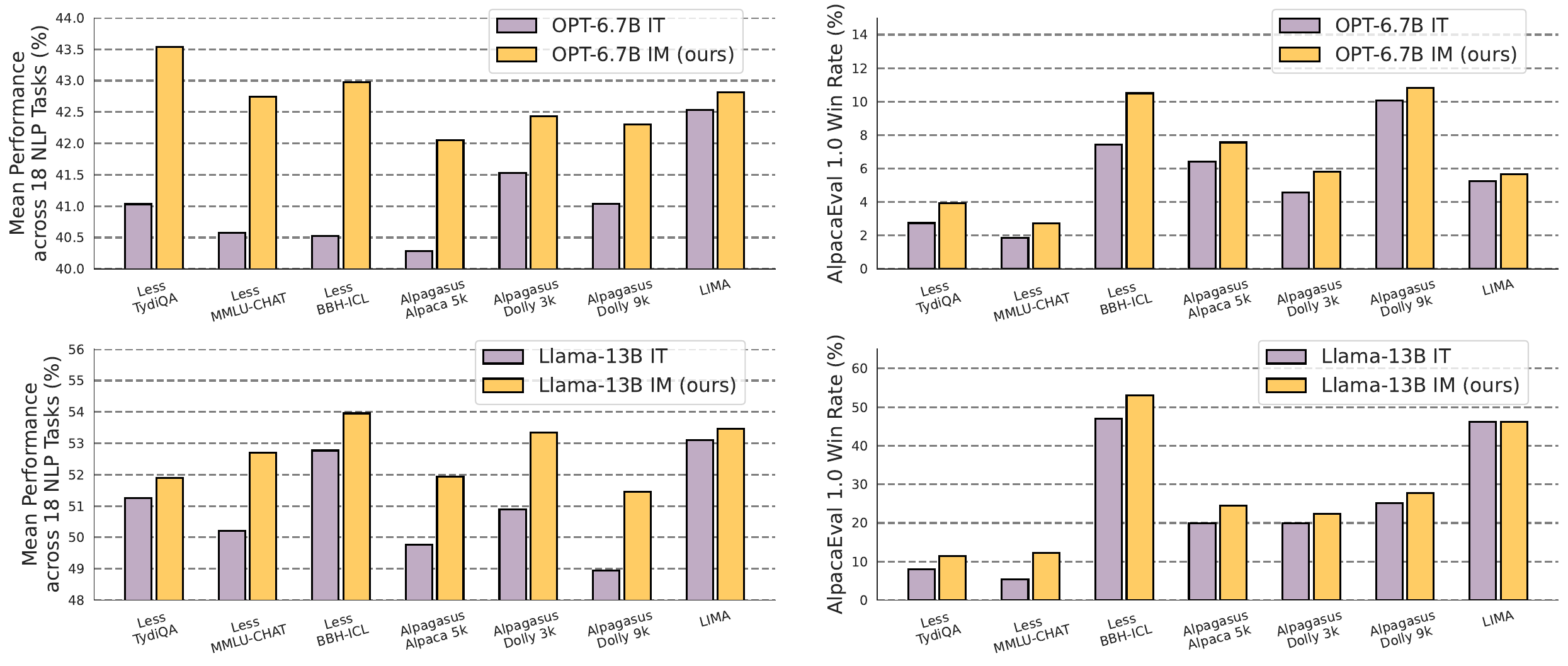}
\caption{
Comparison of \textsc{Instruction Tuning} (\sft) and \textsc{instruction modelling} (\im) methods using \textsc{OPT-6.7B} (\textbf{Top Row}) and \textsc{Llama-2-13B-Base} (\textbf{Bottom Row}) trained on 7 instruction tuning datasets. 
(\textbf{Left}) The mean performance across 18 \nlp tasks.
(\textbf{Right}) The win rate on the AlpacaEval 1.0 benchmark.
}
\label{neurips2024:fig:overview_13b_opt}
\end{figure}
\paragraph{\#1. The advantage of our proposed method persists with different language models and sizes.}
As shown in Figure~\ref{neurips2024:fig:overview_13b_opt}, our analysis demonstrates that our proposed method \im consistently outperforms the \sft across different models and sizes, including \textsc{OPT-6.7B} and \textsc{Llama-2-13B-Base}, on 18 traditional \nlp tasks and the AlpacaEval 1.0 benchmark.
These findings underline the effectiveness of our approach irrespective of the underlying language model or its scale.

\begin{figure*}[!ht]
\centering
\includegraphics[width=\textwidth]{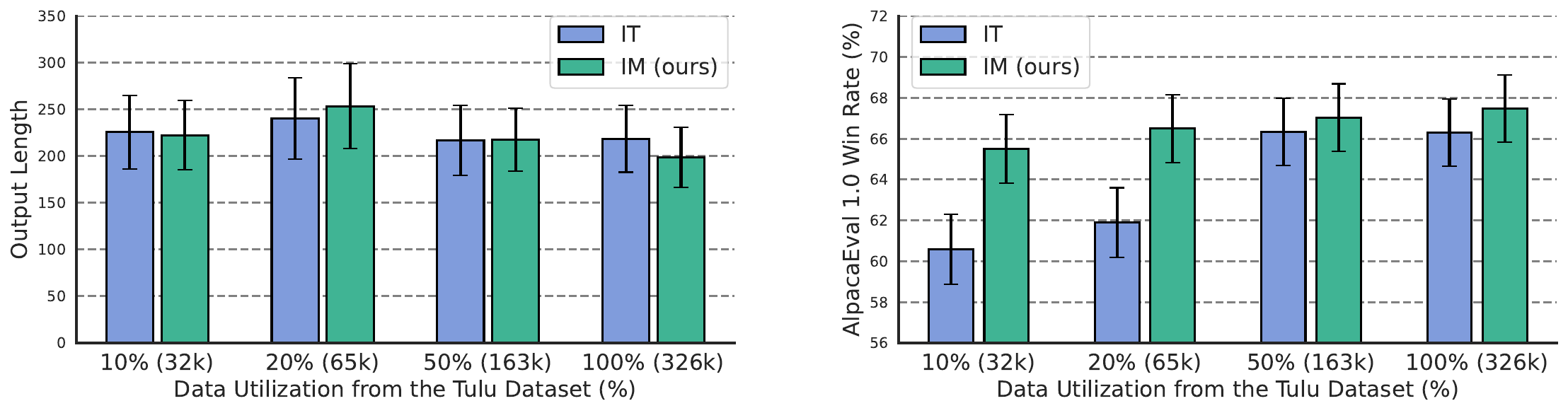}
\caption{
(\textbf{Left}) 
Output length comparison between our approach \textsc{instruction modelling} (\im) and \textsc{Instruction Tuning} (\sft) across various data utilisation levels from the \tulu dataset, as evaluated on the AlpacaEval dataset.
(\textbf{Right}) 
Performance comparison (measured by win rate) between \im and \sft on the AlpacaEval 1.0 across various data utilisation levels from the \tulu dataset.
Importantly, the increase in win rate does not correlate with output length—suggesting that longer responses alone do not explain the performance gains. This highlights that \im's advantages likely stem from differences in learning dynamics or data utilization, rather than verbosity. 
% This analysis suggests that the improvement provided by \im is not necessarily associated with the increased output lengths. See more length analysis in Appendix \ref{neurips2024:sec:analysis_output_length_sah}.
}
\label{neurips2024:fig:tulu_length_alpacaeval}
\end{figure*}

\paragraph{\#2. Relationship between the model output length and the win rate.}
In this analysis, we explore the potential connection between win rates on the AlpacaEval and the increased output length \cite{alpaca_eval,zhao2024long,jain2024neftune}.
As shown in Figure \ref{neurips2024:fig:tulu_length_alpacaeval}, our result reveals that our approach \im does not necessarily generate longer outputs than \sft across different data utilisation levels from the \tulu dataset. 
Specifically, the output lengths for both approaches are similar despite varying levels of data utilisation.
Furthermore, \im consistently outperforms the \sft, suggesting that improvements in performance as measured by win rates on the AlpacaEval 1.0 are not dependent on the output length.
We provide additional analysis on other instruction tuning datasets under the \sah in Appendix \ref{neurips2024:sec:analysis_output_length_sah}.

\begin{table}[!ht]
\centering
\caption{Performance comparison of \im and \im+\neftune on AlpacaEval 1.0 and various NLP benchmarks. 
Green and red arrows indicate performance changes against the baseline (\im).
This analysis shows that adding \neftune to \im could further improve model performance.
}
\label{neurips2024:table:im_and_neftune}
\resizebox{\textwidth}{!}{
\begin{tabular}{lccccccc}
\toprule
 & \lima & \makecell{\texttt{Less}\\\texttt{Tydiqa}} & \makecell{\texttt{Less}\\\texttt{MMLU Chat}} & \makecell{\texttt{Less}\\\texttt{BBH ICL}} & \makecell{\texttt{Alpagasus}\\\texttt{Alpaca 5k}} & \makecell{\texttt{Alpagasus}\\\texttt{Dolly 9k}} & \makecell{\texttt{Alpagasus}\\\texttt{Dolly 3k}} \\
\midrule
\multicolumn{8}{c}{\textbf AlpacaEval 1.0 Win Rate} \\
\midrule
\textbf{\im}                  & 32.94          & 10.10           & 9.78           & 44.15          & 19.52           & 30.77          & 15.11 \\
\textbf{\im+\neftune}         & 30.77\daa{2.17} & 23.41\uaa{13.31} & 12.45\uaa{2.67} & 48.25\uaa{4.10} & 32.07\uaa{12.55} & 38.28\uaa{7.51} & 23.35\uaa{8.24} \\
\midrule
\multicolumn{8}{c}{\textbf Mean Performance Across 18 NLP Tasks} \\
\midrule
\textbf{\im}                  & 49.60          & 48.70           & 47.84          & 49.15          & 47.47          & 48.00          & 48.95 \\
\textbf{\im+\neftune}         & 49.47\daa{0.13} & 49.44\uaa{0.74}  & 47.73\daa{0.11} & 48.62\daa{0.53} & 48.70\uaa{1.23} & 48.63\uaa{0.63} & 49.54\uaa{0.59} \\
\bottomrule
\end{tabular}
}
\end{table}

\paragraph{\#3. Our proposed method \im could further improve the model performance with \neftune.}
Table \ref{neurips2024:table:im_and_neftune} demonstrates the combined effects of our proposed method \im and \neftune on performance across various \nlp tasks and the AlpacaEval 1.0 benchmark. 
The integration of \neftune with \im generally further improves the win rates in AlpacaEval 1.0, showing notable improvements in several datasets such as a $13.31$\% increase on \tydiqa and a $12.55$\% boost on \alpagasusalpaca (in absolute). 
However, this combination leads to a performance drop in certain contexts, such as a lower performance on \nlp tasks on \mmluchat and \bbhicl. 
This indicates that while \neftune may enhance model robustness under certain conditions, its benefits are context-dependent, highlighting the need for the careful application of \neftune when used in conjunction with \im to optimise effectiveness across diverse evaluation settings.

\label{neurips2024:sec:win_rate_vs_epoch}
\begin{figure}[ht]
\centering
\includegraphics[width=\textwidth]{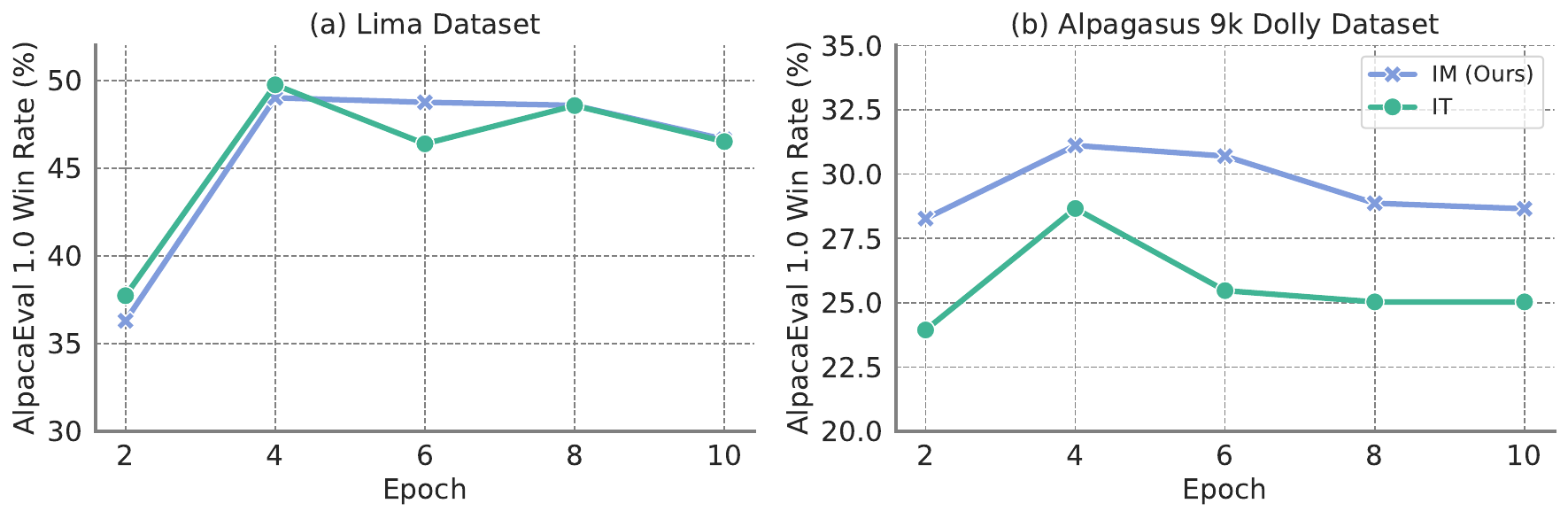}
\caption{
AlpacaEval 1.0 performance trends for \im and \sft approaches on the \lima and \alpagasusdollytwo datasets across different epochs. 
}
\label{neurips2024:fig:win_rate_epoch}
\end{figure}

\paragraph{\#4. The impact of Epochs on the Win Rate.} The figure \ref{neurips2024:fig:win_rate_epoch} illustrates the comparative analysis of AlpacaEval 1.0 scores across different epochs for two datasets, \lima and \alpagasusdollytwo datasets. 
We evaluate the performance of \im and \sft over different numbers of epochs. 
\im consistently surpasses \sft in performance on the \alpagasusdollytwo dataset, while the performance of both approaches is comparable on the \lima dataset.

\paragraph{\#5. Applying KL Divergence Loss for Instruction Tuning.}
\label{neurips2024:sec:kl}
Here we first briefly introduce the Kullback-Leibler (KL) divergence, and then introduce the experimental details. Kullback-Leibler (KL) divergence is commonly employed as a regularisation method in the fine-tuning of LMs, helping to mitigate overfitting by constraining the fine-tuned model to remain similar to the pre-trained model \citep{ouyang2022training}. 
Specifically, the KL divergence is added to the fine-tuning objective as a per-token regularisation term between the fine-tuned LM $\pi_\theta(x)$, and the pre-trained LM, $\pi^\text{pre}(x)$. 
For supervised fine-tuning with next token prediction loss, the training objective incorporating KL divergence is computed as follows:
\begin{align}
    \mathcal{L}_\text{KL}(\theta) &= \mathbb{E}_{x\sim \mathcal{D}} \big[ \sum_{t} -\log \pi_\theta(x_t|x_{0:t-1}) + \lambda \sum_{t} \text{KL}(\pi_\theta(x_t|x_{0:t-1})||\pi^\text{pre}(x_t|x_{0:t-1})) \big],
\end{align}
where $\lambda$ is a regularisation parameter that balances the loss due to the next token prediction and the KL divergence, and $\pi(x_t|x_{0:t-1})$ represents the next token distribution of the fine-tuned or pre-trained LM conditioned on the preceding context.

\begin{table}[!th]
\centering
\caption{Performance on 18 NLP tasks and AlpacaEval 2.0, with various values of $\lambda$, trained on the (\textsc{Llama-2-7B-Base}).
}
\resizebox{0.5\textwidth}{!}{
\begin{tabular}{lcc}
\toprule
\textbf                       & \textbf NLP Tasks      &\textbf AlpacaEval 2.0       \\
\midrule
\textsc{Llama-2-7B-Base}   &  49.32            &  0.01           \\
\midrule
$\lambda=0.01$             &  48.81  & 2.58 \\
$\lambda=0.1$              &  48.77  & 2.44 \\
$\lambda=1.0$              &  49.26  & 0.06 \\
\bottomrule
\end{tabular}
}
\label{neurips2024:table:kl_alpha}
\end{table}
Here we perform ablation study on the effect of $\lambda$.
In Table \ref{neurips2024:table:kl}, we set the value of the $\lambda$ as $1.0$. Here we provide additional experiments with different values of $\lambda$. Table \ref{neurips2024:table:kl_alpha} presents the model performance on the \nlp tasks and AlpacaEval 2.0. This aligns our observations in \sect{neurips2024:sec:overfitting}.

\paragraph{\#6. The impact of Epochs on Output Lengths.}
\label{neurips2024:sec:analysis_output_length_sah}

\begin{figure}[ht]
\centering
\includegraphics[width=\textwidth]{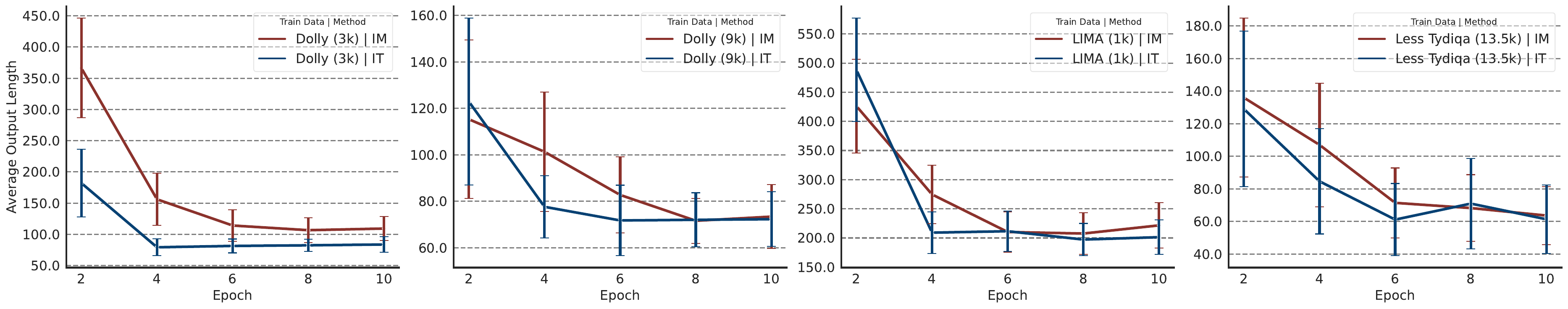}
\caption{
Comparative analysis of output lengths for \im and \sft across different epochs on \alpagasusdollyone, \alpagasusdollytwo, \lima, and \tydiqa datasets.
}
\label{neurips2024:fig:length_sah}
\end{figure}

Figure \ref{neurips2024:fig:length_sah} illustrate the average output length of various models across different epochs. 
We report the output length on four different datasets, including  \alpagasusdollyone, \alpagasusdollytwo, \lima, and \tydiqa.
Each line represents the average output length of a model, with epochs ranging from 2 to 10, and is accompanied by error bars that denote the normalised standard deviation (10\%) of the output lengths. 
Our experimental results show that our approach \im does not consistently increase the output length and that win rates are not necessarily associated with the length of the output. 

\section{Limitations}
% \label{neurips2024:para:limitations}
Here we discuss some potential limitations and the broader impact of our work. 
Several limitations are outlined as follows:
(1) The success of our approach relies on the quality and diversity of the instructions and prompts in the training datasets. Poorly defined or ambiguous instructions may undermine the effectiveness of IM, leading to sub-optimal performance. 
% Future research should explore strategies for optimising prompt design to enhance consistency and reliability in model responses.
(2) It is crucial to ensure that the instructions are ethically sound and free from harmful or biased content. Training on inappropriate or toxic instructions may result in undesirable outputs. 
Previous works \cite{bender-koller-2020-climbing,10.5555/3495724.3495883,10.1145/3442188.3445922} have extensively discussed the risks and potential harms associated with LMs, including the amplification of undesirable biases learned from training data \cite{bender-koller-2020-climbing,austin2021program,carlini2021extracting}.
Our work has the potential to positively impact the community by helping to mitigate overfitting, resulting in models that are more robust and generalise better to new data, especially in low-resource scenarios. This can enhance the reliability and trustworthiness of AI systems in real-world applications.

\section{Summary} 
In conclusion, our study proposes \textsc{Instruction Modelling}, which trains LMs with loss over instructions rather than outputs only. 
Our experimental evaluations demonstrate that our approach largely improves the performance of LMs on both \nlp tasks and open-ended generation benchmarks in some scenarios, especially under the Superficial Alignment Hypothesis and low-resource setting where minimal training data is used for instruction tuning.
Our analysis has shed light on two key factors that influence the effectiveness of our approach, 
(1) the ratio between instruction and output lengths, and
(2) the quantity of training data,
providing practical insights for optimising instruction-based training methods.
Additionally, our analysis reveals the mechanisms behind the effectiveness of \im, particularly its ability to reduce overfitting, showing that applying instruction losses in some scenarios can lead to more robust and adaptable LMs.

\chapter{Benchmarking Complex Reasoning in Language Models}
\label{chapter:downstream_tasks}

\section{Introduction}
Neural networks have been successful in a wide array of perceptual tasks, but it is often stated that they are incapable of solving tasks that require higher-level reasoning~\citep{ding2020object}. %
Since spatial reasoning is ubiquitous in many scenarios such as %
autonomous navigation \citep{vogel2010learning,wang2024investigating}, %
situated dialog~\citep{kruijff2007situated,liu-etal-2024-summequal,10.1007/978-3-031-56027-9_1}, and %
robotic manipulation~\citep{yang2020robust,landsiedel2017review,shi-etal-2023-lexical},
grounding spatial references in texts is essential for effective human-machine communication through natural language. 
Navigation tasks require agents to reason about their relative position to objects and how these relations change as they move through the environment~\cite {chen2019touchdown,shi-etal-2022-learning,shi2023and}. 
If we want to develop conversational systems able to assist users in solving tasks where spatial references are involved, we need to make them able to understand and reason about spatial references in natural language.
Such ability can help conversational systems to follow instructions and understand spatial descriptions successfully. However, despite its tremendous applicability, reasoning over spatial relations remains a challenging task for existing conversational systems.

Earlier works in spatial reasoning focused on spatial instruction understanding in a synthetic environment~\citep{bisk2018learning, tan2018source, janner2018representation,ramos2022condita} or in a simulated world with spatial information annotation in texts~\citep{pustejovsky2015semeval}, spatial relation extractions across entities~\citep{petruck2018representing} and visual observations~\citep{anderson2018vision, chen2019touchdown}. 
However, few of the existing datasets are designed to evaluate models' inference over spatial information in texts. A spatial relational inference task often requires a conversational system to infer the spatial relation between two items given a description of a scene. For example, imagine a user asking to a conversational system to recognize the location of an entity based on the description of other entities in a scene. To do so, the conversational system needs to be able to reason about the location of the various entities in the scene using only textual information.

BAbI~\citep{weston2015towards} is the most relevant dataset for this task. It contains 20 synthetic question-answering (QA) tasks to test a variety of reasoning abilities in texts, 
like deduction, co-reference, and counting. 
In particular, the \textit{positional reasoning} task (no.~17) and the \textit{path finding} task (no.~19) are designed to evaluate models' spatial reasoning ability.
These two tasks are arguably the most challenging ones \citep{van2019does}. 
The state-of-the-art model on the bAbI~\citep{le2020self} dataset almost perfectly solves these two spatial reasoning tasks. 
However, in this chapter, we demonstrate that such good performance is attributable to issues with the bAbI dataset rather than the model inference ability.

We find four major issues with bAbI's tasks 17 and 19: 
(1) 
There is a data leakage between the train and test sets; that is, most of the test set samples appear in the training set. Hence, the evaluation results on the test set cannot truly reflect the models' reasoning ability; 
(2) 
Named entities are fixed and only four relations are considered. Each text sample always contains the same four named entities in the training, validation, and test sets. This further biases the learning models towards these four entities. When named entities in the test set are replaced by unseen entities or the number of such entities increases, the model performance decreases dramatically~\cite {chen2020unseen}. Also, relations such as top-left, top-right, lower-left, and lower-right are not taken into consideration;
(3) 
Learning models are required to reason only over one or two sentences in the text descriptions, making such tasks relatively simple. Previous work \citep{palm2018recurrent} pointed out that multi-hop reasoning is not necessary for the bAbI dataset since models only need a single step to solve all the tasks, and;
(4) It is a synthetic dataset with a limited diversity of spatial relation descriptions. Thus, it cannot truly reveal the models' ability to understand textual space descriptions.

In this chapter, we propose a new dataset called \textit{StepGame} to tackle the above-mentioned issues and a novel \textit{Tensor Product-based Memory-Augmented Neural Network} architecture (TP-MANN) for multi-hop spatial reasoning in texts.  

The StepGame dataset is based on crowdsourced descriptions of 
8 potential spatial relations between 2 entities.
These descriptions are then used as templates when generating the dataset. 
To increase the diversity of these templates, crowd workers were asked to diversify their expressions. 
This was done in order to ensure that the crowdsourced templates cover most of the natural ways relations between two entities can be described in the text.
The StepGame dataset is characterized by a combinatorial growth in the number of possible descriptions of scenes, named \textit{stories}, as the number of described relations between two entities increases. 
This combinatorial growth reduces the chances of leaking stories from the training to the validation and test sets. 
Moreover, we use a large number of named entities and 
require multi-hop reasoning to answer \textit{questions} about two entities mentioned in the stories. % 
Experimental results show that existing models (1) fail to achieve performance on the StepGame dataset similar to that achieved on the bAbI dataset, and (2) suffer from a large performance drop as the number of required reasoning steps increases.

The TP-MANN architecture is based on tensor product representations~\citep{smolensky1990tensor} that are used in a recurrent memory module to store, update or delete the relation information among entities inferred from stories. 
This recurrent architecture provides three key benefits: 
(1) it enables the model to make inferences based on the stored memory; (2) it allows multi-hop reasoning and is robust to noise, and; (3) the number of parameters remains unchanged as the number of recurrent layers in the memory module increases.
Experimental results on the StepGame dataset show that 
our model achieves state-of-the-art performance with a substantial improvement and demonstrates a better generalization ability to more complex stories. 
Finally, we also conduct some analysis of our recurrent structure and demonstrate its importance for multi-hop reasoning.

\paragraph{Focus and Scope.}
Chapter \ref{chapter:downstream_tasks} is dedicated to answering \ref{Q4}, which asks how we can rigorously evaluate the performance and effectiveness of our adapted language models on complex downstream tasks. To that end, this chapter introduces StepGame, a synthetic, multi‑hop spatial‑reasoning benchmark designed to stress‑test relational inference abilities that standard pre‑training and fine‑tuning may overlook. It then evaluates performance against baseline LMs across story length, noise, and hop count, and situates the findings within the context of our earlier adaptation strategies.
By explicitly framing StepGame alongside our earlier adaptation strategies, Chapter \ref{chapter:downstream_tasks} not only delivers a concrete evaluation framework for \ref{Q4} but also closes the loop on the thesis’s central narrative: developing LM adaptations is only meaningful if those adaptations translate into genuine downstream reasoning gains.  

% \section{Related Work and Background}
% \input{Chapters/paper_aaai/related_work}

\section{Background}

The Tensor Product Representation (TPR) is a method to create a vector space embedding of complex symbolic structures by tensor product. 
Such representation can be constructed as follows:
\begin{equation}%\footnotesize
    \boldsymbol{M}=\sum_{i} f_i\otimes o_i= \sum_{i} f_i o_i^{^\top} = \sum_{i} (\boldsymbol{f} \otimes \boldsymbol{r})_{ii},
\end{equation}
where $\boldsymbol{M}$ is the TPR, %
$\boldsymbol{f} = (f_1,\dots,f_n)$ is a set of $n$ filler vectors and %
$\boldsymbol{r} = (o_1,\dots,o_n)$ is a set of $n$ role vectors. 
For each role-filler vector pair, which can be considered as an entity-relation pair, we \textit{bind} (or store) them into $\boldsymbol{M}$ by performing their outer product. 
Then, given an unbinding role vector $u_i$,
associated to the filler vector $f_i$, $f_i$ can be recovered by performing:
\begin{equation}%\footnotesize
    \boldsymbol{M} u_i= \left[ \sum_{i} f_i\otimes o_i \right] u_i = \sum_{i} \alpha_{ij} f_i \propto f_i
\end{equation}
where $\alpha_{ij}\neq0$ if and only if $i=j$. It can be proven that the recovery is perfect if the role vectors are orthogonal to each other.
In our model, TPR-like \textit{binding}  and \textit{unbinding} methods are used to store and retrieve information from and to the TPR $\boldsymbol{M}$, which we will call memory.

\section{The StepGame Dataset}
To design a benchmark dataset that explicitly tests models' spatial reasoning ability and tackle the above-mentioned problems, we build a new dataset named StepGame inspired by the spatial reasoning tasks in the bAbI dataset~\citep{weston2015towards}. %
The StepGame is a contextual QA dataset, where the system is required to interpret a story about several entities expressed in natural language and answer a question about the relative position of two of those entities. %
Although this reasoning task is trivial for humans, equipping current NLU models with such a spatial ability remains still a challenge. 
Also, to increase the complexity of this dataset we model several forms of \textit{distracting noises}. Such noises aim to make the task more difficult and force machine learning models that are trained on this dataset to be more robust in their inference process.  
\begin{figure}[!t]
  \centering
  \includegraphics[width=\textwidth]{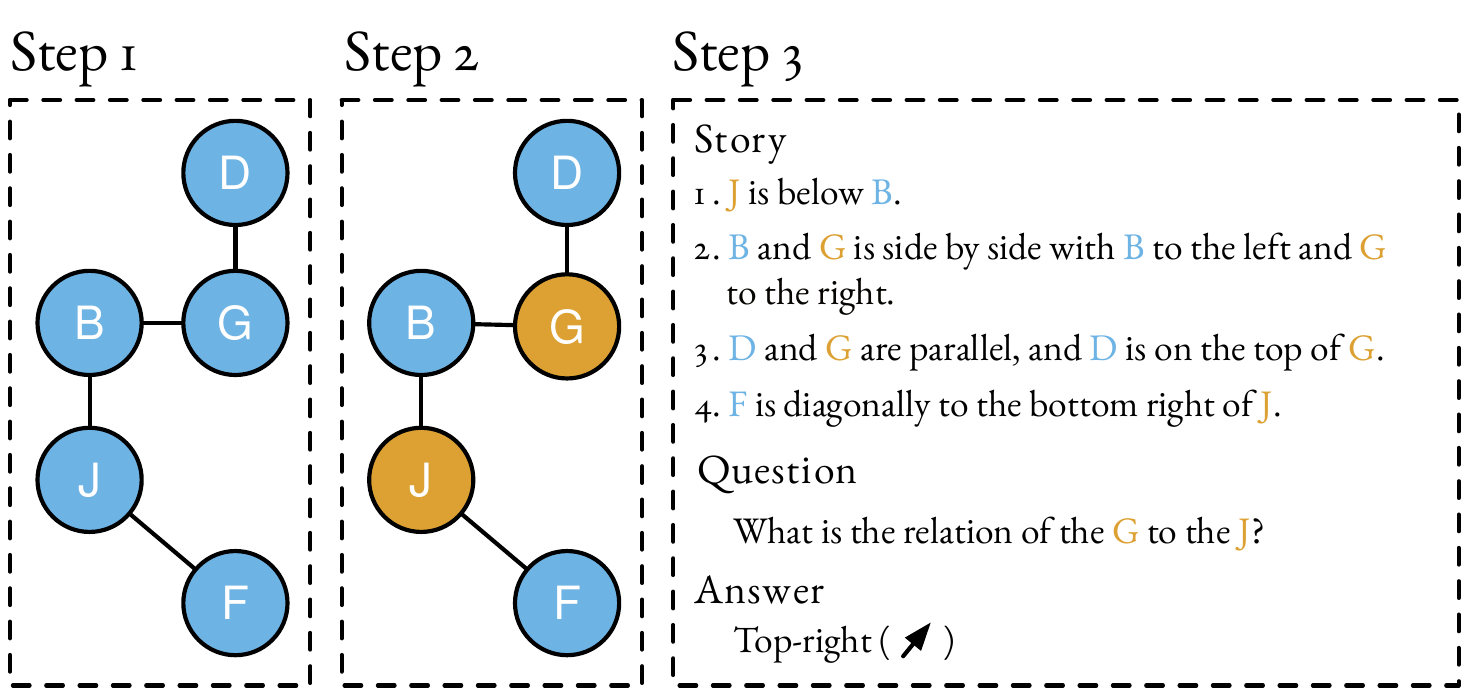}
  \caption{An example of generating a StepGame sample with $k=4$.}
  \label{aaai:generation_step}
\end{figure}

\subsection{Template Collection}

The aim of this crowdsourcing task is to find out all possible ways we can describe the positional relationship between two entities. 
The crowd workers from Amazon Mechanical Turk were provided with an image visually describing the spatial relations of two entities and a request to describe these entities' relations. 
This crowdsourcing task was performed in multiple runs. 
In the first run, we provided crowd workers with an image and two entities (e.g., A and B) and they were asked to describe their positional relation. 
From the data collected in this round, we then manually removed bad answers, and showed the remaining good ones as positive examples to crowdworkers in the next run. However, crowd workers were instructed to avoid repeating them as an answer to our request. We repeated this process until no new templates could be collected. 
In total, after performing a manual generalization where templates discovered for a relation were translated to the other relations, we collected 23 templates for left and right relations, 27 templates for top and down relations, and 26 templates for top-left, top-right, down-left, and down-right relations.

\subsection{Data Generation}
\label{aaai:data_generation}

The task defined by the StepGame dataset is composed of several story-question pairs written in natural language. In its basic form, the story describes a set of $k$ spatial relations among $k+1$ entities, and it is structured as a list of $k$ sentences each talking about $2$ entities. The relations are $k$ and the entities $k+1$ because they define a chain-like shape. The question requests the relative position of two entities among the $k+1$ ones mentioned in the story. To each story-question pair an answer is associated. This answer can take $9$ possible values: \textit{top-left}, \textit{top-right}, \textit{top}, \textit{left}, \textit{overlap}, \textit{right}, \textit{down-left}, \textit{down-right}, and \textit{down}, each representing a relative position.
The number of edges between the two entities in the question ($\leq k$) determines the number of hops a model has to perform in order to get to the correct answer.

To generate a story, we follow three steps, as depicted in Figure~\ref{aaai:generation_step}. Given a value $k$ and a set of entities $\mathcal{E}$:

\paragraph{Step 1.} We generate a sequence of entities by sampling a set of $k+1$ unique entities from $\mathcal{E}$. 
Then, for each pair of entities in the sequence, $k$ spatial relations are sampled. These spatial relations can take any of the 8 possible values: top, down, left, right, top-left, top-right, down-left, and down-right. 
Because the sampling is unconstrained, entities can overlap with each other.
This step results in a sequence of linked entities that from now on we will call a chain. 

\paragraph{Step 2.} Two of the chain's entities are then selected at random to be used in the question.

\paragraph{Step 3.} From the chain generated in Step 1, 
we translate the $k$ relations into $k$ sentence descriptions in natural language. Each description is based on a randomly sampled crowdsourced template. We then shuffle these $k$ sentences to avoid potential distributional biases. These shuffled $k$ sentence descriptions are called a story. From the entities selected in Step 2, we then generate a question also in natural language. Finally, using the chain and the selected entities, we infer the answer to each story-question pair.

\definecolor{wong_1}{rgb}{0.9019, 0.6235, 0}
\definecolor{wong_2}{rgb}{0, 0, 0}
\definecolor{wong_3}{rgb}{0.3372,0.7058,0.9137}
\definecolor{wong_4}{rgb}{0,0.6196,0.4509}

Given this generation process, we can quickly calculate the complexity of the task before using the templates. This is possible because entities can overlap. 
Given $k$ relations, $k+1$ entities sampled from $\mathcal{E}$ in any order (${\color{wong_1}\bullet}$), %
8 possible relations between pairs of entities with 2 ways of describing them (${\color{wong_2}\bullet}$), e.g., A is on the left of B or B is on the right of A, %
a random order of the $k$ sentences in the story (${\color{wong_3}\bullet}$), and %
a question about 2 entities with 2 ways of describing it (${\color{wong_4}\bullet}$), 
the number of examples that we can generate is equal to: 
\begin{equation}
{\color{wong_1}(k+1)!\binom{|\mathcal{E}|}{k+1}} \cdot
{\color{wong_2}16^k}\cdot
{\color{wong_3}\frac{k!}{2}}\cdot
{\color{wong_4}2\binom{k+1}{2}}.
\end{equation}
The complexity of the dataset grows exponentially with $k$. The StepGame dataset uses $|\mathcal{E}| = 26$. For $k=1$ we have 10,400 possible samples, for $k=2$ we have more than 23 million samples, and so on. The sample complexity of the problem guarantees that when generating the dataset the probability of leaking samples from the training set to the test set diminishes with the increase of $k$. Please note that these calculations do not include templates. If we were to consider also the templates, the number of variations of the StepGame would be even larger.

\begin{figure}[!t]
  \centering
  \includegraphics[width=\textwidth]{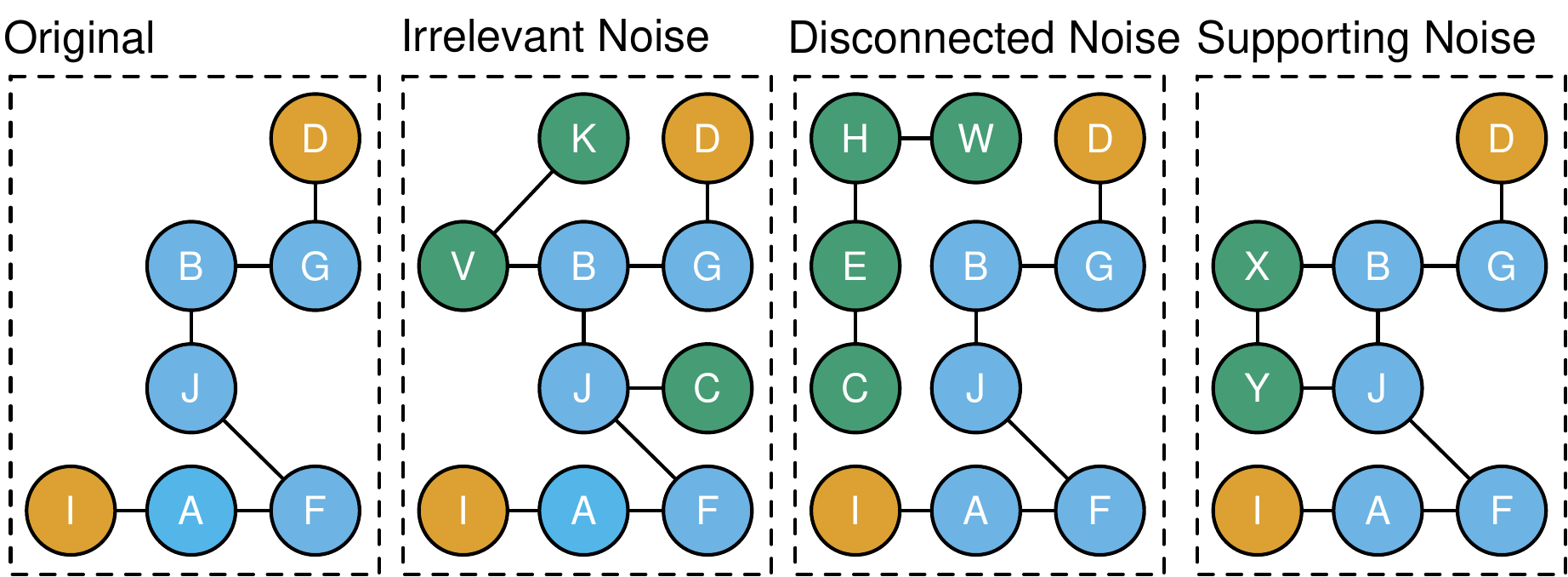}
  \caption{On the left-hand side we have the original chain. Orange entities are those targeted by the question. Besides, we show the same chain with the addition of noise. In green, we represent irrelevant, disconnected and supporting entities.}
  \label{aaai:fig:example_noise}
%   \Description{Examples of noise}
\end{figure}

\subsection{Distracting Noise}
To make the StepGame more challenging we also include noisy examples in the test set.
We assume that when models trained on the non-noisy dataset make mistakes on the noisy test set, these models have failed to learn how to infer spatial relations. 
We generate three kinds of distracting noise: \textit{disconnected}, \textit{irrelevant}, and \textit{supporting}.
Examples of all kinds of noise are provided in Figure~\ref{aaai:fig:example_noise}. 
The irrelevant noise extends the original chain by branching it out with new entities and relations. 
The disconnected noise adds to the original chain, which is a new independent chain with new entities and relations.
The supporting noise adds to the original chain of new entities and relations that may provide alternative reasoning paths. We only add supporting noise into chains with more than 2 entities.
All kinds of noise have no impact on the correct answer.
The type and amount of noise added to each chain are randomly determined. The detailed statistics for each type of distracting noise are provided in the Appendix.

\subsection{Limitations}
StepGame brings welcome rigour to spatial-reasoning evaluation, yet several characteristics restrict how broadly its scores can be interpreted. 
The stories are produced by filling a finite set of 23--27 crowdsourced templates for each spatial relation, so—even after shuffling sentences—the language variety is narrower than in naturally written text; phenomena such as coreference, ellipsis and domain-specific jargon seldom appear, and high accuracy therefore does not automatically imply robustness to real-world prose.

The benchmark’s ontology is intentionally compact: every question must be answered with one of nine 2-D relations (\textit{top-left}, \textit{top-right}, \textit{top}, \textit{left}, \textit{overlap}, \textit{right}, \textit{down-left}, \textit{down-right}, \textit{down}). As a result, competence on distance judgements, containment, 3-D orientation or relative motion cannot be assessed.

Entities are abstractions (single capital letters drawn from a pool of 26) and carry no lexical or commonsense meaning. This design isolates relational reasoning but prevents any examination of how spatial language interacts with semantic knowledge such as object affordances or typical locations.

\section{Tensor-Product based Memory-Augmented Neural Network}
\label{aaai:sec:method}

\begin{figure*}[!t]
  \centering
  \includegraphics[width=\textwidth]{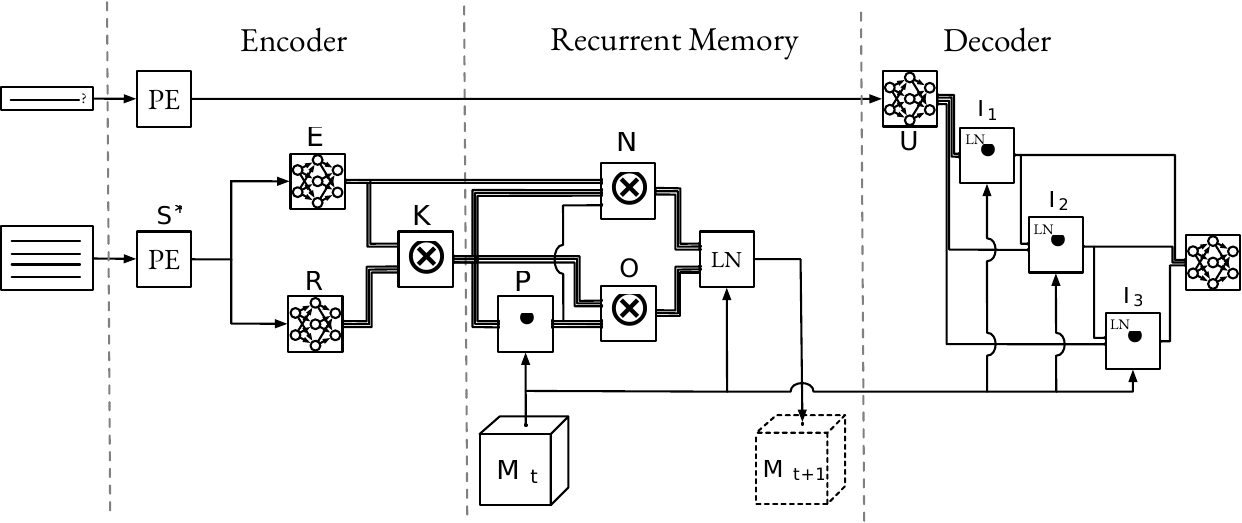}
  \caption{The TP-MANN architecture. PE stands for positional encoder, the sign in the box below the symbol $E$ represents a feed-forward neural network, the $\otimes$ sign represents the outer-product operator, the $\bullet$ sign represents the inner product operator, and LN represents a layer normalization. The $\otimes$, $\bullet$, and LN boxes implement the formulae as presented in Section \ref{aaai:sec:method}. Lines indicate the flow of information. Those without an arrow indicate which symbols are taken as input and are output by their box.}
  \label{aaai:fig:model}
\end{figure*}

In this section, we introduce the proposed TP-MANN model, as shown in Figure~\ref{aaai:fig:model}. %
The model comprises three major components: a question and story encoder, a recurrent memory module, and a relation decoder. %
The encoder learns to represent entities and relations for each sentence in a story. 
The recurrent memory module learns to store entity-relation pair representations in the memory independently. It also updates the entity-relation pair representations based on the current memory and stores the inferred information. 
The decoder learns to represent the question and using the information stored in the memory recurrently infers the spatial relation of the two entities mentioned in the question.

It also has been shown that learned representations in the TPR-like memory could be orthogonal~\citep{schlag2018learning}. We use an example to illustrate the inspiration behind this architecture. A person may experience that when she goes back to her hometown and sees an old tree, her happy childhood memory of playing with her friends under that tree might be recalled. However, this memory may not reminisce unless triggered by the old tree's appearance.
In our model, unbinding vectors in the decoder module play the role of the old tree in the example, where unbinding vectors are learned based on the target questions. 
The decoder module unbinds relevant memories given a question via a recurrent mechanism. 
Moreover, although memories are stored separately, there are integration processes in brains that %at the point of 
retrieving information via a recursive mechanism. This allows episodes in memories to interact with each other 
\citep{kumaran2012generalization, schapiro2017complementary, koster2018big}.

\paragraph{Encoder.}
The input of the encoder is a story and a question.
Given an input story $\text{Story}=(S_1, \dots, S_N)$ with $N$ sentences and a question $q$ both described by words in a vocabulary $\mathcal{V}$. Each sentence $S_i=(s_1, \dots, s_n)$ is mapped to learnable embeddings $(s^*_1, \dots, s^*_n)$. 
Then, a positional encoding (PE) is applied to each word embedding and then averaged together
$s^*_i=\frac{1}{n}\sum_{j=1}^{n} s^*_j \cdot p_j$, where $\{p_1,\dots,p_n\}$ are learnable position vectors, and $\cdot$ is the element-wise product. This operation defines $S^* \in \mathbb{R}^{N \times d}$, where each row of $S^*$ represents an encoded sentence and $d$ is the dimension of a word embedding. For the input question, we convert it to a vector $q^* \in \mathbb{R}^{d}$ in the same way. 
For each sentence of the story in $S^*$, we learn entity and relation representations as:
\begin{align}%\footnotesize
    E_i &= f_{E_i}(S^*),i=1,2, \\
    R_j &= f_{r_j}(S^*),j=1,2,3,
\end{align}
where $f_{e_i}$ are feed-forward neural networks that output entity representations $E_i \in \mathbb{R}^{N \times d_e}$ and 
$f_{r_j}$ are feed-forward neural networks that output relation representations $R_j \in \mathbb{R}^{N \times d_r}$. Finally, we define three search keys $K$ as:
\begin{align}%\footnotesize
    K_1 &= E_1 \otimes R_1, \\
    K_2 &= E_1 \otimes R_2, \\
    K_3 &= E_2 \otimes R_3, 
\end{align}
where $K_1, K_2, K_3 \in \mathbb{R}^{N \times d_e \times d_r}$. Keys will be used to manipulate the memory in the next module and retrieve potential existing associations for each entity-relation pair.

\paragraph{Recurrent Memory Module.} %
To allow stored information to interact with each other, we use a recurrent architecture with $T$ recurrent layers to update the TPR-like memory representation $\boldsymbol{M} \in \mathbb{R}^{d_e \times d_r \times d_e}$, where $\boldsymbol{M}$ contains trainable parameters. 
Through this recurrent architecture, existing episodes stored in memory can interact with new inferences to generate new episodes.
Different from many models such as Transformer~\citep{shi2022attention,vaswani2017attention} and graph-based models~\citep{kipf2016semi,velivckovic2017graph} where adding more layers in the model leads to a larger number of trainable parameters, our model will not increase the number of trainable parameters as the number of recurrent-layers increases.

At each layer $t$, given the keys $K$s, we extract pseudo-entities $P$s for each sentence in $S^*$. 
In the first layer ($t=0$), since there is no previous information existing in memory $\boldsymbol{M}_0$, the model just converts each sentence in $S^*$ as an episode and stores them in it ($\boldsymbol{M}_{1}$).
Then at the later layers ($t>0$), pseudo-entities $P$s build bridges between episodes in the current memory $\boldsymbol{M}_t$ and allow them to interact with potential entity-relation associations.
\begin{equation}%\footnotesize
    P_{jt} = K_j \otimes M_t,  j=1,2,3,
\end{equation}
where $P_{jt} \in \mathbb{R}^{m \times d_e}$.
% replace the previous association.
We then construct the memory episodes that need to be updated or removed. This is done after the first storage at $t=0$ so that all story information is already available in $\boldsymbol{M}$. 
These old episodes, $O_{jt} \in \mathbb{R}^{d_e \times d_r \times d_e}$, will be updated or removed to avoid memory conflicts that may occur when receiving new information:
\begin{equation}%\footnotesize
    O_{jt} = K_j \otimes P_{jt},  j=1,2,3
\end{equation}
Afterwards, new episodes, $N_1, N_{2t}$ and $N_3 \in \mathbb{R}^{d_e \times d_r \times d_e}$, will be added into the memory:
\begin{align}
    N_1 = K_1 \otimes E_2, \\
    N_{2t} = K_2 \otimes P_{1t}, \\
    N_3 = K_3 \otimes E_1. 
\end{align}
Then we apply this change to the memory by removing (subtracting) old episodes and adding up the new ones to the now-dated memory $\boldsymbol{M}_t$:
\begin{multline}
    \boldsymbol{M}_{t+1}=\text{LN}(\boldsymbol{M}_{t}+N_1+N_{2t}+N_3-O_{1t}-O_{2t}-O_{3t}),
\end{multline}
where $LN$ is a layer normalization. 

\paragraph{Decoder.} %
The prediction is computed based on the constructed memory $\boldsymbol{M}$ at the last layer and a question vector $q$. To do this we follow the same procedure designed by \citep{schlag2018learning}:
\begin{align}%\footnotesize
    U_j=f_j^{u}(q), j=1,2,3,4,
\end{align}
where $f_1^{u}$ is a feed-forward neural network that outputs a $d_e$-dimensional unbinding vector, and $f_2^{u},f_3^{u},f_4^{u}$ are feed-forward neural networks that output $d_r$-dimensional unbinding vectors. Then, the information stored in $\boldsymbol{M}$ will be retrieved in a recurrent way based on unbinding vectors learned from the question:
\begin{align}%\footnotesize
    I_1=\text{LN}(\boldsymbol{M}_T\cdot U_1)\cdot U_2, \\%\footnotesize
    I_2=\text{LN}(\boldsymbol{M}_T\cdot I_1)\cdot U_3, \\%\footnotesize
    I_3=\text{LN}(\boldsymbol{M}_T\cdot I_2)\cdot U_4, \\%\footnotesize
    \hat{v}=\text{softmax}(W_o \cdot \sum_{j=1}^{3} I_j).
\end{align}
A linear projection of trainable parameters $W_o \in \mathbb{R}^{\*|V| \times d_e}$ and a softmax function are used to map the extracted information into $\hat{v} \in \mathbb{R}^{\*|V|}$. Hence, the decoder module outputs a probability distribution over the terms of the vocabulary $\*V$.

\section{Experiments and Results}
In this section we aim to address the following research questions:
(\textbf{RQ1}) What is the degree of data leakage in the datasets?
(\textbf{RQ2}) How does our model behave with respect to state-of-the-art NLU models in spatial reasoning tasks? 
(\textbf{RQ3}) How do these models behave when tested on examples more challenging than those used for training? 
(\textbf{RQ4}) What is the effect of the number of recurrent-layers in the recurrent memory module?
Before answering these questions, we first present the material and baselines used in our experiments.

\subsection{Material and Baselines}
In the following experiments, we will use two datasets, the bAbI dataset and the StepGame dataset.
For the bAbI dataset, we only focus on task 17 and task 19 and use the original train and test splits made of $10\,000$ samples for the training set and $1\,000$ for the validation and test sets.
For the StepGame dataset, we generate a training set made of samples varying $k$ from 1 to 5 at steps of 1, and a test set with $k$ varying from 1 to 10. Moreover, the test set will also contain distracting noise. The final dataset consists of, for each $k$ value, $10\,000$ training samples, $1\,000$ validation samples, and $10\,000$ test samples. 

We compare our model against five baselines: 
Recurrent Relational Networks (RRN)~\citep{palm2018recurrent}, %
Relational Network (RN)~\citep{santoro2017simple}, %
TPR-RNN~\citep{schlag2018learning}, % 
Self-attentive Associative Memory (STM)~\citep{le2020self}, and 
Universal Transformer (UT)~\citep{dehghani2018universal}. 
Additionally, we quote the model performance of \textsc{Text-Davinci-003} and GPT-4 from the previous work \cite{Li_Hogg_Cohn_2024}, where 100 examples are used for testing.
Each model is trained and validated on each dataset independently following the hyper-parameter ranges and procedures provided in their original papers.
All training details, including those for our model, are reported in the Appendix.

\subsection{Training-Test Leakage}

To answer \textbf{RQ1} 
we have calculated the degree of data leakage present in bAbI and the StepGame datasets.
For task 17, we counted how many samples in the test set appear in the training set: 23.2\% of the test samples are also in the training set.
For task 19, for each sample we extracted the relevant sentences in the stories (i.e., those sentences necessary to answer the question correctly) and questions. Then we counted how many such pairs in the test set appear in the training set: 
80.2\% of the pairs overlap with pairs in the training set.
For the StepGame dataset, for each sample, we extracted the sentences in the stories and questions. The sentences in the story are sorted in lexicographical order. Then we counted how many such pairs in the test set appeared also in the training set before adding distracting noise and using the templates: 
1.09\% of the pairs overlap with triples in the training set. 
However, such overlap is all produced by the samples with $k=1$, which due to their limited number have a higher chance of being included in the test set.
If we remove those examples, the overlap between training and test sets drops to 0\%.

\subsection{Spatial Inference}
\label{aaai:sec:spatial_inference}

To answer \textbf{RQ2} and judge the spatial inference ability of our model and the baselines we train them on the bAbI and the StepGame datasets and compare them by measuring their test accuracy.

\begin{table}[!t]
\centering
\small
\begin{adjustbox}{max width=\textwidth}
\begin{tabular}{p{2.3cm}ccc}
\toprule
\multicolumn{1}{c}{} & Task 17 & Task 19 & Mean \\ \hline
RN    & 97.33$\pm$1.55 & 98.63$\pm$1.79 & 97.98 \\
RRN & 97.80$\pm$2.34 & 49.80$\pm$5.76 & 73.80 \\
STM & 97.80$\pm$1.06 & 99.98$\pm$0.05 & 98.89 \\
UT & 98.60$\pm$3.40 & 93.90$\pm$7.30 & 96.25 \\ 
TPR-RNN & 97.55$\pm$1.99 & 99.95$\pm$0.06 & 98.75 \\
\hline
Ours & \textbf{99.88$\pm$0.10} & \textbf{99.98$\pm$0.04} & \textbf{99.93} \\ 
\bottomrule
\end{tabular}
\end{adjustbox}
\caption{Test accuracy on the task 17 and 19 of the bAbI dataset: Mean$\pm$Std over 5 runs.}
\label{aaai:tbl:babi_17_19}
\end{table}

In Table \ref{aaai:tbl:babi_17_19} we present the results of our model and the baselines on the tasks 17 and 19 of the bAbI dataset. 
The performance of our model is slightly better than the best baseline. However, due to the issues of the bAbI dataset, these results are not enough to firmly answer RQ2. 

\begin{table*}[!th]
\small
\centering
\begin{adjustbox}{max width=\textwidth}
\begin{tabular}{lrrrrrr}
\toprule
Model & \multicolumn{1}{c}{$k$=1} & \multicolumn{1}{c}{$k$=2} & \multicolumn{1}{c}{$k$=3} & \multicolumn{1}{c}{$k$=4} & \multicolumn{1}{c}{$k$=5} & Mean \\ \hline
\rowcolor{Gray} \multicolumn{7}{l}{\textit{*Model Training}} \\
RN~\cite{santoro2017simple}          & 22.64$\pm$0.25 & 17.08$\pm$1.41 & 15.08$\pm$2.58 & 12.84$\pm$2.27 & 11.52$\pm$1.73 & 15.83 \\
RRN~\cite{palm2018recurrent}         & 24.05$\pm$4.48 & 19.98$\pm$4.68 & 16.03$\pm$2.89 & 13.22$\pm$2.51 & 12.31$\pm$2.16 & 17.12 \\
UT~\cite{dehghani2018universal}      & 45.11$\pm$4.16 & 28.36$\pm$4.50 & 17.41$\pm$2.18 & 14.07$\pm$2.87 & 13.45$\pm$1.35 & 23.68 \\
STM~\cite{le2020self}                & 53.42$\pm$3.73 & 35.96$\pm$4.45 & 23.03$\pm$1.83 & 18.45$\pm$1.87 & 15.14$\pm$1.56 & 29.20 \\
TPR-RNN~\cite{schlag2018learning}    & 70.29$\pm$3.03 & 46.03$\pm$2.24 & 36.14$\pm$2.66 & 26.82$\pm$2.64 & 24.77$\pm$2.75 & 40.81 \\ 
Ours                         & \textbf{85.77$\pm$3.18}  & \textbf{60.31$\pm$2.23} & \textbf{50.18$\pm$2.65} & \textbf{37.45$\pm$4.21} & \textbf{31.25$\pm$3.38} & \textbf{52.99} \\
\hline
\rowcolor{Gray} \multicolumn{7}{l}{\textit{*Model Prompting}} \\ 
Base Prompt (d3) \cite{Li_Hogg_Cohn_2024}             & 77 & 42 & 21 & 26 & 25 & 38.2\\
CoT Prompt (d3) \cite{Li_Hogg_Cohn_2024}                  & - & 48 & 53 & 46 & 46 & 48.3\\
Base Prompt (GPT-4) \cite{Li_Hogg_Cohn_2024}             & 100 & 70 &  55 & 45 & 40 & 62.0\\
CoT Prompt (GPT-4) \cite{Li_Hogg_Cohn_2024}             & - & 85 &  85 & 90 & 90 & 87.5\\
\bottomrule
\end{tabular}
\end{adjustbox}
\caption{
Test accuracy on the StepGame dataset: Mean$\pm$Std over 5 runs.
d3 represents \textsc{text-davinci-003}
}
\label{aaai:tbls:step_game_hard}
\vspace{0.5em}
\end{table*}

In Table~\ref{aaai:tbls:step_game_hard} we present the results for the StepGame dataset. 
In this dataset, the training set is without noise but the test set is with distracting noise.
In the table, we break down the performance of the trained models across $k$. In the last column, we report the average performance across $k$.
Our model outperforms all the baseline models. %
Compared to Table~\ref{aaai:tbl:babi_17_19}, the decreased accuracy in Table~\ref{aaai:tbls:step_game_hard} demonstrates the difficulty of spatial reasoning with distracting noise. %
It is not surprising that the performance of all five baseline models decreases when $k$ increases, \ie, when the number of required inference hops increases. 
In Table \ref{aaai:tbls:step_game_hard}, we also incorporate how recent LLMs perform via the prompting.
In the Appendix, we also report test accuracy on test sets without distracting noise.

\begin{table*}[!th]
\small
\centering
\begin{adjustbox}{max width=\textwidth}
\begin{tabular}{lrrrrrr}
\toprule
Model & \multicolumn{1}{c}{$k=6$} & \multicolumn{1}{c}{$k$=7} & \multicolumn{1}{c}{$k$=8} & \multicolumn{1}{c}{$k$=9} & \multicolumn{1}{c}{$k$=10} & Mean \\ 
\hline
\rowcolor{Gray} \multicolumn{7}{l}{\textit{*Model Training}} \\
RN~\cite{santoro2017simple}       & 11.12$\pm$0.96 & 11.53$\pm$0.70 & 11.21$\pm$0.98 & 11.13$\pm$1.00 & 11.34$\pm$0.87 & 11.27 \\
RRN~\cite{palm2018recurrent}      & 11.62$\pm$0.80 & 11.40$\pm$0.76 & 11.83$\pm$0.75 & 11.22$\pm$0.86 & 11.69$\pm$1.40 & 11.56 \\
UT~\cite{dehghani2018universal}   & 12.73$\pm$2.37 & 12.11$\pm$1.52 & 11.40$\pm$0.92 & 11.41$\pm$0.96 & 11.74$\pm$1.07 & 11.88 \\
STM~\cite{le2020self}             & 13.80$\pm$1.95 & 12.63$\pm$1.69 & 11.54$\pm$1.61 & 11.30$\pm$1.13 & 11.77$\pm$0.93 & 12.21 \\
TPR-RNN~\cite{schlag2018learning} & 22.25$\pm$3.12 & 19.88$\pm$2.80 & 15.45$\pm$2.98 & 13.01$\pm$2.28 & 12.65$\pm$2.71 & 16.65 \\ 
Ours                              & \textbf{28.53$\pm$3.59}    & \textbf{26.45$\pm$2.95} & \textbf{23.67$\pm$2.78} & \textbf{22.52$\pm$2.36} & \textbf{21.46$\pm$1.72} & \textbf{24.53} \\ 
\hline
\rowcolor{Gray} \multicolumn{7}{l}{\textit{*Model Prompting}} \\ 
Base Prompt (d3) \cite{Li_Hogg_Cohn_2024}             & 30 & 23 & 23 & 22 & 22 & 24.0\\
CoT Prompt (d3) \cite{Li_Hogg_Cohn_2024}                  & 48 & 40 & 45 & 41 & 32 & 41.2\\
Base Prompt (GPT-4) \cite{Li_Hogg_Cohn_2024}             & 25 & 40 &  35 & 35 & 25 & 32.0 \\
CoT Prompt (GPT-4) \cite{Li_Hogg_Cohn_2024}             & 85 & 90 &  80 & 60 & 65 & 76.0\\
\bottomrule
\end{tabular}
\end{adjustbox}
\caption{Test accuracy on StepGame for larger $k$s (only on the test set). Mean$\pm$Std over 5 runs.
d3 represents \textsc{Text-Davinci-003}.
}
\label{aaai:tbl:step_game_hard_extra}
\end{table*}
\subsection{Systematic Generalization}
To answer \textbf{RQ3} 
we generate new StepGame test sets with $k \in \{6,7,8,9,10\}$ with distracting noise. 
We then test all the models jointly trained on the StepGame train set with $k \in \{1,2,3,4,5\}$ as in the Section~\ref{aaai:sec:spatial_inference}. We can consider this experiment as a zero-shot learning setting for larger $k$s. 

In Table~\ref{aaai:tbl:step_game_hard_extra} we present the performance of different models on this generalization task. Not surprisingly, the performance of all models degrades monotonically as we increase $k$. RN, RRN, UT and SAM fail to generalize to the test sets with higher $k$ values, while our model is more robust and outperforms the baseline models by a large margin. This demonstrates the better generalization ability of our model, which performs well on longer stories never seen during training. 

\subsection{Inference Analysis}

\begin{figure}[!t]
\centering
% \begin{adjustbox}{max width=\textwidth}
    \begin{subfigure}
        \centering
        \includegraphics[width=.407\linewidth]{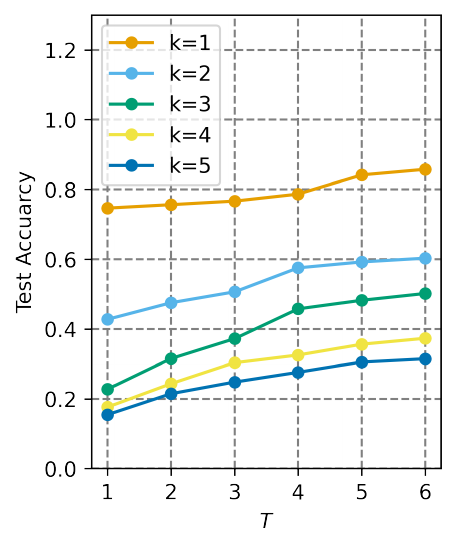}
        % \label{aaai:fig:sub1}
    \end{subfigure}
    % \hfill
    \begin{subfigure}
        \centering
        \includegraphics[width=.422\linewidth]{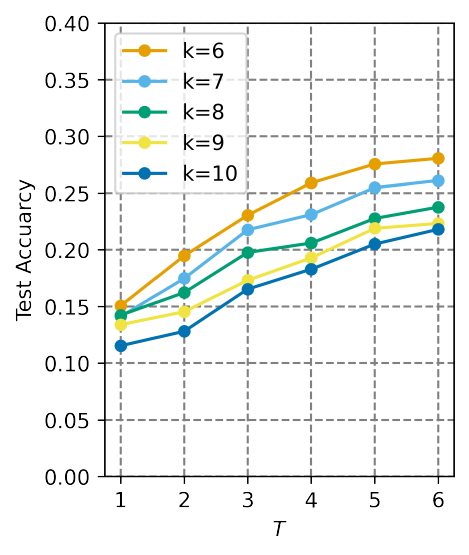}
        % \label{aaai:fig:sub2}
    \end{subfigure}
% \end{adjustbox}
\caption{Analysis of TP-MANN's number of recurrent layers ($T$). The x-axis is $T$ with which the model has been trained. Each line represents a different value of $k$ of the StepGame dataset.}
\label{aaai:fig:inference_analysis}
\end{figure}

To answer \textbf{RQ4}, 
we conduct an analysis of the hyper-parameter $T$, the number of recurrent layers in our model. 
We jointly train TP-MANN on the StepGame dataset with $k$ between 1 and 5 with the number of $T$ between 1 and 6 and report the breakdown test accuracy for each value of $k$. 
These results are shown in the left-hand side figure of Figure~\ref{aaai:fig:inference_analysis}. 
The test sets with higher $k$ benefit more from a higher number of recurrent layers than those with lower $k$, indicating that recurrent layers are critical for multi-hop reasoning.
We also analyze how the recurrent layer structure affects systematic generalization. 
To do this we also test on a StepGame test set with $k$ between 6 and 10 with noise. These $k$s are larger than the largest $k$ used during training.
These results are shown in the right-hand side figure in Figure~\ref{aaai:fig:inference_analysis}.
Here we see that as $T$ increases, the performance of the model improves.
This analysis further corroborates that our recurrent structure supports multi-hop inference. 
It is worth noting, that the number of trainable parameters in our model remains unchanged as $T$ increases. 
Interestingly, we find that the number of recurrent layers needed to solve the task is less than the length of the stories $k$ suggesting that the inference process may happen in parallel.

\section{Summary}
In this paper, we proposed a new dataset named StepGame that requires robust multi-hop spatial reasoning ability to be solved and mitigates the issues observed in the bAbI dataset. The StepGame dataset features greater complexity through its combinatorial growth in possible scene descriptions, diverse spatial relations, and various forms of distracting noise that test models' robustness. We introduced TP-MANN, a tensor product-based memory-augmented neural network architecture. Our experimental results demonstrate that TP-MANN significantly outperforms existing baseline models, particularly as the complexity of spatial reasoning increases. Further analysis also demonstrated the importance of a recurrent memory module for multi-hop reasoning, showing that performance improves with additional recurrent layers without increasing model parameters. The ability to handle complex spatial relationships while maintaining computational efficiency makes TP-MANN particularly promising for real-world applications requiring spatial reasoning capabilities. Most importantly, our work highlights the continuing challenge of developing AI systems that can perform robust spatial reasoning in natural language contexts, suggesting that this remains an important area for future research.
\chapter{Conclusion}
\label{chapterlabel4}

This thesis has made several important contributions to improving how LMs can be adapted for downstream tasks through post-training methods. 
Each of the core chapters addressed key challenges in making LMs more robust, efficient, and capable of following instructions.
In Chapter \ref{chapter:continued_pretraining}, we demonstrated that continued pre-training can serve as a powerful semi-supervised learning method \citep{shi-etal-2023-rethinking}. Our analysis revealed that task-adaptive pre-training often outperforms more sophisticated self-training approaches, especially when dealing with domain shifts or limited unlabelled data. We then introduced Prompt-based Continued Pre-training (PCP), which improves upon conventional continued pre-training by incorporating both task-related texts and prompt templates \citep{shi_dont_2023}. PCP showed consistent improvements over existing methods across multiple tasks and settings, even with only hundreds of unlabelled examples.
In Chapter \ref{chapter:peft}, we tackled the efficiency challenges of prompt tuning by proposing Decomposed Prompt Tuning (DePT) \citep{shi2023dept}. This method decomposes trainable soft prompts into shorter prompts and low-rank matrices, significantly reducing computational costs while maintaining or improving performance. Our experiments demonstrated that DePT becomes increasingly efficient as the model size grows, making it particularly well-suited for large LMs.
Chapter \ref{chapter:sft} introduced Instruction Modelling (IM) \citep{shi2024instruction}, a novel approach that trains LMs by applying loss computation to both instructions and outputs. This method showed substantial improvements in both traditional NLP tasks and open-ended generation benchmarks. Our analysis revealed that IM is particularly effective for datasets with long instructions paired with brief outputs, or in low-resource settings.
Finally, in Chapter \ref{chapter:downstream_tasks}, we developed novel benchmarks and evaluation methods through the StepGame dataset \citep{stepGame2022shi}, designed specifically to test robust multi-hop spatial reasoning in texts. This contribution highlighted significant limitations in current LMs' spatial reasoning abilities and provided a framework for evaluating progress in this area.

\section{Limitations}  
While each preceding chapter closed with a short reflection on method-specific caveats, it is important to examine, in aggregate, the broader limitations that cut across all of the contributions of this thesis. Below, we elaborate on six inter‑connected dimensions to give a more nuanced account of the boundaries within which our findings should be interpreted.

\paragraph{1. Scientific Scope and Generalisability.}
\begin{itemize}
  \item \textbf{Architectural breadth.} Our empirical studies were conducted on encoder‑only, encoder--decoder, and \emph{decoder‑only} transformers, including Llama‑family models up to 13\,B parameters. While this range is diverse, it may not faithfully predict the behaviour of frontier‑scale decoder‑only LLMs (\eg 70\,B--1\,T parameters) or emerging Mixture‑of‑Experts designs. Methodological insights such as PCP or DePT therefore require careful re‑validation at larger scales and on alternative architectures such as retrieval‑augmented or sparse models.
  \item \textbf{Task diversity.} Experiments focus on text‑only benchmarks (classification, QA, reasoning) in English. Performance and stability on generation‑heavy, multilingual or multimodal workloads (\eg long‑form dialogue, code synthesis, or vision-language tasks) remain untested.
  \item \textbf{Domain coverage.} Continued‑pre‑training corpora were sampled from publicly available datasets that, while useful, under‑represent domains with strict privacy requirements (health, law) or low‑resource languages. The applicability of our techniques to those domains is uncertain.
\end{itemize}

\paragraph{2. Technical Robustness \& Safety.}
\begin{itemize}
  \item \textbf{Interpretability gap.} Efficiency‑oriented methods such as DePT compress fine‑tuning signals but offer no additional transparency into model internals. Failure modes (\eg spurious correlations, shortcut learning) can therefore remain hidden.
  \item \textbf{Out‑of‑distribution stability.} Robustness was assessed only under moderate domain shift; resistance to adversarial prompts, prompt‑injection attacks, or extreme distributional drift was not systematically evaluated.
  \item \textbf{Security \& privacy.} No dedicated analysis was performed on memorisation of sensitive data or susceptibility to data‑exfiltration attacks—an increasingly salient concern for real‑world LLM deployment.
\end{itemize}

\paragraph{3. Environmental Sustainability.}
Although PCP and DePT reduce \emph{relative} resource usage, absolute energy consumption for pre‑training and fine‑tuning remains substantial. The experiments consumed \textasciitilde{}125\,kWh and 70\,kg\,CO$_2$e, but scaled industrial adoption would multiply that footprint. Life‑cycle assessments, including hardware manufacture and end‑of‑life, were outside the scope of this thesis.

\paragraph{4. Ethical and Societal Considerations.}
Despite improvements in efficiency and instruction‑following, our methods do not inherently reduce social biases present in pre‑training data; indeed, greater adaptation efficiency may propagate such biases more rapidly. Moreover, enhanced instruction‑following could be misused for harmful content generation or persuasive manipulation if safety alignment fails to keep pace, and the absence of structured human‑in‑the‑loop oversight during deployment leaves a responsibility gap between offline benchmarking and real‑time operation.

\paragraph{5. Evaluation Coverage and Metric Fidelity.}
Our evaluation leans heavily on leaderboard‑centric datasets such as GLUE, SuperGLUE, and AlpacaEval, many of which are nearing saturation and risk over‑fitting research incentives; they also omit human‑centric metrics like trust, cognitive load, and long‑term utility, while synthetic datasets such as StepGame—though useful for controlled spatial‑reasoning tests—may not generalise to complex naturalistic narratives.

\paragraph{6. Reproducibility, Hyper‑parameter Sensitivity \& Scalability.}
Methods like DePT entail a two‑tier learning‑rate schedule whose optimal values vary with model size and task, demanding computationally intensive hyper‑parameter searches that may be prohibitive for practitioners with limited compute; furthermore, experiments were run on A100‑class GPUs with $>$40\,GB memory, limiting applicability to edge settings, and although all code is released, corpus licensing constraints and storage limits on full training logs hinder complete reproducibility.

Recognising these limitations clarifies where caution is warranted when extrapolating our conclusions and highlights key directions for future work. Addressing these gaps will require \emph{inter-disciplinary collaboration}, integrating advances in scalable optimisation, robust alignment, green AI, and human-centred evaluation, to create language-model systems that are truly dependable and societally beneficial.

\section{Open Challenges and Future Work}
The development and deployment of LMs for specific downstream applications reveals several critical challenges that bridge research and engineering concerns.

\paragraph{Task-Specific Data Generation at Scale presents a fundamental challenge.} While recent work shows promise in using LMs to generate training data, creating high-quality datasets for specific downstream tasks remains difficult. This challenge manifests in several ways. 
First, automatically evaluating the quality of generated data requires reliable metrics that can capture both surface-level correctness and deeper semantic properties. 
Second, ensuring diversity in generated examples - particularly covering edge cases and rare scenarios - requires sophisticated sampling strategies. Current approaches like chain-of-thought generation and self-consistency checking provide some quality control, but may miss subtle domain-specific requirements. 
The challenge extends to personalisation: how can we efficiently adapt data generation to specific user preferences or business requirements while maintaining quality? 
This requires frameworks that can take natural language specifications of requirements and automatically translate them into data generation strategies.

\paragraph{Building Effective Agentic Frameworks requires solving several interrelated challenges.} 
While current LMs demonstrate impressive capabilities in tool use and reasoning, developing reliable agent architectures remains complex. 
Key challenges include: designing robust planning mechanisms that can decompose tasks appropriately, implementing reliable error recovery when tools fail or return unexpected results, and managing context effectively over long interactions. The recent emergence of workflow patterns like orchestrator-worker architectures and evaluator-optimiser loops shows promise, but integrating these patterns into production systems while maintaining reliability and performance remains challenging.

\paragraph{Engineering for Real-World Deployment presents practical challenges in adapting powerful but general-purpose LMs to specific business needs.} This includes:
\begin{itemize}
    \item \textbf{Requirements Discovery.} Better methods are needed for efficiently mapping business requirements to LM capabilities. This involves not just understanding what's technically possible, but also identifying where LMs can provide the most value in existing workflows.
    \item \textbf{Domain Customisation.} While LMs have broad knowledge, adapting them efficiently to specific domains requires solving challenges around knowledge integration and fine-tuning strategies. Current approaches often rely on extensive prompt engineering or full model fine-tuning, neither of which scales well.
    \item \textbf{System Integration.} Integrating LMs into existing systems requires solving challenges around API design, latency management, and cost optimisation. This includes developing better patterns for caching, batching, and failover that are specific to LM-based systems.
\end{itemize}

These challenges are amplified by the rapid pace of model development. Organisations need frameworks that allow them to efficiently migrate between model versions while maintaining customisations and integrations. This suggests the need for better abstraction layers and testing frameworks specifically designed for LM-based systems.

Addressing these challenges requires advances in both technical capabilities and engineering best practices. Success will likely come from combining theoretical advances in areas like prompt optimisation and data generation with practical patterns for building reliable, maintainable systems.

% This line manually adds the Bibliography to the table of contents.
% The fact that \include is the last thing before this ensures that it
% is on a clear page, and adding it like this means that it doesn't
% get a chapter or appendix number.

\clearpage
\phantomsection
\addcontentsline{toc}{chapter}{Bibliography}
\bibliographystyle{plain}
\bibliography{main}

\begin{thebibliography}{100}

\bibitem{achiam2023gpt}
Josh Achiam, Steven Adler, Sandhini Agarwal, Lama Ahmad, Ilge Akkaya, Florencia~Leoni Aleman, Diogo Almeida, Janko Altenschmidt, Sam Altman, Shyamal Anadkat, et~al.
\newblock Gpt-4 technical report.
\newblock {\em arXiv preprint arXiv:2303.08774}, 2023.

\bibitem{adler2024nemotron}
Bo~Adler, Niket Agarwal, Ashwath Aithal, Dong~H Anh, Pallab Bhattacharya, Annika Brundyn, Jared Casper, Bryan Catanzaro, Sharon Clay, Jonathan Cohen, et~al.
\newblock Nemotron-4 340b technical report.
\newblock {\em arXiv preprint arXiv:2406.11704}, 2024.

\bibitem{alsentzer-etal-2019-publicly}
Emily Alsentzer, John Murphy, William Boag, Wei-Hung Weng, Di~Jindi, Tristan Naumann, and Matthew McDermott.
\newblock Publicly available clinical {BERT} embeddings.
\newblock In {\em Proceedings of the 2nd Clinical Natural Language Processing Workshop}, pages 72--78, Minneapolis, Minnesota, USA, 2019. Association for Computational Linguistics.

\bibitem{anderson2018vision}
Peter Anderson, Qi~Wu, Damien Teney, Jake Bruce, Mark Johnson, Niko S{\"{u}}nderhauf, Ian~D. Reid, Stephen Gould, and Anton van~den Hengel.
\newblock Vision-and-language navigation: Interpreting visually-grounded navigation instructions in real environments.
\newblock In {\em 2018 {IEEE} Conference on Computer Vision and Pattern Recognition, {CVPR} 2018, Salt Lake City, UT, USA, June 18-22, 2018}, pages 3674--3683. {IEEE} Computer Society, 2018.

\bibitem{arazo2020pseudo}
Eric Arazo, Diego Ortego, Paul Albert, Noel~E O’Connor, and Kevin McGuinness.
\newblock Pseudo-labeling and confirmation bias in deep semi-supervised learning.
\newblock In {\em ArXiv preprint}, volume abs/1908.02983, 2019.

\bibitem{aribandi2022ext}
Vamsi Aribandi, Yi~Tay, Tal Schuster, Jinfeng Rao, Huaixiu~Steven Zheng, Sanket~Vaibhav Mehta, Honglei Zhuang, Vinh~Q. Tran, Dara Bahri, Jianmo Ni, Jai~Prakash Gupta, Kai Hui, Sebastian Ruder, and Donald Metzler.
\newblock Ext5: Towards extreme multi-task scaling for transfer learning.
\newblock In {\em The Tenth International Conference on Learning Representations, {ICLR} 2022, Virtual Event, April 25-29, 2022}. OpenReview.net, 2022.

\bibitem{artetxe-etal-2018-robust}
Mikel Artetxe, Gorka Labaka, and Eneko Agirre.
\newblock A robust self-learning method for fully unsupervised cross-lingual mappings of word embeddings.
\newblock In {\em Proceedings of the 56th Annual Meeting of the Association for Computational Linguistics (Volume 1: Long Papers)}, pages 789--798, Melbourne, Australia, 2018. Association for Computational Linguistics.

\bibitem{asai-etal-2022-attempt}
Akari Asai, Mohammadreza Salehi, Matthew Peters, and Hannaneh Hajishirzi.
\newblock {ATTEMPT}: Parameter-efficient multi-task tuning via attentional mixtures of soft prompts.
\newblock In {\em Proceedings of the 2022 Conference on Empirical Methods in Natural Language Processing}, pages 6655--6672, Abu Dhabi, United Arab Emirates, 2022. Association for Computational Linguistics.

\bibitem{austin2021program}
Jacob Austin, Augustus Odena, Maxwell Nye, Maarten Bosma, Henryk Michalewski, David Dohan, Ellen Jiang, Carrie Cai, Michael Terry, Quoc Le, et~al.
\newblock Program synthesis with large language models.
\newblock {\em ArXiv preprint}, abs/2108.07732, 2021.

\bibitem{azar2024general}
Mohammad~Gheshlaghi Azar, Zhaohan~Daniel Guo, Bilal Piot, Remi Munos, Mark Rowland, Michal Valko, and Daniele Calandriello.
\newblock A general theoretical paradigm to understand learning from human preferences.
\newblock In {\em International Conference on Artificial Intelligence and Statistics}, pages 4447--4455. PMLR, 2024.

\bibitem{bai2023qwen}
Jinze Bai, Shuai Bai, Yunfei Chu, Zeyu Cui, Kai Dang, Xiaodong Deng, Yang Fan, Wenbin Ge, Yu~Han, Fei Huang, et~al.
\newblock Qwen technical report.
\newblock {\em arXiv preprint arXiv:2309.16609}, 2023.

\bibitem{bai2022training}
Yuntao Bai, Andy Jones, Kamal Ndousse, Amanda Askell, Anna Chen, Nova DasSarma, Dawn Drain, Stanislav Fort, Deep Ganguli, Tom Henighan, et~al.
\newblock Training a helpful and harmless assistant with reinforcement learning from human feedback.
\newblock {\em ArXiv preprint}, abs/2204.05862, 2022.

\bibitem{doi:10.1073/pnas.2107022118}
Vijay Balasubramanian.
\newblock Brain power.
\newblock {\em Proceedings of the National Academy of Sciences}, 118(32):e2107022118, 2021.

\bibitem{beltagy-etal-2019-scibert}
Iz~Beltagy, Kyle Lo, and Arman Cohan.
\newblock {S}ci{BERT}: A pretrained language model for scientific text.
\newblock In {\em Proceedings of the 2019 Conference on Empirical Methods in Natural Language Processing and the 9th International Joint Conference on Natural Language Processing (EMNLP-IJCNLP)}, pages 3615--3620, Hong Kong, China, 2019. Association for Computational Linguistics.

\bibitem{zaken2021bitfit}
Elad Ben~Zaken, Yoav Goldberg, and Shauli Ravfogel.
\newblock {B}it{F}it: Simple parameter-efficient fine-tuning for transformer-based masked language-models.
\newblock In {\em Proceedings of the 60th Annual Meeting of the Association for Computational Linguistics (Volume 2: Short Papers)}, pages 1--9, Dublin, Ireland, 2022. Association for Computational Linguistics.

\bibitem{10.1145/3442188.3445922}
Emily~M. Bender, Timnit Gebru, Angelina McMillan-Major, and Shmargaret Shmitchell.
\newblock On the dangers of stochastic parrots: Can language models be too big?
\newblock In {\em Proceedings of the 2021 ACM Conference on Fairness, Accountability, and Transparency}, FAccT '21, page 610–623, New York, NY, USA, 2021. Association for Computing Machinery.

\bibitem{bender-koller-2020-climbing}
Emily~M. Bender and Alexander Koller.
\newblock Climbing towards {NLU}: {On} meaning, form, and understanding in the age of data.
\newblock In {\em Proceedings of the 58th Annual Meeting of the Association for Computational Linguistics}, pages 5185--5198, Online, 2020. Association for Computational Linguistics.

\bibitem{bengio2009curriculum}
Yoshua Bengio, J{\'{e}}r{\^{o}}me Louradour, Ronan Collobert, and Jason Weston.
\newblock Curriculum learning.
\newblock In Andrea~Pohoreckyj Danyluk, L{\'{e}}on Bottou, and Michael~L. Littman, editors, {\em Proceedings of the 26th Annual International Conference on Machine Learning, {ICML} 2009, Montreal, Quebec, Canada, June 14-18, 2009}, volume 382 of {\em {ACM} International Conference Proceeding Series}, pages 41--48. {ACM}, 2009.

\bibitem{bentivogli2009fifth_rte4}
Luisa Bentivogli, Peter Clark, Ido Dagan, and Danilo Giampiccolo.
\newblock The fifth {PASCAL} recognizing textual entailment challenge.
\newblock In {\em TAC}, 2009.

\bibitem{berthelot2019remixmatch}
David Berthelot, Nicholas Carlini, Ekin~D. Cubuk, Alex Kurakin, Kihyuk Sohn, Han Zhang, and Colin Raffel.
\newblock Remixmatch: Semi-supervised learning with distribution matching and augmentation anchoring.
\newblock In {\em 8th International Conference on Learning Representations, {ICLR} 2020, Addis Ababa, Ethiopia, April 26-30, 2020}. OpenReview.net, 2020.

\bibitem{berthelot2019mixmatch}
David Berthelot, Nicholas Carlini, Ian~J. Goodfellow, Nicolas Papernot, Avital Oliver, and Colin Raffel.
\newblock Mixmatch: {A} holistic approach to semi-supervised learning.
\newblock In Hanna~M. Wallach, Hugo Larochelle, Alina Beygelzimer, Florence d'Alch{\'{e}}{-}Buc, Emily~B. Fox, and Roman Garnett, editors, {\em Advances in Neural Information Processing Systems 32: Annual Conference on Neural Information Processing Systems 2019, NeurIPS 2019, December 8-14, 2019, Vancouver, BC, Canada}, pages 5050--5060, 2019.

\bibitem{berthelot2021adamatch}
David Berthelot, Rebecca Roelofs, Kihyuk Sohn, Nicholas Carlini, and Alexey Kurakin.
\newblock Adamatch: {A} unified approach to semi-supervised learning and domain adaptation.
\newblock In {\em The Tenth International Conference on Learning Representations, {ICLR} 2022, Virtual Event, April 25-29, 2022}. OpenReview.net, 2022.

\bibitem{bisk2018learning}
Yonatan Bisk, Kevin~J. Shih, Yejin Choi, and Daniel Marcu.
\newblock Learning interpretable spatial operations in a rich 3d blocks world.
\newblock In Sheila~A. McIlraith and Kilian~Q. Weinberger, editors, {\em Proceedings of the Thirty-Second {AAAI} Conference on Artificial Intelligence, (AAAI-18), the 30th innovative Applications of Artificial Intelligence (IAAI-18), and the 8th {AAAI} Symposium on Educational Advances in Artificial Intelligence (EAAI-18), New Orleans, Louisiana, USA, February 2-7, 2018}, pages 5028--5036. {AAAI} Press, 2018.

\bibitem{Bisk2020piqa}
Yonatan Bisk, Rowan Zellers, Ronan LeBras, Jianfeng Gao, and Yejin Choi.
\newblock {PIQA:} reasoning about physical commonsense in natural language.
\newblock In {\em The Thirty-Fourth {AAAI} Conference on Artificial Intelligence, {AAAI} 2020, The Thirty-Second Innovative Applications of Artificial Intelligence Conference, {IAAI} 2020, The Tenth {AAAI} Symposium on Educational Advances in Artificial Intelligence, {EAAI} 2020, New York, NY, USA, February 7-12, 2020}, pages 7432--7439. {AAAI} Press, 2020.

\bibitem{bowman-etal-2015-large}
Samuel~R. Bowman, Gabor Angeli, Christopher Potts, and Christopher~D. Manning.
\newblock A large annotated corpus for learning natural language inference.
\newblock In {\em Proceedings of the 2015 Conference on Empirical Methods in Natural Language Processing}, pages 632--642, Lisbon, Portugal, 2015. Association for Computational Linguistics.

\bibitem{10.5555/3495724.3495883}
Tom~B. Brown, Benjamin Mann, Nick Ryder, Melanie Subbiah, Jared Kaplan, Prafulla Dhariwal, Arvind Neelakantan, Pranav Shyam, Girish Sastry, Amanda Askell, Sandhini Agarwal, Ariel Herbert{-}Voss, Gretchen Krueger, Tom Henighan, Rewon Child, Aditya Ramesh, Daniel~M. Ziegler, Jeffrey Wu, Clemens Winter, Christopher Hesse, Mark Chen, Eric Sigler, Mateusz Litwin, Scott Gray, Benjamin Chess, Jack Clark, Christopher Berner, Sam McCandlish, Alec Radford, Ilya Sutskever, and Dario Amodei.
\newblock Language models are few-shot learners.
\newblock In Hugo Larochelle, Marc'Aurelio Ranzato, Raia Hadsell, Maria{-}Florina Balcan, and Hsuan{-}Tien Lin, editors, {\em Advances in Neural Information Processing Systems 33: Annual Conference on Neural Information Processing Systems 2020, NeurIPS 2020, December 6-12, 2020, virtual}, 2020.

\bibitem{cai-lapata-2019-semi}
Rui Cai and Mirella Lapata.
\newblock Semi-supervised semantic role labeling with cross-view training.
\newblock In {\em Proceedings of the 2019 Conference on Empirical Methods in Natural Language Processing and the 9th International Joint Conference on Natural Language Processing (EMNLP-IJCNLP)}, pages 1018--1027, Hong Kong, China, 2019. Association for Computational Linguistics.

\bibitem{bojar-etal-2014-findings}
Chris Callison-Burch, Philipp Koehn, Christof Monz, and Josh Schroeder.
\newblock Findings of the 2009 {W}orkshop on {S}tatistical {M}achine {T}ranslation.
\newblock In {\em Proceedings of the Fourth Workshop on Statistical Machine Translation}, pages 1--28, Athens, Greece, 2009. Association for Computational Linguistics.

\bibitem{carlini2021extracting}
Nicholas Carlini, Florian Tramer, Eric Wallace, Matthew Jagielski, Ariel Herbert-Voss, Katherine Lee, Adam Roberts, Tom~B Brown, Dawn Song, Ulfar Erlingsson, et~al.
\newblock Extracting training data from large language models.
\newblock In {\em USENIX Security Symposium}, volume~6, 2021.

\bibitem{cer-etal-2017-semeval}
Daniel Cer, Mona Diab, Eneko Agirre, I{\~n}igo Lopez-Gazpio, and Lucia Specia.
\newblock {S}em{E}val-2017 task 1: Semantic textual similarity multilingual and crosslingual focused evaluation.
\newblock In {\em Proceedings of the 11th International Workshop on Semantic Evaluation ({S}em{E}val-2017)}, pages 1--14, Vancouver, Canada, 2017. Association for Computational Linguistics.

\bibitem{cer2017semeval_sts-b}
Daniel Cer, Mona Diab, Eneko Agirre, I{\~n}igo Lopez-Gazpio, and Lucia Specia.
\newblock {S}em{E}val-2017 task 1: Semantic textual similarity multilingual and crosslingual focused evaluation.
\newblock In {\em Proceedings of the 11th International Workshop on Semantic Evaluation ({S}em{E}val-2017)}, pages 1--14, Vancouver, Canada, 2017. Association for Computational Linguistics.

\bibitem{chang2008importance}
Ming-Wei Chang, Lev Ratinov, Dan Roth, and Vivek Srikumar.
\newblock Importance of semantic representation: dataless classification.
\newblock In {\em Proceedings of the 23rd national conference on Artificial intelligence-Volume 2}, pages 830--835, 2008.

\bibitem{chapelle2009semi}
Olivier Chapelle, Bernhard Scholkopf, and Alexander Zien.
\newblock Semi-supervised learning (chapelle, o. et al., eds.; 2006)[book reviews].
\newblock {\em IEEE Transactions on Neural Networks}, 20(3):542--542, 2009.

\bibitem{codealpaca}
Sahil Chaudhary.
\newblock Code alpaca: An instruction-following llama model for code generation.
\newblock https://github.com/sahil280114/codealpaca, 2023.

\bibitem{DST2022}
Baixu Chen, Junguang Jiang, Ximei Wang, Pengfei Wan, Jianmin Wang, and Mingsheng Long.
\newblock Debiased self-training for semi-supervised learning.
\newblock In {\em Advances in Neural Information Processing Systems}, NIPS'22, 2022.

\bibitem{chen2020unseen}
Chin-Hui Chen, Yi-Fu Fu, Hsiao-Hua Cheng, and Shou-De Lin.
\newblock Unseen filler generalization in attention-based natural language reasoning models.
\newblock In {\em 2020 IEEE Second International Conference on Cognitive Machine Intelligence (CogMI)}, pages 42--51. IEEE, 2020.

\bibitem{chen2019touchdown}
Howard Chen, Alane Suhr, Dipendra Misra, Noah Snavely, and Yoav Artzi.
\newblock {TOUCHDOWN:} natural language navigation and spatial reasoning in visual street environments.
\newblock In {\em {IEEE} Conference on Computer Vision and Pattern Recognition, {CVPR} 2019, Long Beach, CA, USA, June 16-20, 2019}. Computer Vision Foundation / {IEEE}, 2019.

\bibitem{chen-etal-2020-mixtext}
Jiaao Chen, Zichao Yang, and Diyi Yang.
\newblock {M}ix{T}ext: Linguistically-informed interpolation of hidden space for semi-supervised text classification.
\newblock In {\em Proceedings of the 58th Annual Meeting of the Association for Computational Linguistics}, pages 2147--2157, Online, 2020. Association for Computational Linguistics.

\bibitem{chen2024alpagasus}
Lichang Chen, Shiyang Li, Jun Yan, Hai Wang, Kalpa Gunaratna, Vikas Yadav, Zheng Tang, Vijay Srinivasan, Tianyi Zhou, Heng Huang, and Hongxia Jin.
\newblock Alpagasus: Training a better alpaca model with fewer data.
\newblock In {\em The Twelfth International Conference on Learning Representations}, 2024.

\bibitem{chen2021codex}
Mark Chen, Jerry Tworek, Heewoo Jun, Qiming Yuan, Henrique~Ponde de~Oliveira~Pinto, Jared Kaplan, Harri Edwards, Yuri Burda, Nicholas Joseph, Greg Brockman, Alex Ray, Raul Puri, Gretchen Krueger, Michael Petrov, Heidy Khlaaf, Girish Sastry, Pamela Mishkin, Brooke Chan, Scott Gray, Nick Ryder, Mikhail Pavlov, Alethea Power, Lukasz Kaiser, Mohammad Bavarian, Clemens Winter, Philippe Tillet, Felipe~Petroski Such, Dave Cummings, Matthias Plappert, Fotios Chantzis, Elizabeth Barnes, Ariel Herbert-Voss, William~Hebgen Guss, Alex Nichol, Alex Paino, Nikolas Tezak, Jie Tang, Igor Babuschkin, Suchir Balaji, Shantanu Jain, William Saunders, Christopher Hesse, Andrew~N. Carr, Jan Leike, Josh Achiam, Vedant Misra, Evan Morikawa, Alec Radford, Matthew Knight, Miles Brundage, Mira Murati, Katie Mayer, Peter Welinder, Bob McGrew, Dario Amodei, Sam McCandlish, Ilya Sutskever, and Wojciech Zaremba.
\newblock Evaluating large language models trained on code.
\newblock {\em ArXiv preprint}, abs/2107.03374, 2021.

\bibitem{chen2015microsoft}
Xinlei Chen, Hao Fang, Tsung-Yi Lin, Ramakrishna Vedantam, Saurabh Gupta, Piotr Doll{\'a}r, and C~Lawrence Zitnick.
\newblock Microsoft coco captions: Data collection and evaluation server.
\newblock {\em ArXiv preprint}, abs/1504.00325, 2015.

\bibitem{chevalier2023adapting}
Alexis Chevalier, Alexander Wettig, Anirudh Ajith, and Danqi Chen.
\newblock Adapting language models to compress contexts.
\newblock In Houda Bouamor, Juan Pino, and Kalika Bali, editors, {\em Proceedings of the 2023 Conference on Empirical Methods in Natural Language Processing}, pages 3829--3846, Singapore, 2023. Association for Computational Linguistics.

\bibitem{vicuna2023}
Wei-Lin Chiang, Zhuohan Li, Zi~Lin, Ying Sheng, Zhanghao Wu, Hao Zhang, Lianmin Zheng, Siyuan Zhuang, Yonghao Zhuang, Joseph~E. Gonzalez, Ion Stoica, and Eric~P. Xing.
\newblock Vicuna: An open-source chatbot impressing gpt-4 with 90\%* chatgpt quality, 2023.

\bibitem{chowdhery2022palm}
Aakanksha Chowdhery, Sharan Narang, Jacob Devlin, Maarten Bosma, Gaurav Mishra, Adam Roberts, Paul Barham, Hyung~Won Chung, Charles Sutton, Sebastian Gehrmann, et~al.
\newblock Palm: Scaling language modeling with pathways.
\newblock {\em ArXiv preprint}, abs/2204.02311, 2022.

\bibitem{NIPS2017_d5e2c0ad}
Paul~F Christiano, Jan Leike, Tom Brown, Miljan Martic, Shane Legg, and Dario Amodei.
\newblock Deep reinforcement learning from human preferences.
\newblock In I.~Guyon, U.~Von Luxburg, S.~Bengio, H.~Wallach, R.~Fergus, S.~Vishwanathan, and R.~Garnett, editors, {\em Advances in Neural Information Processing Systems}, volume~30. Curran Associates, Inc., 2017.

\bibitem{JMLR:v25:23-0870}
Hyung~Won Chung, Le~Hou, Shayne Longpre, Barret Zoph, Yi~Tay, William Fedus, Yunxuan Li, Xuezhi Wang, Mostafa Dehghani, Siddhartha Brahma, Albert Webson, Shixiang~Shane Gu, Zhuyun Dai, Mirac Suzgun, Xinyun Chen, Aakanksha Chowdhery, Alex Castro-Ros, Marie Pellat, Kevin Robinson, Dasha Valter, Sharan Narang, Gaurav Mishra, Adams Yu, Vincent Zhao, Yanping Huang, Andrew Dai, Hongkun Yu, Slav Petrov, Ed~H. Chi, Jeff Dean, Jacob Devlin, Adam Roberts, Denny Zhou, Quoc~V. Le, and Jason Wei.
\newblock Scaling instruction-finetuned language models.
\newblock {\em Journal of Machine Learning Research}, 25(70):1--53, 2024.

\bibitem{clark-etal-2019-boolq}
Christopher Clark, Kenton Lee, Ming-Wei Chang, Tom Kwiatkowski, Michael Collins, and Kristina Toutanova.
\newblock {B}ool{Q}: Exploring the surprising difficulty of natural yes/no questions.
\newblock In {\em Proceedings of the 2019 Conference of the North {A}merican Chapter of the Association for Computational Linguistics: Human Language Technologies, Volume 1 (Long and Short Papers)}, pages 2924--2936, Minneapolis, Minnesota, 2019. Association for Computational Linguistics.

\bibitem{clark-etal-2018-semi}
Kevin Clark, Minh-Thang Luong, Christopher~D. Manning, and Quoc Le.
\newblock Semi-supervised sequence modeling with cross-view training.
\newblock In {\em Proceedings of the 2018 Conference on Empirical Methods in Natural Language Processing}, pages 1914--1925, Brussels, Belgium, 2018. Association for Computational Linguistics.

\bibitem{Clark2018ThinkYH}
Peter Clark, Isaac Cowhey, Oren Etzioni, Tushar Khot, Ashish Sabharwal, Carissa Schoenick, and Oyvind Tafjord.
\newblock Think you have solved question answering? try arc, the ai2 reasoning challenge.
\newblock {\em ArXiv preprint}, abs/1803.05457, 2018.

\bibitem{cobbe2021training}
Karl Cobbe, Vineet Kosaraju, Mohammad Bavarian, Jacob Hilton, Reiichiro Nakano, Christopher Hesse, and John Schulman.
\newblock Training verifiers to solve math word problems, 2021.

\bibitem{DollyV2}
Mike Conover, Matt Hayes, Ankit Mathur, Jianwei Xie, Jun Wan, Sam Shah, Ali Ghodsi, Patrick Wendell, Matei Zaharia, and Reynold Xin.
\newblock Free dolly: Introducing the world's first truly open instruction-tuned llm, 2023.

\bibitem{dagan2005pascal_rte1}
Ido Dagan, Oren Glickman, and Bernardo Magnini.
\newblock The {PASCAL} recognising textual entailment challenge.
\newblock In {\em the First International Conference on Machine Learning Challenges: Evaluating Predictive Uncertainty Visual Object Classification, and Recognizing Textual Entailment}, 2005.

\bibitem{dao2023flashattention2}
Tri Dao.
\newblock Flashattention-2: Faster attention with better parallelism and work partitioning, 2023.

\bibitem{de2019commitmentbank}
Marie-Catherine De~Marneffe, Mandy Simons, and Judith Tonhauser.
\newblock The commitmentbank: Investigating projection in naturally occurring discourse.
\newblock In {\em proceedings of Sinn und Bedeutung}, volume~23, pages 107--124, 2019.

\bibitem{dehghani2018universal}
Mostafa Dehghani, Stephan Gouws, Oriol Vinyals, Jakob Uszkoreit, and Lukasz Kaiser.
\newblock Universal transformers.
\newblock In {\em 7th International Conference on Learning Representations, {ICLR} 2019, New Orleans, LA, USA, May 6-9, 2019}. OpenReview.net, 2019.

\bibitem{devlin2018bert}
Jacob Devlin, Ming-Wei Chang, Kenton Lee, and Kristina Toutanova.
\newblock {BERT}: Pre-training of deep bidirectional transformers for language understanding.
\newblock In {\em Proceedings of the 2019 Conference of the North {A}merican Chapter of the Association for Computational Linguistics: Human Language Technologies, Volume 1 (Long and Short Papers)}, pages 4171--4186, Minneapolis, Minnesota, 2019. Association for Computational Linguistics.

\bibitem{ding2020object}
David Ding, Felix Hill, Adam Santoro, and Matt Botvinick.
\newblock Object-based attention for spatio-temporal reasoning: Outperforming neuro-symbolic models with flexible distributed architectures.
\newblock {\em ArXiv preprint}, abs/2012.08508, 2020.

\bibitem{dolan-brockett-2005-automatically}
William~B. Dolan and Chris Brockett.
\newblock Automatically constructing a corpus of sentential paraphrases.
\newblock In {\em Proceedings of the Third International Workshop on Paraphrasing ({IWP}2005)}, 2005.

\bibitem{dolan2005automatically_mrpc}
William~B. Dolan and Chris Brockett.
\newblock Automatically constructing a corpus of sentential paraphrases.
\newblock In {\em Proceedings of the Third International Workshop on Paraphrasing ({IWP}2005)}, 2005.

\bibitem{dong-de-melo-2019-robust}
Xin Dong and Gerard de~Melo.
\newblock A robust self-learning framework for cross-lingual text classification.
\newblock In {\em Proceedings of the 2019 Conference on Empirical Methods in Natural Language Processing and the 9th International Joint Conference on Natural Language Processing (EMNLP-IJCNLP)}, pages 6306--6310, Hong Kong, China, 2019. Association for Computational Linguistics.

\bibitem{dubey2024llama}
Abhimanyu Dubey, Abhinav Jauhri, Abhinav Pandey, Abhishek Kadian, Ahmad Al-Dahle, Aiesha Letman, Akhil Mathur, Alan Schelten, Amy Yang, Angela Fan, et~al.
\newblock The llama 3 herd of models.
\newblock {\em arXiv preprint arXiv:2407.21783}, 2024.

\bibitem{dunn2017searchqa}
Matthew Dunn, Levent Sagun, Mike Higgins, V~Ugur Guney, Volkan Cirik, and Kyunghyun Cho.
\newblock Searchqa: A new q\&a dataset augmented with context from a search engine.
\newblock {\em ArXiv preprint}, abs/1704.05179, 2017.

\bibitem{fisch-etal-2019-mrqa}
Adam Fisch, Alon Talmor, Robin Jia, Minjoon Seo, Eunsol Choi, and Danqi Chen.
\newblock {MRQA} 2019 shared task: Evaluating generalization in reading comprehension.
\newblock In {\em Proceedings of the 2nd Workshop on Machine Reading for Question Answering}, pages 1--13, Hong Kong, China, 2019. Association for Computational Linguistics.

\bibitem{gao-etal-2021-making}
Tianyu Gao, Adam Fisch, and Danqi Chen.
\newblock Making pre-trained language models better few-shot learners.
\newblock In {\em Proceedings of the 59th Annual Meeting of the Association for Computational Linguistics and the 11th International Joint Conference on Natural Language Processing (Volume 1: Long Papers)}, pages 3816--3830, Online, 2021. Association for Computational Linguistics.

\bibitem{gera2022zero}
Ariel Gera, Alon Halfon, Eyal Shnarch, Yotam Perlitz, Liat Ein-Dor, and Noam Slonim.
\newblock Zero-shot text classification with self-training.
\newblock In {\em Proceedings of the 2022 Conference on Empirical Methods in Natural Language Processing}, pages 1107--1119, Abu Dhabi, United Arab Emirates, 2022. Association for Computational Linguistics.

\bibitem{giampiccolo-etal-2007-third}
Danilo Giampiccolo, Bernardo Magnini, Ido Dagan, and Bill Dolan.
\newblock The third {PASCAL} recognizing textual entailment challenge.
\newblock In {\em Proceedings of the {ACL}-{PASCAL} Workshop on Textual Entailment and Paraphrasing}, pages 1--9, Prague, 2007. Association for Computational Linguistics.

\bibitem{giampiccolo2007third_rte3}
Danilo Giampiccolo, Bernardo Magnini, Ido Dagan, and Bill Dolan.
\newblock The third {PASCAL} recognizing textual entailment challenge.
\newblock In {\em Proceedings of the {ACL}-{PASCAL} Workshop on Textual Entailment and Paraphrasing}, pages 1--9, Prague, 2007. Association for Computational Linguistics.

\bibitem{goel2022pars}
Arushi Goel, Yunlong Jiao, and Jordan Massiah.
\newblock Pars: Pseudo-label aware robust sample selection for learning with noisy labels.
\newblock {\em ArXiv preprint}, abs/2201.10836, 2022.

\bibitem{goyal2017making}
Yash Goyal, Tejas Khot, Douglas Summers{-}Stay, Dhruv Batra, and Devi Parikh.
\newblock Making the {V} in {VQA} matter: Elevating the role of image understanding in visual question answering.
\newblock In {\em 2017 {IEEE} Conference on Computer Vision and Pattern Recognition, {CVPR} 2017, Honolulu, HI, USA, July 21-26, 2017}, pages 6325--6334. {IEEE} Computer Society, 2017.

\bibitem{grandvalet2004semi}
Yves Grandvalet and Yoshua Bengio.
\newblock Semi-supervised learning by entropy minimization.
\newblock In {\em Advances in Neural Information Processing Systems 17 [Neural Information Processing Systems, {NIPS} 2004, December 13-18, 2004, Vancouver, British Columbia, Canada]}, pages 529--536, 2004.

\bibitem{grangier2024projected}
David Grangier, Angelos Katharopoulos, Pierre Ablin, and Awni Hannun.
\newblock Projected language models: A large model pre-segmented into smaller ones.
\newblock In {\em ICML 2024 Workshop on Foundation Models in the Wild}, 2024.

\bibitem{gu-etal-2022-ppt}
Yuxian Gu, Xu~Han, Zhiyuan Liu, and Minlie Huang.
\newblock {PPT}: Pre-trained prompt tuning for few-shot learning.
\newblock In {\em Proceedings of the 60th Annual Meeting of the Association for Computational Linguistics (Volume 1: Long Papers)}, pages 8410--8423, Dublin, Ireland, 2022. Association for Computational Linguistics.

\bibitem{gunasekar2023textbooks}
Suriya Gunasekar, Yi~Zhang, Jyoti Aneja, Caio C{\'e}sar~Teodoro Mendes, Allie Del~Giorno, Sivakanth Gopi, Mojan Javaheripi, Piero Kauffmann, Gustavo de~Rosa, Olli Saarikivi, et~al.
\newblock Textbooks are all you need.
\newblock {\em ArXiv preprint}, abs/2306.11644, 2023.

\bibitem{guo2021diff}
Demi Guo, Alexander Rush, and Yoon Kim.
\newblock Parameter-efficient transfer learning with diff pruning.
\newblock In {\em Proceedings of the 59th Annual Meeting of the Association for Computational Linguistics and the 11th International Joint Conference on Natural Language Processing (Volume 1: Long Papers)}, pages 4884--4896, Online, 2021. Association for Computational Linguistics.

\bibitem{gururangan-etal-2019-variational}
Suchin Gururangan, Tam Dang, Dallas Card, and Noah~A. Smith.
\newblock Variational pretraining for semi-supervised text classification.
\newblock In {\em Proceedings of the 57th Annual Meeting of the Association for Computational Linguistics}, pages 5880--5894, Florence, Italy, 2019. Association for Computational Linguistics.

\bibitem{gururangan-etal-2020-dont}
Suchin Gururangan, Ana Marasovi{\'c}, Swabha Swayamdipta, Kyle Lo, Iz~Beltagy, Doug Downey, and Noah~A. Smith.
\newblock Don{'}t stop pretraining: Adapt language models to domains and tasks.
\newblock In {\em Proceedings of the 58th Annual Meeting of the Association for Computational Linguistics}, pages 8342--8360, Online, 2020. Association for Computational Linguistics.

\bibitem{hambardzumyan-etal-2021-warp}
Karen Hambardzumyan, Hrant Khachatrian, and Jonathan May.
\newblock {WARP}: {W}ord-level {A}dversarial {R}e{P}rogramming.
\newblock In {\em Proceedings of the 59th Annual Meeting of the Association for Computational Linguistics and the 11th International Joint Conference on Natural Language Processing (Volume 1: Long Papers)}, pages 4921--4933, Online, 2021. Association for Computational Linguistics.

\bibitem{hartvigsen-etal-2022-toxigen}
Thomas Hartvigsen, Saadia Gabriel, Hamid Palangi, Maarten Sap, Dipankar Ray, and Ece Kamar.
\newblock {T}oxi{G}en: A large-scale machine-generated dataset for adversarial and implicit hate speech detection.
\newblock In {\em Proceedings of the 60th Annual Meeting of the Association for Computational Linguistics (Volume 1: Long Papers)}, pages 3309--3326, Dublin, Ireland, 2022. Association for Computational Linguistics.

\bibitem{he2023preserving}
Guande He, Jianfei Chen, and Jun Zhu.
\newblock Preserving pre-trained features helps calibrate fine-tuned language models.
\newblock {\em ArXiv preprint}, abs/2305.19249, 2023.

\bibitem{10.1007/978-3-030-99736-6_20}
Mariya Hendriksen, Maurits Bleeker, Svitlana Vakulenko, Nanne van Noord, Ernst Kuiper, and Maarten de~Rijke.
\newblock Extending clip for category-to-image retrieval in e-commerce.
\newblock In {\em Advances in Information Retrieval: 44th European Conference on IR Research, ECIR 2022, Stavanger, Norway, April 10–14, 2022, Proceedings, Part I}, Berlin, Heidelberg, 2022.

\bibitem{hendrycks2021aligning}
Dan Hendrycks, Collin Burns, Steven Basart, Andrew Critch, Jerry Li, Dawn Song, and Jacob Steinhardt.
\newblock Aligning {AI} with shared human values.
\newblock In {\em 9th International Conference on Learning Representations, {ICLR} 2021, Virtual Event, Austria, May 3-7, 2021}. OpenReview.net, 2021.

\bibitem{hendrycks2021measuring}
Dan Hendrycks, Collin Burns, Steven Basart, Andy Zou, Mantas Mazeika, Dawn Song, and Jacob Steinhardt.
\newblock Measuring massive multitask language understanding.
\newblock In {\em 9th International Conference on Learning Representations, {ICLR} 2021, Virtual Event, Austria, May 3-7, 2021}. OpenReview.net, 2021.

\bibitem{hoffmann2022training}
Jordan Hoffmann, Sebastian Borgeaud, Arthur Mensch, Elena Buchatskaya, Trevor Cai, Eliza Rutherford, Diego de~Las Casas, Lisa~Anne Hendricks, Johannes Welbl, Aidan Clark, et~al.
\newblock Training compute-optimal large language models.
\newblock {\em ArXiv preprint}, abs/2203.15556, 2022.

\bibitem{hong2024orpo}
Jiwoo Hong, Noah Lee, and James Thorne.
\newblock Orpo: Monolithic preference optimization without reference model.
\newblock {\em arXiv preprint arXiv:2403.07691}, 2(4):5, 2024.

\bibitem{honovich-etal-2023-unnatural}
Or~Honovich, Thomas Scialom, Omer Levy, and Timo Schick.
\newblock Unnatural instructions: Tuning language models with (almost) no human labor.
\newblock In Anna Rogers, Jordan Boyd-Graber, and Naoaki Okazaki, editors, {\em Proceedings of the 61st Annual Meeting of the Association for Computational Linguistics (Volume 1: Long Papers)}, pages 14409--14428, Toronto, Canada, 2023. Association for Computational Linguistics.

\bibitem{hosking2024human}
Tom Hosking, Phil Blunsom, and Max Bartolo.
\newblock Human feedback is not gold standard.
\newblock In {\em The Twelfth International Conference on Learning Representations}, 2024.

\bibitem{10.1162/tacl_a_00517}
Zejiang Hou, Julian Salazar, and George Polovets.
\newblock Meta-learning the difference: Preparing large language models for efficient adaptation.
\newblock {\em Transactions of the Association for Computational Linguistics}, 10:1249--1265, 2022.

\bibitem{houlsby2019parameter}
Neil Houlsby, Andrei Giurgiu, Stanislaw Jastrzebski, Bruna Morrone, Quentin de~Laroussilhe, Andrea Gesmundo, Mona Attariyan, and Sylvain Gelly.
\newblock Parameter-efficient transfer learning for {NLP}.
\newblock In Kamalika Chaudhuri and Ruslan Salakhutdinov, editors, {\em Proceedings of the 36th International Conference on Machine Learning, {ICML} 2019, 9-15 June 2019, Long Beach, California, {USA}}, volume~97 of {\em Proceedings of Machine Learning Research}, pages 2790--2799. {PMLR}, 2019.

\bibitem{howard-ruder-2018-universal}
Jeremy Howard and Sebastian Ruder.
\newblock Universal language model fine-tuning for text classification.
\newblock In {\em Proceedings of the 56th Annual Meeting of the Association for Computational Linguistics (Volume 1: Long Papers)}, pages 328--339, Melbourne, Australia, 2018. Association for Computational Linguistics.

\bibitem{overfitting2023}
Jeremy Howard and Jonathan Whitaker.
\newblock Can llms learn from a single example?, 2023.

\bibitem{hu2021lora}
Edward~J. Hu, Yelong Shen, Phillip Wallis, Zeyuan Allen{-}Zhu, Yuanzhi Li, Shean Wang, Lu~Wang, and Weizhu Chen.
\newblock Lora: Low-rank adaptation of large language models.
\newblock In {\em The Tenth International Conference on Learning Representations, {ICLR} 2022, Virtual Event, April 25-29, 2022}. OpenReview.net, 2022.

\bibitem{hu2004mining_cr}
Minqing Hu and Bing Liu.
\newblock Mining and summarizing customer reviews.
\newblock In {\em ACM SIGKDD international conference on Knowledge discovery and data mining}, 2004.

\bibitem{Mathew2024Instruction}
Mathew Huerta-Enochian.
\newblock Instruction fine-tuning: Does prompt loss matter?, 2024.

\bibitem{inan2023llama}
Hakan Inan, Kartikeya Upasani, Jianfeng Chi, Rashi Rungta, Krithika Iyer, Yuning Mao, Michael Tontchev, Qing Hu, Brian Fuller, Davide Testuggine, et~al.
\newblock Llama guard: Llm-based input-output safeguard for human-ai conversations.
\newblock {\em arXiv preprint arXiv:2312.06674}, 2023.

\bibitem{ivison-peters-2022-hyperdecoders}
Hamish Ivison and Matthew Peters.
\newblock Hyperdecoders: Instance-specific decoders for multi-task {NLP}.
\newblock In {\em Findings of the Association for Computational Linguistics: EMNLP 2022}, pages 1715--1730, Abu Dhabi, United Arab Emirates, 2022. Association for Computational Linguistics.

\bibitem{ivison2023camels}
Hamish Ivison, Yizhong Wang, Valentina Pyatkin, Nathan Lambert, Matthew Peters, Pradeep Dasigi, Joel Jang, David Wadden, Noah~A Smith, Iz~Beltagy, et~al.
\newblock Camels in a changing climate: Enhancing lm adaptation with tulu 2, 2023.

\bibitem{jain2024neftune}
Neel Jain, Ping yeh Chiang, Yuxin Wen, John Kirchenbauer, Hong-Min Chu, Gowthami Somepalli, Brian~R. Bartoldson, Bhavya Kailkhura, Avi Schwarzschild, Aniruddha Saha, Micah Goldblum, Jonas Geiping, and Tom Goldstein.
\newblock {NEFT}une: Noisy embeddings improve instruction finetuning.
\newblock In {\em The Twelfth International Conference on Learning Representations}, 2024.

\bibitem{janner2018representation}
Michael Janner, Karthik Narasimhan, and Regina Barzilay.
\newblock Representation learning for grounded spatial reasoning.
\newblock {\em Transactions of the Association for Computational Linguistics}, 6:49--61, 2018.

\bibitem{jha2023limit}
Aditi Jha, Sam Havens, Jeremey Dohmann, Alex Trott, and Jacob Portes.
\newblock Limit: Less is more for instruction tuning across evaluation paradigms.
\newblock {\em ArXiv preprint}, abs/2311.13133, 2023.

\bibitem{kaplan2020scaling}
Jared Kaplan, Sam McCandlish, Tom Henighan, Tom~B Brown, Benjamin Chess, Rewon Child, Scott Gray, Alec Radford, Jeffrey Wu, and Dario Amodei.
\newblock Scaling laws for neural language models.
\newblock {\em ArXiv preprint}, abs/2001.08361, 2020.

\bibitem{mahabadi2021parameter}
Rabeeh Karimi~Mahabadi, Sebastian Ruder, Mostafa Dehghani, and James Henderson.
\newblock Parameter-efficient multi-task fine-tuning for transformers via shared hypernetworks.
\newblock In {\em Proceedings of the 59th Annual Meeting of the Association for Computational Linguistics and the 11th International Joint Conference on Natural Language Processing (Volume 1: Long Papers)}, pages 565--576, Online, 2021. Association for Computational Linguistics.

\bibitem{khashabi-etal-2018-looking}
Daniel Khashabi, Snigdha Chaturvedi, Michael Roth, Shyam Upadhyay, and Dan Roth.
\newblock Looking beyond the surface: A challenge set for reading comprehension over multiple sentences.
\newblock In {\em Proceedings of the 2018 Conference of the North {A}merican Chapter of the Association for Computational Linguistics: Human Language Technologies, Volume 1 (Long Papers)}, pages 252--262, New Orleans, Louisiana, 2018. Association for Computational Linguistics.

\bibitem{khashabi-etal-2020-unifiedqa}
Daniel Khashabi, Sewon Min, Tushar Khot, Ashish Sabharwal, Oyvind Tafjord, Peter Clark, and Hannaneh Hajishirzi.
\newblock {UNIFIEDQA}: Crossing format boundaries with a single {QA} system.
\newblock In {\em Findings of the Association for Computational Linguistics: EMNLP 2020}, pages 1896--1907, Online, 2020. Association for Computational Linguistics.

\bibitem{Khot_Sabharwal_Clark_2018}
Tushar Khot, Ashish Sabharwal, and Peter Clark.
\newblock Scitail: A textual entailment dataset from science question answering.
\newblock {\em Proceedings of the AAAI Conference on Artificial Intelligence}, 32(1), Apr. 2018.

\bibitem{khot2022decomposed}
Tushar Khot, Harsh Trivedi, Matthew Finlayson, Yao Fu, Kyle Richardson, Peter Clark, and Ashish Sabharwal.
\newblock Decomposed prompting: A modular approach for solving complex tasks.
\newblock In {\em The Eleventh International Conference on Learning Representations}, 2022.

\bibitem{kipf2017semi}
Thomas~N. Kipf and Max Welling.
\newblock Semi-supervised classification with graph convolutional networks.
\newblock In {\em 5th International Conference on Learning Representations, {ICLR} 2017, Toulon, France, April 24-26, 2017, Conference Track Proceedings}. OpenReview.net, 2017.

\bibitem{kipf2016semi}
Thomas~N. Kipf and Max Welling.
\newblock Semi-supervised classification with graph convolutional networks.
\newblock In {\em 5th International Conference on Learning Representations, {ICLR} 2017, Toulon, France, April 24-26, 2017, Conference Track Proceedings}. OpenReview.net, 2017.

\bibitem{koster2018big}
Raphael Koster, Martin~J Chadwick, Yi~Chen, David Berron, Andrea Banino, Emrah D{\"u}zel, Demis Hassabis, and Dharshan Kumaran.
\newblock Big-loop recurrence within the hippocampal system supports integration of information across episodes.
\newblock {\em Neuron}, 99(6):1342--1354, 2018.

\bibitem{kruijff2007situated}
Geert-Jan~M Kruijff, Hendrik Zender, Patric Jensfelt, and Henrik~I Christensen.
\newblock Situated dialogue and spatial organization: What, where… and why?
\newblock {\em International Journal of Advanced Robotic Systems}, 2007.

\bibitem{kumaran2012generalization}
Dharshan Kumaran and James~L McClelland.
\newblock Generalization through the recurrent interaction of episodic memories: a model of the hippocampal system.
\newblock {\em Psychological review}, 2012.

\bibitem{kwiatkowski-etal-2019-natural}
Tom Kwiatkowski, Jennimaria Palomaki, Olivia Redfield, Michael Collins, Ankur Parikh, Chris Alberti, Danielle Epstein, Illia Polosukhin, Jacob Devlin, Kenton Lee, Kristina Toutanova, Llion Jones, Matthew Kelcey, Ming-Wei Chang, Andrew~M. Dai, Jakob Uszkoreit, Quoc Le, and Slav Petrov.
\newblock Natural questions: A benchmark for question answering research.
\newblock {\em Transactions of the Association for Computational Linguistics}, 7:452--466, 2019.

\bibitem{DBLP:conf/iclr/LaineA17}
Samuli Laine and Timo Aila.
\newblock Temporal ensembling for semi-supervised learning.
\newblock In {\em 5th International Conference on Learning Representations, {ICLR} 2017, Toulon, France, April 24-26, 2017, Conference Track Proceedings}. OpenReview.net, 2017.

\bibitem{lambert2024t}
Nathan Lambert, Jacob Morrison, Valentina Pyatkin, Shengyi Huang, Hamish Ivison, Faeze Brahman, Lester James~V Miranda, Alisa Liu, Nouha Dziri, Shane Lyu, et~al.
\newblock T$\backslash$" ulu 3: Pushing frontiers in open language model post-training.
\newblock {\em arXiv preprint arXiv:2411.15124}, 2024.

\bibitem{landsiedel2017review}
Christian Landsiedel, Verena Rieser, Matthew Walter, and Dirk Wollherr.
\newblock A review of spatial reasoning and interaction for real-world robotics.
\newblock {\em Advanced Robotics}, 2017.

\bibitem{le2020self}
Hung Le, Truyen Tran, and Svetha Venkatesh.
\newblock Self-attentive associative memory.
\newblock In {\em Proceedings of the 37th International Conference on Machine Learning, {ICML} 2020, 13-18 July 2020, Virtual Event}, volume 119 of {\em Proceedings of Machine Learning Research}, pages 5682--5691. {PMLR}, 2020.

\bibitem{le2023bloom}
Teven Le~Scao, Angela Fan, Christopher Akiki, Ellie Pavlick, Suzana Ili{\'c}, Daniel Hesslow, Roman Castagn{\'e}, Alexandra~Sasha Luccioni, Fran{\c{c}}ois Yvon, Matthias Gall{\'e}, et~al.
\newblock Bloom: A 176b-parameter open-access multilingual language model.
\newblock {\em arXiv preprint}, 2023.

\bibitem{le-scao-rush-2021-many}
Teven Le~Scao and Alexander Rush.
\newblock How many data points is a prompt worth?
\newblock In {\em Proceedings of the 2021 Conference of the North American Chapter of the Association for Computational Linguistics: Human Language Technologies}, pages 2627--2636, Online, 2021. Association for Computational Linguistics.

\bibitem{lee2013pseudo}
Dong-Hyun Lee et~al.
\newblock Pseudo-label: The simple and efficient semi-supervised learning method for deep neural networks.
\newblock In {\em Workshop on challenges in representation learning, ICML}, page 896, 2013.

\bibitem{lee2020biobert}
Jinhyuk Lee, Wonjin Yoon, Sungdong Kim, Donghyeon Kim, Sunkyu Kim, Chan~Ho So, and Jaewoo Kang.
\newblock Biobert: a pre-trained biomedical language representation model for biomedical text mining.
\newblock {\em ArXiv preprint}, abs/1901.08746, 2019.

\bibitem{leike2018scalable}
Jan Leike, David Krueger, Tom Everitt, Miljan Martic, Vishal Maini, and Shane Legg.
\newblock Scalable agent alignment via reward modeling: a research direction.
\newblock {\em ArXiv preprint}, abs/1811.07871, 2018.

\bibitem{lester-etal-2021-power}
Brian Lester, Rami Al-Rfou, and Noah Constant.
\newblock The power of scale for parameter-efficient prompt tuning.
\newblock In {\em Proceedings of the 2021 Conference on Empirical Methods in Natural Language Processing}, pages 3045--3059, Online and Punta Cana, Dominican Republic, 2021. Association for Computational Linguistics.

\bibitem{levesque2012winograd_wnli}
Hector Levesque, Ernest Davis, and Leora Morgenstern.
\newblock The winograd schema challenge.
\newblock In {\em Thirteenth International Conference on the Principles of Knowledge Representation and Reasoning}, 2012.

\bibitem{ea01b9c0db064caca6986b925d75f2bb}
{Hector J.} Levesque, Ernest Davis, and Leora Morgenstern.
\newblock The winograd schema challenge.
\newblock In {\em 13th International Conference on the Principles of Knowledge Representation and Reasoning, KR 2012}, Proceedings of the International Conference on Knowledge Representation and Reasoning, pages 552--561. Institute of Electrical and Electronics Engineers Inc., 2012.
\newblock 13th International Conference on the Principles of Knowledge Representation and Reasoning, KR 2012 ; Conference date: 10-06-2012 Through 14-06-2012.

\bibitem{li-etal-2021-semi-supervised}
Changchun Li, Ximing Li, and Jihong Ouyang.
\newblock Semi-supervised text classification with balanced deep representation distributions.
\newblock In {\em Proceedings of the 59th Annual Meeting of the Association for Computational Linguistics and the 11th International Joint Conference on Natural Language Processing (Volume 1: Long Papers)}, pages 5044--5053, Online, 2021. Association for Computational Linguistics.

\bibitem{Li_Hogg_Cohn_2024}
Fangjun Li, David~C. Hogg, and Anthony~G. Cohn.
\newblock Advancing spatial reasoning in large language models: An in-depth evaluation and enhancement using the stepgame benchmark.
\newblock {\em Proceedings of the AAAI Conference on Artificial Intelligence}, 38(17):18500--18507, Mar. 2024.

\bibitem{li_dividemix_2020}
Junnan Li, Richard Socher, and Steven C.~H. Hoi.
\newblock Dividemix: Learning with noisy labels as semi-supervised learning.
\newblock In {\em 8th International Conference on Learning Representations, {ICLR} 2020, Addis Ababa, Ethiopia, April 26-30, 2020}. OpenReview.net, 2020.

\bibitem{li-etal-2022-learning-transfer}
Junyi Li, Tianyi Tang, Jian-Yun Nie, Ji-Rong Wen, and Xin Zhao.
\newblock Learning to transfer prompts for text generation.
\newblock In {\em Proceedings of the 2022 Conference of the North American Chapter of the Association for Computational Linguistics: Human Language Technologies}, pages 3506--3518, Seattle, United States, 2022. Association for Computational Linguistics.

\bibitem{li-etal-2021-task-adaptive}
Shiyang Li, Semih Yavuz, Wenhu Chen, and Xifeng Yan.
\newblock Task-adaptive pre-training and self-training are complementary for natural language understanding.
\newblock In {\em Findings of the Association for Computational Linguistics: EMNLP 2021}, pages 1006--1015, Punta Cana, Dominican Republic, 2021. Association for Computational Linguistics.

\bibitem{li2024selfalignment}
Xian Li, Ping Yu, Chunting Zhou, Timo Schick, Omer Levy, Luke Zettlemoyer, Jason~E Weston, and Mike Lewis.
\newblock Self-alignment with instruction backtranslation.
\newblock In {\em The Twelfth International Conference on Learning Representations}, 2024.

\bibitem{li-liang-2021-prefix}
Xiang~Lisa Li and Percy Liang.
\newblock Prefix-tuning: Optimizing continuous prompts for generation.
\newblock In {\em Proceedings of the 59th Annual Meeting of the Association for Computational Linguistics and the 11th International Joint Conference on Natural Language Processing (Volume 1: Long Papers)}, pages 4582--4597, Online, 2021. Association for Computational Linguistics.

\bibitem{alpaca_eval}
Xuechen Li, Tianyi Zhang, Yann Dubois, Rohan Taori, Ishaan Gulrajani, Carlos Guestrin, Percy Liang, and Tatsunori~B. Hashimoto.
\newblock Alpacaeval: An automatic evaluator of instruction-following models.
\newblock \url{https://github.com/tatsu-lab/alpaca_eval}, 5 2023.

\bibitem{lialin2023scaling}
Vladislav Lialin, Vijeta Deshpande, and Anna Rumshisky.
\newblock Scaling down to scale up: A guide to parameter-efficient fine-tuning.
\newblock {\em ArXiv preprint}, abs/2303.15647, 2023.

\bibitem{lin2023unlocking}
Bill~Yuchen Lin, Abhilasha Ravichander, Ximing Lu, Nouha Dziri, Melanie Sclar, Khyathi Chandu, Chandra Bhagavatula, and Yejin Choi.
\newblock The unlocking spell on base llms: Rethinking alignment via in-context learning.
\newblock {\em ArXiv preprint}, abs/2312.01552, 2023.

\bibitem{lin-etal-2022-truthfulqa}
Stephanie Lin, Jacob Hilton, and Owain Evans.
\newblock {T}ruthful{QA}: Measuring how models mimic human falsehoods.
\newblock In {\em Proceedings of the 60th Annual Meeting of the Association for Computational Linguistics (Volume 1: Long Papers)}, pages 3214--3252, Dublin, Ireland, 2022. Association for Computational Linguistics.

\bibitem{liu-etal-2023-compositional}
Fangyu Liu, Qianchu Liu, Shruthi Bannur, Fernando P{\'e}rez-Garc{\'\i}a, Naoto Usuyama, Sheng Zhang, Tristan Naumann, Aditya Nori, Hoifung Poon, Javier Alvarez-Valle, Ozan Oktay, and Stephanie~L. Hyland.
\newblock Compositional zero-shot domain transfer with text-to-text models.
\newblock {\em Transactions of the Association for Computational Linguistics}, 11:1097--1113, 2023.

\bibitem{NEURIPS2022_0cde695b}
Haokun Liu, Derek Tam, Mohammed Muqeeth, Jay Mohta, Tenghao Huang, Mohit Bansal, and Colin~A Raffel.
\newblock Few-shot parameter-efficient fine-tuning is better and cheaper than in-context learning.
\newblock In S.~Koyejo, S.~Mohamed, A.~Agarwal, D.~Belgrave, K.~Cho, and A.~Oh, editors, {\em Advances in Neural Information Processing Systems}, volume~35, pages 1950--1965. Curran Associates, Inc., 2022.

\bibitem{liu-etal-2024-summequal}
Junyuan Liu, Zhengyan Shi, and Aldo Lipani.
\newblock {S}umm{EQ}u{AL}: Summarization evaluation via question answering using large language models.
\newblock In Bhavana Dalvi~Mishra, Greg Durrett, Peter Jansen, Ben Lipkin, Danilo Neves~Ribeiro, Lionel Wong, Xi~Ye, and Wenting Zhao, editors, {\em Proceedings of the 2nd Workshop on Natural Language Reasoning and Structured Explanations (@ACL 2024)}, pages 46--55, Bangkok, Thailand, August 2024. Association for Computational Linguistics.

\bibitem{liu2024statistical}
Tianqi Liu, Yao Zhao, Rishabh Joshi, Misha Khalman, Mohammad Saleh, Peter~J Liu, and Jialu Liu.
\newblock Statistical rejection sampling improves preference optimization.
\newblock In {\em The Twelfth International Conference on Learning Representations}, 2024.

\bibitem{liu2023makes}
Wei Liu, Weihao Zeng, Keqing He, Yong Jiang, and Junxian He.
\newblock What makes good data for alignment? a comprehensive study of automatic data selection in instruction tuning.
\newblock {\em ArXiv preprint}, abs/2312.15685, 2023.

\bibitem{liu2021gpt}
Xiao Liu, Yanan Zheng, Zhengxiao Du, Ming Ding, Yujie Qian, Zhilin Yang, and Jie Tang.
\newblock Gpt understands, too.
\newblock {\em ArXiv preprint}, abs/2103.10385, 2021.

\bibitem{liu2019roberta}
Yinhan Liu, Myle Ott, Naman Goyal, Jingfei Du, Mandar Joshi, Danqi Chen, Omer Levy, Mike Lewis, Luke Zettlemoyer, and Veselin Stoyanov.
\newblock Roberta: A robustly optimized bert pretraining approach.
\newblock {\em ArXiv preprint}, abs/1907.11692, 2019.

\bibitem{logeswaran-etal-2019-zero}
Lajanugen Logeswaran, Ming-Wei Chang, Kenton Lee, Kristina Toutanova, Jacob Devlin, and Honglak Lee.
\newblock Zero-shot entity linking by reading entity descriptions.
\newblock In {\em Proceedings of the 57th Annual Meeting of the Association for Computational Linguistics}, pages 3449--3460, Florence, Italy, 2019. Association for Computational Linguistics.

\bibitem{lu2024ai}
Chris Lu, Cong Lu, Robert~Tjarko Lange, Jakob Foerster, Jeff Clune, and David Ha.
\newblock The ai scientist: Towards fully automated open-ended scientific discovery.
\newblock {\em arXiv preprint arXiv:2408.06292}, 2024.

\bibitem{luo2022biogpt}
Renqian Luo, Liai Sun, Yingce Xia, Tao Qin, Sheng Zhang, Hoifung Poon, and Tie-Yan Liu.
\newblock Biogpt: generative pre-trained transformer for biomedical text generation and mining.
\newblock {\em Briefings in bioinformatics}, 23(6):bbac409, 2022.

\bibitem{maas-etal-2011-learning}
Andrew~L. Maas, Raymond~E. Daly, Peter~T. Pham, Dan Huang, Andrew~Y. Ng, and Christopher Potts.
\newblock Learning word vectors for sentiment analysis.
\newblock In {\em Proceedings of the 49th Annual Meeting of the Association for Computational Linguistics: Human Language Technologies}, pages 142--150, Portland, Oregon, USA, 2011. Association for Computational Linguistics.

\bibitem{karimi2021compacter}
Rabeeh~Karimi Mahabadi, James Henderson, and Sebastian Ruder.
\newblock Compacter: Efficient low-rank hypercomplex adapter layers.
\newblock In Marc'Aurelio Ranzato, Alina Beygelzimer, Yann~N. Dauphin, Percy Liang, and Jennifer~Wortman Vaughan, editors, {\em Advances in Neural Information Processing Systems 34: Annual Conference on Neural Information Processing Systems 2021, NeurIPS 2021, December 6-14, 2021, virtual}, pages 1022--1035, 2021.

\bibitem{mao-etal-2022-unipelt}
Yuning Mao, Lambert Mathias, Rui Hou, Amjad Almahairi, Hao Ma, Jiawei Han, Scott Yih, and Madian Khabsa.
\newblock {U}ni{PELT}: A unified framework for parameter-efficient language model tuning.
\newblock In {\em Proceedings of the 60th Annual Meeting of the Association for Computational Linguistics (Volume 1: Long Papers)}, pages 6253--6264, Dublin, Ireland, 2022. Association for Computational Linguistics.

\bibitem{margatina-etal-2022-importance}
Katerina Margatina, Loic Barrault, and Nikolaos Aletras.
\newblock On the importance of effectively adapting pretrained language models for active learning.
\newblock In {\em Proceedings of the 60th Annual Meeting of the Association for Computational Linguistics (Volume 2: Short Papers)}, pages 825--836, Dublin, Ireland, 2022. Association for Computational Linguistics.

\bibitem{10.1145/2507157.2507163}
Julian~J. McAuley and Jure Leskovec.
\newblock Hidden factors and hidden topics: understanding rating dimensions with review text.
\newblock In Qiang Yang, Irwin King, Qing Li, Pearl Pu, and George Karypis, editors, {\em Seventh {ACM} Conference on Recommender Systems, RecSys '13, Hong Kong, China, October 12-16, 2013}, pages 165--172. {ACM}, 2013.

\bibitem{mcclosky-etal-2006-effective}
David McClosky, Eugene Charniak, and Mark Johnson.
\newblock Effective self-training for parsing.
\newblock In {\em Proceedings of the Human Language Technology Conference of the {NAACL}, Main Conference}, pages 152--159, New York City, USA, 2006. Association for Computational Linguistics.

\bibitem{mihaylov-etal-2018-suit}
Todor Mihaylov, Peter Clark, Tushar Khot, and Ashish Sabharwal.
\newblock Can a suit of armor conduct electricity? a new dataset for open book question answering.
\newblock In {\em Proceedings of the 2018 Conference on Empirical Methods in Natural Language Processing}, pages 2381--2391, Brussels, Belgium, 2018. Association for Computational Linguistics.

\bibitem{mikolov2013efficient}
Tomas Mikolov, Kai Chen, Greg Corrado, and Jeffrey Dean.
\newblock Efficient estimation of word representations in vector space.
\newblock {\em Proceedings of Workshop at ICLR}, 2013.

\bibitem{min-etal-2022-metaicl}
Sewon Min, Mike Lewis, Luke Zettlemoyer, and Hannaneh Hajishirzi.
\newblock {M}eta{ICL}: Learning to learn in context.
\newblock In {\em Proceedings of the 2022 Conference of the North American Chapter of the Association for Computational Linguistics: Human Language Technologies}, pages 2791--2809, Seattle, United States, 2022. Association for Computational Linguistics.

\bibitem{min-etal-2022-rethinking}
Sewon Min, Xinxi Lyu, Ari Holtzman, Mikel Artetxe, Mike Lewis, Hannaneh Hajishirzi, and Luke Zettlemoyer.
\newblock Rethinking the role of demonstrations: What makes in-context learning work?
\newblock In {\em Proceedings of the 2022 Conference on Empirical Methods in Natural Language Processing}, pages 11048--11064, Abu Dhabi, United Arab Emirates, 2022. Association for Computational Linguistics.

\bibitem{mishra-etal-2022-cross}
Swaroop Mishra, Daniel Khashabi, Chitta Baral, and Hannaneh Hajishirzi.
\newblock Cross-task generalization via natural language crowdsourcing instructions.
\newblock In {\em Proceedings of the 60th Annual Meeting of the Association for Computational Linguistics (Volume 1: Long Papers)}, pages 3470--3487, Dublin, Ireland, 2022. Association for Computational Linguistics.

\bibitem{miyato2018virtual}
Takeru Miyato, Shin-ichi Maeda, Masanori Koyama, and Shin Ishii.
\newblock Virtual adversarial training: a regularization method for supervised and semi-supervised learning.
\newblock {\em ArXiv preprint}, abs/1704.03976, 2017.

\bibitem{nayak2022learning}
Nihal~V Nayak, Peilin Yu, and Stephen Bach.
\newblock Learning to compose soft prompts for compositional zero-shot learning.
\newblock In {\em The Eleventh International Conference on Learning Representations}, 2022.

\bibitem{openai2023gpt4}
OpenAI.
\newblock Gpt-4 technical report.
\newblock {\em ArXiv preprint}, abs/2303.08774, 2023.

\bibitem{ott2019fairseq}
Myle Ott, Sergey Edunov, Alexei Baevski, Angela Fan, Sam Gross, Nathan Ng, David Grangier, and Michael Auli.
\newblock fairseq: A fast, extensible toolkit for sequence modeling.
\newblock In {\em Proceedings of the 2019 Conference of the North {A}merican Chapter of the Association for Computational Linguistics (Demonstrations)}, pages 48--53, Minneapolis, Minnesota, 2019. Association for Computational Linguistics.

\bibitem{ouyang2022training}
Long Ouyang, Jeffrey Wu, Xu~Jiang, Diogo Almeida, Carroll Wainwright, Pamela Mishkin, Chong Zhang, Sandhini Agarwal, Katarina Slama, Alex Gray, John Schulman, Jacob Hilton, Fraser Kelton, Luke Miller, Maddie Simens, Amanda Askell, Peter Welinder, Paul Christiano, Jan Leike, and Ryan Lowe.
\newblock Training language models to follow instructions with human feedback.
\newblock In Alice~H. Oh, Alekh Agarwal, Danielle Belgrave, and Kyunghyun Cho, editors, {\em Advances in Neural Information Processing Systems}, 2022.

\bibitem{NEURIPS2022_b1efde53}
Long Ouyang, Jeffrey Wu, Xu~Jiang, Diogo Almeida, Carroll Wainwright, Pamela Mishkin, Chong Zhang, Sandhini Agarwal, Katarina Slama, Alex Ray, John Schulman, Jacob Hilton, Fraser Kelton, Luke Miller, Maddie Simens, Amanda Askell, Peter Welinder, Paul~F Christiano, Jan Leike, and Ryan Lowe.
\newblock Training language models to follow instructions with human feedback.
\newblock In S.~Koyejo, S.~Mohamed, A.~Agarwal, D.~Belgrave, K.~Cho, and A.~Oh, editors, {\em Advances in Neural Information Processing Systems}, volume~35, pages 27730--27744. Curran Associates, Inc., 2022.

\bibitem{pal2024smaug}
Arka Pal, Deep Karkhanis, Samuel Dooley, Manley Roberts, Siddartha Naidu, and Colin White.
\newblock Smaug: Fixing failure modes of preference optimisation with dpo-positive.
\newblock {\em arXiv preprint arXiv:2402.13228}, 2024.

\bibitem{palm2018recurrent}
Rasmus~Berg Palm, Ulrich Paquet, and Ole Winther.
\newblock Recurrent relational networks.
\newblock In Samy Bengio, Hanna~M. Wallach, Hugo Larochelle, Kristen Grauman, Nicol{\`{o}} Cesa{-}Bianchi, and Roman Garnett, editors, {\em Advances in Neural Information Processing Systems 31: Annual Conference on Neural Information Processing Systems 2018, NeurIPS 2018, December 3-8, 2018, Montr{\'{e}}al, Canada}, pages 3372--3382, 2018.

\bibitem{pang2004sentimental_subj}
Bo~Pang and Lillian Lee.
\newblock A sentimental education: Sentiment analysis using subjectivity summarization based on minimum cuts.
\newblock In {\em Proceedings of the 42nd Annual Meeting of the Association for Computational Linguistics ({ACL}-04)}, pages 271--278, Barcelona, Spain, 2004.

\bibitem{pang2005seeing_mr}
Bo~Pang and Lillian Lee.
\newblock Seeing stars: Exploiting class relationships for sentiment categorization with respect to rating scales.
\newblock In {\em Proceedings of the 43rd Annual Meeting of the Association for Computational Linguistics ({ACL}{'}05)}, pages 115--124, Ann Arbor, Michigan, 2005. Association for Computational Linguistics.

\bibitem{pang2024iterative}
Richard~Yuanzhe Pang, Weizhe Yuan, Kyunghyun Cho, He~He, Sainbayar Sukhbaatar, and Jason Weston.
\newblock Iterative reasoning preference optimization.
\newblock {\em arXiv preprint arXiv:2404.19733}, 2024.

\bibitem{paperno-etal-2016-lambada}
Denis Paperno, Germ{\'a}n Kruszewski, Angeliki Lazaridou, Ngoc~Quan Pham, Raffaella Bernardi, Sandro Pezzelle, Marco Baroni, Gemma Boleda, and Raquel Fern{\'a}ndez.
\newblock The {LAMBADA} dataset: Word prediction requiring a broad discourse context.
\newblock In {\em Proceedings of the 54th Annual Meeting of the Association for Computational Linguistics (Volume 1: Long Papers)}, pages 1525--1534, Berlin, Germany, 2016. Association for Computational Linguistics.

\bibitem{papineni-etal-2002-bleu}
Kishore Papineni, Salim Roukos, Todd Ward, and Wei-Jing Zhu.
\newblock {B}leu: a method for automatic evaluation of machine translation.
\newblock In {\em Proceedings of the 40th Annual Meeting of the Association for Computational Linguistics}, pages 311--318, Philadelphia, Pennsylvania, USA, 2002. Association for Computational Linguistics.

\bibitem{petruck2018representing}
Miriam R.~L. Petruck and Michael~J. Ellsworth.
\newblock Representing spatial relations in {F}rame{N}et.
\newblock In {\em Proceedings of the First International Workshop on Spatial Language Understanding}, pages 41--45, New Orleans, 2018. Association for Computational Linguistics.

\bibitem{pilehvar-camacho-collados-2019-wic}
Mohammad~Taher Pilehvar and Jose Camacho-Collados.
\newblock {W}i{C}: the word-in-context dataset for evaluating context-sensitive meaning representations.
\newblock In {\em Proceedings of the 2019 Conference of the North {A}merican Chapter of the Association for Computational Linguistics: Human Language Technologies, Volume 1 (Long and Short Papers)}, pages 1267--1273, Minneapolis, Minnesota, 2019. Association for Computational Linguistics.

\bibitem{pustejovsky2015semeval}
James Pustejovsky, Parisa Kordjamshidi, Marie-Francine Moens, Aaron Levine, Seth Dworman, and Zachary Yocum.
\newblock {S}em{E}val-2015 task 8: {S}pace{E}val.
\newblock In {\em Proceedings of the 9th International Workshop on Semantic Evaluation ({S}em{E}val 2015)}, pages 884--894, Denver, Colorado, 2015. Association for Computational Linguistics.

\bibitem{putta2024agent}
Pranav Putta, Edmund Mills, Naman Garg, Sumeet Motwani, Chelsea Finn, Divyansh Garg, and Rafael Rafailov.
\newblock Agent q: Advanced reasoning and learning for autonomous ai agents.
\newblock {\em arXiv preprint arXiv:2408.07199}, 2024.

\bibitem{radford2021learning}
Alec Radford, Jong~Wook Kim, Chris Hallacy, Aditya Ramesh, Gabriel Goh, Sandhini Agarwal, Girish Sastry, Amanda Askell, Pamela Mishkin, Jack Clark, Gretchen Krueger, and Ilya Sutskever.
\newblock Learning transferable visual models from natural language supervision.
\newblock In Marina Meila and Tong Zhang, editors, {\em Proceedings of the 38th International Conference on Machine Learning, {ICML} 2021, 18-24 July 2021, Virtual Event}, volume 139 of {\em Proceedings of Machine Learning Research}, pages 8748--8763. {PMLR}, 2021.

\bibitem{radford2018improving}
Alec Radford and Karthik Narasimhan.
\newblock Improving language understanding by generative pre-training.
\newblock {\em arXiv preprint}, 2018.

\bibitem{radford2019language}
Alec Radford, Jeffrey Wu, Rewon Child, David Luan, Dario Amodei, Ilya Sutskever, et~al.
\newblock Language models are unsupervised multitask learners.
\newblock {\em OpenAI blog}, 1(8), 2019.

\bibitem{rafailov2023direct}
Rafael Rafailov, Archit Sharma, Eric Mitchell, Christopher~D Manning, Stefano Ermon, and Chelsea Finn.
\newblock Direct preference optimization: Your language model is secretly a reward model.
\newblock In {\em Thirty-seventh Conference on Neural Information Processing Systems}, 2023.

\bibitem{t5}
Colin Raffel, Noam Shazeer, Adam Roberts, Katherine Lee, Sharan Narang, Michael Matena, Yanqi Zhou, Wei Li, and Peter~J. Liu.
\newblock Exploring the limits of transfer learning with a unified text-to-text transformer.
\newblock {\em J. Mach. Learn. Res.}, 21:140:1--140:67, 2020.

\bibitem{rajpurkar-etal-2016-squad}
Pranav Rajpurkar, Jian Zhang, Konstantin Lopyrev, and Percy Liang.
\newblock {SQ}u{AD}: 100,000+ questions for machine comprehension of text.
\newblock In {\em Proceedings of the 2016 Conference on Empirical Methods in Natural Language Processing}, pages 2383--2392, Austin, Texas, 2016. Association for Computational Linguistics.

\bibitem{rajpurkar2016squad}
Pranav Rajpurkar, Jian Zhang, Konstantin Lopyrev, and Percy Liang.
\newblock {SQ}u{AD}: 100,000+ questions for machine comprehension of text.
\newblock In {\em Proceedings of the 2016 Conference on Empirical Methods in Natural Language Processing}, pages 2383--2392, Austin, Texas, 2016. Association for Computational Linguistics.

\bibitem{ramos2022condita}
Jerome Ramos, To~Eun Kim, Zhengxiang Shi, Xiao Fu, Fanghua Ye, Yue Feng, and Aldo Lipani.
\newblock Condita: A state machine like architecture for multimodal task bots.
\newblock {\em Alexa Prize TaskBot Challenge Proceedings}, 2022.

\bibitem{reddy-etal-2019-coqa}
Siva Reddy, Danqi Chen, and Christopher~D. Manning.
\newblock {C}o{QA}: A conversational question answering challenge.
\newblock {\em Transactions of the Association for Computational Linguistics}, 7:249--266, 2019.

\bibitem{roziere2023code}
Baptiste Roziere, Jonas Gehring, Fabian Gloeckle, Sten Sootla, Itai Gat, Xiaoqing~Ellen Tan, Yossi Adi, Jingyu Liu, Romain Sauvestre, Tal Remez, et~al.
\newblock Code llama: Open foundation models for code.
\newblock {\em arXiv preprint arXiv:2308.12950}, 2023.

\bibitem{ruckle-etal-2021-adapterdrop}
Andreas R{\"u}ckl{\'e}, Gregor Geigle, Max Glockner, Tilman Beck, Jonas Pfeiffer, Nils Reimers, and Iryna Gurevych.
\newblock {AdapterDrop}: {O}n the efficiency of adapters in transformers.
\newblock In {\em Proceedings of the 2021 Conference on Empirical Methods in Natural Language Processing}, pages 7930--7946, Online and Punta Cana, Dominican Republic, 2021. Association for Computational Linguistics.

\bibitem{10.1145/3474381}
Keisuke Sakaguchi, Ronan~Le Bras, Chandra Bhagavatula, and Yejin Choi.
\newblock Winogrande: An adversarial winograd schema challenge at scale.
\newblock In {\em The Thirty-Fourth {AAAI} Conference on Artificial Intelligence, {AAAI} 2020, The Thirty-Second Innovative Applications of Artificial Intelligence Conference, {IAAI} 2020, The Tenth {AAAI} Symposium on Educational Advances in Artificial Intelligence, {EAAI} 2020, New York, NY, USA, February 7-12, 2020}, pages 8732--8740. {AAAI} Press, 2020.

\bibitem{sakaguchi2019winogrande}
Keisuke Sakaguchi, Ronan~Le Bras, Chandra Bhagavatula, and Yejin Choi.
\newblock Winogrande: An adversarial winograd schema challenge at scale.
\newblock In {\em The Thirty-Fourth {AAAI} Conference on Artificial Intelligence, {AAAI} 2020, The Thirty-Second Innovative Applications of Artificial Intelligence Conference, {IAAI} 2020, The Tenth {AAAI} Symposium on Educational Advances in Artificial Intelligence, {EAAI} 2020, New York, NY, USA, February 7-12, 2020}, pages 8732--8740. {AAAI} Press, 2020.

\bibitem{sanhmultitask}
Victor Sanh, Albert Webson, Colin Raffel, Stephen~H. Bach, Lintang Sutawika, Zaid Alyafeai, Antoine Chaffin, Arnaud Stiegler, Arun Raja, Manan Dey, M~Saiful Bari, Canwen Xu, Urmish Thakker, Shanya~Sharma Sharma, Eliza Szczechla, Taewoon Kim, Gunjan Chhablani, Nihal~V. Nayak, Debajyoti Datta, Jonathan Chang, Mike~Tian{-}Jian Jiang, Han Wang, Matteo Manica, Sheng Shen, Zheng~Xin Yong, Harshit Pandey, Rachel Bawden, Thomas Wang, Trishala Neeraj, Jos Rozen, Abheesht Sharma, Andrea Santilli, Thibault F{\'{e}}vry, Jason~Alan Fries, Ryan Teehan, Teven~Le Scao, Stella Biderman, Leo Gao, Thomas Wolf, and Alexander~M. Rush.
\newblock Multitask prompted training enables zero-shot task generalization.
\newblock In {\em The Tenth International Conference on Learning Representations, {ICLR} 2022, Virtual Event, April 25-29, 2022}. OpenReview.net, 2022.

\bibitem{santoro2017simple}
Adam Santoro, David Raposo, David G.~T. Barrett, Mateusz Malinowski, Razvan Pascanu, Peter~W. Battaglia, and Tim Lillicrap.
\newblock A simple neural network module for relational reasoning.
\newblock In Isabelle Guyon, Ulrike von Luxburg, Samy Bengio, Hanna~M. Wallach, Rob Fergus, S.~V.~N. Vishwanathan, and Roman Garnett, editors, {\em Advances in Neural Information Processing Systems 30: Annual Conference on Neural Information Processing Systems 2017, December 4-9, 2017, Long Beach, CA, {USA}}, pages 4967--4976, 2017.

\bibitem{schapiro2017complementary}
Anna~C Schapiro, Nicholas~B Turk-Browne, Matthew~M Botvinick, and Kenneth~A Norman.
\newblock Complementary learning systems within the hippocampus: a neural network modelling approach to reconciling episodic memory with statistical learning.
\newblock {\em Philosophical Transactions of the Royal Society B: Biological Sciences}, 2017.

\bibitem{schick-schutze-2021-exploiting}
Timo Schick and Hinrich Sch{\"u}tze.
\newblock Exploiting cloze-questions for few-shot text classification and natural language inference.
\newblock In {\em Proceedings of the 16th Conference of the European Chapter of the Association for Computational Linguistics: Main Volume}, pages 255--269, Online, 2021. Association for Computational Linguistics.

\bibitem{schick-schutze-2021-just}
Timo Schick and Hinrich Sch{\"u}tze.
\newblock It{'}s not just size that matters: Small language models are also few-shot learners.
\newblock In {\em Proceedings of the 2021 Conference of the North American Chapter of the Association for Computational Linguistics: Human Language Technologies}, pages 2339--2352, Online, 2021. Association for Computational Linguistics.

\bibitem{schlag2018learning}
Imanol Schlag and J{\"{u}}rgen Schmidhuber.
\newblock Learning to reason with third order tensor products.
\newblock In Samy Bengio, Hanna~M. Wallach, Hugo Larochelle, Kristen Grauman, Nicol{\`{o}} Cesa{-}Bianchi, and Roman Garnett, editors, {\em Advances in Neural Information Processing Systems 31: Annual Conference on Neural Information Processing Systems 2018, NeurIPS 2018, December 3-8, 2018, Montr{\'{e}}al, Canada}, pages 10003--10014, 2018.

\bibitem{schulman2017proximal}
John Schulman, Filip Wolski, Prafulla Dhariwal, Alec Radford, and Oleg Klimov.
\newblock Proximal policy optimization algorithms.
\newblock {\em arXiv preprint arXiv:1707.06347}, 2017.

\bibitem{sennrich-etal-2016-edinburgh}
Rico Sennrich, Barry Haddow, and Alexandra Birch.
\newblock {E}dinburgh neural machine translation systems for {WMT} 16.
\newblock In {\em Proceedings of the First Conference on Machine Translation: Volume 2, Shared Task Papers}, pages 371--376, Berlin, Germany, 2016. Association for Computational Linguistics.

\bibitem{shi-etal-2022-learning}
Zhengxiang Shi, Yue Feng, and Aldo Lipani.
\newblock Learning to execute actions or ask clarification questions.
\newblock In Marine Carpuat, Marie-Catherine de~Marneffe, and Ivan~Vladimir Meza~Ruiz, editors, {\em Findings of the Association for Computational Linguistics: NAACL 2022}, pages 2060--2070, Seattle, United States, July 2022. Association for Computational Linguistics.

\bibitem{shi_dont_2023}
Zhengxiang Shi and Aldo Lipani.
\newblock Don{\textquoteright}t stop pretraining? make prompt-based fine-tuning powerful learner.
\newblock In {\em Thirty-seventh Conference on Neural Information Processing Systems}, 2023.

\bibitem{shi2023rethink_data}
Zhengxiang Shi and Aldo Lipani.
\newblock Rethink the effectiveness of text data augmentation: An empirical analysis.
\newblock {\em arXiv preprint arXiv:2306.07664}, 2023.

\bibitem{shi2023dept}
Zhengxiang Shi and Aldo Lipani.
\newblock De{PT}: Decomposed prompt tuning for parameter-efficient fine-tuning.
\newblock In {\em The Twelfth International Conference on Learning Representations}, 2024.

\bibitem{shi2022attention}
Zhengxiang Shi, Pin Ni, Meihui Wang, To~Eun Kim, and Aldo Lipani.
\newblock Attention-based ingredient phrase parser.
\newblock {\em arXiv preprint arXiv:2210.02535}, 2022.

\bibitem{shi2023and}
Zhengxiang Shi, Jerome Ramos, To~Eun Kim, Xi~Wang, Hossein~A Rahmani, and Aldo Lipani.
\newblock When and what to ask through world states and text instructions: Iglu nlp challenge solution.
\newblock {\em arXiv preprint arXiv:2305.05754}, 2023.

\bibitem{shi-etal-2023-lexical}
Zhengxiang Shi, Procheta Sen, and Aldo Lipani.
\newblock Lexical entrainment for conversation systems.
\newblock In {\em Findings of the Association for Computational Linguistics: EMNLP 2023}. Association for Computational Linguistics, 2023.

\bibitem{shi-etal-2023-rethinking}
Zhengxiang Shi, Francesco Tonolini, Nikolaos Aletras, Emine Yilmaz, Gabriella Kazai, and Yunlong Jiao.
\newblock Rethinking semi-supervised learning with language models.
\newblock In {\em Findings of the Association for Computational Linguistics: ACL 2023}, pages 5614--5634, Toronto, Canada, 2023. Association for Computational Linguistics.

\bibitem{10.1007/978-3-031-56027-9_1}
Zhengxiang Shi, Xi~Wang, and Aldo Lipani.
\newblock Self contrastive learning for session-based recommendation.
\newblock In Nazli Goharian, Nicola Tonellotto, Yulan He, Aldo Lipani, Graham McDonald, Craig Macdonald, and Iadh Ounis, editors, {\em Advances in Information Retrieval}, pages 3--20, Cham, 2024. Springer Nature Switzerland.

\bibitem{stepGame2022shi}
Zhengxiang Shi, Qiang Zhang, and Aldo Lipani.
\newblock Stepgame: {A} new benchmark for robust multi-hop spatial reasoning in texts.
\newblock In {\em Thirty-Sixth {AAAI} Conference on Artificial Intelligence, {AAAI} 2022, Thirty-Fourth Conference on Innovative Applications of Artificial Intelligence, {IAAI} 2022, The Twelveth Symposium on Educational Advances in Artificial Intelligence, {EAAI} 2022 Virtual Event, February 22 - March 1, 2022}, pages 11321--11329. {AAAI} Press, 2022.

\bibitem{shi2024understanding}
Zhengyan Shi, Sander Land, Acyr Locatelli, Matthieu Geist, and Max Bartolo.
\newblock Understanding likelihood over-optimisation in direct alignment algorithms.
\newblock {\em arXiv preprint arXiv:2410.11677}, 2024.

\bibitem{shi2024instruction}
Zhengyan Shi, Adam~X. Yang, Bin Wu, Laurence Aitchison, Emine Yilmaz, and Aldo Lipani.
\newblock Instruction tuning with loss over instructions.
\newblock In {\em The Thirty-eighth Annual Conference on Neural Information Processing Systems}, 2024.

\bibitem{shin-etal-2020-autoprompt}
Taylor Shin, Yasaman Razeghi, Robert~L. Logan~IV, Eric Wallace, and Sameer Singh.
\newblock {A}uto{P}rompt: {E}liciting {K}nowledge from {L}anguage {M}odels with {A}utomatically {G}enerated {P}rompts.
\newblock In {\em Proceedings of the 2020 Conference on Empirical Methods in Natural Language Processing (EMNLP)}, pages 4222--4235, Online, 2020. Association for Computational Linguistics.

\bibitem{singh2024aya}
Shivalika Singh, Freddie Vargus, Daniel Dsouza, B{\"o}rje~F Karlsson, Abinaya Mahendiran, Wei-Yin Ko, Herumb Shandilya, Jay Patel, Deividas Mataciunas, Laura OMahony, et~al.
\newblock Aya dataset: An open-access collection for multilingual instruction tuning.
\newblock {\em ArXiv preprint}, abs/2402.06619, 2024.

\bibitem{smolensky1990tensor}
Paul Smolensky.
\newblock Tensor product variable binding and the representation of symbolic structures in connectionist systems.
\newblock {\em Artificial intelligence}, 46(1-2):159--216, 1990.

\bibitem{socher-etal-2013-recursive}
Richard Socher, Alex Perelygin, Jean Wu, Jason Chuang, Christopher~D. Manning, Andrew Ng, and Christopher Potts.
\newblock Recursive deep models for semantic compositionality over a sentiment treebank.
\newblock In {\em Proceedings of the 2013 Conference on Empirical Methods in Natural Language Processing}, pages 1631--1642, Seattle, Washington, USA, 2013. Association for Computational Linguistics.

\bibitem{socher2013recursive_sst-2}
Richard Socher, Alex Perelygin, Jean Wu, Jason Chuang, Christopher~D. Manning, Andrew Ng, and Christopher Potts.
\newblock Recursive deep models for semantic compositionality over a sentiment treebank.
\newblock In {\em Proceedings of the 2013 Conference on Empirical Methods in Natural Language Processing}, pages 1631--1642, Seattle, Washington, USA, 2013. Association for Computational Linguistics.

\bibitem{sohn2020fixmatch}
Kihyuk Sohn, David Berthelot, Nicholas Carlini, Zizhao Zhang, Han Zhang, Colin Raffel, Ekin~Dogus Cubuk, Alexey Kurakin, and Chun{-}Liang Li.
\newblock Fixmatch: Simplifying semi-supervised learning with consistency and confidence.
\newblock In Hugo Larochelle, Marc'Aurelio Ranzato, Raia Hadsell, Maria{-}Florina Balcan, and Hsuan{-}Tien Lin, editors, {\em Advances in Neural Information Processing Systems 33: Annual Conference on Neural Information Processing Systems 2020, NeurIPS 2020, December 6-12, 2020, virtual}, 2020.

\bibitem{NEURIPS2020_1f89885d}
Nisan Stiennon, Long Ouyang, Jeffrey Wu, Daniel Ziegler, Ryan Lowe, Chelsea Voss, Alec Radford, Dario Amodei, and Paul~F Christiano.
\newblock Learning to summarize with human feedback.
\newblock In H.~Larochelle, M.~Ranzato, R.~Hadsell, M.F. Balcan, and H.~Lin, editors, {\em Advances in Neural Information Processing Systems}, volume~33, pages 3008--3021. Curran Associates, Inc., 2020.

\bibitem{su-etal-2022-transferability}
Yusheng Su, Xiaozhi Wang, Yujia Qin, Chi-Min Chan, Yankai Lin, Huadong Wang, Kaiyue Wen, Zhiyuan Liu, Peng Li, Juanzi Li, Lei Hou, Maosong Sun, and Jie Zhou.
\newblock On transferability of prompt tuning for natural language processing.
\newblock In {\em Proceedings of the 2022 Conference of the North American Chapter of the Association for Computational Linguistics: Human Language Technologies}, pages 3949--3969, Seattle, United States, 2022. Association for Computational Linguistics.

\bibitem{sun2019fine}
Chi Sun, Xipeng Qiu, Yige Xu, and Xuanjing Huang.
\newblock How to fine-tune bert for text classification?
\newblock In {\em ArXiv preprint}, volume abs/1905.05583, 2019.

\bibitem{sung2022vl}
Yi-Lin Sung, Jaemin Cho, and Mohit Bansal.
\newblock Vl-adapter: Parameter-efficient transfer learning for vision-and-language tasks.
\newblock In {\em ArXiv preprint}, volume abs/2112.06825, 2021.

\bibitem{sunglst}
Yi-Lin Sung, Jaemin Cho, and Mohit Bansal.
\newblock {LST}: Ladder side-tuning for parameter and memory efficient transfer learning.
\newblock In Alice~H. Oh, Alekh Agarwal, Danielle Belgrave, and Kyunghyun Cho, editors, {\em Advances in Neural Information Processing Systems}, 2022.

\bibitem{NEURIPS2021_cb2653f5}
Yi{-}Lin Sung, Varun Nair, and Colin Raffel.
\newblock Training neural networks with fixed sparse masks.
\newblock In Marc'Aurelio Ranzato, Alina Beygelzimer, Yann~N. Dauphin, Percy Liang, and Jennifer~Wortman Vaughan, editors, {\em Advances in Neural Information Processing Systems 34: Annual Conference on Neural Information Processing Systems 2021, NeurIPS 2021, December 6-14, 2021, virtual}, pages 24193--24205, 2021.

\bibitem{suzgun-etal-2023-challenging}
Mirac Suzgun, Nathan Scales, Nathanael Sch{\"a}rli, Sebastian Gehrmann, Yi~Tay, Hyung~Won Chung, Aakanksha Chowdhery, Quoc Le, Ed~Chi, Denny Zhou, and Jason Wei.
\newblock Challenging {BIG}-bench tasks and whether chain-of-thought can solve them.
\newblock In Anna Rogers, Jordan Boyd-Graber, and Naoaki Okazaki, editors, {\em Findings of the Association for Computational Linguistics: ACL 2023}, pages 13003--13051, Toronto, Canada, 2023. Association for Computational Linguistics.

\bibitem{tan2018source}
Hao Tan and Mohit Bansal.
\newblock Source-target inference models for spatial instruction understanding.
\newblock In Sheila~A. McIlraith and Kilian~Q. Weinberger, editors, {\em Proceedings of the Thirty-Second {AAAI} Conference on Artificial Intelligence, (AAAI-18), the 30th innovative Applications of Artificial Intelligence (IAAI-18), and the 8th {AAAI} Symposium on Educational Advances in Artificial Intelligence (EAAI-18), New Orleans, Louisiana, USA, February 2-7, 2018}, pages 5504--5511. {AAAI} Press, 2018.

\bibitem{alpaca}
Rohan Taori, Ishaan Gulrajani, Tianyi Zhang, Yann Dubois, Xuechen Li, Carlos Guestrin, Percy Liang, and Tatsunori~B. Hashimoto.
\newblock Stanford alpaca: An instruction-following llama model, 2023.

\bibitem{tarvainen2017mean}
Antti Tarvainen and Harri Valpola.
\newblock Mean teachers are better role models: Weight-averaged consistency targets improve semi-supervised deep learning results.
\newblock In Isabelle Guyon, Ulrike von Luxburg, Samy Bengio, Hanna~M. Wallach, Rob Fergus, S.~V.~N. Vishwanathan, and Roman Garnett, editors, {\em Advances in Neural Information Processing Systems 30: Annual Conference on Neural Information Processing Systems 2017, December 4-9, 2017, Long Beach, CA, {USA}}, pages 1195--1204, 2017.

\bibitem{touvron2023llama2}
Hugo Touvron, Louis Martin, Kevin Stone, Peter Albert, Amjad Almahairi, Yasmine Babaei, Nikolay Bashlykov, Soumya Batra, Prajjwal Bhargava, Shruti Bhosale, et~al.
\newblock Llama 2: Open foundation and fine-tuned chat models.
\newblock {\em arXiv preprint arXiv:2307.09288}, 2023.

\bibitem{touvron2023llama}
Hugo Touvron, Louis Martin, Kevin Stone, Peter Albert, Amjad Almahairi, Yasmine Babaei, Nikolay Bashlykov, Soumya Batra, Prajjwal Bhargava, Shruti Bhosale, et~al.
\newblock Llama 2: Open foundation and fine-tuned chat models.
\newblock {\em ArXiv preprint}, abs/2307.09288, 2023.

\bibitem{trischler-etal-2017-newsqa}
Adam Trischler, Tong Wang, Xingdi Yuan, Justin Harris, Alessandro Sordoni, Philip Bachman, and Kaheer Suleman.
\newblock {N}ews{QA}: A machine comprehension dataset.
\newblock In {\em Proceedings of the 2nd Workshop on Representation Learning for {NLP}}, pages 191--200, Vancouver, Canada, 2017. Association for Computational Linguistics.

\bibitem{van2019does}
Betty van Aken, Benjamin Winter, Alexander L{\"{o}}ser, and Felix~A. Gers.
\newblock How does {BERT} answer questions?: {A} layer-wise analysis of transformer representations.
\newblock In Wenwu Zhu, Dacheng Tao, Xueqi Cheng, Peng Cui, Elke~A. Rundensteiner, David Carmel, Qi~He, and Jeffrey~Xu Yu, editors, {\em Proceedings of the 28th {ACM} International Conference on Information and Knowledge Management, {CIKM} 2019, Beijing, China, November 3-7, 2019}, pages 1823--1832. {ACM}, 2019.

\bibitem{NIPS2017_3f5ee243}
Ashish Vaswani, Noam Shazeer, Niki Parmar, Jakob Uszkoreit, Llion Jones, Aidan~N Gomez, \L~ukasz Kaiser, and Illia Polosukhin.
\newblock Attention is all you need.
\newblock In I.~Guyon, U.~Von Luxburg, S.~Bengio, H.~Wallach, R.~Fergus, S.~Vishwanathan, and R.~Garnett, editors, {\em Advances in Neural Information Processing Systems}, volume~30. Curran Associates, Inc., 2017.

\bibitem{vaswani2017attention}
Ashish Vaswani, Noam Shazeer, Niki Parmar, Jakob Uszkoreit, Llion Jones, Aidan~N. Gomez, Lukasz Kaiser, and Illia Polosukhin.
\newblock Attention is all you need.
\newblock In Isabelle Guyon, Ulrike von Luxburg, Samy Bengio, Hanna~M. Wallach, Rob Fergus, S.~V.~N. Vishwanathan, and Roman Garnett, editors, {\em Advances in Neural Information Processing Systems 30: Annual Conference on Neural Information Processing Systems 2017, December 4-9, 2017, Long Beach, CA, {USA}}, pages 5998--6008, 2017.

\bibitem{velivckovic2017graph}
Petar Velickovic, Guillem Cucurull, Arantxa Casanova, Adriana Romero, Pietro Li{\`{o}}, and Yoshua Bengio.
\newblock Graph attention networks.
\newblock In {\em 6th International Conference on Learning Representations, {ICLR} 2018, Vancouver, BC, Canada, April 30 - May 3, 2018, Conference Track Proceedings}. OpenReview.net, 2018.

\bibitem{vogel2010learning}
Adam Vogel and Daniel Jurafsky.
\newblock Learning to follow navigational directions.
\newblock In {\em Proceedings of the 48th Annual Meeting of the Association for Computational Linguistics}, pages 806--814, Uppsala, Sweden, 2010. Association for Computational Linguistics.

\bibitem{voorhees2000building_trec}
Ellen~M Voorhees and Dawn~M Tice.
\newblock Building a question answering test collection.
\newblock In {\em the 23rd annual international ACM SIGIR conference on Research and development in information retrieval}, 2000.

\bibitem{vu-etal-2022-spot}
Tu~Vu, Brian Lester, Noah Constant, Rami Al-Rfou{'}, and Daniel Cer.
\newblock {SP}o{T}: Better frozen model adaptation through soft prompt transfer.
\newblock In {\em Proceedings of the 60th Annual Meeting of the Association for Computational Linguistics (Volume 1: Long Papers)}, pages 5039--5059, Dublin, Ireland, 2022. Association for Computational Linguistics.

\bibitem{wang2019superglue}
Alex Wang, Yada Pruksachatkun, Nikita Nangia, Amanpreet Singh, Julian Michael, Felix Hill, Omer Levy, and Samuel~R. Bowman.
\newblock Superglue: {A} stickier benchmark for general-purpose language understanding systems.
\newblock In Hanna~M. Wallach, Hugo Larochelle, Alina Beygelzimer, Florence d'Alch{\'{e}}{-}Buc, Emily~B. Fox, and Roman Garnett, editors, {\em Advances in Neural Information Processing Systems 32: Annual Conference on Neural Information Processing Systems 2019, NeurIPS 2019, December 8-14, 2019, Vancouver, BC, Canada}, pages 3261--3275, 2019.

\bibitem{wang-etal-2018-glue}
Alex Wang, Amanpreet Singh, Julian Michael, Felix Hill, Omer Levy, and Samuel~R. Bowman.
\newblock {GLUE:} {A} multi-task benchmark and analysis platform for natural language understanding.
\newblock In {\em 7th International Conference on Learning Representations, {ICLR} 2019, New Orleans, LA, USA, May 6-9, 2019}. OpenReview.net, 2019.

\bibitem{wang2024investigating}
Meihui Wang, James Haworth, Huanfa Chen, Yunzhe Liu, and Zhengxiang Shi.
\newblock Investigating the potential of crowdsourced street-level imagery in understanding the spatiotemporal dynamics of cities: A case study of walkability in inner london.
\newblock {\em Cities}, 153:105243, 2024.

\bibitem{wang2021selftuning}
Ximei Wang, Jinghan Gao, Mingsheng Long, and Jianmin Wang.
\newblock Self-tuning for data-efficient deep learning.
\newblock In Marina Meila and Tong Zhang, editors, {\em Proceedings of the 38th International Conference on Machine Learning, {ICML} 2021, 18-24 July 2021, Virtual Event}, volume 139 of {\em Proceedings of Machine Learning Research}, pages 10738--10748. {PMLR}, 2021.

\bibitem{wang2023selfconsistency}
Xuezhi Wang, Jason Wei, Dale Schuurmans, Quoc~V Le, Ed~H. Chi, Sharan Narang, Aakanksha Chowdhery, and Denny Zhou.
\newblock Self-consistency improves chain of thought reasoning in language models.
\newblock In {\em The Eleventh International Conference on Learning Representations}, 2023.

\bibitem{wang2022usb}
Yidong Wang, Hao Chen, Yue Fan, Wang SUN, Ran Tao, Wenxin Hou, Renjie Wang, Linyi Yang, Zhi Zhou, Lan-Zhe Guo, Heli Qi, Zhen Wu, Yu-Feng Li, Satoshi Nakamura, Wei Ye, Marios Savvides, Bhiksha Raj, Takahiro Shinozaki, Bernt Schiele, Jindong Wang, Xing Xie, and Yue Zhang.
\newblock {USB}: A unified semi-supervised learning benchmark for classification.
\newblock In {\em Thirty-sixth Conference on Neural Information Processing Systems Datasets and Benchmarks Track}, 2022.

\bibitem{wang-etal-2023-self-instruct}
Yizhong Wang, Yeganeh Kordi, Swaroop Mishra, Alisa Liu, Noah~A. Smith, Daniel Khashabi, and Hannaneh Hajishirzi.
\newblock Self-instruct: Aligning language models with self-generated instructions.
\newblock In Anna Rogers, Jordan Boyd-Graber, and Naoaki Okazaki, editors, {\em Proceedings of the 61st Annual Meeting of the Association for Computational Linguistics (Volume 1: Long Papers)}, pages 13484--13508, Toronto, Canada, 2023. Association for Computational Linguistics.

\bibitem{wang-etal-2022-super}
Yizhong Wang, Swaroop Mishra, Pegah Alipoormolabashi, Yeganeh Kordi, Amirreza Mirzaei, Atharva Naik, Arjun Ashok, Arut~Selvan Dhanasekaran, Anjana Arunkumar, David Stap, Eshaan Pathak, Giannis Karamanolakis, Haizhi Lai, Ishan Purohit, Ishani Mondal, Jacob Anderson, Kirby Kuznia, Krima Doshi, Kuntal~Kumar Pal, Maitreya Patel, Mehrad Moradshahi, Mihir Parmar, Mirali Purohit, Neeraj Varshney, Phani~Rohitha Kaza, Pulkit Verma, Ravsehaj~Singh Puri, Rushang Karia, Savan Doshi, Shailaja~Keyur Sampat, Siddhartha Mishra, Sujan Reddy~A, Sumanta Patro, Tanay Dixit, and Xudong Shen.
\newblock Super-{N}atural{I}nstructions: Generalization via declarative instructions on 1600+ {NLP} tasks.
\newblock In {\em Proceedings of the 2022 Conference on Empirical Methods in Natural Language Processing}, pages 5085--5109, Abu Dhabi, United Arab Emirates, 2022. Association for Computational Linguistics.

\bibitem{wang2023multitask}
Zhen Wang, Rameswar Panda, Leonid Karlinsky, Rogerio Feris, Huan Sun, and Yoon Kim.
\newblock Multitask prompt tuning enables parameter-efficient transfer learning.
\newblock In {\em The Eleventh International Conference on Learning Representations}, 2023.

\bibitem{warstadt2019neural_cola}
Alex Warstadt, Amanpreet Singh, and Samuel~R. Bowman.
\newblock Neural network acceptability judgments.
\newblock {\em Transactions of the Association for Computational Linguistics}, 7:625--641, 2019.

\bibitem{wei2021finetuned}
Jason Wei, Maarten Bosma, Vincent~Y. Zhao, Kelvin Guu, Adams~Wei Yu, Brian Lester, Nan Du, Andrew~M. Dai, and Quoc~V. Le.
\newblock Finetuned language models are zero-shot learners.
\newblock In {\em The Tenth International Conference on Learning Representations, {ICLR} 2022, Virtual Event, April 25-29, 2022}. OpenReview.net, 2022.

\bibitem{weston2015towards}
Jason Weston, Antoine Bordes, Sumit Chopra, and Tom{\'{a}}s Mikolov.
\newblock Towards ai-complete question answering: {A} set of prerequisite toy tasks.
\newblock In Yoshua Bengio and Yann LeCun, editors, {\em 4th International Conference on Learning Representations, {ICLR} 2016, San Juan, Puerto Rico, May 2-4, 2016, Conference Track Proceedings}, 2016.

\bibitem{wiebe2005annotating_mpqa}
Janyce Wiebe, Theresa Wilson, and Claire Cardie.
\newblock Annotating expressions of opinions and emotions in language.
\newblock {\em Language resources and evaluation}, 39(2-3), 2005.

\bibitem{williams-etal-2018-broad}
Adina Williams, Nikita Nangia, and Samuel Bowman.
\newblock A broad-coverage challenge corpus for sentence understanding through inference.
\newblock In {\em Proceedings of the 2018 Conference of the North {A}merican Chapter of the Association for Computational Linguistics: Human Language Technologies, Volume 1 (Long Papers)}, pages 1112--1122, New Orleans, Louisiana, 2018. Association for Computational Linguistics.

\bibitem{williams2018broad_mnli}
Adina Williams, Nikita Nangia, and Samuel Bowman.
\newblock A broad-coverage challenge corpus for sentence understanding through inference.
\newblock In {\em Proceedings of the 2018 Conference of the North {A}merican Chapter of the Association for Computational Linguistics: Human Language Technologies, Volume 1 (Long Papers)}, pages 1112--1122, New Orleans, Louisiana, 2018. Association for Computational Linguistics.

\bibitem{wingate-etal-2022-prompt}
David Wingate, Mohammad Shoeybi, and Taylor Sorensen.
\newblock Prompt compression and contrastive conditioning for controllability and toxicity reduction in language models.
\newblock In {\em Findings of the Association for Computational Linguistics: EMNLP 2022}, pages 5621--5634, Abu Dhabi, United Arab Emirates, 2022. Association for Computational Linguistics.

\bibitem{pmlr-v202-wu23d}
Bin Wu, Jinyuan Fang, Xiangxiang Zeng, Shangsong Liang, and Qiang Zhang.
\newblock Adaptive compositional continual meta-learning.
\newblock In Andreas Krause, Emma Brunskill, Kyunghyun Cho, Barbara Engelhardt, Sivan Sabato, and Jonathan Scarlett, editors, {\em Proceedings of the 40th International Conference on Machine Learning}, volume 202 of {\em Proceedings of Machine Learning Research}, pages 37358--37378. PMLR, 2023.

\bibitem{10.1145/3609225}
Bin Wu, Zaiqiao Meng, and Shangsong Liang.
\newblock Dynamic bayesian contrastive predictive coding model for personalized product search.
\newblock {\em ACM Trans. Web}, 17(4), 2023.

\bibitem{10.1145/3485447.3512036}
Bin Wu, Zaiqiao Meng, Qiang Zhang, and Shangsong Liang.
\newblock Meta-learning helps personalized product search.
\newblock In {\em Proceedings of the ACM Web Conference 2022}, WWW '22, page 2277–2287, New York, NY, USA, 2022. Association for Computing Machinery.

\bibitem{wu2024understanding}
Bin Wu, Zhengyan Shi, Hossein~A Rahmani, Varsha Ramineni, and Emine Yilmaz.
\newblock Understanding the role of user profile in the personalization of large language models.
\newblock {\em arXiv preprint arXiv:2406.17803}, 2024.

\bibitem{xia2024less}
Mengzhou Xia, Sadhika Malladi, Suchin Gururangan, Sanjeev Arora, and Danqi Chen.
\newblock Less: Selecting influential data for instruction tuning, 2024.

\bibitem{10.5555/3495724.3496249}
Qizhe Xie, Zihang Dai, Eduard~H. Hovy, Thang Luong, and Quoc Le.
\newblock Unsupervised data augmentation for consistency training.
\newblock In Hugo Larochelle, Marc'Aurelio Ranzato, Raia Hadsell, Maria{-}Florina Balcan, and Hsuan{-}Tien Lin, editors, {\em Advances in Neural Information Processing Systems 33: Annual Conference on Neural Information Processing Systems 2020, NeurIPS 2020, December 6-12, 2020, virtual}, 2020.

\bibitem{xie2020self}
Qizhe Xie, Minh{-}Thang Luong, Eduard~H. Hovy, and Quoc~V. Le.
\newblock Self-training with noisy student improves imagenet classification.
\newblock In {\em 2020 {IEEE/CVF} Conference on Computer Vision and Pattern Recognition, {CVPR} 2020, Seattle, WA, USA, June 13-19, 2020}, pages 10684--10695. {IEEE}, 2020.

\bibitem{xu2024wizardlm}
Can Xu, Qingfeng Sun, Kai Zheng, Xiubo Geng, Pu~Zhao, Jiazhan Feng, Chongyang Tao, Qingwei Lin, and Daxin Jiang.
\newblock Wizard{LM}: Empowering large pre-trained language models to follow complex instructions.
\newblock In {\em The Twelfth International Conference on Learning Representations}, 2024.

\bibitem{xu2023rethinking}
Yang Xu, Yongqiang Yao, Yufan Huang, Mengnan Qi, Maoquan Wang, Bin Gu, and Neel Sundaresan.
\newblock Rethinking the instruction quality: Lift is what you need, 2023.

\bibitem{xu2021dash}
Yi~Xu, Lei Shang, Jinxing Ye, Qi~Qian, Yu{-}Feng Li, Baigui Sun, Hao Li, and Rong Jin.
\newblock Dash: Semi-supervised learning with dynamic thresholding.
\newblock In Marina Meila and Tong Zhang, editors, {\em Proceedings of the 38th International Conference on Machine Learning, {ICML} 2021, 18-24 July 2021, Virtual Event}, volume 139 of {\em Proceedings of Machine Learning Research}, pages 11525--11536. {PMLR}, 2021.

\bibitem{xue2023to}
Fuzhao Xue, Yao Fu, Wangchunshu Zhou, Zangwei Zheng, and Yang You.
\newblock To repeat or not to repeat: Insights from scaling {LLM} under token-crisis.
\newblock In {\em Thirty-seventh Conference on Neural Information Processing Systems}, 2023.

\bibitem{xue-etal-2021-mt5}
Linting Xue, Noah Constant, Adam Roberts, Mihir Kale, Rami Al-Rfou, Aditya Siddhant, Aditya Barua, and Colin Raffel.
\newblock m{T}5: A massively multilingual pre-trained text-to-text transformer.
\newblock In {\em Proceedings of the 2021 Conference of the North American Chapter of the Association for Computational Linguistics: Human Language Technologies}, pages 483--498, Online, 2021. Association for Computational Linguistics.

\bibitem{yang2024bayesianrm}
Adam~X Yang, Maxime Robeyns, Thomas Coste, Zhengyan Shi, Jun Wang, Haitham Bou-Ammar, and Laurence Aitchison.
\newblock Bayesian reward models for llm alignment.
\newblock {\em arXiv preprint arXiv:2402.13210}, 2024.

\bibitem{yang2023theory}
Adam~X Yang, Maxime Robeyns, Edward Milsom, Ben Anson, Nandi Schoots, and Laurence Aitchison.
\newblock A theory of representation learning gives a deep generalisation of kernel methods.
\newblock In {\em International Conference on Machine Learning}, pages 39380--39415. PMLR, 2023.

\bibitem{yang2024bayesian}
Adam~X. Yang, Maxime Robeyns, Xi~Wang, and Laurence Aitchison.
\newblock Bayesian low-rank adaptation for large language models.
\newblock In {\em The Twelfth International Conference on Learning Representations}, 2023.

\bibitem{yang2020robust}
Tsung-Yen Yang, Andrew Lan, and Karthik Narasimhan.
\newblock Robust and interpretable grounding of spatial references with relation networks.
\newblock In {\em Findings of the Association for Computational Linguistics: EMNLP 2020}, pages 1908--1923, Online, 2020. Association for Computational Linguistics.

\bibitem{yang-etal-2018-hotpotqa}
Zhilin Yang, Peng Qi, Saizheng Zhang, Yoshua Bengio, William Cohen, Ruslan Salakhutdinov, and Christopher~D. Manning.
\newblock {H}otpot{QA}: A dataset for diverse, explainable multi-hop question answering.
\newblock In {\em Proceedings of the 2018 Conference on Empirical Methods in Natural Language Processing}, pages 2369--2380, Brussels, Belgium, 2018. Association for Computational Linguistics.

\bibitem{yarowsky-1995-unsupervised}
David Yarowsky.
\newblock Unsupervised word sense disambiguation rivaling supervised methods.
\newblock In {\em 33rd Annual Meeting of the Association for Computational Linguistics}, pages 189--196, Cambridge, Massachusetts, USA, 1995. Association for Computational Linguistics.

\bibitem{zellers-etal-2019-hellaswag}
Rowan Zellers, Ari Holtzman, Yonatan Bisk, Ali Farhadi, and Yejin Choi.
\newblock {H}ella{S}wag: Can a machine really finish your sentence?
\newblock In {\em Proceedings of the 57th Annual Meeting of the Association for Computational Linguistics}, pages 4791--4800, Florence, Italy, 2019. Association for Computational Linguistics.

\bibitem{zhang2021flexmatch}
Bowen Zhang, Yidong Wang, Wenxin Hou, Hao Wu, Jindong Wang, Manabu Okumura, and Takahiro Shinozaki.
\newblock Flexmatch: Boosting semi-supervised learning with curriculum pseudo labeling.
\newblock In Marc'Aurelio Ranzato, Alina Beygelzimer, Yann~N. Dauphin, Percy Liang, and Jennifer~Wortman Vaughan, editors, {\em Advances in Neural Information Processing Systems 34: Annual Conference on Neural Information Processing Systems 2021, NeurIPS 2021, December 6-14, 2021, virtual}, pages 18408--18419, 2021.

\bibitem{zhang2022differentiable}
Ningyu Zhang, Luoqiu Li, Xiang Chen, Shumin Deng, Zhen Bi, Chuanqi Tan, Fei Huang, and Huajun Chen.
\newblock Differentiable prompt makes pre-trained language models better few-shot learners.
\newblock In {\em The Tenth International Conference on Learning Representations, {ICLR} 2022, Virtual Event, April 25-29, 2022}. OpenReview.net, 2022.

\bibitem{zhang2018record}
Sheng Zhang, Xiaodong Liu, Jingjing Liu, Jianfeng Gao, Kevin Duh, and Benjamin Van~Durme.
\newblock Record: Bridging the gap between human and machine commonsense reading comprehension.
\newblock {\em ArXiv preprint}, abs/1810.12885, 2018.

\bibitem{zhang2022opt}
Susan Zhang, Stephen Roller, Naman Goyal, Mikel Artetxe, Moya Chen, Shuohui Chen, Christopher Dewan, Mona Diab, Xian Li, Xi~Victoria Lin, et~al.
\newblock Opt: Open pre-trained transformer language models.
\newblock {\em ArXiv preprint}, abs/2205.01068, 2022.

\bibitem{10.5555/2969239.2969312}
Xiang Zhang, Junbo~Jake Zhao, and Yann LeCun.
\newblock Character-level convolutional networks for text classification.
\newblock In Corinna Cortes, Neil~D. Lawrence, Daniel~D. Lee, Masashi Sugiyama, and Roman Garnett, editors, {\em Advances in Neural Information Processing Systems 28: Annual Conference on Neural Information Processing Systems 2015, December 7-12, 2015, Montreal, Quebec, Canada}, pages 649--657, 2015.

\bibitem{zhang-etal-2019-paws}
Yuan Zhang, Jason Baldridge, and Luheng He.
\newblock {PAWS}: Paraphrase adversaries from word scrambling.
\newblock In {\em Proceedings of the 2019 Conference of the North {A}merican Chapter of the Association for Computational Linguistics: Human Language Technologies, Volume 1 (Long and Short Papers)}, pages 1298--1308, Minneapolis, Minnesota, 2019. Association for Computational Linguistics.

\bibitem{zhao2024long}
Hao Zhao, Maksym Andriushchenko, Francesco Croce, and Nicolas Flammarion.
\newblock Long is more for alignment: A simple but tough-to-beat baseline for instruction fine-tuning, 2024.

\bibitem{zhao2023slic}
Yao Zhao, Rishabh Joshi, Tianqi Liu, Misha Khalman, Mohammad Saleh, and Peter~J Liu.
\newblock Slic-hf: Sequence likelihood calibration with human feedback.
\newblock {\em arXiv preprint arXiv:2305.10425}, 2023.

\bibitem{pmlr-v139-zhao21c}
Zihao Zhao, Eric Wallace, Shi Feng, Dan Klein, and Sameer Singh.
\newblock Calibrate before use: Improving few-shot performance of language models.
\newblock In Marina Meila and Tong Zhang, editors, {\em Proceedings of the 38th International Conference on Machine Learning, {ICML} 2021, 18-24 July 2021, Virtual Event}, volume 139 of {\em Proceedings of Machine Learning Research}, pages 12697--12706. {PMLR}, 2021.

\bibitem{zheng2023judging}
Lianmin Zheng, Wei-Lin Chiang, Ying Sheng, Siyuan Zhuang, Zhanghao Wu, Yonghao Zhuang, Zi~Lin, Zhuohan Li, Dacheng Li, Eric Xing, Hao Zhang, Joseph~E. Gonzalez, and Ion Stoica.
\newblock Judging {LLM}-as-a-judge with {MT}-bench and chatbot arena.
\newblock In {\em Thirty-seventh Conference on Neural Information Processing Systems Datasets and Benchmarks Track}, 2023.

\bibitem{zhou2025riot}
Chenyi Zhou, Zhengyan Shi, Yuan Yao, Lei Liang, Huajun Chen, and Qiang Zhang.
\newblock Riot: Efficient prompt refinement with residual optimization tree.
\newblock In {\em Proceedings of the Association for Computational Linguistics (ACL)}, 2025.

\bibitem{zhou2023lima}
Chunting Zhou, Pengfei Liu, Puxin Xu, Srini Iyer, Jiao Sun, Yuning Mao, Xuezhe Ma, Avia Efrat, Ping Yu, LILI YU, Susan Zhang, Gargi Ghosh, Mike Lewis, Luke Zettlemoyer, and Omer Levy.
\newblock {LIMA}: Less is more for alignment.
\newblock In {\em Thirty-seventh Conference on Neural Information Processing Systems}, 2023.

\bibitem{zhu2024can}
Yuchen Zhu, Daniel~Augusto de~Souza, Zhengyan Shi, Mengyue Yang, Pasquale Minervini, Alexander D'Amour, and Matt~J Kusner.
\newblock When can proxies improve the sample complexity of preference learning?
\newblock 2024.

\bibitem{ziegler2019fine}
Daniel~M Ziegler, Nisan Stiennon, Jeffrey Wu, Tom~B Brown, Alec Radford, Dario Amodei, Paul Christiano, and Geoffrey Irving.
\newblock Fine-tuning language models from human preferences.
\newblock {\em arXiv preprint arXiv:1909.08593}, 2019.

\end{thebibliography}

\clearpage
\appendix
\addcontentsline{toc}{chapter}{Appendices}
% \chapter{Rethinking Semi-supervised Learning with Language Models}
% \input{Chapters/paper_acl_findings/appendix}
\chapter{The power of Continued Pre-training}
\clearpage
\section*{Appendix Overview}
The appendix is structured as follows:
\paragraph{Appendix \sect{neurips:sec:dataset}} provides a brief description for each dataset.
\paragraph{Appendix \sect{neurips:sec:templates}} provides details of templates and label words used for each dataset.
\paragraph{Appendix \sect{neurips:appendix:baseline_models}} presents a brief description of state-of-the-art four semi-supervised (self-training) approaches.
% \paragraph{Appendix \sect{neurips:sec:supplementary_experiments}} provides the supplementary experimental results to investigate the potential reasons for the ineffectiveness of \textsc{Cls}-based fine-tuning on the sentence pair tasks.
\paragraph{Appendix \sect{neurips:sec:implementation_details}} provides implementation details and hyperparameters for all comparison methods used in our experiments.

% \input{paper/broader_impact}

% \input{Tables/table_neurips/back_translation}

% \subsection{The effect of different pre-training sizes on sentence pair tasks}
% \label{neurips:sec:larger_pretraining_size}
% \subsection{Training self training from the \pcp checkpoints}
% \label{neurips:sec:train_st}

% \section{Datasets}
% \label{sec:dataset}

\section{Datasets}
\label{neurips:sec:dataset}

\subsection{Data Description for Task Adaptive Pre-training vs Self Training}
\label{appendix:description}
In this section, we briefly introduce \imdb, \sst, \ag, \amazon, %\yelp, 
and \yahoo datasets. Table \ref{table:dataset_example} lists examples for each dataset.

\begin{table*}[ht!]
\centering
\small
\caption{Examples for datasets.}
\label{table:dataset_example}
\begin{adjustbox}{max width=\textwidth}
\begin{tabular}{>{\raggedright}p{3cm}p{15cm}}
\toprule
\textbf Dataset     & \textbf Example  \\ 
\midrule
\imdb           & I watched this movie after seeing other comments on IMDb, even convincing my wife that it was a "unique horror movie." I wanted to like this movie, but was unable to.The "love story" was good, but the horror aspect was quite bad. If the story was just about a young man who fell in love with a girl suffering from parasomnia, then it would have been a better movie.The care centre stretched credulity well past the limits, in fact it was quite ridiculous. The doctor happily ignors privacy laws and professionalism. A nurse goes into a room for a routine feeding of a dangerous patient (without security escort), and drops the tray and runs out of the room screaming for no apparent reason. The forensic patient (and the film's villain) is tied up in a standing position fully clothed - apparently for years? None of it makes much sense.The movie even had some actors that I've liked in other things, such as the detectives, but still I can't recommend this movie. \\
\midrule
\sst            & a rewarding work of art for only the most patient and challenge-hungry moviegoers. \\
\midrule
\ag             & Teen flies in plane \#39;s landing gearA homeless teenager who hid in the landing gear of a passenger plane survived a 700-kilometre flight across south-western China but his companion fell and probably died, state media reported on Friday. \\
\midrule
\amazon         & THIS is MUSIC at its BESTRob Dougan has done it. He's crafted musical perfection, or close to it anyway. I have finally found the music I've been waiting for my whole life in this album - Rob D you are a genius. I think a lot of us wanted to know more about this guy as soon as we heard the track playing to the "Woman in the Red Dress" scene. Now I know why the Wachowski brothers have enlisted his musical talents to flesh out their movies.I know I should be trying to write a more helpful, objective review but I can do nothing but wax poetic for Rob Dougan and his debut album. He has mixed classical melodies with awesome electric beats and it all comes together in an audio orgy. Just buy the album already and let's get Rob some more mainstream recognition. \\
\midrule
\yahoo          & Does anybody know a great deal about angels? I'm looking for names, if they're good or bad, what they look like, etc.  The more detail the better.  All religions accepted \\
\bottomrule
\end{tabular}
\end{adjustbox}
\end{table*}
\paragraph{\imdb.} 
The \imdb dataset \cite{maas-etal-2011-learning} contains a collection of $50\,000$ reviews from the Internet Movie Database, with no more than 30 reviews per movie.
This dataset contains an equal number of positive and negative reviews, yielding a 33\% Marco-$F_1$ score for random guessing. There are $25\,000$ and $25\,000$ for training and testing, respectively. We follow the previous work \cite{wang2022usb} to split the dataset by selecting $12\,500$ samples and $1\,000$ samples per class from the train set to form a train and validation set, respectively.

\paragraph{\sst.} The \sst dataset \cite{wang-etal-2018-glue} consists of sentences from movie reviews and human annotations of their sentiment. The task is to predict the sentiment of a given sentence. Similar to \imdb, this is also a binary classification task. There are $67\,349$ and $872$ for training and testing. We select $60\,000$ and $7\,349$  samples from the train set to form a train and validation set, respectively, where the validation set contains $3\,675$ and $3\,674$ samples for two classes, respectively.

\paragraph{\ag.}
The \ag topic classification dataset is constructed by the previous work \cite{10.5555/2969239.2969312}, where 4 classes are used. Each class contains $30\,000$ training samples and $1\,900$ test samples. We follow the previous work \cite{wang2022usb} to split the dataset by selecting $25\,000$ samples and $2\,500$ samples per class from the train set samples to form a train and validation set, respectively.

\paragraph{\amazon.}
The \amazon dataset \cite{10.1145/2507157.2507163} is a sentiment classification dataset, with five classes. There are $600\,000$ train samples and $130\,000$ test samples per class.  We follow the previous work \cite{wang2022usb} to split the dataset by selecting $50\,000$ samples and $5\,000$ samples per class from the train set samples to form a train and validation set, respectively.

\paragraph{\yahoo.}
The \yahoo dataset \cite{chang2008importance} is a topic classification dataset, with ten classes. There are $140\,000$ train samples and $6\,000$ test samples per class.  We follow the previous work \cite{wang2022usb} to split the dataset by selecting $50\,000$ samples and $5\,000$ samples per class from the train set samples to form a train and validation set, respectively.

\begin{figure*}[t!]
  \centering
  \includegraphics[width=0.6\textwidth]{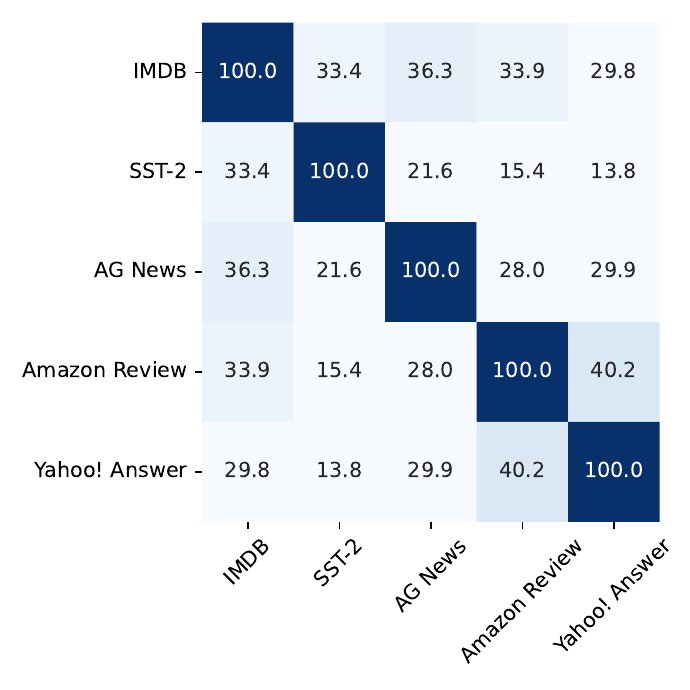}
  \caption{Vocabulary overlap (\%) across datasets.}
  \label{fig:overlap}
\end{figure*}

\paragraph{Dataset Similarity}
\label{appendix:dataset_similarity}
We provide an analysis of the vocabulary overlap of the datasets, as shown in Figure \ref{fig:overlap}. Additionally, in Table \ref{table:overlap_imdb_and_amazon}, we provide some examples to illustrate the overlap between \imdb and \amazon.

As shown in Table \ref{table:dataset_example}, although both the \sst and \imdb datasets are sentiment analysis tasks for movie reviews, the \sst datasets contain shorter and vaguer sentences than the \imdb dataset.

\begin{table*}
\centering
\small
\caption{Similarity analysis between \imdb and \amazon with four examples that highlight the overlap.}
\resizebox{\textwidth}{!}{
\begin{tabular}{>{\raggedright}p{8cm}p{8cm}}
\toprule 
\textbf{\imdb}  & \textbf{\amazon}\\
\midrule 
I loved this \textbf{movie} since I was 7 and I saw it on the opening day. It was so \textbf{touching} and beautiful. I strongly recommend seeing for all. It's a \textbf{movie} to watch with your family by far. My MPAA rating: PG-13 for thematic elements, prolonged scenes of disastor, nudity/sexuality and some language. 
& 
This is a very \textbf{touching}, spiritual \textbf{movie}! When I first saw this film, [...]. I was deeply moved by this motion picture, and the DVD brings the story to your own home. The bonus materials could be better, but the main part of the DVD is the actual \textbf{movie}. Great, great, great film... [...] \\
\midrule
Pacino is over-the-top but to good effect as he's clearly having loads of \textbf{fun}. Beatty is \textbf{great} [...] The lighting, velvet overtones and smog/smoke combine to create a \textbf{great} effect.There are some really \textbf{funny} cameos [...] \textbf{Highly recommended}. 4.5/5 stars. [...] 
& 
Makes a \textbf{great} gift! We bought this book for my dad for Father's Day this year, and thought he would have \textbf{fun} reading it since he has four granddaughters. He loved it and has even selected stories to read to the girls during over-nights with Grandpa and Grandma. I \textbf{highly recommend} it as a \textbf{great} gift. \\
\midrule
The late [...] scripted this tale of \textbf{terror} and it was absolutely one of the \textbf{scariest movies} I ever saw as a kid. (I had to walk MILES just to see a \textbf{movie}, and it was usually dark when I emerged from the theater; seeing a horror \textbf{movie} was always unnerving [...] & Movia ... please .... This \textbf{movie} is a masterpiece of \textbf{terror} \& suspence \& Beautifully filmed \& acted.Comparisons to reality are not allowed when reviewing films of this caliber. Your reaction (though it MAY be \textbf{sarcastic}) is EXACT proof of it's genius! Watch it again...and this time....bask in all it's glory! \\
\midrule
Fabulous actors, beautiful scenery, stark reality [...] I tried to buy the video for several years, finally bought it used from a video store that went out of business. But Yippee! The DVD is now for sale, I purchased it on amazon.com. Not cheap, but \textbf{well worth} it to me. [...]
& 
\textbf{Well worth} the import price. My first impression of this album was a good one, but as time went on it came to grow on me more and more. This is certainly one of the better Costes albums. The mixing is nothing revolutionary, but it is well done and all tracks flow into each other very well. [...]. \\
\bottomrule
\end{tabular}}
\label{table:overlap_imdb_and_amazon}
\end{table*}

\subsection{Data Description for Prompt-based Continued Pre-training}

In this work, we use 21 popular datasets from previous few-shot learning and semi-supervised learning research.

For experiments in \cref{neurips:sec:comparsion_with_conventional_continued_pretraining}, we adhere to the approach in \cite{gao-etal-2021-making} and utilise 16 different datasets\footnote{\url{https://github.com/princeton-nlp/LM-BFF/blob/main/data/download_dataset.sh}}, including SST-2~\cite{socher2013recursive_sst-2}, SST-5~\cite{socher2013recursive_sst-2}, MR~\cite{pang2005seeing_mr}, CR~\cite{hu2004mining_cr}, MPQA~\cite{wiebe2005annotating_mpqa}, Subj~\cite{pang2004sentimental_subj}, TREC~\cite{voorhees2000building_trec}, CoLA~\cite{warstadt2019neural_cola}, 
MNLI~\cite{williams2018broad_mnli}, SNLI~\cite{bowman-etal-2015-large}, QNLI~\cite{rajpurkar2016squad}, RTE~\cite{dagan2005pascal_rte1,giampiccolo2007third_rte3,bentivogli2009fifth_rte4}, MRPC~\cite{dolan2005automatically_mrpc}, QQP\footnote{\url{https://www.quora.com/q/quoradata/}}, 
and STS-B~\cite{cer2017semeval_sts-b}. 
Consistent with prior research \cite{gao-etal-2021-making}, our validation set comprises 16 examples per class from the aforementioned datasets. Additionally, we use 16 examples per class for the training set and the entire training set as the unlabeled set in the semi-supervised setting. We also utilise the full training set for training purposes in the fully supervised setting. For sentence pair tasks, we select at most 10k examples for continued pre-training to reduce the computational costs.

For experiments in \cref{neurips:sec:comparsion_with_self_training}, we follow the setup in \cite{shi-etal-2023-rethinking} and utilise 5 different datasets, including \imdb \cite{maas-etal-2011-learning}, \ag \cite{10.5555/2969239.2969312}, \yelp \footnote{\url{https://www.yelp.com/dataset}}, \yahoo \cite{chang2008importance}, and \amazon \cite{10.1145/2507157.2507163}.
Refer to the dataset statistics in Table~\ref{neurips:appendix:datasets}. Our validation set comprises 1,000 examples for each dataset.

\begin{table*}[ht!]
\centering
\resizebox{\columnwidth}{!}{%
\begin{tabular}{lrrrrll}
\toprule
\multicolumn{7}{c}{\textbf \textit{Single Sentence Tasks}} \\
\midrule
\textbf Dataset & $|\mathcal{Y}|$ & $L$ & \textbf \#Train & \textbf \#Test & \textbf Type & \textbf Labels (classification tasks) \\
\midrule
SST-2 & 2 & 19 & 6,920 & 872 & Sentiment & positive, negative \\ \rowcolor{Gray}
SST-5 & 5 & 18 & 8,544 & 2,210 & Sentiment & v. pos., positive, neutral, negative, v. neg. \\ 
MR & 2 & 20 & 8,662& 2,000 & Sentiment & positive, negative \\ \rowcolor{Gray}
CR & 2 & 19 & 1,775 & 2,000 & Sentiment & positive, negative \\ 
MPQA & 2 & 3 & 8,606 & 2,000 & Opinion Polarity & positive, negative \\ \rowcolor{Gray}
Subj & 2 & 23 & 8,000 & 2,000 & Subjectivity & subjective, objective \\ 
TREC & 6 & 10 & 5,452 & 500 & Question cls. & abbr., entity, description, human, loc., num.\\ \rowcolor{Gray}
CoLA & 2 & 8 & 8,551 & 1,042 & Acceptability & grammatical, not\_grammatical\\ 
\imdb   & 2 & 149 & 8,000 & 1,000 & Movie Review & positive, negative \\ \rowcolor{Gray}
\ag     & 2 & 37 & 8,000 & 1,000 & News Topic & world, sports, business, sci/tech \\ 
\yelp   & 2 & 134 & 8,000 & 1,000 & Review Sentiment & 1, 2, 3, 4, 5 \\ \rowcolor{Gray}
\amazon & 2 & 79 & 8,000 & 1,000 & Review Sentiment & 1, 2, 3, 4, 5 \\ 
\begin{tabular}[l]{@{}c@{}}\yahoo\end{tabular}  & \begin{tabular}[l]{@{}c@{}}2\end{tabular} & \begin{tabular}[l]{@{}c@{}}32\end{tabular} & \begin{tabular}[l]{@{}c@{}}8,000\end{tabular} & \begin{tabular}[l]{@{}c@{}}1,000\end{tabular} & \begin{tabular}[l]{@{}c@{}}Topic Classification\end{tabular} & \begin{tabular}[l]{@{}l@{}}culture, science, health, education, computer,\\sports, business, music, family, politics\end{tabular} \\
\midrule
\multicolumn{7}{c}{\textbf \textit{Sentence Pair Tasks}} \\
\midrule
\textbf Dataset & $|\mathcal{Y}|$ & $L$ & \textbf \#Train & \textbf \#Test & \textbf Type & \textbf Labels (classification tasks) \\
\midrule
MNLI & 3 & 22/11 & 392,702 & 9,815 & NLI & entailment, neutral, contradiction\\ \rowcolor{Gray} 
SNLI & 3 & 14/8 &  549,367 & 9,842 & NLI & entailment, neutral, contradiction \\
QNLI & 2 & 11/30  & 104,743 & 5,463 & NLI & entailment, not\_entailment \\ \rowcolor{Gray} 
RTE & 2 &  49/10 & 2,490 & 277 & NLI &  entailment, not\_entailment \\
MRPC & 2 & 22/21  & 3,668 & 408 & Paraphrase & equivalent, not\_equivalent \\ \rowcolor{Gray} 
QQP & 2 & 12/12 & 363,846 & 40,431 & Paraphrase & equivalent, not\_equivalent  \\
STS-B & $\mathcal{R}$ & 11/11  & 5,749 & 1,500  & Sent. Similarity & - \\ 
\bottomrule
\end{tabular}
}
\caption{The datasets evaluated in this work. $|\mathcal{Y}|$: \# of classes for classification tasks (with one exception: STS-B is a real-valued regression task over the interval $[0, 5]$). $L$: average \# of words in input sentence(s). Note that we only sample  examples from the original training set in our few-shot experiments.}
\label{neurips:appendix:datasets}
\end{table*}

\newcommand\ttt[1]{\texttt{#1}}
\newcommand{\sent}{\ttt{<}$S_1$\ttt{>}}
\newcommand{\firstsent}{\ttt{<}$S_1$\ttt{>}}
\newcommand{\secondsent}{\ttt{<}$S_2$\ttt{>}}
\newcommand{\mask}{\texttt{[MASK]}}

\begin{table*}[!ht]
\centering
\resizebox{\textwidth}{!}{%
\begin{tabular}{lll}
\toprule
\multicolumn{3}{c}{\textbf \textit{Single Sentence Tasks}} \\
\midrule
\tf{Task} & \tf{Template} & \tf{Label words}\\
\midrule
SST-2   &  {\sent} It was {\mask} . & positive: great, negative: terrible\\ \rowcolor{Gray}
SST-5   &  {\sent} It was {\mask} . & v.positive: great, positive: good, neutral: okay,\\ \rowcolor{Gray}
        &                           & negative: bad, v.negative: terrible \\ 
MR      & {\sent} It was {\mask} .  & positive: great, negative: terrible\\ \rowcolor{Gray}
CR      & {\sent} It was {\mask} .  & positive: great, negative: terrible\\
MPQA    & {\sent} is {\mask} .      & positive: positive, negative: negative\\ \rowcolor{Gray}
Subj    & {\sent} This is {\mask} . & subjective: subjective, objective: objective \\ 
TREC    & {\mask} : {\sent}         & abbreviation: Expression, entity: Entity, description: Description \\ 
        &                           & human: Human, location: Location, numeric: Number \\ \rowcolor{Gray}
COLA    & {\sent} This is {\mask} . & grammatical: correct, not\_grammatical: incorrect \\ 
\imdb   & {\sent} It was {\mask} .  & positive: great, negative: terrible \\  \rowcolor{Gray}
\ag     & {\sent} It was {\mask} .  & World: world, Sports:sports, Business: business, Sci/Tech: tech  \\ 
\yelp   & {\sent} It was {\mask} .  & 0: 0, 1: 1, 2: 2, 3: 3, 4: 4  \\  \rowcolor{Gray}
\amazon & {\sent} It was {\mask} .  & 0: 0, 1: 1, 2: 2, 3: 3, 4: 4  \\  
\yahoo  & {\sent} It was {\mask} .  & culture: culture, science: science, health: health, education: education \\
        &                           & computer: computer, sports: sports, business: business \\ 
        &                           & music: music, family: family, politics: politics \\ 
\midrule
\multicolumn{3}{c}{\textbf \textit{Sentence Pair Tasks}} \\
\midrule
\tf{Task} & \tf{Template} & \tf{Label words}\\
\midrule
MNLI  & {\firstsent} ? {\mask} , {\secondsent} & entailment: Yes, neutral: Maybe, contradiction: No \\ \rowcolor{Gray} 
SNLI  & {\firstsent} ? {\mask} , in this case {\secondsent} & entailment: Yes, neutral: Maybe, contradiction: No\\
QNLI  & {\firstsent} ? {\mask} , {\secondsent} & entailment: Yes, not\_entailment: No \\ \rowcolor{Gray} 
RTE   & {\firstsent} ? {\mask} , I think that {\secondsent} & entailment: Clearly, not\_entailment: Yet \\
MRPC  & {\firstsent} {\mask} , {\secondsent} & equivalent: Yes, not\_equivalent: No\\ \rowcolor{Gray}
QQP   & {\firstsent} {\mask} , {\secondsent} & equivalent: Yes, not\_equivalent: No\\ 
STS-B & {\firstsent} {\mask} , {\secondsent} & $y_u$: Yes, $y_l$: No \\
\bottomrule
\end{tabular}
}
\caption{Templates and label words used for ``Prompt-based \ft (hard)''. We use the STS-2 and STS-B template for all single sentence tasks and sentence pair tasks using ``Prompt-based \ft (soft)'', respectively.
}
\label{neurips:table:template}
\end{table*}

\section{Templates for Prompt-based Fine-tuning}
\label{neurips:sec:templates}
Here we introduce the templates used in two state-of-the-art prompt-based \ft approaches for each dataset.
For ``Prompt-based \ft (hard)'', we use high-quality manual or auto-generated prompts and label words for each task from previous works \cite{schick-schutze-2021-exploiting,gao-etal-2021-making}.
For ``Prompt-based \ft (soft)'', we use the STS-2 template for all single sentence tasks and STS-B template for all sentence pair tasks, while the label words for each task follow the description in Table \ref{neurips:table:template}.

\section{Self Training Frameworks}
\label{neurips:appendix:baseline_models}
\paragraph{\vat.} \vat \cite{miyato2018virtual} proposed a regularization technique that forces pairs of data points that are very close in the input space to be close to each other in the output space. \vat adds a small perturbation to the input data and forces the model to produce similar predictions.

\paragraph{\fixmatch.} \fixmatch \cite{sohn2020fixmatch} generates artificial labels using both consistency regularization and pseudo-labelling, where the artificial labels are produced based on weakly-augmented unlabelled data. These artificial labels are then used as targets to train the model on strongly-augmented unlabelled data. \fixmatch only retains an artificial label if the model assigns a high probability to one of the possible classes.

\paragraph{\dash.} \dash \cite{xu2021dash} extends \fixmatch by introducing a mechanism with a dynamically adjusted threshold of loss to select a subset of training examples from the unlabelled data for performing \ssl.

\paragraph{\flexmatch.} \flexmatch \cite{zhang2021flexmatch} also extends \fixmatch by introducing the concept of curriculum learning \cite{bengio2009curriculum} to flexibly adjust thresholds for different classes at each time step and select unlabelled data and their pseudo labels that are more likely to be informative.

\paragraph{\adamatch.} \adamatch \cite{berthelot2021adamatch} aims to solve domain adaptation problems in \ssl and build a high-accuracy model that trains on and tests on different data distributions. \adamatch builds on \fixmatch and introduces a relative confidence threshold and a modified distribution alignment from \cite{berthelot2019remixmatch}.

\section{Implementation Details}
\label{neurips:sec:implementation_details}

Our code is implemented using Pytorch\footnote{\url{https://pytorch.org/}} and Huggingface\footnote{\url{https://huggingface.co/}}.
The semi-supervised approaches are implemented upon the repository\footnote{\url{https://github.com/amzn/pretraining-or-self-training}}.
Below, we provide a comprehensive list of the hyperparameters used in our code.
For fine-tuning, as shown in Table \ref{neurips:table:pft_hyperparameters}, we conduct a grid search for learning rates within the set \{1e-5, 2e-5, 5e-5\}, and choose a batch size of 8. In each trial, we train the model for 1,000 steps, evaluate performance every 100 steps, and select the best checkpoint based on optimal performance on the evaluation set. The best performance is determined by the relevant evaluation metric.
For continued pre-training, we utilise the same set of hyperparameters for both \tapt and \pcp, as shown in Table \ref{neurips:table:tapt_hyperparameters}.
The learning rate and unlabeled data size are closely linked and need to be adjusted simultaneously. As a general guideline, we suggest decreasing the learning rate as the unlabeled data size decreases. In contrast to its predecessor, \textsc{Bert} \cite{devlin2018bert}, which uses the next sentence prediction objective,  \roberta \cite{liu2019roberta} is trained solely with the masked language model (MLM) objective, specifically the cross-entropy loss on predicting randomly masked tokens. RoBERTa dynamically alters the masking pattern applied to training examples, typically employing a masking probability of 0.15. 
Additionally, Table \ref{neurips:table:self_training} lists the hyperparameters for self-training approaches, where a grid search for learning rates within the set \{1e5, 2e-5, 5e-5\} is conducted.
\clearpage
\begin{table*}[!ht]
    \centering
    \small
    \begin{tabular}{cc}
        \toprule
        \textbf{Hyperparameter} & \textbf{Assignment}  \\
        \midrule
        number of steps & 1000 steps (evaluate every 100 steps)\\
        \midrule
        batch size & 8 \\
        \midrule
        maximum learning rate & 1e-05, 2e-5, 5e-5 \\
        \midrule
        maximum sequence length & 128, 256 \\
        \midrule
        learning rate optimizer & AdamW \\
        \midrule
        Adam epsilon & 1e-6 \\
        \midrule
        Adam beta weights & 0.9, 0.98\\
        \midrule
        learning rate scheduler & Warmup linear \\
        \midrule
        Weight decay & 0.01 \\
        \midrule
        Warmup proportion & 0.06 \\
        \bottomrule
    \end{tabular}
    \caption{Hyperparameters for hard and soft prompt-based fine-tuning.} 
    \label{neurips:table:pft_hyperparameters}
\end{table*}
\begin{table*}[!ht]
    \centering
    \small
    \begin{tabular}{cc}
        \toprule
        \textbf{Hyperparameter} & \textbf{Assignment}  \\
        \midrule
        number of steps & 100 epochs\\
        \midrule
        batch size & 256 \\
        \midrule
        maximum learning rate & 1e-05, 1e-4 \\
        \midrule
        learning rate optimizer & AdamW \\
        \midrule
        Adam epsilon & 1e-6 \\
        \midrule
        Adam beta weights & 0.9, 0.98\\
        \midrule
        learning rate scheduler & Warmup linear \\
        \midrule
        Weight decay & 0.01 \\
        \midrule
        Warmup proportion & 0.06 \\
        \midrule
        Masking Probability & 0.15 \\
        \bottomrule
    \end{tabular}
    \caption{Hyperparameters for both conventional continued pre-training (\tapt) and prompt-based conventional fine-tuning (\pcp).
    } 
    \label{neurips:table:tapt_hyperparameters}
\end{table*}

\begin{table*}[!ht]
    \centering
    \small
    \begin{tabular}{cc}
        \toprule
        \textbf{Hyperparameter} & \textbf{Assignment}  \\
        \midrule
        number of steps & $12\,800$ or $25\,600$ steps \\
        \midrule
        batch size & 16 \\
        \midrule
        learning rate & 1e-05, 2e-05, 5e-05 \\
        \midrule
        learning rate optimizer & AdamW \\
        \midrule
        maximum sequence length & 256 \\
        \midrule
        learning rate scheduler & Warmup linear \\
        \midrule
        Warmup proportion & 0.05 \\
        \midrule
        learning rate decay & linear \\
        \bottomrule
    \end{tabular}
    \caption{Hyperparameters for self training. Algorithm-specific hyperparameters will be released in configuration files with the code.} 
    \label{neurips:table:self_training}
\end{table*}

\chapter{Decomposed Prompt Tuning for Parameter-Efficient Fine-tuning}
\clearpage
\section*{Appendix Overview}
The appendix is structured as follows:
\paragraph{Appendix \sect{iclr:sec:sub_figure}} provides a visualization of the model performance against the number of trainable parameters on the GLUE and SuperGLUE benchmarks.
\paragraph{Appendix \sect{iclr:sec:llama}} presents the additional experimental results, including using a larger size of language models (\llama and TB-3B) and testing the impact of different lengths of soft prompts.
\paragraph{Appendix \sect{iclr:sec:dataset}} provides a brief description of all datasets used in this work.
\paragraph{Appendix \sect{iclr:sec:implementation_details}} provides implementation details and hyperparameters for all comparison methods used in our experiments.

\section{Model performance against the parameter-efficiency}
\label{iclr:sec:sub_figure}
\begin{figure*}[!th]
% \vspace{-2mm}
\includegraphics[width=\textwidth]{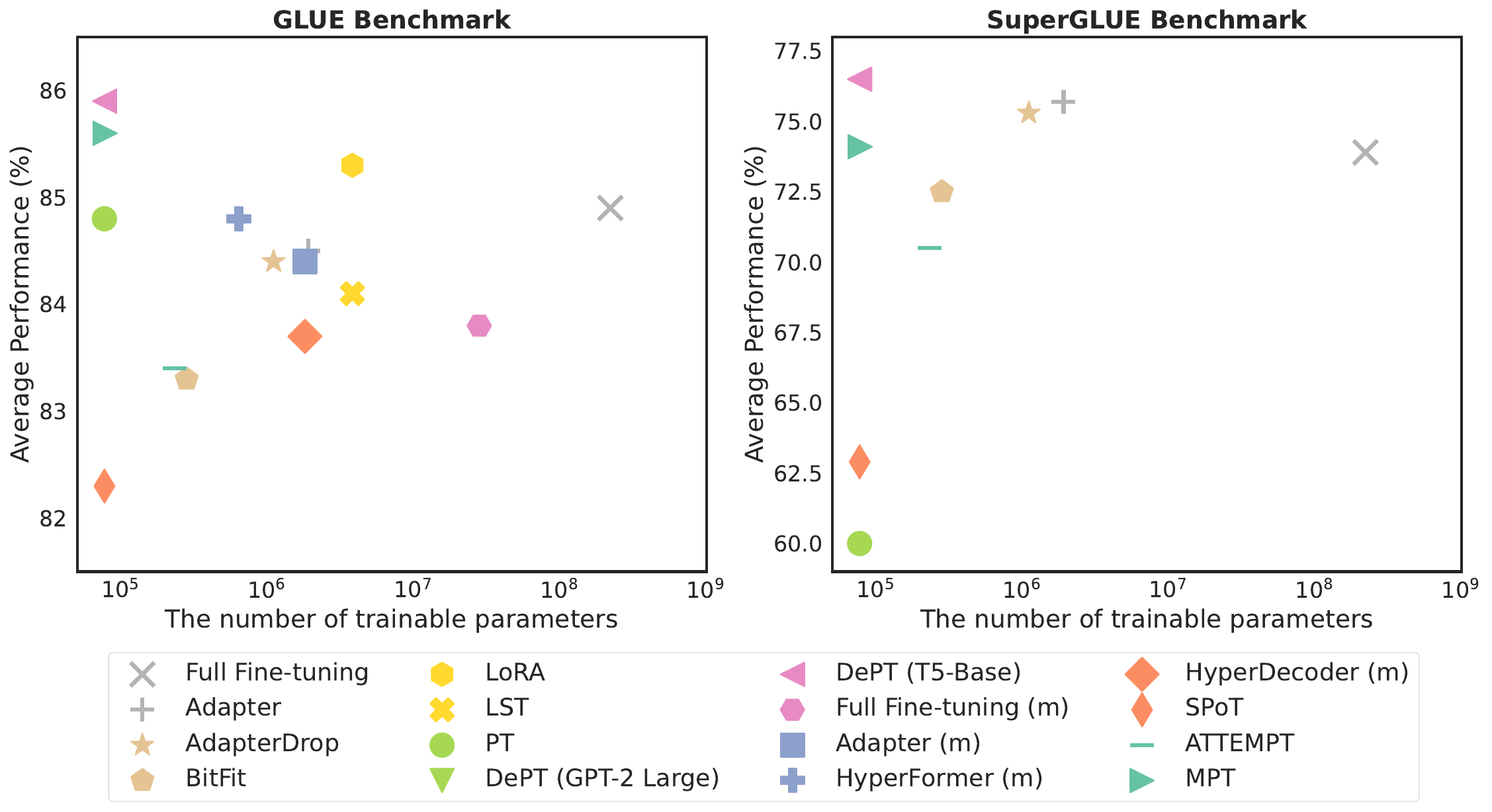}
\caption{
The average performance against the number of trainable parameters on the GLUE and SuperGLUE benchmark using the \tfbase model.
}
\label{iclr:fig:performance_vs_efficient}
\end{figure*}

We visualize the experimental results in Table \ref{iclr:table:main_results_glue}, as shown in Figure \ref{iclr:fig:performance_vs_efficient}. 
 The visualization shows that our proposed method \dept outperforms other \peft approaches and full fine-tuning baselines on the GLUE and SuperGLUE benchmark (y-axis) while updating only a small number of trainable parameters (x-axis).

\section{Additional Experiments}
\label{iclr:sec:llama}

\paragraph{Different prompt lengths.}
We have performed additional experiments regarding different prompt lengths, as shown in the Table below. Specifically, we have increased the size of trainable parameters in both \dept and \pt by a factor of two. We use the \tfbase as the backbone. 
As shown in Table \ref{iclr:table:different_prompt_length}, we report the average performance on the Glue dataset as well as inference speed, measured in inference samples per second. Our findings indicate that \dept (m=120, r=60) outperforms \pt in terms of inference speed by 34\%. We believe that this performance advantage can be further enhanced by reducing the value of $m$, which represents the length of the soft prompt. To provide a concrete example, on the SST-2 dataset, \dept can achieve an inference speed of 77.2 samples per second, while \pt can only infer 57.4 samples per second. This suggests the advantage of \dept over \pt increases as the model size increases.

\begin{table}[!ht]
\centering
% \footnotesize
% \small
\resizebox{0.75\columnwidth}{!}{
\begin{tabular}{lcc}
\toprule
\textbf Method                & \textbf Average Glue Performance   & \textbf Inference samples per second \\   
\midrule
\dept (m=120, r=60)        &  86.0 &	 54.8 \\  
PT (m=200)                 &  85.2 &	 40.8 \\
\bottomrule
\end{tabular}
}
\caption{
The impact of using longer soft prompt length. 
Test results using \tfbase on the Glue Benchmark.
}
\label{iclr:table:different_prompt_length}
\end{table}

\section{Dataset}
\label{iclr:sec:dataset}
In this work, we use 23 popular datasets from previous few-shot learning and \peft research.
We limit the maximum training data number of Yelp-2 to 100k samples.
We train MNLI with longer steps, 200k steps in total.
For the GLUE dataset, we use the HuggingFace dataset\footnote{\url{https://huggingface.co/datasets/glue}}.
For the Super GLUE dataset, we use the HuggingFace dataset\footnote{\url{https://huggingface.co/datasets/super_glue}}.
For MRQA 2019 Shared Task and other datasets, we use the HuggingFace dataset\footnote{\url{https://huggingface.co/lucadiliello}}.

\begin{table}[!ht]
\centering
\resizebox{\columnwidth}{!}{%
\begin{tabular}{lrrrrrl}
\toprule
\multicolumn{7}{c}{\textbf \textit{GLUE Benchmark}} \\
\midrule
\textbf Dataset & \textbf Source & \textbf Target & \textbf \#Train & \textbf \#Valid & \textbf \#Test & \textbf Type  \\
\midrule
MNLI  & 31.8 & 1.0  & 392,702 & 9,832 & 9,815  & NLI \\ \rowcolor{Gray} 
QQP   & 24.1 & 1.0  & 362,846 & 1,000 & 40,431 & Paraphrase \\
QNLI  & 38.4 & 1.0  & 103,743 & 1,000 & 5,463  & NLI \\ \rowcolor{Gray} 
SST-2 & 10.4 & 1.0     & 66,349  & 1,000 & 872    & Sentiment  \\ 
STS-B & 21.9 & 1.0  & 5,749 & 750 & 750  & Sent. Similarity \\ \rowcolor{Gray}
MRPC  & 45.9 & 1.0  & 3,668 & 204 & 204 & Paraphrase \\ 
RTE   & 54.4 & 1.0  & 2,490 & 138 & 139 & NLI \\ \rowcolor{Gray} 
CoLA  & 8.7 & 1.0  & 8,551 & 521 & 522 & Acceptability \\ 
\midrule
\multicolumn{7}{c}{\textbf \textit{SuperGLUE Benchmark}} \\
\midrule
\textbf Dataset & \textbf Source & \textbf Target & \textbf \#Train & \textbf \#Valid & \textbf \#Test & \textbf Type \\
\midrule
MultiRC     & 286.1 & 1.0 & 27,243 & 2,424 & 2,424 & Question Answering \\ \rowcolor{Gray}
BoolQ       & 108.3 & 1.0 & 9,427  & 1,635 & 1,635 & Question Answering \\ 
WiC         & 18.4 & 1.0 & 5,428  & 319   & 319 & Word Sense Disambiguation \\ \rowcolor{Gray}
WSC         & 28.1 & 1.0 & 554    & 52    & 52 & Common Sense Reasoning \\ 
CB          & 64.6 & 1.0 & 250    & 28    & 28 & NLI \\ \rowcolor{Gray}
ReCoRD      & 210.7 & 1.5 & 137,484 & 1,370 & 15,176 & Common Sense Reasoning \\
\midrule
\multicolumn{7}{c}{\textbf \textit{MRQA 2019 Shared Task}} \\
\midrule
\textbf Dataset & \textbf Source & \textbf Target & \textbf \#Train & \textbf \#Valid & \textbf \#Test & \textbf Type \\
\midrule
NaturalQuestions & 242.7 & 4.5 & 103,071 & 1,000 & 12836 & Question Answering \\ \rowcolor{Gray}
HotpotQA         & 225.7 & 2.6 & 71,928  & 1,000 & 5,901 & Question Answering \\ 
SearchQA         & 942.8 & 2.0 & 116,384 & 1,000 & 16,980 & Question Answering \\ \rowcolor{Gray}
NewsQA           & 615.5 & 5.1 & 73,160  & 1,000 & 4,212 &  Question Answering \\ 
\midrule
\multicolumn{7}{c}{\textbf \textit{Other Datasets}} \\
\midrule
\textbf Dataset & Source & Target & \textbf \#Train & \textbf \#Valid & \textbf \#Test & \textbf Type \\
\midrule
WinoGrande   & 23.8 & 1.0 & 39,398  & 1,000 & 1,267  & Common Sense Reasoning \\ \rowcolor{Gray}
YelpPolarity & 134.0 & 1.0 & 100,000 & 1,000 & 38,000 & Sentiment \\ 
SciTail      & 30.8 & 1.0 & 23,596  & 652   & 652    & NLI \\ \rowcolor{Gray}
PAWS         & 44.7 & 1.0 & 4,9401  & 8,000 & 8,000  & Sent. Similarity \\ 
\midrule
\multicolumn{7}{c}{{\textbf \textit{Vision Language Tasks}} (\#Images \& \#Texts)} \\
\midrule
Visual Question Answering   & - & - & 113.2K/605.1K & 5.0K/26.7K & 5.0K/26.3K  & Question Answering \\ \rowcolor{Gray}
MS CoCo Caption & - & - & 113.2K/566.8K & 5.0K/5.0K & 5.0K/5.0K & Caption Generation \\ 

\bottomrule
\end{tabular}
}
\caption{The datasets evaluated in this work. 
Source indicates the average length of the source sentences in the training set.
Target indicates the average length of the target sentences in the training set.
STS-B is a real-valued regression task over the interval $[0, 5]$). 
% $L$: average \# of words in input sentence(s). 
Note that we only sample examples from the original training set in our few-shot experiments.}
\label{iclr:appendix:datasets}
\end{table}

\section{Implementation Details}
\label{iclr:sec:implementation_details}
\begin{table*}[!th]
    \centering
    \small
    \begin{tabular}{cc}
        \toprule
        \textbf{Hyperparameter} & \textbf{Assignment}  \\
        \midrule
        number of steps & 30,000 steps (evaluate every 1,000 steps)\\
        \midrule
        batch size & 16 \\
        \midrule
        maximum learning rate ($\alpha_1$) & 3e-1, 4e-1, 5e-1 \\
        \midrule
        maximum learning rate ($\alpha_2$) & 1e-04, 5e-4, 1e-03 \\
        \midrule
        length of the soft prompt ($m$) & 20, 40, 60, 80 \\
        \midrule
        maximum sequence length & 256 \\
        \midrule
        learning rate optimizer & AdamW \\
        \midrule
        Adam epsilon & 1e-6 \\
        \midrule
        Adam beta weights & 0.9, 0.98\\
        \midrule
        learning rate scheduler & Warmup linear \\
        \midrule
        Weight decay & 0.01 \\
        \midrule
        Warmup proportion & 0.06 \\
        \bottomrule
    \end{tabular}
    \caption{Hyperparameters for Prompt Tuning and \dept.} 
    \label{iclr:table:pft_hyperparameters}
\end{table*}
Our code is implemented using Pytorch\footnote{\url{https://pytorch.org/}}, Huggingface Transformers\footnote{\url{https://github.com/huggingface/transformers}},
and Huggingface \peft \footnote{\url{https://github.com/huggingface/peft}}.
Below, we provide a comprehensive list of the hyperparameters used in our code.
In our work, we mainly cite the experimental results from the previous works \cite{asai-etal-2022-attempt,wang2023multitask,sunglst}. 
In addition, we train LoRA with up to 200k steps.
We search the learning rate within the set
\{5e-4, 1e-4, 5e-5, 1e-5\}. 
We set the rank as 35.
We choose a batch size of 32. 
We find that training LoRA on the MRQA dataset presents challenges, despite conducting a thorough search for optimal learning rates and training steps. The reasons for these difficulties remain uncertain.
For prompt tuning and \dept, as shown in Table \ref{iclr:table:pft_hyperparameters}, we conduct a grid search for learning rates.
For the soft prompt, we search the learning rate within the set \{3e-1, 4e-1, 5e-1\}.
For the low-rank matrice pairs, we search the learning rate within the set
\{1e-04, 5e-4, 1e-03, 5e-03\}.
We choose a batch size of 16. 
We typically use the max sequence length as 256 except for the SuperGLUE-MultiRC, where the max sequence length is 348.
In each trial, we train the model for 30,000 steps, evaluate performance every 1,000 steps, and select the best checkpoint based on optimal performance on the evaluation set. 
For the large dataset with more than 100,000 training examples, we follow the prior work \citep{vu-etal-2022-spot} to train the vanilla \pt and our proposed method \dept with up to 300,000 steps.
Training more steps helps improve the performance of the vanilla \pt for the large dataset.
The best performance is determined by the relevant evaluation metric.
We train the T5 model from the original checkpoint rather than the LM-adapted 1.1 version \citep{lester-etal-2021-power}.

\section{Further Discussion}
\label{iclr:sec:appendix_related}
\paragraph{Intuition.}
The intuition of \dept is that (1) given the same number of trainable parameters, allowing some updates for word embeddings will improve the performance; and (2) a shorter soft prompt will improve the efficiency. 
To illustrate, the previous study \citep{wingate-etal-2022-prompt} has shown that a soft prompt can interpolate between many token embeddings, enabling the representation of more abstract concepts compared to relying on a single discrete token. 
However, the soft prompt in the \pt is consistently added at the beginning of the frozen word embedding. In contrast, we propose \dept, which decomposes the long soft prompt into a short soft prompt and a pair of low-rank matrices. This approach can 
(1) reduce the length of the soft prompt for better efficiency; and 
(2) permit representation updates within the frozen word embedding, thereby increasing the adaptability of input representations that were previously unavailable.

\paragraph{Related works with similar titles.}
The meaning of “compose” and the method are fundamentally different between previous works \citep{khot2022decomposed,nayak2022learning} and our work. Specifically, Decomposed Promptin \citep{khot2022decomposed} focuses on in-context learning, without the need to update parameters. Decomposed Prompting aligns closely with the work of chain-of-thoughts and self-consistency. In addition, CSP \citep{nayak2022learning} treats the attributes and objects that are composed to define classes as learnable tokens within the vocabulary. In contrast, our proposed method DePT does not train soft prompts associated with any vocabulary token, nor does it add additional tokens to the vocabulary. The main goal of DePT is to improve the efficiency of Prompt Tuning (PT) due to the increased input sequence issue.

\chapter{Instruction Tuning With Loss Over Instructions}
\clearpage
\section*{Appendix Overview}
The appendix is structured as follows:
\paragraph{Appendix \sect{neurips2024:sec:sft_dataset}} provides a brief description (with statistical summaries) for instruction tuning datasets.
\paragraph{Appendix \sect{neurips2024:sec:evaluation}} provides details of evaluation benchmarks and settings.
% \paragraph{Appendix \sect{neurips2024:appendix:baseline_models}} presents a brief description of state-of-the-art four semi-supervised (self-training) approaches.
\paragraph{Appendix \sect{neurips2024:sec:implementation_details}} provides the experimental settings, implementation details and hyperparameters for all comparison methods used in our experiments.
% \paragraph{Appendix \sect{neurips2024:sec:train_test_loss}} provides the supplementary experimental results to investigate the effect of our approach on training and testing losses.
% \paragraph{Appendix \sect{neurips2024:sec:win_rate_vs_epoch}} provides the supplementary experimental results to investigate the relationship between the win rate on the AlpacaEval 1.0 and the number of epochs.
% \paragraph{Appendix \sect{neurips2024:sec:kl}} provides the mathematical formula for the Kullback-Leibler (KL) divergence used in our paper.
% \paragraph{Appendix \sect{neurips2024:sec:analysis_output_length_sah}} provides the supplementary experimental results to investigate the relationship between the output length and the number of epochs.

\section{Instruction Tuning Dataset}
\label{neurips2024:sec:sft_dataset}
\begin{figure}[ht]
\centering
\includegraphics[width=\textwidth]{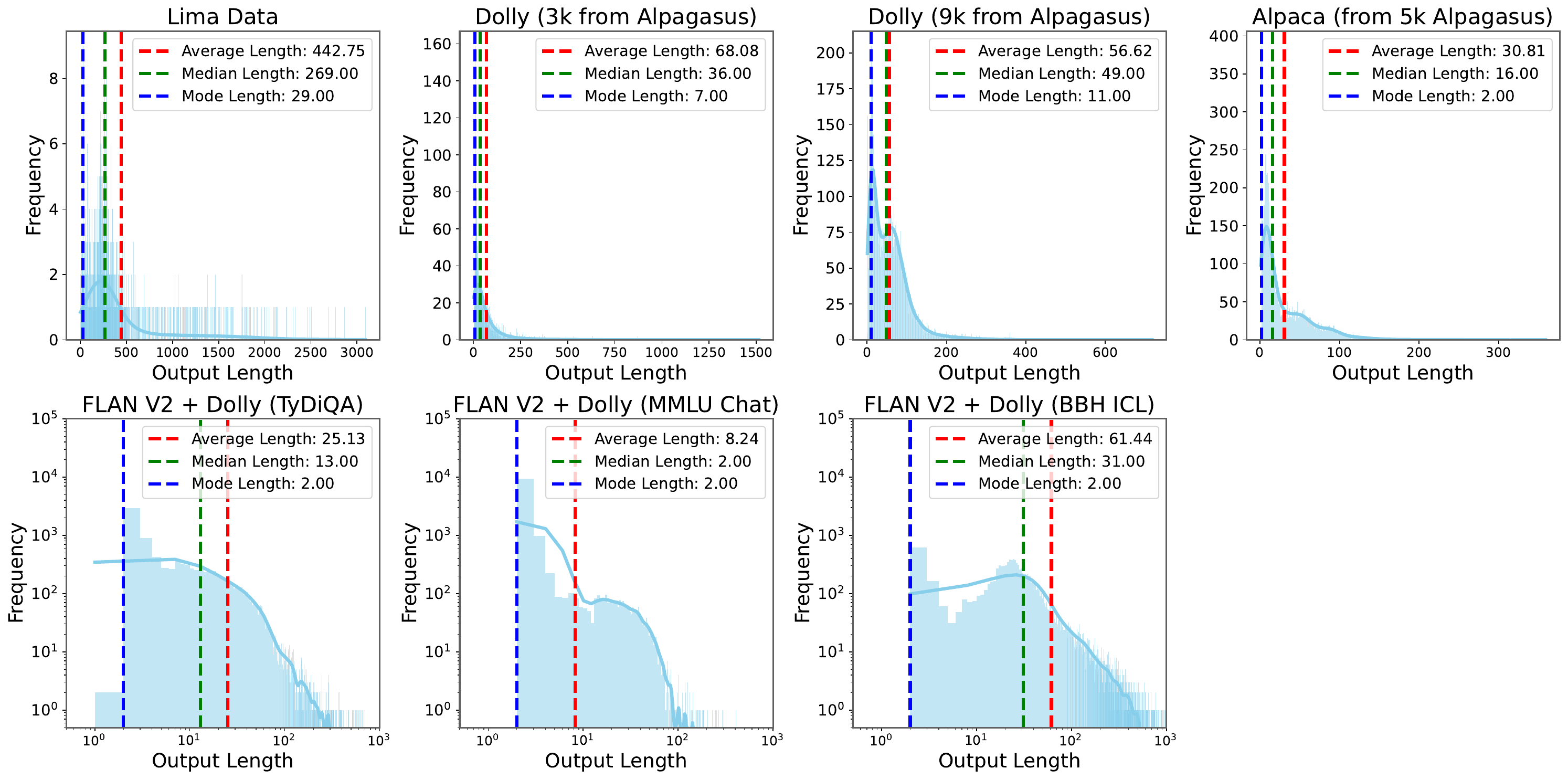}
\caption{
Distribution of output lengths of instruction tuning datasets.
This figure presents histograms for the distribution of output lengths across seven datasets, including \lima, \alpagasusdollyone, \alpagasusdollytwo, \alpagasusalpaca, \mmluchat, \tydiqa, and \bbhicl. 
Each subplot displays the frequency of output lengths with key statistical indicators: the average (red dashed line), median (green dashed line), and mode (blue dashed line) of each dataset. 
The last three subplots employ a logarithmic scale on both axes to better illustrate data spread.
}
\label{neurips2024:fig:length_distribution}
\end{figure}

In this work, we use 13 popular datasets from previous instruction-tuning research.
For the \wizardlm, \sharegpt, \science, and \codealpaca datasets, we directly use the subset provided by the previous work \cite{ivison2023camels}.
Refer to the dataset statistics in Table~\ref{neurips2024:table:new_datasets_summary}. 
In addition, we provide an analysis of the output length distribution for \lima, \alpagasusdollyone, \alpagasusdollytwo, \alpagasusalpaca, \mmluchat, \tydiqa, and \bbhicl datasets, as shown in Figure \ref{neurips2024:fig:length_distribution}.

\begin{table}[!th]
\centering
\caption{Statistical summary for various instruction tuning datasets. 
The table includes sample sizes, the average total length of instructions and outputs, the average output length, and the average instruction length with their standard deviations, and ratio calculations.}
\label{neurips2024:table:new_datasets_summary}
\resizebox{\textwidth}{!}{%
\begin{tabular}{lrrrrrrrr}
\toprule
\textbf Dataset & \textbf Size & \textbf Total & \textbf Output & \textbf Output Std & \textbf Instruction & \textbf Instruction Std & \textbf Output/Instruction & \textbf Instruction/Output \\
\midrule
\lima & 1,030 & 484.47 & 442.75 & 491.34 & 41.72 & 79.28 & 10.6124 & 0.0942 \\ \rowcolor{Gray}
\mmluchat & 13,533 & 225.19 & 8.24 & 16.42 & 216.95 & 301.64 & 0.0380 & 26.3316 \\
\tydiqa & 13,533 & 172.44 & 25.13 & 42.62 & 147.31 & 235.37 & 0.1706 & 5.862 \\ \rowcolor{Gray}
\bbhicl & 13,533 & 262.03 & 61.44 & 92.55 & 200.60 & 196.79 & 0.3063 & 3.265 \\
\alpagasusdollyone & 2,996 & 111.91 & 68.08 & 106.38 & 43.83 & 107.53 & 1.5530 & 0.6439 \\ \rowcolor{Gray}
\alpagasusdollytwo & 9,229 & 73.40 & 56.62 & 48.91 & 16.79 & 11.33 & 3.3727 & 0.2965 \\
\alpagasusalpaca & 5,305 & 48.29 & 30.81 & 34.44 & 17.48 & 12.45 & 1.7631 & 0.5672 \\ \rowcolor{Gray}
\tulu  & 326,181 & 541.16 & 343.56 & 575.32 & 197.60 & 345.99 & 1.7387 & 0.5751 \\
\tulu (10\%) & 32,618 & 517.45 & 338.96 & 562.74 & 178.49 & 345.72 & 1.8991 & 0.5266 \\ \rowcolor{Gray}
\tulu (50\%) & 163,090 & 515.63 & 340.67 & 571.06 & 174.97 & 343.45 & 1.9470 & 0.5136 \\
\tulu (20\%) & 65,236 & 504.56 & 336.89 & 562.46 & 167.68 & 331.24 & 2.0092 & 0.4977 \\ \rowcolor{Gray}
\wizardlm & 30,000 & 350.05 & 258.35 & 182.98 & 91.71 & 86.09 & 2.8170 & 0.3550 \\
\sharegpt & 50,000 & 1035.39 & 831.15 & 757.10 & 204.24 & 344.51 & 4.0695 & 0.2457 \\ \rowcolor{Gray}
% \flan & 50,000 & 357.92 & 16.60 & 38.59 & 341.31 & 359.46 & 0.0487 & 20.555 \\
\science & 7,544 & 1196.08 & 46.46 & 57.34 & 1149.62 & 905.99 & 0.0404 & 24.7417 \\ 
\alpaca & 52,002 & 63.77 & 45.18 & 44.97 & 18.59 & 12.42 & 2.4302 & 0.4115 \\ \rowcolor{Gray}
\codealpaca & 20,022 & 49.74 & 27.40 & 27.35 & 22.34 & 10.67 & 1.2262 & 0.8156 \\
\bottomrule
\end{tabular}
}
\end{table}

\section{Evaluation Datasets and Details}
\label{neurips2024:sec:evaluation}

We use the open-source repositories, LM-Evaluation Harness\footnote{\url{https://github.com/EleutherAI/lm-evaluation-harness}} and Huggingface Dataset\footnote{\url{https://huggingface.co/docs/datasets}} as the evaluation tools. 
We describe our evaluation setup below:

\paragraph{MMLU.} We evaluate the model using the dataset at the huggingface dataset \footnote{\url{https://huggingface.co/datasets/hails/mmlu_no_train}}.
We follow the protocol outlined in HuggingFace Open LLM Leaderboard \footnote{\url{https://huggingface.co/spaces/HuggingFaceH4/open_llm_leaderboard}}. 
The evaluation uses multiple-choice questions formatted as the question followed by four choices (A, B, C, D) and prompting for an answer. 
We calculate the mean accuracy (acc) across test examples.

\paragraph{BBH.} The model evaluation utilizes the dataset at the huggingface dataset \footnote{\url{https://huggingface.co/datasets/lukaemon/bbh}}, specifically tested on the `test` split without the use of few-shot examples. 
We follow the setup in previous works \cite{ivison2023camels,suzgun-etal-2023-challenging}.
The evaluation metric is the exact match score, averaged (mean) to assess performance.
Generation is constrained to a maximum of 1024 tokens, with termination upon encountering specific delimiters such as "</s>", "Q", or double newlines. 
The generation is greedy decoding (temperature set to 0.0) and does not use sampling. 
Answer extraction employs regex patterns to identify responses immediately following "the answer is" and captures only the first occurrence.

\paragraph{GSM8K.} We evaluate using the dataset at the huggingface dataset \footnote{\url{https://huggingface.co/datasets/gsm8k}}, focusing on arithmetic problem-solving in the `test` split.
We follow the HuggingFace Open LLM Leaderboard to 8 few-shot examples. 
Exact match is the chosen metric, with case insensitivity and select regex-based filtering of common punctuation and formatting characters to ensure precise validation of numerical answers. 
The primary focus is on extracting and comparing the final numerical answer to the model's output using a strict regex-based match setup.

\paragraph{HumanEval.} 
We evaluate using the dataset and the evaluation code from the previous work \cite{ivison2023camels}.
We report the performance of the pass@1.
We perform the decoding using two different temperatures, 0.1 and 0.7. We report the better pass@1 from these two decoding results.

\paragraph{ARC.} The evaluation setup for the dataset at the huggingface dataset \footnote{\url{https://huggingface.co/datasets/allenai/ai2_arc}} utilizes a multiple-choice format.
We follow the HuggingFace Open LLM Leaderboard to 25 few-shot examples. 
The performance metric used is mean normalized accuracy (acc\_norm).

\paragraph{CoQA.} We conduct the model evaluation on the dataset at the huggingface dataset \footnote{\url{https://huggingface.co/datasets/EleutherAI/coqa}}. 
We follow the HuggingFace Open LLM Leaderboard to 0 few-shot examples. 
The output generation terminates upon encountering a new line followed by "Q:". 
The mean F1 score is used as the evaluation metric.

\paragraph{PIQA.} Evaluation on the dataset at the huggingface dataset \footnote{\url{https://huggingface.co/datasets/piqa}} involves a multiple-choice. 
The evaluation incorporates 10 few-shot examples, according to the LIMIT \cite{jha2023limit}. 
Performance is measured using the mean normalized accuracy (acc\_norm).

\paragraph{OpenBookQA.} The dataset at the huggingface dataset \footnote{\url{https://huggingface.co/datasets/openbookqa}} is evaluated in a multiple-choice format. 
The mean normalized accuracy (acc\_norm) is used as the evaluation metric.

\paragraph{LAMBADA.} The evaluation of the model on the dataset at the huggingface dataset \footnote{\url{https://huggingface.co/datasets/lambada}} is performed using a loglikelihood output type. 
The mean accuracy is used as the evaluation metric.

\paragraph{HellaSwag.} In the `hellaswag` dataset at the huggingface dataset \footnote{\url{https://huggingface.co/datasets/hellaswag}}, model evaluation is conducted using a multiple-choice format. 
We follow the HuggingFace Open LLM Leaderboard to 10 few-shot examples. 
The mean normalized accuracy (acc\_norm) is used as the evaluation metric.

\paragraph{The Winograd Schema Challenge.} The evaluation is conducted using a multiple-choice format on the `test` split at the huggingface dataset \footnote{\url{https://huggingface.co/datasets/winograd_wsc}}. 
The mean accuracy is used as the evaluation metric.

\paragraph{Winogrande.} The `winogrande` dataset is assessed using a multiple-choice format at the huggingface dataset \footnote{\url{https://huggingface.co/datasets/winogrande}}. 
We follow the HuggingFace Open LLM Leaderboard to 5 few-shot examples. 
The mean accuracy is used as the evaluation metric.

\paragraph{LAMBADA.} For this dataset, evaluation is conducted using the loglikelihood output type on the `test` split at the huggingface dataset \footnote{\url{https://huggingface.co/datasets/EleutherAI/lambada_openai}}. 
This variant focuses on predicting the last word of text passages in English. 
The mean accuracy is used as the evaluation metric.

\paragraph{Translation Benchmarks WMT.} The evaluation of the translation capabilities is performed on the WMT 2014\footnote{\url{https://huggingface.co/datasets/wmt14}} and WMT 2016\footnote{\url{https://huggingface.co/datasets/wmt16}} datasets at the huggingface dataset.
Here we use the `ter` score as the evaluation metric.

\paragraph{TruthfulQA.} 
We use the dataset at the huggingface dataset \footnote{\url{https://huggingface.co/datasets/truthful_qa}}.
We follow the setup at the HuggingFace Open LLM Leaderboard using the 6 few-shot examples. 
The mean accuracy is used as the evaluation metric.

\paragraph{ToxiGen.} 
We use the dataset at the huggingface dataset \footnote{\url{https://huggingface.co/datasets/skg/toxigen-data}}.
The task is assessed using a multiple-choice framework to evaluate the model's capability to identify hateful content in text statements. 
The mean accuracy is used as the evaluation metric.

\paragraph{Hendrycks Ethics.} 
We use the dataset at the huggingface dataset \footnote{\url{https://huggingface.co/datasets/EleutherAI/hendrycks_ethics}}, with a multiple-choice format.
The model aims to detect whether described actions in various contexts are ethically wrong. 
The prompt format integrates a specific scenario followed by a structured question: "Is this wrong?" and then prompts for an answer with options 'no' or 'yes'. 
The mean accuracy is used as the evaluation metric.

\section{Implementation Details}
\label{neurips2024:sec:implementation_details}
\paragraph{Experimental Design for Figure \ref{neurips2024:fig:insight} Left.}
Here we present a detailed experimental design for Figure \ref{neurips2024:fig:insight} Left. 
We perform experiments on a variety of datasets, including \lima, \alpagasusdollyone, \alpagasusdollytwo, \alpagasusalpaca, \mmluchat, \tydiqa, \bbhicl, \tulu, \codealpaca, \alpaca, \science, \wizardlm, and \sharegpt.
Furthermore, to evaluate the effectiveness of \im on datasets with different instruction-to-output length ratios, we select three subsets from \tulu. Each subset contains 3,000 training examples, with instruction-to-output length ratios of approximately 5, 10, and 15, respectively.

\paragraph{Experimental Design for Figure \ref{neurips2024:fig:insight} Right.}
Here we provide a detailed experimental design for Figure \ref{neurips2024:fig:insight} Right.
We strategically sampled varying sizes of training examples from the \tulu dataset to investigate the effectiveness of \im with different sizes training examples. 
Starting with approximately 320,000 examples in the \tulu dataset, we creates subsets of data ranging from as few as 1,000 to as many as 35,000 examples. 
These subsets were selected randomly, ensuring a representative mix across different scales. 
We adhered to a fixed instruction-to-output length ratio of approximately 10 to maintain consistency in training conditions across all samples. 
We train the \textsc{Llama-2-7B-Base} on all these subsets and evaluate them respectively.

\begin{table}[h]
\centering
\caption{Hyperparameters and configurations for supervised fine-tuning.} 
\label{neurips2024:table:hyperparameters}
\resizebox{0.5\textwidth}{!}{
\begin{tabular}{cc}
\toprule
\textbf{Hyperparameter} & \textbf{Assignment}  \\
\midrule
GPUs & 2 or 4 A100 80G GPUs \\
\midrule
Batch size per GPU & 1 \\
\midrule
Total batch size & 128 \\
\midrule
Number of epochs & 2, 3, or 10 \\
\midrule
Maximum sequence length & 2048 \\
\midrule
Learning rate & $2 \times 10^{-5}$ \\
\midrule
Learning rate optimizer & AdamW \\
\midrule
Adam epsilon & 1e-6 \\
\midrule
Adam beta weights & 0.9, 0.98 \\
\midrule
Learning rate scheduler & Linear with warmup \\
\midrule
Warmup proportion & 0.03 \\
\midrule
Weight decay & 0 \\
\midrule
Mixed precision & bf16 \\
\midrule
Gradient accumulation steps & Calculated dynamically \\
\bottomrule
\end{tabular}}
\end{table}

\paragraph{Implementation Details.}
In our study, we fine-tune the LLaMA-2-7B, LLaMA-2-13B and OPT-6.7 model using four A100 80G GPUs, 
with a per-GPU batch size of 1 and a total batch size of 128, employing a learning rate of 2e-5.
Training typically proceeds for 2 epochs with a maximum sequence length of 2048 tokens. 
We utilise gradient accumulation, calculated to effectively distribute training steps across the available hardware, resulting in larger batch sizes despite hardware limitations. 
We employ mixed precision (bf16), linear learning rate scheduling with a warm-up ratio of 0.03, and a weight decay of 0. 
To optimise our training, we use DeepSpeed with a stage 3 configuration without offloading. 
Our setup also includes the usage of Flash Attention \cite{dao2023flashattention2} and slow tokenization to enhance training efficiency and compatibility. 
Our code is implemented using Open-Instruct\footnote{\url{https://github.com/allenai/open-instruct}}, Pytorch\footnote{\url{https://pytorch.org/}} and Huggingface\footnote{\url{https://huggingface.co/}}.
Table \ref{neurips2024:table:hyperparameters} lists the hyperparameters.

\end{document}